%% file: main.tex
\documentclass[11pt]{article}

\usepackage[preprint]{acl}

\usepackage{times}
\usepackage{latexsym}
\usepackage{amsmath}
\usepackage{natbib}
\usepackage{booktabs}
\usepackage{multirow}
\usepackage{makecell}
\usepackage{enumitem}
\usepackage{todonotes}
\usepackage{longtable}
\usepackage{float}
\usepackage{pifont}
\usepackage{amssymb}
\usepackage{pgfplots}
\pgfplotsset{compat=1.18}
\usetikzlibrary{arrows.meta}
\usepackage{tcolorbox}
\tcbuselibrary{breakable}
\usepackage{amssymb}
\usepackage{listings}
\usepackage{algorithm}
\usepackage{algpseudocode}
\usepackage{nicefrac}

\usepackage[scaled=0.92]{helvet}

\usepackage[T1]{fontenc}

\usepackage[utf8]{inputenc}
\usepackage{stfloats}
\usepackage{microtype}
\usepackage{subcaption}

\usepackage{inconsolata}

\usepackage{graphicx}

\DeclareMathOperator*{\argmin}{arg\,min}
\title{Are Multilingual Models Improving? Revisiting Cross-Lingual Transfer}

\author{Prasoon Bajpai\textsuperscript{1}\thanks{\texttt{prbajpai@google.com} \quad \textsuperscript{\dag}\texttt{vpiratla@google.com}},
  Eleftheria Briakou\textsuperscript{1},
  Colin Cherry\textsuperscript{1},
  Preethi Jyothi\textsuperscript{1,2},
  Vihari Piratla\textsuperscript{1,\dag}\\
  \textsuperscript{1}Google DeepMind \\
  \textsuperscript{2}Indian Institute of Technology Bombay}

\newcommand{\eclektic}{ECLeKTic}
\newcommand{\newmgsm}{MGSMv2}
\newcommand{\mmlu}{MMLU-ProX-Lite}
\begin{document}
\maketitle

\begin{abstract}
Cross-lingual transfer is a model's ability to generalize capabilities from well-represented source languages to under-represented target languages. Existing measures of a model's transfer strength conflate improvements in transfer with general improvements to accuracy in the source language. We advocate for an alternate metric that reliably captures transfer strength called Hardness Adjusted Transfer (HAT) Score, and use it to derive multiple insights on factors influencing transfer strength. Our analysis across twenty diverse language models and three popular mainstream multilingual benchmarks argues that 1) transfer in small models is not broken, 2) we are making slower than expected progress in cross-lingual transfer with model size, and 3) we have made clear progress over time.\footnote{Code and project details are available at \url{https://prasoon1207.github.io/hat_xlt}.}

\end{abstract}

\section{Introduction}

Most large language models (LLMs) officially support multiple non-English languages~\citep{comanici2025gemini,singh2025openai, anthropic_claude}. However, significant performance gaps across languages are known to exist~\citep{cohere2024languagegap,hu2025quantifying}, thus putting certain demographics at a technological disadvantage. Cross-lingual transfer (XLT), the ability of an LLM to transfer performance from a source language (such as English) to other target languages, is critical for achieving cross-lingual performance parity~\citep{pires-etal-2019-multilingual,hu2020xtrememassivelymultilingualmultitask,philippy-etal-2023-towards}. Improvements on multilingual benchmarks are commonly interpreted as evidence of improved XLT~\citep{hu2020xtrememassivelymultilingualmultitask}. In this work, we examine this assumption carefully and question whether gains in target language performance necessarily signal improved XLT.


\begin{figure*}[h!]
    \centering
    \includegraphics[width=0.85\linewidth]{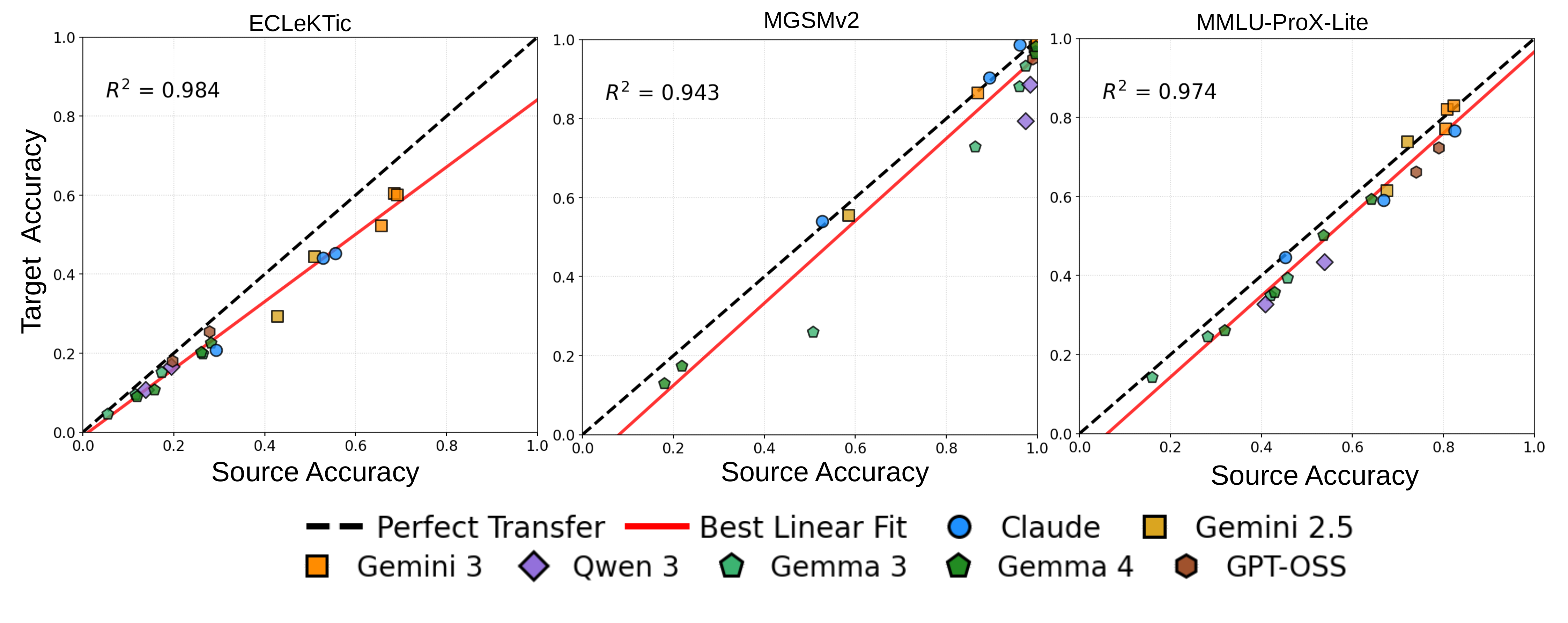}
    \caption{\textbf{Target accuracy scales predictably from source accuracy.} Evaluation with twenty diverse LLMs across multilingual benchmarks \eclektic{}, \newmgsm{}, and \mmlu{} reveals a linear trend.  
    For \newmgsm\  and \mmlu, we set English as the source language, while for \eclektic{}, the source language varies per question as defined by the dataset. Target is a subset of non-source languages. Because the improvements in source and target accuracies are coupled, traditional XLT metrics conflate general source-language improvements with transfer. 
    Please see Section~\ref{sec:xlt_progress} for more details and context.}

    \label{fig:accuracy_line}
\end{figure*}

Transfer strength is characterized by the fraction of performance translating from source to target languages. Therefore, we may declare improvements in transfer if target accuracy improved for the same value of source accuracy. In practice, however, source and target accuracies are linearly coupled with goodness of fit ($R^2$) exceeding 0.94 as shown in Figure~\ref{fig:accuracy_line}, i.e., target accuracy follows predictably from source accuracy. The surprising linear trend is further discussed in Section~\ref{sec:xlt_progress}. The coupled performance growth calls to question the standard measures of XLT such as (1) the absolute performance on target language, or (2) relative performance with source language, because they do not adjust for general improvements to source accuracy.

Both source and target accuracies may increase with general improvements such as factuality, reasoning, or model size, but increased target accuracy, a typical measure of cross-lingual transfer, need not imply progress on transfer. 
Source-target performance gap is another standard measure of transfer. We observe from the left plot of Figure~\ref{fig:accuracy_line} that both target accuracy and performance gap increase as source accuracy increases. While increasing target accuracy implies progress on XLT, increasing gap indicates worsening XLT. Together, they paint a conflicting narrative on XLT progress, which warrants a closer study.  



We argue that a well-calibrated transfer metric should measure improvements in expected target accuracies while adjusting for improvements in source accuracies. More precisely, in order to measure XLT, it is important to disentangle the effect of target improvements that simply follow from source improvements and measure when the model's accuracy lies above the expected correlation line (shown in Figure~\ref{fig:accuracy_line}).

Motivated by these observations, this paper rethinks how we define progress in cross-lingual transfer. Our contributions are summarized below.

\begin{itemize}[noitemsep]
    \item We show that existing measures of cross-lingual gaps do not reflect correct progress on XLT, and derive an alternate measure from first principles called the \textbf{Hardness Adjusted Transfer (HAT)} score. 
    \item Leveraging our calibrated XLT measure, we present insights on XLT influencing model choices such as thinking, parameter counts and language scripts. 
    \item We show a linear trend between source and target accuracies across twenty models and three multilingual datasets spanning factuality and reasoning. This strong linear trend is interesting given the diversity in model class, size and thinking budget. We analyze progress over time and find that \mmlu{} and \newmgsm{} are both nearly solved in terms of XLT, highlighting the need for more challenging multilingual benchmarks. 
\end{itemize}

\section{Related Work}

\paragraph{Multilingual Generalization.}
Recent LLM-era benchmarks, including XTREME~\citep{hu2020xtrememassivelymultilingualmultitask}, XSQuAD~\citep{Artetxe_2020}, MGSM~\citep{shi2022languagemodelsmultilingualchainofthought}, MMLU-ProX~\citep{xuan2025mmluproxmultilingualbenchmarkadvanced}, ECLeKTic~\citep{goldman2025eclekticnovelchallengeset}, and MuBench~\citep{han2025mubenchassessmentmultilingualcapabilities}, shift multilingual evaluation towards assessing generative capabilities such as factual question answering and reasoning. This broader evaluation landscape revealed inconsistencies in model performance across languages of different resources and scripts ~\citep{Xu_2025,hengle2024multilingualneedlehaystackinvestigating, han2025mubenchassessmentmultilingualcapabilities, xuan2025mmluproxmultilingualbenchmarkadvanced}. Complementary analyses suggest that these failures reflect differences in internal alignment, shared multilingual computation, and, in some cases, English-mediated latent processing~\citep{zhao2024largelanguagemodelshandle, ifergan2024beneathsurfaceconsistencyexploring, tang-etal-2024-language}.

\paragraph{Metrics for XLT.} Typically, XLT is measured via target accuracy~\citep{xuan2025mmluproxmultilingualbenchmarkadvanced, hu2020xtrememassivelymultilingualmultitask} or the cross-lingual transfer gap (XLT gap)~\citep{hu2020xtrememassivelymultilingualmultitask, chi2021infoxlminformationtheoreticframeworkcrosslingual}, i.e., the absolute difference between source and target accuracy. Recently, \eclektic{} used a ``Transfer score'' that measures target accuracy conditioned on correct source prediction~\citep{goldman2025eclekticnovelchallengeset}. Complementary white-box work probes XLT through the model’s internal structure: cross-lingual neuron overlap asks whether languages share feature-selective neurons~\citep{wang2024probingemergencecrosslingualalignment}, neuron activation overlap measures whether parallel inputs recruit similar circuits~\citep{hu2025largelanguagemodelscrosslingual}, and sub-network similarity compares the language-specific computation used for transfer~\citep{yun2023xsnscrosslingualtransferprediction}. These techniques require model access and rely on the interpretability of internal activations, making them complementary to black-box, behavioural metrics such as ours.

\paragraph{XLT as Unsupervised Domain Adaptation.}

We are motivated by recent works modelling XLT as an unsupervised domain adaptation problem~\citep{piratla2025rethinking}, where source and target languages define domains with shared task semantics. Several prior works on domain adaptation highlight that target performance cannot be interpreted independently of source competence ~\citep{10.1007/s10994-009-5152-4, mansour2023domainadaptationlearningbounds, miller2021accuracy}. We take inspiration from these studies to propose a more principled metric to measure XLT.

\section{Target and Source Accuracies are Coupled}
\label{sec:xlt_progress}

  For all the models and datasets described in Section~\ref{sec:experimental_setup}, we plotted the source and target accuracy values in Figure~\ref{fig:accuracy_line}. 
  Also shown in each plot is the best linear fit in red and the line of perfect transfer, shown as a dashed line. The figure helps illuminate the nature of cross-lingual gaps at a glance, (1) target accuracy improves with source accuracy, and (2) target accuracy is often worse than that of source but not by much.  
  
  We observe that source and target accuracies follow a strong linear trend for all datasets and models (with the goodness of fit $R^2$ exceeding 0.94). This  phenomenon is reminiscent of out-of-distribution accuracy falling on a line with in-distribution accuracy~\citep{objectnet,miller2021accuracy,salaudeen2024imagenot}. 
  It is surprising that the strong linear trend continues to hold despite no clean separation of train-test splits and across various model architectures.  

Since source and target accuracies are coupled, we cannot declare progress in XLT through source or target accuracy in isolation. An ideal measure of improvement in transfer must discount the improvements to target accuracy arising from general improvements in source accuracy. In the next section, we derive the existing XLT measures from first principles, explain why they are misaligned for measuring XLT progress, and introduce our proposed metric.

\section{Proposed XLT Metric: HAT Score}
\label{sec:proposed_xlt_metric}
Let $S, T_\ell \in [0, 1]$ denote the model's instance-level source and target language $\ell$ pass-rates, respectively. Pass-rate is the fraction of times the model responds correctly, for instance the pass-rate (@10) of 0.3 indicates the model is correct only three out of ten times on a particular prompt. 
When a model achieves language parity, the model is expected to achieve comparable pass-rates in both source and target, i.e., $\mathbb{E}[T_\ell\mid S=s]=s, \forall \ell, s\in [0, 1]$.

We will formally define standard measures of XLT and illustrate how they can conflate general capability gains with cross-lingual transfer progress.

\paragraph{(1) Average Target Accuracy.}
The average target accuracy for a given model $M$ on a dataset $D$ is expressed as $\mathbb{E}_{D,\ell}[T_\ell|M]$, which can also be written as a  weighted aggregation of conditional expectations as shown below.
\begin{align}
  &\text{Avg. } \text{target accuracy}\triangleq \mathbb{E}_{D,\ell}[T_\ell\mid M]\nonumber\\
  &= \mathbb{E}_S[\mathbb{E}_{D,\ell}[T_\ell\mid S, M]]\nonumber\\
                            &= \int_0^1 \mathbb{E}_{D,\ell}[T_\ell\mid S=s, M]p_{S}(s\mid M)\,ds
                            \label{eqn:avg_tgt}
\end{align}
For the sake of argument, consider two models: $M_1, M_2$ with comparable source-conditional target accuracies, i.e., $\mathbb{E}[T_\ell\mid S=s, M_1] = \mathbb{E}[T_\ell\mid S=s, M_2], \; \forall s\in [0, 1]$ but with the difference that $p_{S}(s\mid M_1)$ is shifted to the right of $p_{S}(s\mid M_2)$. Since the target accuracy is known to increase with source accuracy (Figure~\ref{fig:accuracy_line}), the target accuracy for $M_1$ will be greater than that of $M_2$, despite comparable conditional accuracies.

\paragraph{(2) XLT Gap.}
Another standard measure of XLT progress is the difference between averaged source and target accuracies (XLT gap). This measure also suffers from the same issues as average target accuracy, as we show below.
\begin{align*}
  &\text{XLT gap}\triangleq \mathbb{E}_{D,\ell}[S-T_\ell\mid M]\\
             & =\int_0^1 (s - \mathbb{E}_{D,\ell}[T_\ell\mid S=s,M])p_S(s\mid M)\,ds.
\end{align*}



\noindent Following the same argument outlined for the average target accuracy, if we consider two models $M_1, M_2$ such that $s - \mathbb{E}[T_\ell \mid S = s, M_1] = s - \mathbb{E}[T_\ell \mid S = s, M_2]$ but with higher source accuracy for $M_1$, i.e., $p_S(s\mid M_1)$ is tilted right. The XLT gap, given that the conditional gap $s - \mathbb{E}[T_\ell \mid S = s, M]$  varies with $s$, will show variations between $M_1$ and $M_2$, despite their comparable conditional accuracies.


\paragraph{(3) Transfer Score.} 
The transfer score was proposed by \citet{goldman2025eclekticnovelchallengeset} to quantify the proportion of target questions answered correctly among questions that were answered correctly in the source language. 

We express the score using our framework by deriving the fraction of examples correct in source and fraction correct in both source and target. 
For any given value of $S=s$ and target language $\ell$, the fraction of correct source responses is $s$, and the fraction of examples correct in both source and target is $s \times \mathbb{E}[T_\ell \mid S=s, M]$. Integrating over the full spectrum of $S$, the theoretical transfer score ($\tau_\ell$) is formulated as:
\begin{align*}
    \tau_\ell &\triangleq \frac{\int_0^1 p_S(s \mid M) \mathbb{E}[T_\ell \mid S=s, M] s \, ds}{\int_0^1 p_S(s \mid M) s \, ds}.\\
    \tau &= \mathbb{E}_\ell[\tau_\ell]
\end{align*}
We illustrate the (erroneous) sensitivity of $\tau_\ell$ to source accuracy with an example. Consider a model with perfect transfer, i.e., $\mathbb{E}[T_\ell\mid S=s, M]=s, \quad \forall s\in [0,1]$. Depending on the distribution $p_S(s)$, the $\tau_\ell$ value can vary between \nicefrac{2}{3} for uniformly distributed $p_S(s)$ and 1 when source accuracy is near perfect.



We propose an alternate XLT metric called the \textbf{Hardness Adjusted Transfer Score (HAT Score)}, designed to better isolate transfer strength by controlling for source-language difficulty.


\subsection{HAT Score and Transfer Profile}
\label{sec:hat_score_intro}

\paragraph{HAT Score.}
We propose to replace $p(s)$ in Equation~\eqref{eqn:avg_tgt} with a uniform distribution and define our new metric as the normalized expected target accuracy shown below.
\begin{equation}
    \text{HAT Score} = 2\int_{0}^1 \mathbb{E}_{D,\ell}[T_\ell \mid S=s, M] \, ds
\label{eqn:hat}
\end{equation}
The multiplier two ensures that the HAT score is bounded in the range $[0, 1]$ and is comparable to other metrics. 
\paragraph{Empirical Estimation of HAT score.} 
HAT score computation, unlike standard measures, requires an estimate for the conditional: $\mathbb{E}[T_\ell\mid S,M]$. For any given example $x^{(i)}$ from the dataset, we denote the instantiation of the same prompt in source and target languages as $x_s^{(i)}, x_\ell^{(i)}$, respectively, along with their corresponding ground-truth outputs $y_s^{(i)}, y_\ell^{(i)}$. We approximate $S$ and $T_\ell$ for a given example with pass-rates across $N=10$ promptings of model $M$ with their corresponding prompts $x_s^{(i)}, x_\ell^{(i)}$. 

We use a linear regressor for fitting $T_\ell$ from $S$ because (a) it works reasonably well, with the goodness of fit ($R^2$) typically being above 0.6, and (b) it is a low-variance estimator that can handle sparse observations better. We further impose a constraint on the estimator that when source language performance hits the minimum value of zero, i.e., $S=0$, then $T$ must follow suit, i.e., $T_\ell=0$. More formally:

{\small
\begin{align}
&\left((x_s^{(i)}, y_s^{(i)}), (x_\ell^{(i)}, y_\ell^{(i)})\right)\sim D \nonumber \\
s^{(i)} &= \sum_{n=1}^N\frac{\mathbb{J}(M(x_s^{(i)})_n, y_s^{(i)})}{N} \nonumber \\ 
t^{(i)} &= \mathbb{E}_\ell\left[\sum_{n=1}^N\frac{\mathbb{J}(M(x_\ell^{(i)})_n, y_\ell^{(i)})}{N}\right] \nonumber \\
\hat{m} &= \argmin_m \sum_i \Big(t^{(i)} - ms^{(i)}\Big)^2 \\
\text{HAT Score} &= 2\int_{0}^1 \hat{m}s\, ds = \hat{m}
\label{eqn:hat_estimator}
\end{align}
}
\noindent where $M$ denotes the model, $\mathbb{J}$ denotes the judge evaluator we use, and $M(x)_n$ denotes the model's response for the $n^{\text{th}}$ pass of input $x$.\footnote{For the linear fit, we used the standard numpy library in Python~\citep{harris2020array}.}



In Appendix~\ref{sec:appendix_estimator_choice}, we compare our constrained linear model with unconstrained linear regression and Kernel Ridge Regression (KRR), and show that it achieves comparable goodness of fit while preserving model rankings across benchmarks.

\paragraph{Transfer Profile.} A handy graphic is to plot the expected target accuracy: $\mathbb{E}_{D,\ell}[T_\ell\mid S=s, M]$ on the Y-axis and source accuracy $s$ on the X-axis. We refer to the regression curve defined by ($s$, $\mathbb{E}[T_\ell\mid S=s, M]$) as the transfer profile. If the model has perfect transfer, then the transfer profile aligns with the $y=x$ line. If the transfer profile for model $M_1$ falls strictly below that of model $M_2$, we may claim better transfer for model $M_2$. 

\section{Experimental Details}
\label{sec:experimental_setup}

\subsection{Models}
\label{sec:experiments_models}
We evaluate 20 models spanning five families: \texttt{Gemini} ~\citep{comanici2025gemini, Google2025Gemini3} (5), \texttt{Gemma} ~\citep{gemmateam2025gemma3technicalreport, gemma42026} (8), \texttt{Qwen} (2) \cite{yang2025qwen3technicalreport}, \texttt{GPT-OSS} (2) \cite{openai2025gptoss120bgptoss20bmodel}, and \texttt{Claude} ~\citep{anthropic_claude} (3), to ensure our findings generalize across architectures, scales, and access paradigms (open-weight vs. proprietary API). The \texttt{Gemma-3} series (1B, 4B, 12B, 27B) provides a dense scaling ladder for studying how XLT evolves with model size within a single architecture. \texttt{Gemma-4} extends this with both dense (E2B, E4B, 31B) and mixture-of-experts (26B-A4B) variants. \texttt{Gemini-3-Flash} is evaluated at three reasoning depths (default, low, minimal) to isolate the effect of thinking budget on XLT. The \texttt{Claude} family (\texttt{Haiku 4.5}, \texttt{Sonnet 4.6}, \texttt{Opus 4.7}) and \texttt{GPT-OSS} (20B, 120B) provide further independent scaling axes along model size and thinking budget. Our model selection also spans old-to-new timelines between 2024 to the latest 2026 \texttt{Gemma-4} release.   

\subsection{Datasets}
\label{sec:experiments_datasets}
We evaluate on three multilingual benchmarks, each targeting a distinct capability: \eclektic{} (cross-lingual factoid QA) ~\citep{goldman2025eclekticnovelchallengeset}, \newmgsm{} (multilingual math reasoning), and \mmlu{} (multilingual multiple-choice probing conceptual knowledge and requires reasoning)~\citep{xuan2025mmluproxmultilingualbenchmarkadvanced}. The original MGSM benchmark~\citep{shi2022languagemodelsmultilingualchainofthought} is known to contain translation and validation-related artifacts, challenging the accurate estimation of cross-lingual gaps. \citet{peter2025mindgapnottranslation} released a version fixing many issues with MGSM, which we use for our evaluation and refer to as \newmgsm{}. 

\paragraph{Source and Target Definitions for \newmgsm{}, \mmlu{}.}
For both \newmgsm{} and \mmlu{}, we consider English as the source language and the target split comprises all available non-English languages in the dataset. Despite no clean train-test separation with LLMs, we believe English as source language is justified given the dominance of reasoning and/or conceptual knowledge focus in these two datasets.
\newmgsm{} contains ten non-English languages while \mmlu{} contains twenty eight. Please see Appendix~\ref{sec:appendix_experimental_details} for the full list.

\paragraph{Source and Target for \eclektic{}.}
Source-target separation is more clearly specified for \eclektic{}. The dataset collects questions from Wikipedia pages that are only available in one language. For each factual question, the source is the language of the original Wikipedia article while target comprises human-verified translations of the question and response into eleven other languages. 

\subsection{Experimental Setup}
\paragraph{Response Validation.}
Multiple choice questions in \mmlu{} can be validated with an exact match. On the other hand, \eclektic{} and \newmgsm{} require an LLM-based judge~\citep{gu2025surveyllmasajudge} for validation. We mark responses via majority voting with a model ensemble: \texttt{Gemini-3-Flash (Low)}, \texttt{GPT-OSS-20B}, and \texttt{Claude-Sonnet-4.6}. For response validation, we prompt the judge model with the question, response and correct answer; this is a much simpler task than answering the original question. We employed an ensemble of judge models instead of one to suppress any self-preferential bias. We provide complete details of the evaluation protocol, including the answer extraction and answer scoring prompts in Appendix \ref{sec:appendix_experimental_details}. An empirical evaluation of the judge system's reliability against a deterministic exact-match subset of ECLeKTic is provided in Appendix \ref{sec:appendix_judge_validation} 



\paragraph{Inference Related Parameters.} When validating different models, we retain a common temperature (0.7) for all. We also retain the default thinking budget (unless explicitly specified) and sampling strategy.  We provide dataset statistics and further experimental details in Appendix \ref{sec:appendix_experimental_details}.

\subsection{Statistical Significance}
\label{sec:statistical_significance}
All metrics are accompanied by 95\% confidence intervals estimated using the bootstrap method~\citep{bootstrap_efron}. We resample the observations with replacement $1000$ times to estimate the distribution of any aggregated measure, which yields their mean and confidence interval. 


\begin{figure}[htb]
    \centering
    \begin{subfigure}[c]{0.85\linewidth} 
        \centering
        \includegraphics[width=\linewidth]{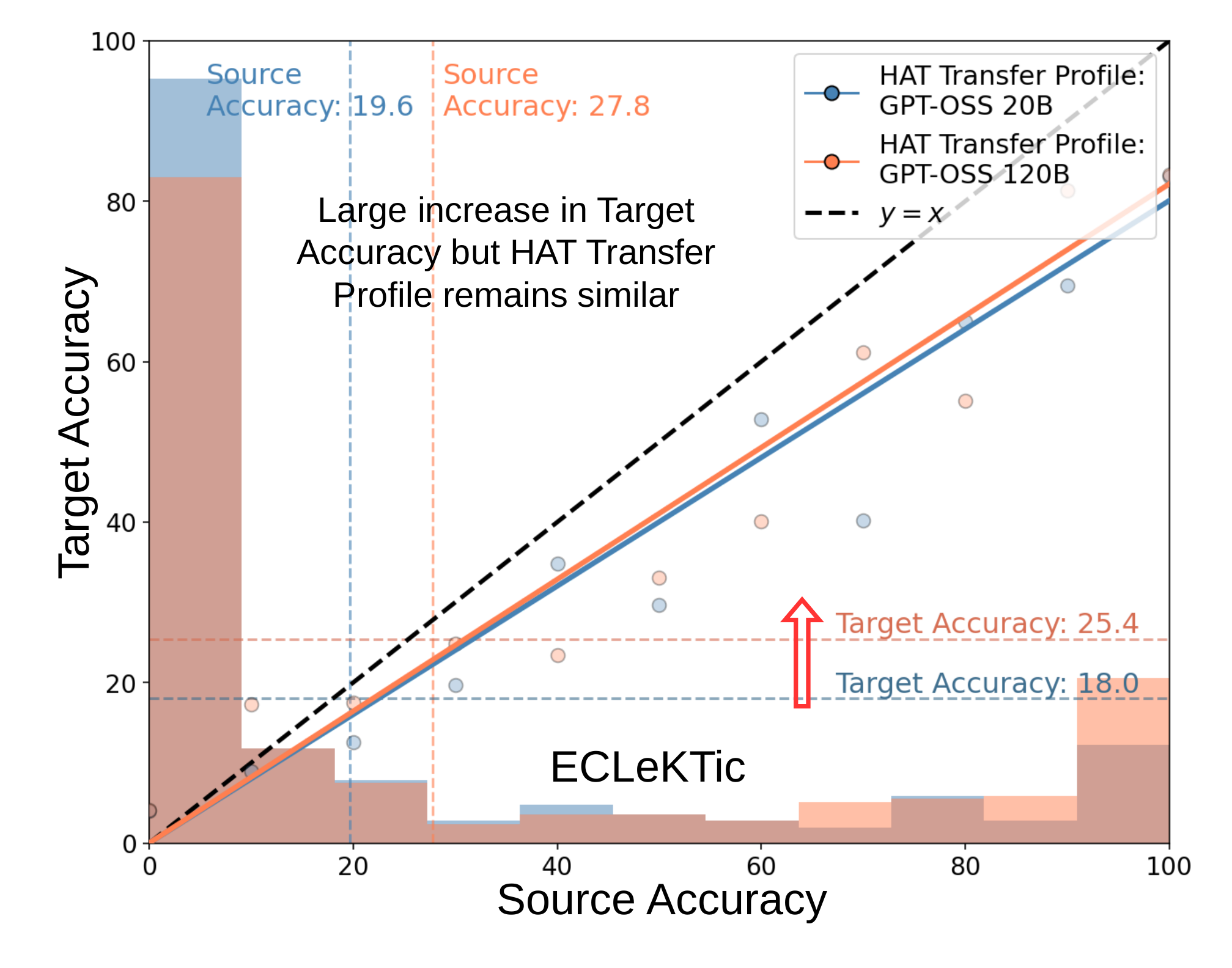}
        \caption{Target accuracy indicates progress despite overlapping transfer profiles. HAT score remains unchanged, thereby better representing progress.}
        \label{fig:metric_comparison_a}
    \end{subfigure}
    
    \vspace{0.2cm} 
    
    \begin{subfigure}[c]{0.85\linewidth} 
        \centering
        \includegraphics[width=\linewidth]{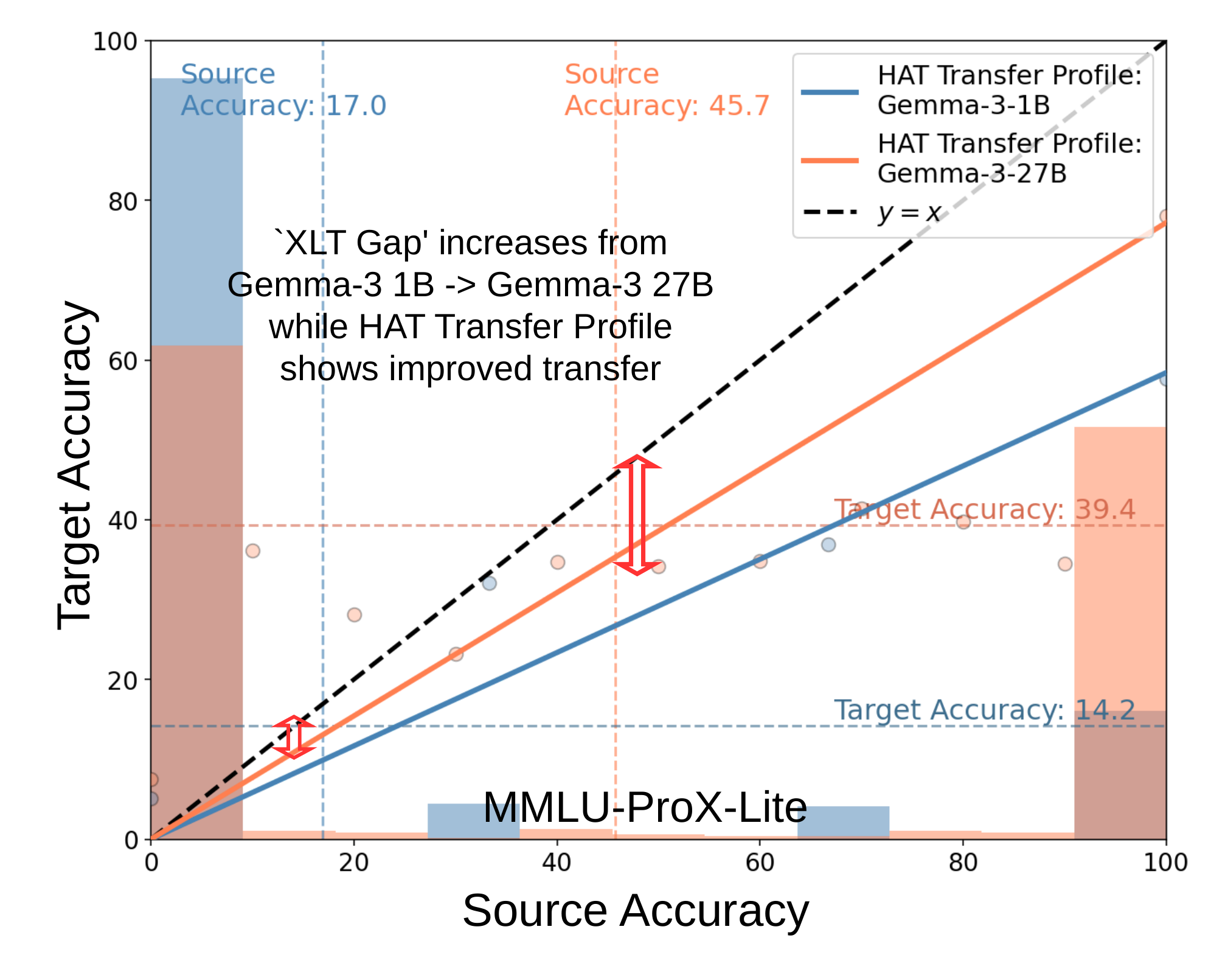}
        \caption{XLT gap falsely indicates worsening transfer although there is a clear improvement in the transfer profile.}
        \label{fig:metric_comparison_b}
    \end{subfigure}
    
    \caption{\textbf{Improvements in source accuracy inflate traditional metrics like (a) average target accuracy and (b) XLT gap}. Please refer to Section \ref{sec:why_hat} for more details.}
    \label{fig:metric_comparison}
\end{figure}

\section{Results}

We establish the reliability of HAT score in Section~\ref{sec:why_hat}.  
We then utilize HAT score to address interesting questions related to progress. We track the improvements on XLT with model size in Section~\ref{sec:xlt_model_scale}, over time in Section~\ref{sec:xlt_over_time}, and conclude with an overall comparison across all models in Section~\ref{sec:xlt_overall}. We present  comprehensive cross-lingual evaluation results in Tables~\ref{tab:eclektic_full_metrics} (\eclektic), \ref{tab:mmlu_full_metrics} (\mmlu) and \ref{tab:mgsm2_full_metrics} (\newmgsm) of Appendix~\ref{sec:full_results}

\subsection{HAT Score Reliability}
\label{sec:why_hat}

\begin{figure}[htb]
    \centering
    \begin{subfigure}[c]{0.85\linewidth} 
    \includegraphics[width=\linewidth]{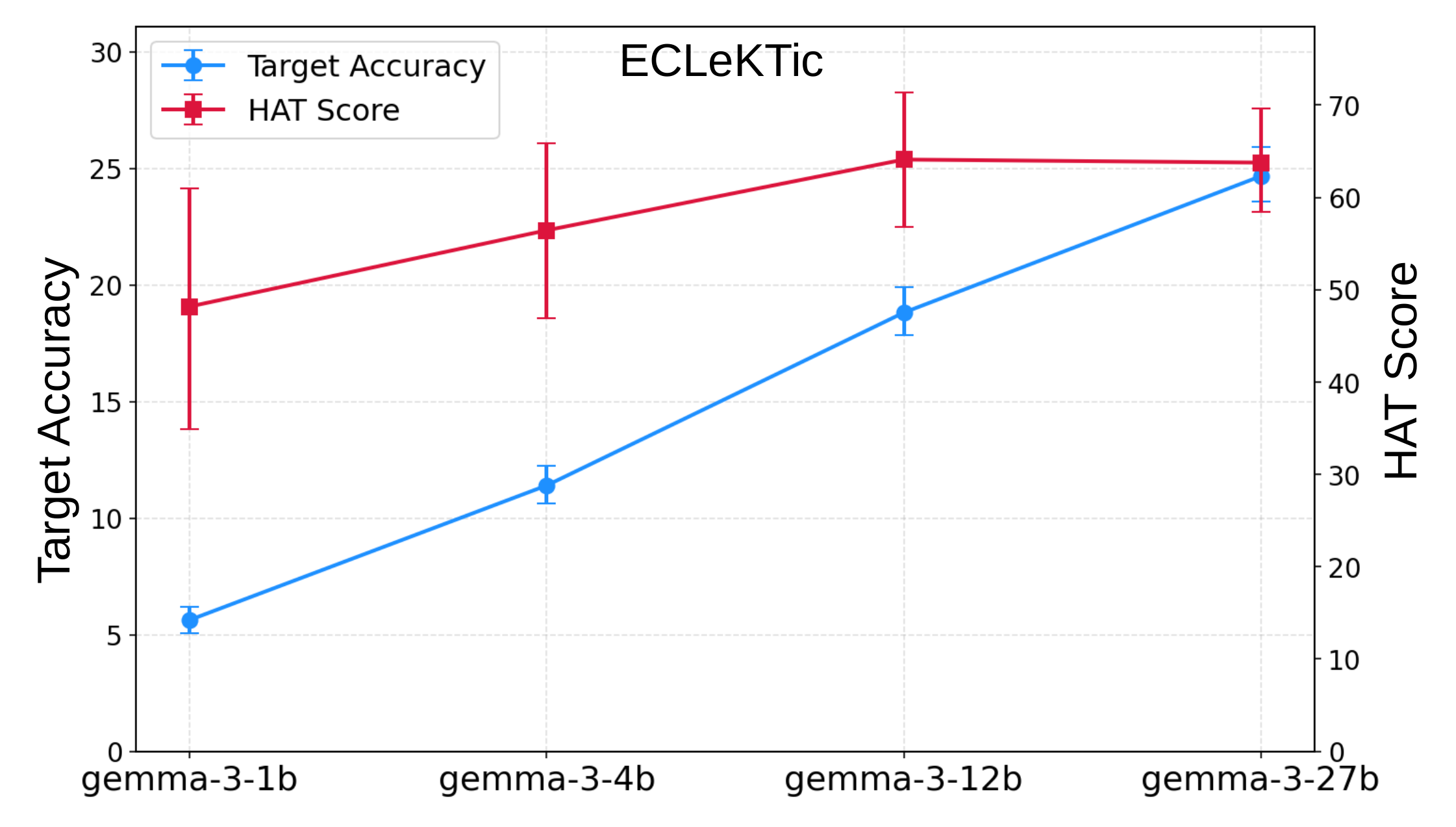}
        \caption{ECLeKTic: Target accuracy inflates while HAT score plateaus.}
        \label{fig:hat_scale_size_a}
    \end{subfigure}
    
    \vspace{0.2cm} 
    
    \begin{subfigure}[c]{0.85\linewidth} 
        \centering
        \includegraphics[width=\linewidth]{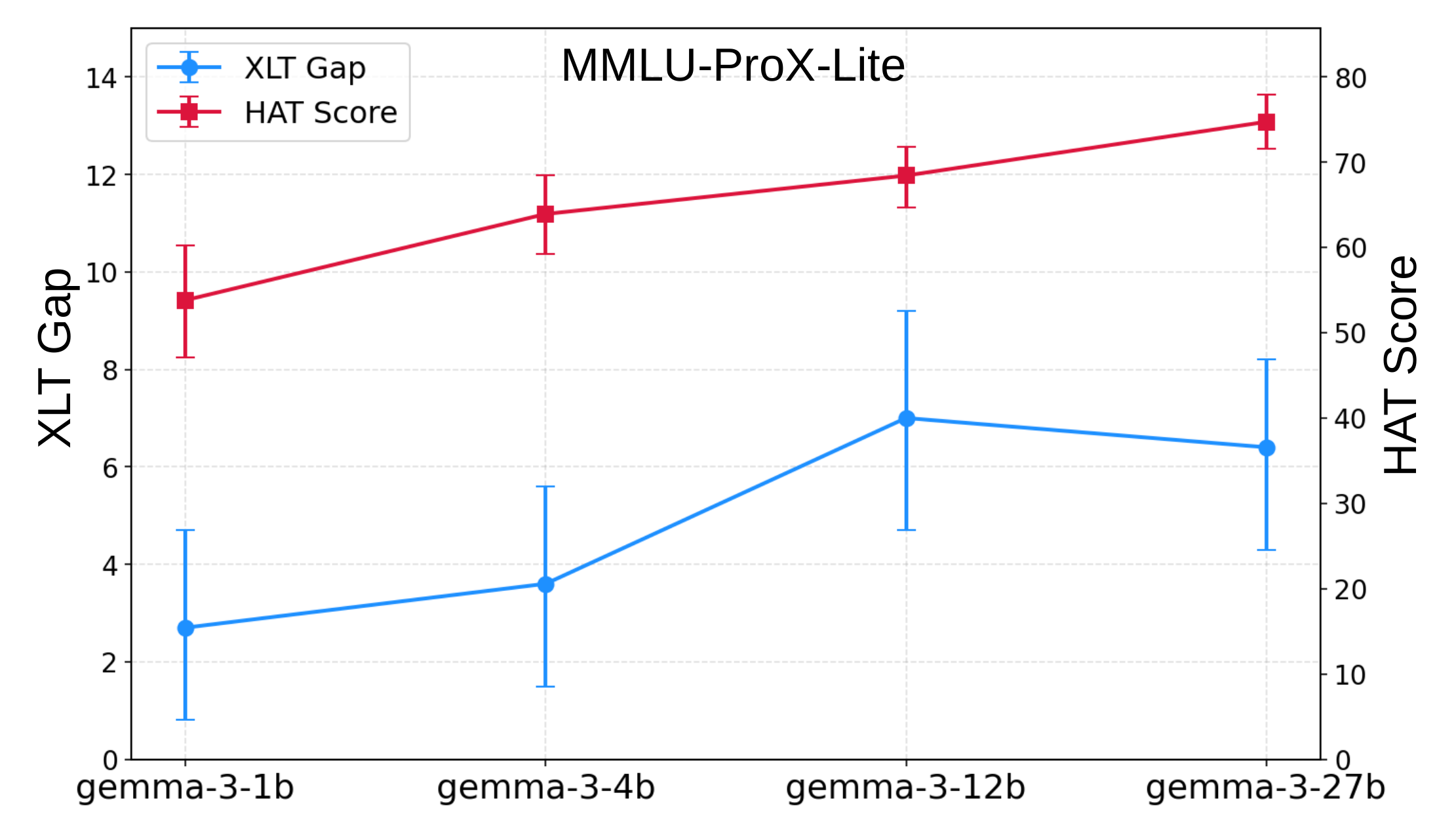}
        \caption{MMLU-ProX-Lite: XLT gap falsely implies regression as models scale.}
        \label{fig:hat_scale_size_b}
    \end{subfigure}
    
    \caption{\textbf{Scaling ladder with HAT score.} Traditional metrics distort XLT progress with model size: (a) target accuracy falsely inflates on ECLeKTic, and (b) the XLT gap falsely shows regression on MMLU-ProX-Lite. The HAT score corrects the confounders, revealing that actual transfer progress is much more gradual (Section~\ref{sec:xlt_model_scale}).}
    \label{fig:hat_scale_size}
\end{figure}

In Figure~\ref{fig:metric_comparison}, we present two comparisons between different pairs of models. In each of the plots, we show the transfer profile $\mathbb{E}[T\mid S]$  for both the models. At the bottom of each plot, we show frequency histograms of source accuracies. 

\paragraph{HAT Score is Preferable over Target Accuracy.} In Figure ~\ref{fig:metric_comparison} (a), we contrast \texttt{GPT-OSS 20B} with \texttt{GPT-OSS 120B} on \eclektic{}. Target accuracy for the 120B model (25.4\%) is significantly larger than that of the 20B model (18.0\%) despite overlapping transfer profiles, i.e., \emph{no improvement in XLT}. The increase in target accuracy with 120B is due to right-shifted source accuracies as visualized by the frequency histogram. HAT score for the two models is close to 0.82, thus indicating no improvement in transfer.

\paragraph{HAT Score is Preferable over Transfer Score.}
In the same Figure~\ref{fig:metric_comparison_a}, transfer score unlike HAT score showed a significant jump from 53.5 to 61 between \textsc{GPT-OSS} models despite overlapping transfer profiles. 
We contrast transfer and HAT Scores for various models on \eclektic{} in Table~\ref{tab:combined-scores} of Appendix~\ref{sec:comparison-eclektic}, where we discuss more differences in detail. 


\paragraph{HAT Score is Preferable over XLT gap.}
 Figure~\ref{fig:metric_comparison} (b) contrasts the \texttt{Gemma-3-1B} model with the \texttt{Gemma-3-27B} model on \mmlu{}. The 27B model clearly improves over the 1B model, which is apparent from its steeper slope. However, the XLT gap increases from 2.7\% for the 1B model to 6.4\% for the 27B model, falsely indicating worsened XLT. The HAT score, which represents the area under the transfer profile, correctly captures this progress with increasing from 58.4 to 77.2.

In all the examples above, HAT score is robust to the shifting frequency histogram of source accuracies. Overall, HAT score is a more reliable indicator of XLT progress.

\subsection{XLT Progress with Model Size Scaling}
\label{sec:xlt_model_scale}
We compare progress with model size among models that belong to the same family to control for any unrelated changes to training recipes. Figure~\ref{fig:hat_scale_size} compares across the \texttt{Gemma-3} class of models. Progress indicated by HAT score in contrast to target accuracy is much slower. This seemingly slow progress is in part because cross-lingual transfer in the smallest model is not as bad as indicated by target accuracy. Nevertheless, we continue to see some progress with model size. This observation aligns with the general trend of improved generalization with larger models~\citep{zhang2016understanding}. 

\paragraph{XLT in Small Models.} Our validation includes some relatively small-sized models in the range of 1B to 4B size: \texttt{Gemma-3-1B, Gemma-3-4B, Qwen-3-4B, Gemma-4-E2B}. We observe from Figures~\ref{fig:hat_scale_size},~\ref{fig:all_hat} that language transfer is not broken at very small scales. The HAT score for the four models on the three datasets is in the 50-80 range meaning the target accuracy is 50-80\% of the source accuracy. Target accuracy, on the other hand, as shown in Figure~\ref{fig:hat_scale_size} provides an overly pessimistic estimate for small models.

\subsection{XLT Progress over Time}
\label{sec:xlt_over_time}

\begin{figure*}
    \centering
    \includegraphics[width=0.32\linewidth]{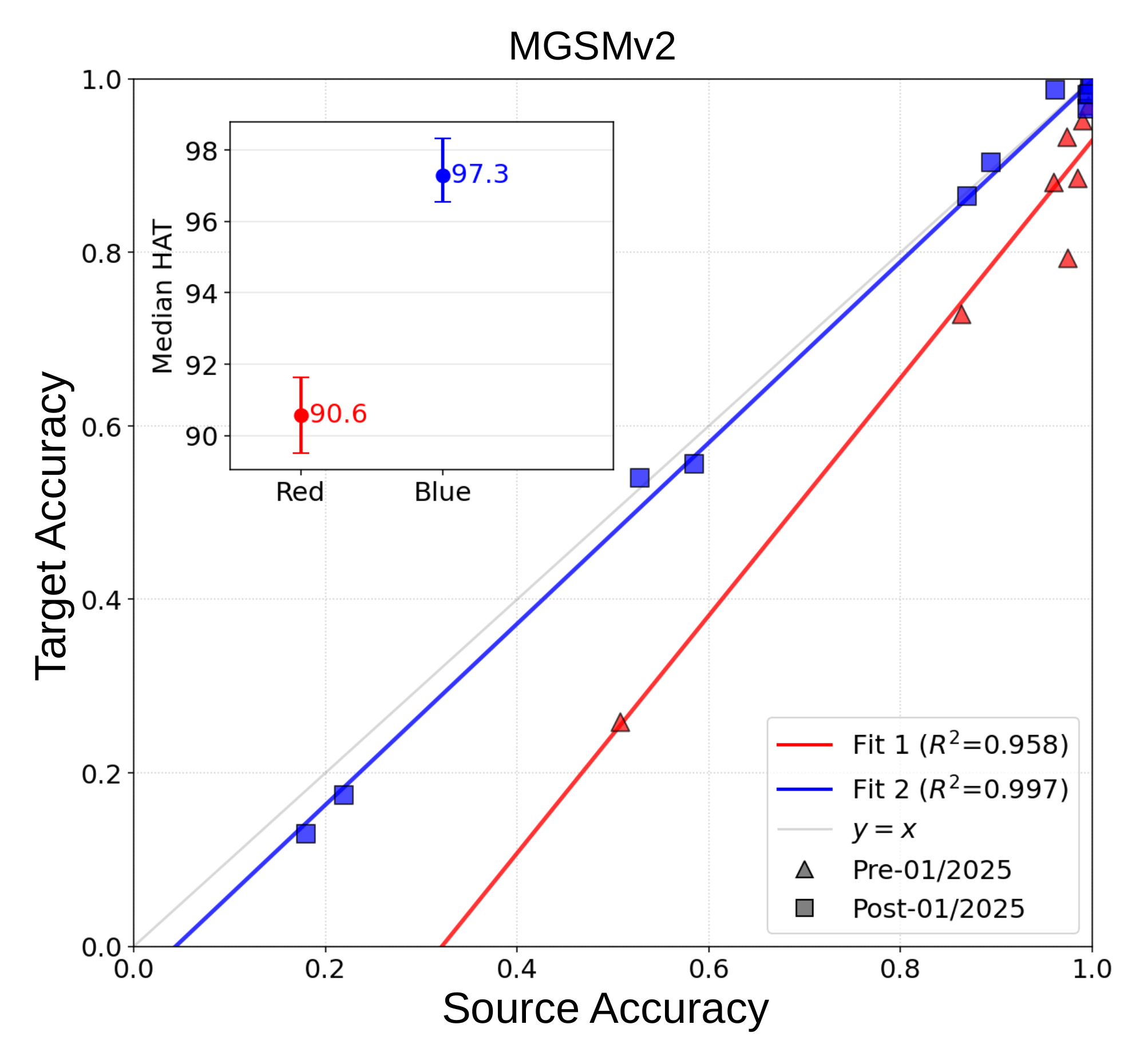}
    \includegraphics[width=0.32\linewidth]{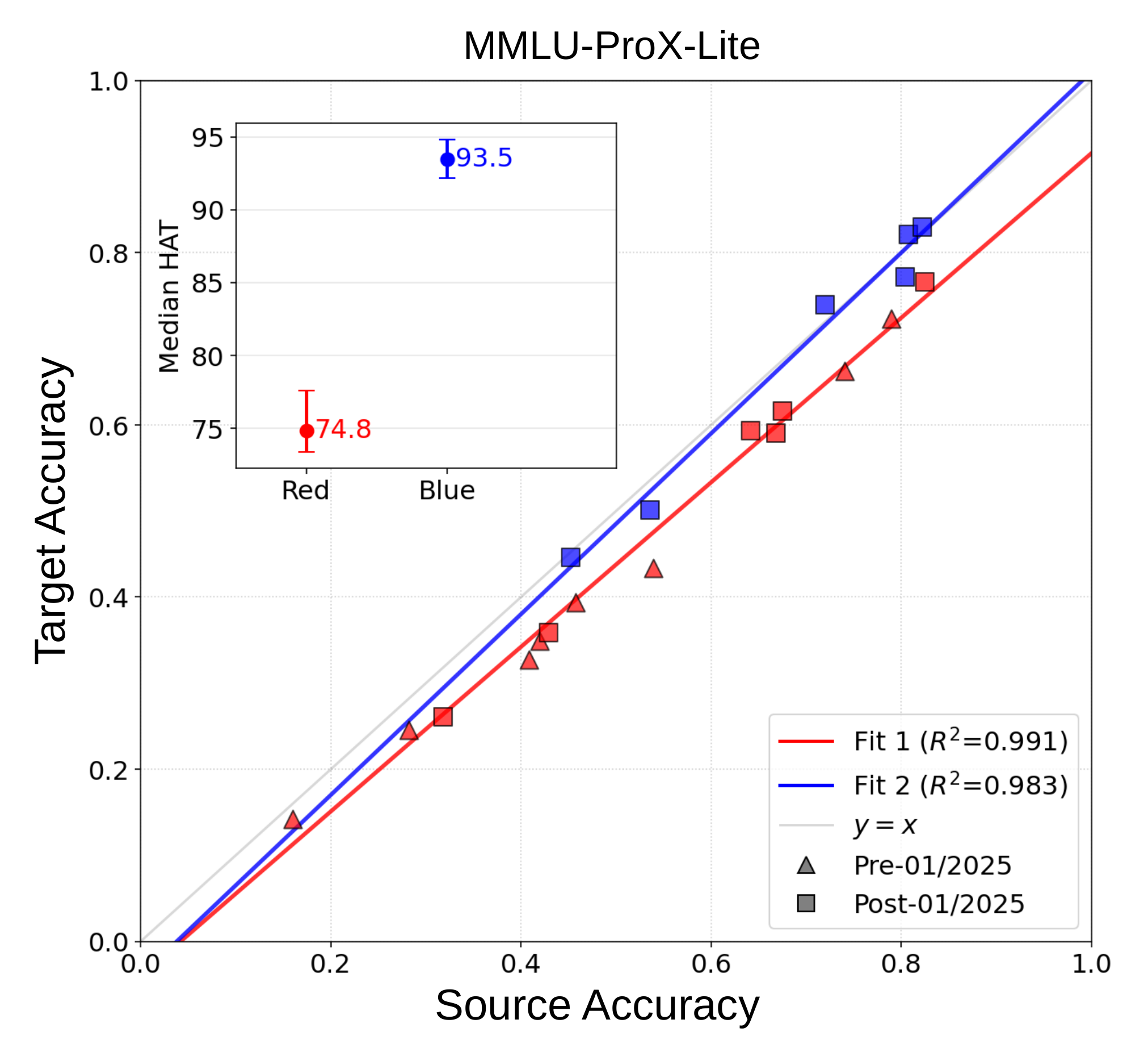}
    \includegraphics[width=0.32\linewidth]{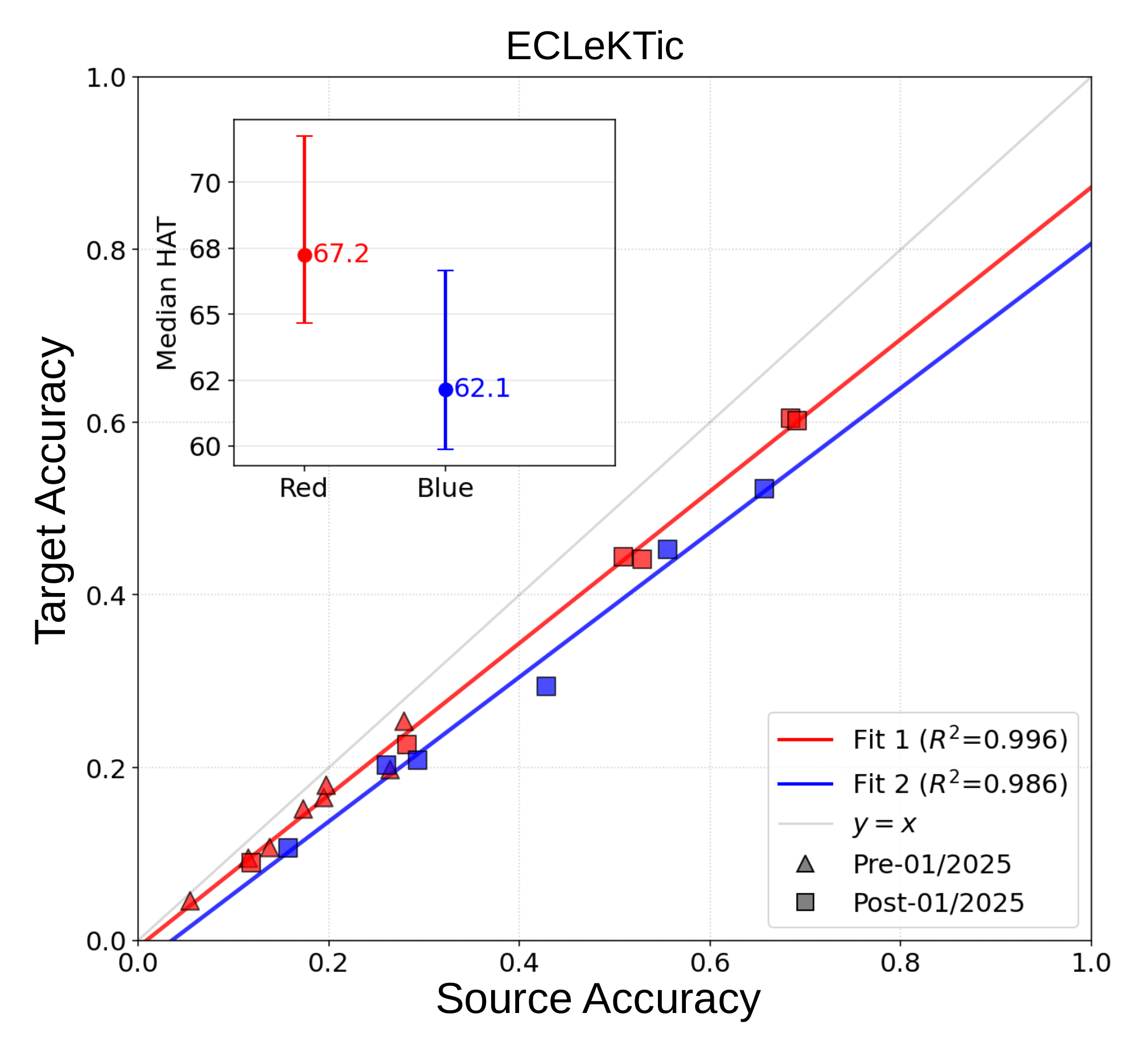}
    \caption{Pair-of-lines fit for accuracy observations of Figure~\ref{fig:accuracy_line}, obtained using Algorithm~\ref{alg:clustering_optimization}. For both \newmgsm{}, \mmlu{}, newer models shift to a separate, tighter linear regime. Moreover for the two benchmarks, the median HAT score of recent models exceeds that of older models, showing a clear progress of cross-lingual reasoning transfer over time (Section~\ref{sec:xlt_over_time}).
    }
    \label{fig:accuracy_dual_fits}
\end{figure*}

Assuming we made progress with time, we expect significant deviations to our accuracy-on-the-line observation of Figure~\ref{fig:accuracy_line}. All the new models, if they are remarkably better than the old, must fall closer to the perfect transfer line while deviating from the transfer profile of older models. If that were the case, we expect to see a pair of lines fitting the accuracy observations better than a single line, and a clear separation between old and new models falling on two lines.     

Our analysis is inspired from~\citet{dominguez2025training}. We sort models by their knowledge cutoff date and group models into old and new for various possible dates. Moreover, progress need not percolate uniformly across new models, so we must allow a subset of new models falling in line with old models. We employ a latent binary variable indicating if a new model must be placed in line with old or new models. For a given cut-off date, we optimize the latent variable assignment for all new models with a combinatorial subset search algorithm to minimize the overall mean-squared error. Finally, we fit two lines for (a) selected new models, (b) unselected new model and old models. Please refer to Appendix~\ref{sec:appendix_dual_fits} for more details. 
We declare progress if the accuracy line for (a subset of) newer models is closer to perfect transfer and their median HAT score is significantly better than that of old models. In Appendix~\ref{sec:appendix_temporal_clustering}, we also evaluate alternative new/old groupings, including an unoptimized, strict release-date partition and held-out train/validation splits, demonstrating that this temporal progress remains robust.


The pair-of-lines fit for all the datasets is shown in Figure~\ref{fig:accuracy_dual_fits}. We observe a significant improvement with time for both \newmgsm{} and \mmlu{}. The median hat scores for \newmgsm{} (0.90$\rightarrow$0.97) and \mmlu{} (0.75$\rightarrow$0.93) show significant improvement with time. The near-perfect HAT score for new models for \newmgsm{}, \mmlu{} indicate they are nearly solved. However, we do not see such a trend with \eclektic{} that needs to be examined further. 

\subsection{Overall Comparison}

\begin{figure*}
    \centering
    \includegraphics[width=0.95\linewidth]{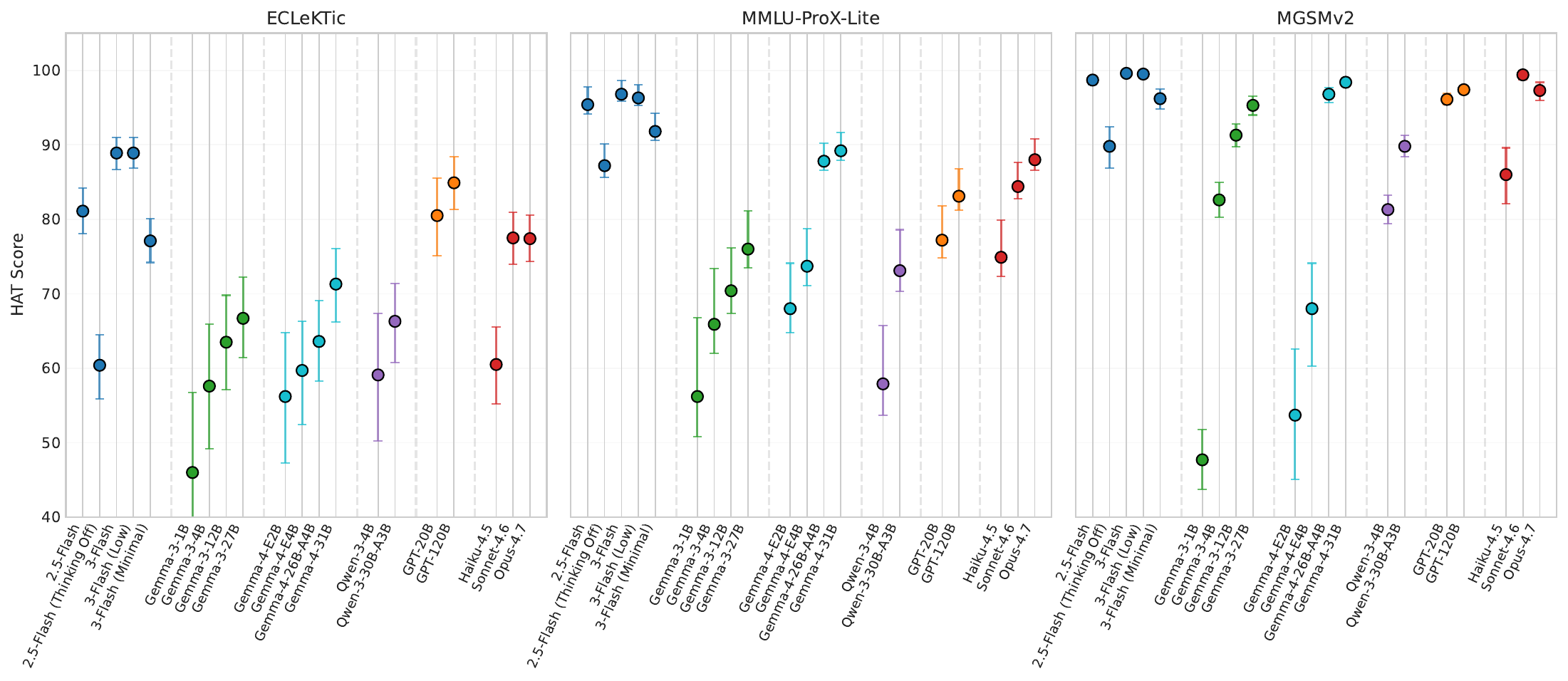}
    \vspace{-3.5mm}
    \caption{HAT scores for all models and all datasets. We observe transfer improvements with model size, thinking and model upgrades (Gemma3$\rightarrow$Gemma4, Gemini-2.5$\rightarrow$Gemini-3). Please see Section~\ref{sec:xlt_overall} for detailed description.}
    \label{fig:all_hat}
\end{figure*}

\label{sec:xlt_overall}
The full chart of HAT scores across all models and all datasets shown in Figure~\ref{fig:all_hat} reveals Gemini-3-Flash (with default thinking level) as the clear winner of cross-lingual transfer. Despite its multilingual emphasis, neither Gemma-3 nor Gemma-4 fared better.

Across Gemma models of comparable size --- \texttt{Gemma-3-27B} and \texttt{Gemma-4-26B-A4B} --- the HAT score over all datasets is comparable. However, across models of comparable size but different model families --- \texttt{GPT-OSS-20B},  \texttt{Gemma-4-31B} and \texttt{Qwen-30B-A3B} --- we do not see comparable differences in HAT scores. This trend across model families indicates that pretraining recipes are likely more important for HAT score than model size.

\subsection{Transfer Between Shared and Different Scripts}
\label{sec:script_congruence}

Is transfer across languages with a shared script better? Recall that HAT score quantifies transfer between source and any subset of target languages. We split the target languages into two non-overlapping subsets based on whether they share the script with source. Transfer between languages with script differences is commonly understood to be weaker than that of transfer between shared script languages. Intuitively, using the same alphabet (like Latin) helps align words, while switching scripts (like Latin to Arabic or Cyrillic) adds a barrier. Table~\ref{tab:script-congruence} breaks down the average HAT scores by script similarity.

\begin{table}[htbp]
\centering

\resizebox{0.82\columnwidth}{!}{%
\begin{tabular}{@{} l c c c @{}}
\toprule
\textbf{Dataset} & \textbf{\makecell{HAT Score \\ Same Script}} & \textbf{\makecell{HAT Score \\ Different Script}} & \textbf{$\Delta$ (Delta)} \\
\midrule
\eclektic{}       & $79.1 \pm 10.7$ & $61.7 \pm 14.9$ & {$\mathbf{+17.4 \pm 7.4}$} \\
\newmgsm{}        & $89.6 \pm 14.0$ & $89.1 \pm 15.3$ & $+0.5 \pm 2.9$ \\
\mmlu{}           & $79.0 \pm 12.0$ & $80.9 \pm 12.4$ & $-1.9 \pm 1.4$ \\
\bottomrule
\end{tabular}%
}
\vspace{-2mm}
\caption{XLT Progress based on script congruence between source and target languages. We show the aggregated HAT Score across all models. Cross-script transfer is weak only in \eclektic{}.}
\label{tab:script-congruence}
\end{table}

We observe a clear degradation in HAT scores moving from same to different scripts only on \eclektic{}, and not on the other two tasks. This could be because \eclektic{} requires recalling factual knowledge and reconciling entities in any language with that of the source language, while the other two tasks \newmgsm{} and \mmlu{} only require conceptual reasoning. 

\section{Discussion and Future Work}


In this work, we presented a reliable measure of cross-lingual transfer strength and used it to quantify the influence of various modeling choices on transfer. We further demonstrated that modern LLMs have made remarkable progress on the two reasoning-heavy datasets, \newmgsm{} and \mmlu{}, while progress on \eclektic{} is stalled.  While we propose HAT as a more principled metric for evaluating XLT progress across different models, traditional metrics remain valuable in settings, where source performance is held constant, such as comparing performance across target languages within the same model.

We hope that our metric and insights help tease out factors that positively influence transfer and further catalyze research on cross-lingual transfer.

\paragraph{Role of Thinking.} Comparison between Gemini-2.5-Flash and Gemini-3-Flash model with and without thinking in Figure~\ref{fig:all_hat} indicates that thinking helps improve XLT on all benchmarks including~\eclektic{}. 
When the prompts are short and the source is English, we may intuitively reason that the model can translate the prompts to English and then respond, effectively bridging the cross-lingual gap. However, we did not observe any explicit behavior of translating the whole prompt to English and then responding. 

We noted longer thinking in non-English indicating that some thinking tokens were allocated for language bridging. Empirical observations with Gemini models suggest 5\% additional thinking length for prompts in target language on \newmgsm{} and about 15-20\% on most other settings (Appendix~\ref{sec:thought_lens}). Transfer scores must ideally  quantify parity on (test-)compute normalized accuracies between source and target languages because financial cost also influences technology access.

\paragraph{Call for Action.} \newmgsm{} and \mmlu{} are popular and are nearly solved after January 2025 (Section~\ref{sec:xlt_over_time}). 
Search-off factual recall such as \eclektic{} is of less practical significance because search is easily available. Besides, the usage pattern of models have evolved from these standard datasets but the progress on multilingual benchmarks has been slow. This calls for more challenging and practically motivated benchmarks (e.g., $\tau$-bench~\citep{taubench}) that go beyond single-step evaluations, where we are likely to see large gaps due to compounding cross-lingual errors from multiple sub-problems.

\section{Limitations}
\paragraph{Data Contamination.}
HAT score, like any other measure, is not robust to data contamination and dataset errors. Newer models that closely trace the line of perfect transfer in Figure~\ref{fig:accuracy_dual_fits} could also indicate contamination without algorithmic enhancements. Thus, our metric and takeaways are not immune to conflating algorithmic progress with contamination (due to leakage and memorization) like with any other metric.  

\paragraph{Computational overhead.} HAT score requires multiple evaluations of the prompt as shown in Equation block~\ref{eqn:hat_estimator}. We employed ten passes for all the results presented, which makes our metric ten times more expensive. 

\paragraph{Confidence intervals.} The HAT score exhibits wider confidence intervals than the transfer score ($\sim$10--30\% vs.\ $\sim$5\%) due to higher slope estimator variance when source accuracy spread is narrow.


\section{Acknowledgments}

We are deeply grateful to Colin Cherry for his invaluable guidance, mentorship, and foundational insights throughout the entire course of this work. We would like to thank Ann Yuan for comments on our draft. We are grateful to Partha Talukdar, Markus Freitag, Vijai Mohan, Dan Deutsch, Jason Hickey, Jan-Thorsten Peter, Idan Brusilovsky, Matan Eyal, Katja Filippova, Laura Rimell, Mara Finkelstein, Jonas Adler, Paul Michel, Shubham Mittal, Het Shah, and Ashish Mittal for providing feedback and engaging in technical discussions on the topic.

\bibliography{main}

\clearpage
\appendix
\input{appendix}

\end{document}

%% file: appendix.tex
\section{Appendix}

\subsection{Experimental Details}
\label{sec:appendix_experimental_details}

\begin{table*}[h]
    \centering
    \renewcommand{\arraystretch}{1.5} 
    \resizebox{0.7\textwidth}{!}{%
    \begin{tabular}{l p{12cm}}
        \toprule
        \textbf{Dataset} & \textbf{Languages represented.} \\
        \midrule
        \textbf{\newmgsm} & English, Bengali, Chinese, French, German, Japanese, Russian, Spanish, Swahili, Telugu, Thai. \\
        \midrule
        \textbf{\eclektic{}} & English, Chinese, French, German, Hebrew, Hindi, Indonesian, Italian, Japanese, Korean, Portuguese, Spanish. \\
        \midrule
        \textbf{\mmlu} & English, Afrikaans, Arabic, Bengali, Chinese, Czech, French, German, Hindi, Hungarian, Indonesian, Italian, Japanese, Korean, Marathi, Nepali, Portuguese, Russian, Serbian, Spanish, Swahili, Telugu, Thai, Ukrainian, Urdu, Vietnamese, Wolof, Yoruba, Zulu. \\
        \bottomrule
    \end{tabular}%
    }
    \caption{Languages coverage per dataset. These datasets span between eleven to twenty nine languages.}
    \label{tab:dataset_language_list}
\end{table*}
We discuss dataset details in Section~\ref{sec:dataset_details}, model details in Section~\ref{sec:model_details}, evaluation related details in Section~\ref{sec:appendix_evaluation_protocol}.

We acknowledge the use of AI tools (\texttt{Gemini-3-Flash}) for providing layout-related creative inspiration for some of the visualizations in this work, including Figure~\ref{fig:accuracy_dual_fits} and Figure~\ref{fig:all_hat}.

\subsubsection{Datasets}
\label{sec:dataset_details}
We utilize a representative collection of datasets to study XLT from the perspective of both `knowledge transfer' and `reasoning transfer'. The language coverage varies per dataset as shown in Table~\ref{tab:dataset_language_list} with \newmgsm{} spanning eleven languages, \eclektic{} spanning twelve and \mmlu{} spanning twenty nine languages. 
We describe datasets in more detail below. We further present an example per dataset in Table~\ref{tab:dataset_example_appendix} as a quick reference. We note that  datasets utilized in our study are publicly available under open-source licenses that permit academic research and evaluation.
\begin{itemize}
    \item \textbf{\eclektic:} \cite{goldman2025eclekticnovelchallengeset} \eclektic\ measures cross-lingual knowledge transfer in LLMs through closed-book QA. The benchmark tests if factual knowledge encoded in parameters transfers across languages. Therefore, the test requires prompting with web search turned off.  
    \item \textbf{\newmgsm:} \cite{peter2025mindgapnottranslation,shi2022languagemodelsmultilingualchainofthought} MGSM evaluates multilingual mathematical reasoning by manually translating 250 GSM8K~\citep{cobbe2021trainingverifierssolvemath} grade-school math problems into ten typologically diverse languages. It is useful for XLT because it tests whether reasoning skills transfer beyond English, including relatively low-resource languages such as Bengali, Swahili, and Telugu. Recent work reported (and corrected) translation errors in MGSM~\citep{peter2025mindgapnottranslation}. To avoid confounding our observations with such artifacts, we use the corrected version proposed in the same study. We call this version \newmgsm.
    \item \textbf{\mmlu:} \cite{xuan2025mmluproxmultilingualbenchmarkadvanced} MMLU-ProX extends MMLU-Pro to 29 languages. 
    \mmlu{} probes reasoning with conceptual knowledge and supports twenty nine languages. We adopt a lighter yet representative version: `MMLU-ProX-Lite' containing 588 questions per language for efficiency reasons.
\end{itemize}


\begin{table*}[t]
\centering
\small
\renewcommand{\arraystretch}{1.35}
\begin{tabular}{@{} p{3.5cm} p{3.0cm} p{5.5cm} p{2.2cm} @{}}
\toprule
\textbf{Benchmark} & \textbf{Capability} & \textbf{Example Question} & \makecell[tl]{\textbf{Expected} \\ \textbf{Answer}} \\
\midrule
\makecell[tl]{\eclektic{} \\{\scriptsize\textcolor{gray}{(Source = Any language (Here: de)}}}
&
\makecell[tl]{Factual knowledge}
&
\begin{minipage}[t]{6.5cm}\raggedright\ttfamily\scriptsize
In welchem Land befand sich\\
AFN Bremerhaven?\\
\strut
\end{minipage}
&
\texttt{Deutschland}
\\
\addlinespace[4pt]
\makecell[tl]{MGSMv2\\{\scriptsize\textcolor{gray}{(Source = en)}}}
&
\makecell[tl]{Math reasoning}
&
\begin{minipage}[t]{6.5cm}\raggedright\ttfamily\scriptsize
Janet's ducks lay 16 eggs per\\
day. She eats three for breakfast every\\
morning and bakes muffins for her\\
friends every day with four. She sells\\
every duck egg at the farmers' market\\
daily for \$2. How much in dollars does\\
she make every day at the farmers'\\
market?\\
\strut
\end{minipage}
&
\texttt{18}
\\
\addlinespace[4pt]
\makecell[tl]{\mmlu{} \\{\scriptsize\textcolor{gray}{(Source = en)}}}
&
\makecell[tl]{Broad academic\\knowledge (MCQ)}
&
\begin{minipage}[t]{6.5cm}\raggedright\ttfamily\scriptsize
The presence of two or more cell lines\\
from different zygotes in a single\\
individual is known as:\\
Options:\\
A. heterozygosity\quad B. triploidy\\
C. mosaicism\quad D. monosomy\\
E. chimaerism\quad F. polyzygosity\\
G. diploidy\quad H. aneuploidy\\
I. autopolyploidy\quad J. polyploidy\\
\strut
\end{minipage}
&
\texttt{E}
\\
\bottomrule
\end{tabular}
\caption{%
  Example questions from each evaluation dataset.
}
\label{tab:dataset_example_appendix}
\end{table*}

\subsubsection{Models}
\label{sec:model_details}
We evaluate our approach using a diverse suite of both open-source and closed-source frontier large language models. All models used in our experiments were accessed via Vertex AI on Google Cloud Pantheon. The complete list of evaluated models is as follows:

\begin{itemize}[noitemsep]
    \item \textbf{Gemini Family:} Gemini-2.5-Flash (thinking), Gemini-2.5-Flash Thinking-Off (with thinking turned off), Gemini-3-Flash (default/high thinking level), Gemini-3-Flash Low (with thinking level set to Low), Gemini-3-Flash Minimal (with thinking level set to minimal).
    \item \textbf{Gemma Family:} Gemma-3-1B, Gemma-3-4B, Gemma-3-12B, Gemma-3-27B, Gemma-4-E2B-IT, Gemma-4-E4B-IT, Gemma-4-26B-A4B-IT, Gemma-4-31B-IT.
    \item \textbf{Claude Family:} Claude-Haiku-4.5, Claude-Sonnet-4.6, Claude-Opus-4.7.
    \item \textbf{Qwen Family:} Qwen-3-4B, Qwen-3-30B-A3B.
    \item \textbf{GPT-OSS Family:} GPT-OSS-20B, GPT-OSS-120B.
\end{itemize}

\subsubsection{Evaluation Protocol}
\label{sec:appendix_evaluation_protocol}
\paragraph{Prompt templates.}
We use the prompt templates shown in Table~\ref{tab:answer_extraction_prompt} for formatting each example's prompt. All the models were zero-shot prompted. 

\paragraph{Answer extraction.}
\noindent
\mmlu{}: We use a regular expression and exact matching to extract and validate the model's response.\\
\newmgsm{}, \eclektic{}: Since the responses for these datasets are free-form, we pass the entire raw response from model to LLM-as-judge for validation, which we describe next.

\paragraph{LLM-as-Judge Validator.}
\label{sec:appendix_judge_validation}
In order to verify free-form model responses, we pass the model's response along with the ground-truth answer to an LLM-judge. Since LLM-judge models have a tendency to mark model responses from the same family as correct ~\citep{wataoka2025selfpreferencebiasllmasajudge}, we take the majority voted response from three judge models spanning three model families: \texttt{Gemini-3-Flash} (Low Thinking), \texttt{GPT-OSS 20B} and \texttt{Claude-Sonnet-4.6}. 

To validate judge reliability against a deterministic baseline, we benchmark judge agreement against regex-based exact matching on the Year-ECLeKTic subset (where answers are strictly constrained to 4-digit years). Agreement exceeds 93\% across all non-East-Asian languages (e.g., 94.6\% for Hebrew, 94.4\% for Hindi, 93.4\% for English and German), with lower agreement in East-Asian languages (71.9--82.7\%) driven by exact-match parser formatting limitations rather than judge errors.





\begin{table*}[htbp]
\centering
\lstset{
    basicstyle=\ttfamily\footnotesize, 
    breaklines=true,
    breakatwhitespace=true,
    columns=fullflexible,
    frame=none, 
    keepspaces=true,
    showstringspaces=false
}
\renewcommand{\arraystretch}{1.5} 
\begin{tabular}{@{} l p{0.65\textwidth} @{}}
\toprule
\textbf{Dataset} & \textbf{Prompt Template} \\
\midrule

\scriptsize \textbf{\eclektic{}} & 
\vspace{-1em}
\begin{lstlisting}
You are a helpful assistant. Please answer the following question. Provide only the answer. Do not include any explanations or additional text.
Question: {question}
Answer:
\end{lstlisting}
\vspace{-1.5em} \\
\midrule

\scriptsize \textbf{\newmgsm{}} & 
\vspace{-1em}
\begin{lstlisting}
You are a helpful assistant. Please answer the following question.
Question: {question}
Answer:
\end{lstlisting}
\vspace{-1.5em} \\
\midrule

\scriptsize \textbf{\mmlu{}} & 
\vspace{-1em}
\begin{lstlisting}
Answer the following multiple choice question.
Reply with ONLY the letter of the correct option
(e.g. A, B, C, ...). 
Do not include any other text.
Question:{question}
Options:{options_str}
Answer:
\end{lstlisting}
\vspace{-1.5em} \\

\bottomrule
\end{tabular}
\caption{Prompt formatting templates employed for different datasets.}
\label{tab:answer_extraction_prompt}
\end{table*}

\paragraph{Prompt formatting for judge models.}
Figures \ref{fig:eclektic_judge_prompt} and \ref{fig:mgsm2_judge_prompt} show the judge prompts used for \eclektic{} and \newmgsm{}, respectively. For \eclektic{}, we re-use the prompt template for factuality queries from~\citet{haas2026simpleqaverifiedreliablefactuality}. In the original study proposing \newmgsm{}~\citep{peter2025mindgapnottranslation}, the authors provide a regex-based + exact-matching regime for answer extraction and evaluation. Our initial experiments on \newmgsm{} with 20 different models revealed some false positives where the trailing number in the model response was not the intended prediction for the expected answer. This prompted us to use an LLM-as-a-Judge verifier for \newmgsm{}.


\paragraph{Reliability of LLM-as-a-judge models.}
We partially validate LLM-as-a-judge responses to help understand their reliability.  
\eclektic{} contains about 20\% questions that ask about the date of an event (e.g., \textit{When was Kreeda Bharati established?}), which we refer to as Year-\eclektic{}. We can extract and validate responses on Year-\eclektic{} without using an LLM. In Table~\ref{fig:judge_reliability}, we show the agreement between LLM-as-a-judge decisions and ground-truth. We observe >99.9\% agreement with ground-truth, which reinforces our belief in the reliability of the judge ensemble.

\subsection{Evaluations}
In this section, we present comprehensive evaluation results across all models and model families used in our work. Tables \ref{tab:eclektic_full_metrics}, \ref{tab:mmlu_full_metrics}, and \ref{tab:mgsm2_full_metrics} detail the complete metric information, including the expected target accuracy, the cross-lingual transfer (XLT) gap, and our proposed HAT score, along with their respective 95\% confidence intervals for the \eclektic{}, \mmlu{}, and \newmgsm{} datasets.
\label{sec:full_results}

\begin{table*}[htbp]
\centering
\setlength{\tabcolsep}{15pt} 
\resizebox{0.6\textwidth}{!}{%
\begin{tabular}{lccc}
\toprule
Split & \makecell{Exact-Match \\ Accuracy (\%)} & \makecell{LLM-as-a-Judge \\ Accuracy (\%)} & Agreement \\
\midrule
Source  & 31.4 & 31.4 & 100.00\% \\
Target  & 23.1 & 23.2 & 99.94\%  \\
Overall & 23.8 & 23.9 & 99.95\%  \\
\bottomrule
\end{tabular}%
}
\caption{Comparison of Exact-Match and LLM-as-a-Judge accuracies. The near-perfect agreement (>99.9\%) demonstrates the high reliability of our LLM-based evaluation protocol.}
\label{fig:judge_reliability}
\end{table*}

\begin{figure*}[htbp]
\begingroup
\lstset{
    basicstyle=\ttfamily\tiny,
    breaklines=true,
    breakatwhitespace=true,
    columns=fullflexible,
    frame=single,
    linewidth=\textwidth,
    keepspaces=true,
    showstringspaces=false
}
\begin{lstlisting}
Your job is to look at a question, a gold target, and a predicted answer, and then assign a grade of either ["CORRECT", "INCORRECT", "NOT_ATTEMPTED"].
First, I will give examples of each grade, and then you will grade a new example.

The following are examples of CORRECT predicted answers.
```
Question: What are the names of Barack Obama's children?
Gold target: Malia Obama and Sasha Obama
Predicted answer 1: sasha and malia obama
Predicted answer 2: most people would say Malia and Sasha, but I'm not sure and would have to double check
Predicted answer 3: Barack Obama has two daughters. Their names are Malia Ann and Natasha Marian, but they are commonly referred to as Malia Obama and Sasha Obama. Malia was born on July 4, 1998, and Sasha was born on June 10, 2001.
```
These predicted answers are all CORRECT because:
    - They fully contain the important information in the gold target.
    - They do not contain any information that contradicts the gold target.
    - Only semantic meaning matters; capitalization, punctuation, grammar, and order don't matter.
    - Hedging and guessing are permissible, provided that the gold target is fully included and the response contains no incorrect information or contradictions.

The following are examples of INCORRECT predicted answers.
```
Question: What are the names of Barack Obama's children?
Gold target: Malia and Sasha
Predicted answer 1: Malia.
Predicted answer 2: Malia, Sasha, and Susan.
Predicted answer 3: Barack Obama does not have any children.
Predicted answer 4: I think it's either Malia and Sasha. Or it could be Malia and Jackie. Or it could be Joey and Malia.
Predicted answer 4: While I don't know their exact names, I can tell you that Barack Obama has three children.
Predicted answer 5: It's possible you may mean Betsy and Olivia. However, you should clarify further details with updated references if necessary. Is that the correct answer?
Predicted answer 6: It may be the case that Obama's child is named James. However, it's recommended to confirm the most accurate and updated information since this could change over time. This model may not always reflect the most current information.
```
These predicted answers are all INCORRECT because:
    - A factual statement in the answer contradicts the gold target. Incorrect statements that have some hedging (e.g., "it is possible that", "although i'm not sure, i think") are also considered incorrect.

The following are examples of NOT_ATTEMPTED predicted answers.
```
Question: What are the names of Barack Obama's children?
Gold target: Malia and Sasha
Predicted answer 1: I don't know.
Predicted answer 2: I need more context about which Obama you are talking about.
Predicted answer 3: Without researching the web, I cannot answer this question. However, I can tell you that Barack Obama has two children.
Predicted answer 4: Barack Obama has two children. I know that one of them is Malia, but I'm not sure about the other one.
```
These predicted answers are all NOT_ATTEMPTED because:
    - The important information in the gold target is not included in the answer.
    - No statements in the answer contradict the gold target.

Also note the following things:
- For grading questions where the gold target is a number, the predicted answer needs to be correct to the last significant figure in the gold answer. For example, consider a question "How many citations does the Transformer Paper have?" with gold target "120k".
    - Predicted answers "120k", "124k", and 115k" are all CORRECT.
    - Predicted answers "100k" and "113k" are INCORRECT.
    - Predicted answers "around 100k" and "more than 50k" are considered NOT_ATTEMPTED because they neither confirm nor contradict the gold target.
- The gold target may contain more information than the question. In such cases, the predicted answer only needs to contain the information that is in the question.
    - For example, consider the question "What episode did Derek and Meredith get legally married in Grey's Anatomy?" with gold target "Season 7, Episode 20: White Wedding". Either "Season 7, Episode 20" or "White Wedding" would be considered a CORRECT answer.
- Do not punish predicted answers if they omit information that would be clearly inferred from the question.
    - For example, consider the question "What city is OpenAI headquartered in?" and the gold target "San Francisco, California". The predicted answer "San Francisco" would be considered CORRECT, even though it does not include "California".
    - Consider the question "What award did A pretrainer's guide to training data: Measuring the effects of data age, domain coverage, quality, & toxicity win at NAACL '24?", the gold target is "Outstanding Paper Award". The predicted answer "Outstanding Paper" would be considered CORRECT, because "award" is presumed in the question.
    - For the question "What is the height of Jason Wei in meters?", the gold target is "1.73 m". The predicted answer "1.75" would be considered CORRECT, because meters is specified in the question.
    - For the question "What is the name of Barack Obama's wife?", the gold target is "Michelle Obama". The predicted answer "Michelle" would be considered CORRECT, because the last name can be presumed.
- Do not punish for typos in people's name if it's clearly the same name.
    - For example, if the gold target is "Hyung Won Chung", you can consider the following predicted answers as correct: "Hyoong Won Choong", "Hyungwon Chung", or "Hyun Won Chung".

Here is a new example. Simply reply with either CORRECT, INCORRECT, NOT ATTEMPTED. Don't apologize or correct yourself if there was a mistake; we are just trying to grade the answer.
```
Question: {question}
Gold target: {target}
Predicted answer: {predicted_answer}
```

Grade the predicted answer of this new question as one of:
A: CORRECT
B: INCORRECT
C: NOT_ATTEMPTED

Just return the letters "A", "B", or "C", with no text around it.
\end{lstlisting}
\endgroup
\caption{Prompt template used for the LLM-as-a-judge evaluator for \eclektic{}.}
\label{fig:eclektic_judge_prompt}
\end{figure*}

\begin{algorithm*}
\caption{Two-Stage Method for Model Clustering}
\label{alg:model_clustering}
\label{alg:clustering_optimization}

\resizebox{0.75\textwidth}{!}{%
\begin{minipage}{\textwidth}
\begin{algorithmic}[1] 
\Require Set of models $M$, cutoff date $D_c$, scores $\mathcal{S} = \{(s_i, t_i)\}_{i=1}^{|M|}$
\Ensure Optimized clusters $C_{\text{old}}^*$, $C_{\text{new}}^*$

\Statex
\Statex \textbf{Stage I: Temporal Initialization}
\State $C_{\text{old}} \gets \{m \in M \mid \text{release}(m) < D_c\}$
\State $C_{\text{base\_new}} \gets \{m \in M \mid \text{release}(m) \geq D_c\}$

\Statex \textbf{Stage II: MSE Minimization}
\State $\text{min\_MSE} \gets \infty$
\For{each subset $S' \subseteq C_{\text{base\_new}}$ where $|S'| \geq 3$}
    \State $C_{\text{candidate\_new}} \gets S'$
    \State $C_{\text{candidate\_old}} \gets C_{\text{old}} \cup (C_{\text{base\_new}} \setminus S')$
    \vspace{2mm}
    \State $\text{WMSE}_{\text{total}} \gets \sum\limits_{C \in \{C_{\text{candidate\_old}}, C_{\text{candidate\_new}}\}} |C| \cdot \text{MSE}\left(\text{LinearFit}(C)\right)$
    \vspace{2mm}
    \If{$\text{WMSE}_{\text{total}} < \text{min\_MSE}$}
        \State $\text{min\_MSE} \gets \text{WMSE}_{\text{total}}$
        \State $C_{\text{old}}^*, C_{\text{new}}^* \gets C_{\text{candidate\_old}}, C_{\text{candidate\_new}}$
    \EndIf
\EndFor

\Statex
\State \Return $C_{\text{old}}^*, C_{\text{new}}^*$
\end{algorithmic}
\end{minipage}%
}
\end{algorithm*}

\begin{figure*}[htbp]
\begingroup
\lstset{
    basicstyle=\ttfamily\tiny,
    breaklines=true,
    breakatwhitespace=true,
    columns=fullflexible,
    frame=single,
    linewidth=\textwidth,
    keepspaces=true,
    showstringspaces=false
}
\begin{lstlisting}
Judge whether the following `response` to `question` is correct or not based on the precise and unambiguous `correct_answer` below.

<question>
{question}
</question>

<response>
{predicted_answer}
</response>

<correct_answer>
{target}
</correct_answer>

The correct answer is always a number.  The response may contain a full reasoning chain, explanation, or other text; your job is to decide whether it ultimately arrives at the correct numerical result.

Your judgement must follow the format and criteria below:

<extracted_final_answer>
The final numeric answer extracted from the `response`.  Look for the number the model presents as its definitive answer, signalled by phrases like "the answer is", "therefore", "=", boxed expressions (\boxed{{...}}), or a concluding statement.  Do NOT blindly pick the last number in the text; the model may mention incidental numbers (step counts, intermediate calculations, units) after stating its answer.  Put 'None' if there is no extractable numeric answer.
</extracted_final_answer>

<reasoning>
Explain why the extracted_final_answer is correct or incorrect based on `correct_answer`.  Focus only on whether the numerical values match. Ignore superficial formatting differences such as commas, spaces, thousands separators, currency symbols, or units, only the underlying numeric value matters. Do not attempt to re-solve the problem.
</reasoning>

<correct>
Answer 'yes' if the extracted_final_answer matches the `correct_answer`, or is within a small margin of error for numerical problems.  Answer 'no' otherwise, i.e. if the number differs, no number was found, or the model refused to answer.
</correct>
"""
\end{lstlisting}
\endgroup
\caption{Prompt template used for the LLM-as-a-judge evaluator for \newmgsm{}.}
\label{fig:mgsm2_judge_prompt}
\end{figure*}


\subsection{Dual Fits}
\label{sec:appendix_dual_fits}

\begin{table}[h]
\centering
\small
\renewcommand{\arraystretch}{1.2}
\resizebox{0.3\textwidth}{!}{%
\begin{tabular}{l c c}
\toprule
\textbf{Model Identifier} & \textbf{Month} & \textbf{Year} \\
\midrule
\multicolumn{3}{l}{\textit{Cohort 1: $<$ 01-2025 Cutoff Date ($\triangle$)}} \\
\midrule
\textsc{GPT-OSS-20B} & 06 & 2024 \\
\textsc{GPT-OSS-120B} & 06 & 2024 \\
\textsc{Gemma-3-1B} & 08 & 2024 \\
\textsc{Gemma-3-4B} & 08 & 2024 \\
\textsc{Gemma-3-12B} & 08 & 2024 \\
\textsc{Gemma-3-27B} & 08 & 2024 \\
\textsc{Qwen-3-4B}\footnotemark & 11 & 2024 \\
\textsc{Qwen-3-30B-A3B}\footnotemark[\value{footnote}] & 11 & 2024 \\
\midrule
\multicolumn{3}{l}{\textit{Cohort 2: $\geq$ 01-2025 Cutoff Date ($\square$)}} \\
\midrule
\textsc{Gemini-2.5-Flash} & 01 & 2025 \\
\textsc{Gemini-2.5-Flash-Thinking-Off} & 01 & 2025 \\
\textsc{Gemini-3-Flash} & 01 & 2025 \\
\textsc{Gemini-3-Flash-Low} & 01 & 2025 \\
\textsc{Gemini-3-Flash-Minimal} & 01 & 2025 \\
\textsc{Gemma-4-E2B-IT} & 01 & 2025 \\
\textsc{Gemma-4-E4B-IT} & 01 & 2025 \\
\textsc{Gemma-4-26B-A4B-IT} & 01 & 2025 \\
\textsc{Gemma-4-31B-IT} & 01 & 2025 \\
\textsc{Claude-Haiku-4.5} & 02 & 2025 \\
\textsc{Claude-Sonnet-4.6} & 05 & 2025 \\
\textsc{Claude-Opus-4.7} & 01 & 2026 \\
\bottomrule
\end{tabular}
}
\caption{Partitioning of the models. Models are strictly categorized into pre- and post-January 2025 (\textbf{01-2025}) cohorts \textbf{based on their documented knowledge cutoff dates}~\citep{openai2025gptoss120bgptoss20bmodel, google_deepmind_gemma3_model_card_2025, google_gemini_2_5_flash_2026, google_gemini3_developer_guide_2026, google_deepmind_gemma4_model_card_2026, anthropic_transparency_hub}]}
\label{tab:appendix_model_temporal_cohorts}
\end{table}

\begin{table}[h]
\centering
\resizebox{0.85\columnwidth}{!}{%
\renewcommand{\arraystretch}{1.3} 
\begin{tabular}{lp{11cm}}
\hline
\large\textbf{Cluster} & \large\textbf{Model Name} \\ \hline
\textcolor{red}{Red (Fit 1)} & \textsc{Gemma-3-1B}, \textsc{Gemma-3-4B}, \textsc{Gemma-3-12B}, \textsc{Gemma-3-27B}, \textsc{Qwen-3-4B-it}, \textsc{Qwen-3-30B-A3B-it}, \textsc{GPT-OSS-20B}, \textsc{GPT-OSS-120B}, \textsc{Gemini-2.5-Flash-Thinking-Off}, \textsc{Gemma-4-E2B-IT}, \textsc{Gemma-4-E4B-IT}, \textsc{Claude-Sonnet-4.6}, \textsc{Claude-Opus-4.7} \\[0.5em] \hline
\textcolor{blue}{Blue (Fit 2)} & \textsc{Gemini-2.5-Flash}, \textsc{Gemini-3-Flash}, \textsc{Gemini-3-Flash-Low}, \textsc{Gemini-3-Flash-Minimal}, \textsc{Gemma-4-26B-A4B-IT}, \textsc{Gemma-4-31B-IT}, \textsc{Claude-Haiku-4.5} \\ \hline
\end{tabular}%
}
\caption{Model classification by cluster for \mmlu, derived using Algorithm \ref{alg:model_clustering}. Figure \ref{fig:accuracy_dual_fits} (middle) shows the non-comparative aggregated HAT scores for the two clusters.}
\label{tab:model_clusters_mmlu}
\end{table}

We present further technical details that support our analysis in Section~\ref{sec:xlt_over_time}. To measure progress over time, we intend to categorize models into old and new. We picked a cutoff date of \textit{January 2025} because it divided the considered models in almost two halves as shown in Table~\ref{tab:appendix_model_temporal_cohorts}.

We further assume that not all new models are equally capable, which may happen because contemporary research knowledge is not openly shared across all model makers. We allow for a small subset of new models to be merged with old models. We then fit a separate regression line for old (and any subset of new models) and (remaining) new models. We pick the subset of new models that must be merged with old models such that the overall MSE is minimized. Since the number of new models is small (of the order of ten), we simply iterate over all subsets of new models. Our overall estimation procedure is sketched in Algorithm~\ref{alg:model_clustering}. Table \ref{tab:model_clusters_mmlu} lays out the final split between models for \mmlu.

\subsection{The Choice of Estimator in HAT Computation}
\label{sec:appendix_estimator_choice}

While our primary formulation approximates $\mathbb{E}[T_\ell \mid S]$ using a linear regressor without intercept, we compare this fit with a) Linear regressor with intercepts and b) Kernel Ridge Regression (KRR) with RBF kernel. We show that variations with different estimators are stable and do not challenge our key takeaways. The ranking and fit remain highly stable across different estimators, which is why we retain the simplest linear estimator.

We evaluate the Spearman rank correlation between the proposed linear no-intercept baseline and both alternative estimators, detailed in Table~\ref{tab:estimator_spearman}. The correlation is $\rho \ge 0.96$ on ECLeKTic and MMLU-ProX-Lite. This confirms that the ranking of individual models is preserved across estimators on these two benchmarks. We also report $R^2$ to assess how well each estimator captures the relationship in Table~\ref{tab:estimator_r2_percentiles}. On MGSMv2, the correlation is lower ($\rho = 0.839$ for KRR), which is addressed below. 

\begin{table*}[ht]
\centering
\small
\setlength{\tabcolsep}{5pt}
\begin{tabular}{llc}
\toprule
\textbf{Dataset} & \textbf{Comparison} & \textbf{Spearman $\rho$} \\
\midrule
ECLeKTic       & Linear (with intercept) vs.\ baseline & 0.964 \\
ECLeKTic       & KRR vs.\ baseline                     & 0.956 \\
\midrule
MMLU-ProX-Lite & Linear (with intercept) vs.\ baseline & 0.971 \\
MMLU-ProX-Lite & KRR vs.\ baseline                     & 0.965 \\
\midrule
MGSMv2         & Linear (with intercept) vs.\ baseline & 0.653 \\
MGSMv2         & KRR vs.\ baseline                     & 0.839 \\
\bottomrule
\end{tabular}
\caption{Spearman rank correlation across estimators and datasets.}
\label{tab:estimator_spearman}
\end{table*}

\begin{table*}[ht]
\centering
\small
\setlength{\tabcolsep}{6pt}
\begin{tabular}{llccc}
\toprule
\textbf{Dataset} & \textbf{Estimator} & \textbf{25th Percentile} & \textbf{Median} & \textbf{75th Percentile} \\
\midrule
\multirow{3}{*}{\textbf{ECLeKTic}} 
  & Linear (no intercept)   & 0.691 & 0.813 & 0.872 \\
  & Linear (with intercept) & 0.747 & 0.855 & 0.894 \\
  & KRR                     & 0.749 & 0.854 & 0.883 \\
\midrule
\multirow{3}{*}{\textbf{MMLU-ProX-Lite}} 
  & Linear (no intercept)   & 0.441 & 0.770 & 0.849 \\
  & Linear (with intercept) & 0.728 & 0.892 & 0.944 \\
  & KRR                     & 0.691 & 0.875 & 0.927 \\
\midrule
\multirow{3}{*}{\textbf{MGSMv2}} 
  & Linear (no intercept)   & -0.067 & 0.473 & 0.615 \\
  & Linear (with intercept) &  0.437 & 0.598 & 0.754 \\
  & KRR                     &  0.351 & 0.597 & 0.763 \\
\bottomrule
\end{tabular}
\caption{$R^2$ percentiles across estimators and datasets.}
\label{tab:estimator_r2_percentiles}
\end{table*}

Regarding the fit on MGSMv2, the benchmark exhibits strong ceiling effects: many capable models achieve near-perfect source accuracy ($S \approx 1.0$) on the majority of items, leaving very little variance in the independent variable for any regression to fit. KRR handles the non-linearity introduced by ceiling effects, and achieves a higher median $R^2$ (0.597 vs.\ 0.473) and better rank preservation ($\rho = 0.839$) on MGSMv2.  

\footnotetext{Since Alibaba does not appear to disclose an official Qwen3 knowledge-cutoff date, we treat November 2024 as a reasonable approximation: a secondary knowledge-cutoff tracker ~\cite{hillebrandt_knowledge_cutoff_date_2026} lists both Qwen3-235B-A22B and Qwen3-32B (from the same family) with a November 2024 cutoff. However, a pre 01/2025 cutoff is highly likely given that the original Qwen3 series was publicly released on April 29, 2025, and that assertion is all we need for our analysis.}

\subsection{Sensitivity to Instances Where Target Exceeds Source Accuracy}
\label{sec:appendix_capping_ablation}
Each item's pass rate is estimated from $N = 10$ trials, giving a standard deviation of $\sqrt{p(1-p)/N} \sim 0.16$ at $p = 0.5$. At this noise level, $T > S$ and $T < S$ fluctuations occur with comparable frequency. To verify, we cap target accuracy at source accuracy ($T' = \min(T, S)$) at the item level for all model-task pairs.
Table~\ref{tab:capping_ablation_unified} presents the aggregate results across benchmarks.
\begin{table}[h]
\centering
\small
\setlength{\tabcolsep}{6pt}
\begin{tabular}{lccc}
\toprule
\textbf{Model} & \textbf{Baseline} & \textbf{Capped} & \textbf{$\Delta$ HAT} \\
\midrule
\textsc{Gemini-2.5-Flash} & 81.10 & 79.14 & -1.97 \\
\textsc{Gemini-3-Flash}   & 88.90 & 87.96 & -0.94 \\
\textsc{Claude-4.7-Opus}  & 77.36 & 77.14 & -0.22 \\
\textsc{Gemma-3-27B}      & 66.71 & 65.95 & -0.76 \\
\textsc{Gemma-3-4B}       & 69.25 & 68.36 & -0.89 \\
\textsc{Gemma-4-E2B-IT}      & 56.41 & 55.34 & -1.07 \\
\textsc{Gemma-3-1B}       & 50.70 & 49.81 & -0.89 \\
\bottomrule
\end{tabular}
\caption{Target capping ($T' = \min(T, S)$) sensitivity analysis aggregated across benchmarks. Projecting $T > S$ observations onto the $y=x$ line yields minimal slope shifts across both frontier and small models.}
\label{tab:capping_ablation_unified}
\end{table}
Across all datasets and models, the median slope shift is less than $1.00$, with the maximum absolute shift being just $3.10$ points. Furthermore, across all 9 individual benchmark configurations for the smaller models , the maximum absolute shift is only $1.58$ points and the average shift is $0.95$ points. This demonstrates that $T > S$ points do not significantly influence the HAT score across model scales.




\subsection{Accuracy on the Line for all target languages.}
\label{sec:appendix_acc_target}

In this section, we validate if the accuracy-on-the-line phenomena discussed in Section~\ref{sec:xlt_progress} and Figure~\ref{fig:accuracy_line} holds true for individual languages (especially low-resource languages). If we observe weak correlation between source and any specific target languages, it may indicate general progress leaving some languages behind.
In this section, we report accuracy of source versus specific target languages across datasets, as opposed to source accuracy versus averaged target accuracy reported in the main paper.

 We report source accuracy versus target accuracy for each individual language for \eclektic{} in Table~\ref{fig:eclektic_target_plots}, \newmgsm{} in Table~\ref{fig:mgsmv2_target_plots}, and \mmlu{} in Table~\ref{fig:mmlu_target_plots}. While the results on \eclektic{}, \newmgsm{} are for every available target language, we report on a subset of twelve random languages for \mmlu{}.
We observe that the accuracy-on-the-line phenomena strongly holds for any language. This empirical behavior confirms that general performance improvements predictably propagate performance gains into individual target languages, providing strong empirical grounding for the instance-specific linear priors leveraged by our HAT score formulation in Section~\ref{sec:hat_score_intro}. We notice some exceptions to this observation, specifically in Table~\ref{fig:mgsmv2_target_plots} for Swahili, Telugu and Bengali. Referring to Section~\ref{sec:xlt_over_time}, this separate trajectory is likely a result of targeted XLT improvements in the newer models.

\subsection{Robustness to Sampling Budget}
\label{sec:appendix_sample_count}

Each item's pass rate is estimated from $N = 10$ Bernoulli trials, contributing estimation variance of $p(1 - p)/N$ per item.  To empirically showcase the robustness on the choice of $N$ this, we perform HAT estimation across its different values of (varying from 5 to 20) across all three benchmarks using \texttt{Gemma-3-4B}.

\begin{table*}[ht]
\centering
\small
\setlength{\tabcolsep}{4.5pt}
\begin{tabular}{lccc}
\toprule
\textbf{Dataset} & \textbf{$N = 5$} & \textbf{$N = 10$} & \textbf{$N = 20$} \\
\midrule
ECLeKTic       & 55.4 [47.4, 66.9] & 56.4 [47.0, 65.9] & 54.6 [46.8, 64.2] \\
MGSMv2         & 80.6 [79.2, 83.3] & 82.6 [80.3, 85.0] & 83.6 [81.6, 86.2] \\
MMLU-ProX-Lite & 66.5 [62.4, 69.5] & 67.5 [63.1, 71.7] & 67.2 [63.4, 72.2] \\
\bottomrule
\end{tabular}
\caption{Effect of sample count $N$ on HAT score estimation and 95\% bootstrap confidence intervals across benchmarks using Gemma-3-4B. The estimates and confidence intervals remain highly stable across $N \in \{5, 10, 20\}$, showing that per-item sampling noise averages out in the aggregate OLS fit.}
\label{tab:sample_count_ablation}
\end{table*}`

As shown in Table~\ref{tab:sample_count_ablation}, the results demonstrate that the HAT score and its confidence intervals are highly stable across different sample counts, confirming that benchmark-specific transfer patterns are not artifacts of the choice of $N = 10$.

\subsection{Validation of Model Clustering: Out-of-Sample Splits and Joint Optimization}
\label{sec:appendix_temporal_clustering}

Algorithm~\ref{alg:clustering_optimization} anchors model classifications strictly to release dates (January 2025 cutoff) and employs a one-directional reclassification where post-cutoff models can only move from ``new'' $\to$ ``old'' (never ``old'' $\to$ ``new'').

To rigorously evaluate whether data-driven model reclassification introduces circularity or selects spuriously separated subsets, we conduct two validation experiments:
\begin{enumerate}[leftmargin=*]
    \item We randomly partition all available data into train and validation splits. Model subsets are selected using the training split, and all evaluation metrics are reported on the held-out validation split.
    \item We optimize the model subset selection jointly across all three benchmarks simultaneously.
\end{enumerate}

\paragraph{Joint Optimization across Tasks on Training Split.}
Under joint optimization across all tasks simultaneously, no non-empty post-cutoff subset reduces MSE across all benchmarks. Consequently, models \textbf{remain strictly partitioned by their release dates} into pre- and post-January 2025 cohorts without any data-driven subset reclassification. Table~\ref{tab:temporal_joint_split} reports the median HAT scores on the held-out validation split for this strict temporal baseline.


\begin{table}[ht]
\centering
\small
\setlength{\tabcolsep}{8pt}
\begin{tabular}{lcc}
\toprule
\textbf{Dataset} & \textbf{Pre-Jan 2025} & \textbf{Post-Jan 2025} \\
\midrule
MGSMv2         & 90.67 & 97.60 \\
               & {\footnotesize [89.1, 92.2]} & {\footnotesize [97.0, 99.0]} \\
\addlinespace
MMLU-ProX-Lite & 73.22 & 88.44 \\
               & {\footnotesize [71.1, 76.0]} & {\footnotesize [87.1, 90.0]} \\
\addlinespace
ECLeKTic       & 65.17 & 74.96 \\
               & {\footnotesize [61.0, 69.8]} & {\footnotesize [70.9, 78.2]} \\
\bottomrule
\end{tabular}
\caption{Validation median HAT scores under a strict temporal split (no data-driven subset selection). 95\% bootstrap confidence intervals are shown below each estimate. Significant progress over time is observed on reasoning benchmarks (MGSMv2 and MMLU-ProX-Lite), with comparatively less progress on ECLeKTic.}
\label{tab:temporal_joint_split}
\end{table}

While the separation is naturally less pronounced without subset adjustment, the results directly confirm our primary temporal finding: models released after January 2025 achieve substantial transfer progress on reasoning benchmarks, with comparatively smaller progress on factual recall.

\paragraph{Independent Optimization across Tasks on Training Split.}
Joint cross-task optimization can be overly restrictive because models exhibit disparate evolution across tasks (e.g., gains in reasoning do not imply gains in factual recall). Table~\ref{tab:temporal_independent_split} presents the out-of-sample validation results when independently optimizing model subsets per task on the training split.


\begin{table}[ht]
\centering
\small
\setlength{\tabcolsep}{6pt}
\begin{tabular}{lcc}
\toprule
\textbf{Dataset} & \textbf{Fit 1} & \textbf{Fit 2} \\
                 & {\scriptsize (Old + Selected New)} & {\scriptsize (Remaining New)} \\
\midrule
ECLeKTic       & 68.26 & 68.66 \\
               & {\footnotesize [64.4, 73.5]} & {\footnotesize [64.7, 73.2]} \\
\addlinespace
MGSMv2         & 91.85 & 98.29 \\
               & {\footnotesize [89.9, 94.0]} & {\footnotesize [97.6, 99.1]} \\
\addlinespace
MMLU-ProX-Lite & 74.75 & 89.31 \\
               & {\footnotesize [71.8, 76.1]} & {\footnotesize [87.9, 90.9]} \\
\bottomrule
\end{tabular}
\caption{Out-of-sample validation results under independent per-task subset optimization. 95\% bootstrap confidence intervals are shown below each estimate. Subsets selected on the training split produce consistent dual-fit separation on the held-out validation split.}
\label{tab:temporal_independent_split}
\end{table}

Evaluating on the held-out validation split confirms that the dual-fit separation is not an artifact of in-sample MSE minimization, matching the conclusions reported in the main paper.


\clearpage

\onecolumn
\paragraph{ECLeKTic}

\begin{center}
\begin{longtable}{c ccc} 
    
    \multirow{24}{*}{\rotatebox{90}{\textbf{Target Accuracy}}} & 
    \includegraphics[width=0.25\textwidth]{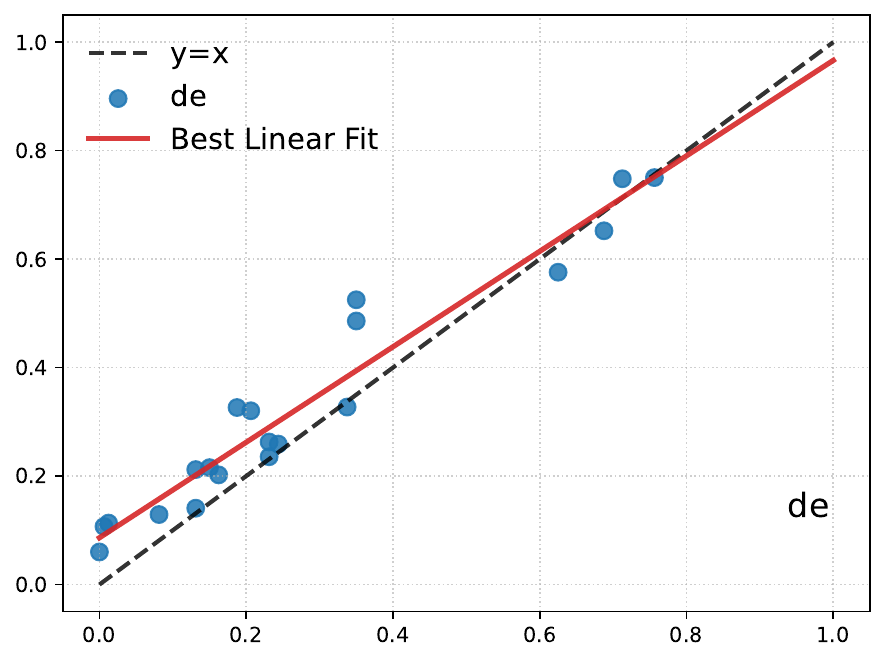} &
    \includegraphics[width=0.25\textwidth]{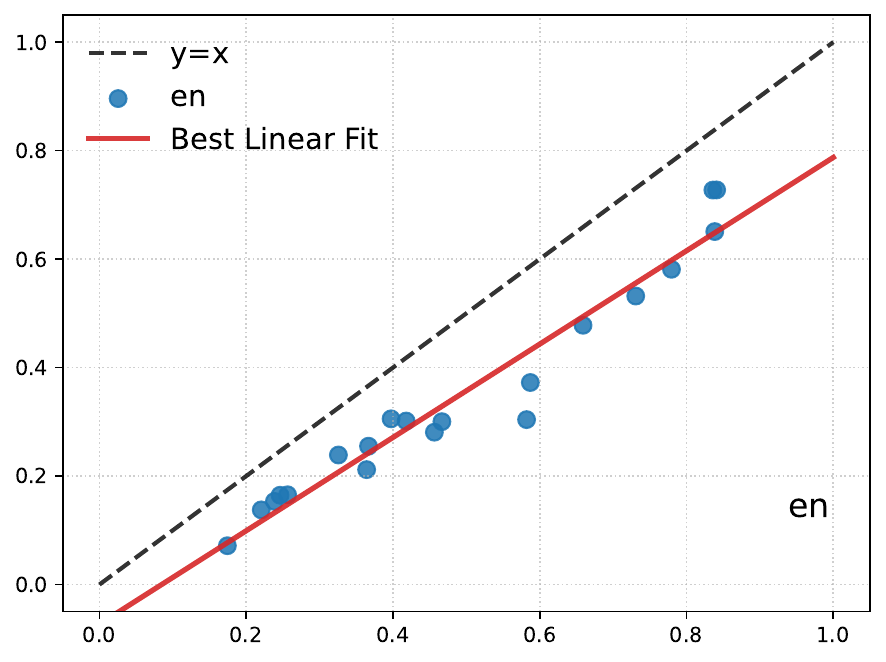} &
    \includegraphics[width=0.25\textwidth]{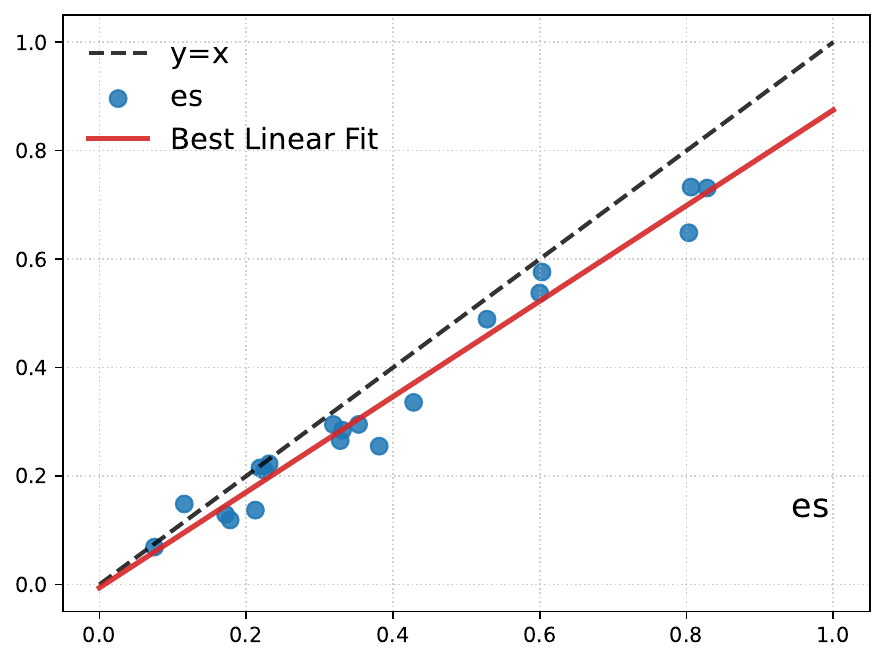} \\
    & \footnotesize (a) German (de) & \footnotesize (b) English (en) & \footnotesize (c) Spanish (es) \\[0.4cm]

     & 
    \includegraphics[width=0.25\textwidth]{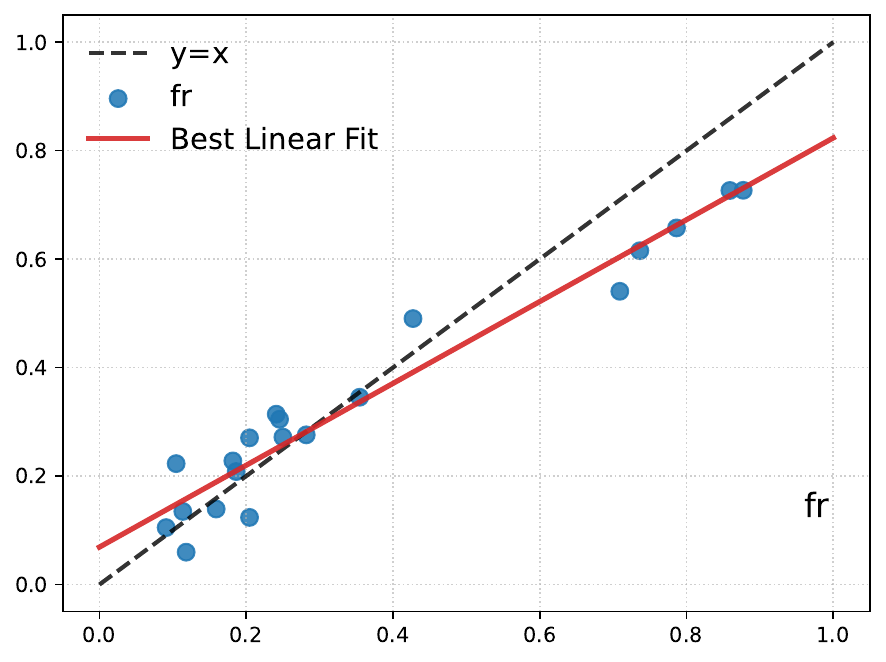} &
    \includegraphics[width=0.25\textwidth]{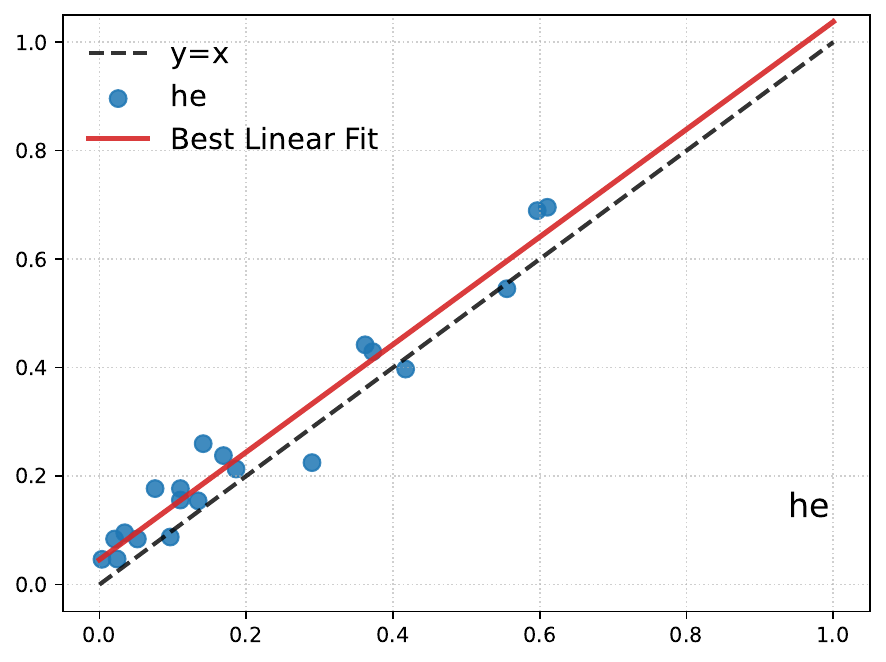} &
    \includegraphics[width=0.25\textwidth]{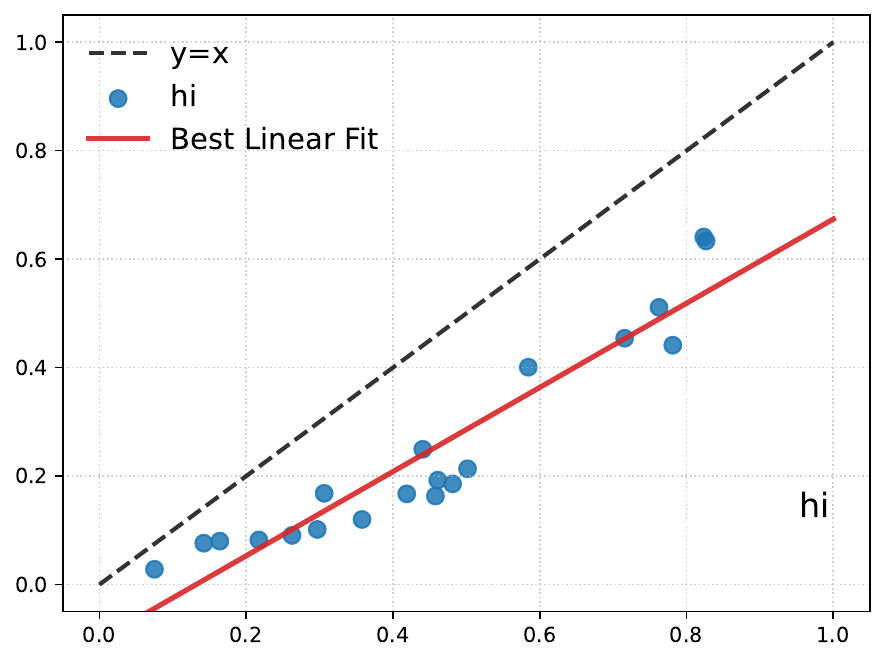} \\
    & \footnotesize (d) French (fr) & \footnotesize (e) Hebrew (he) & \footnotesize (f) Hindi (hi) \\[0.4cm]

     & 
    \includegraphics[width=0.25\textwidth]{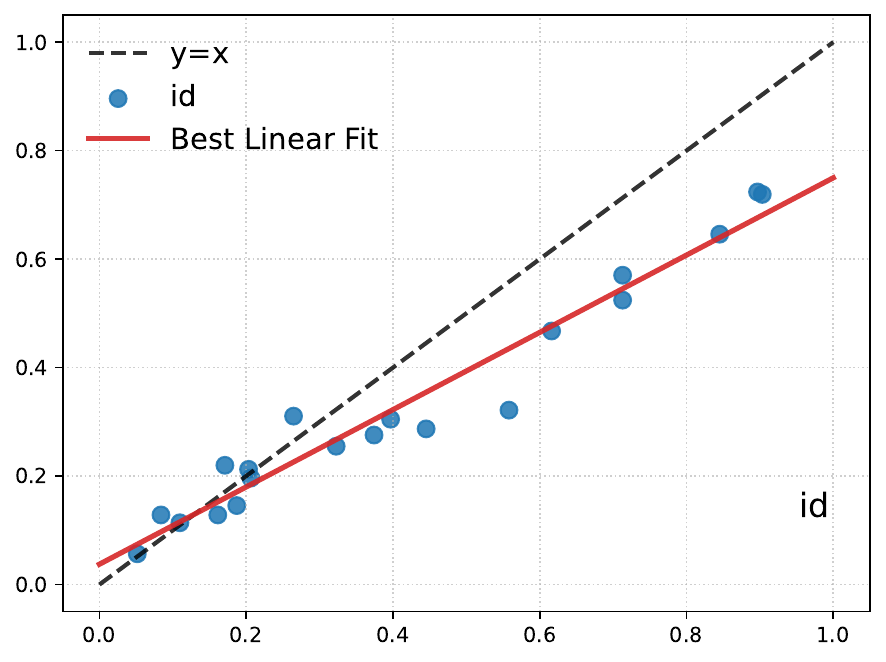} &
    \includegraphics[width=0.25\textwidth]{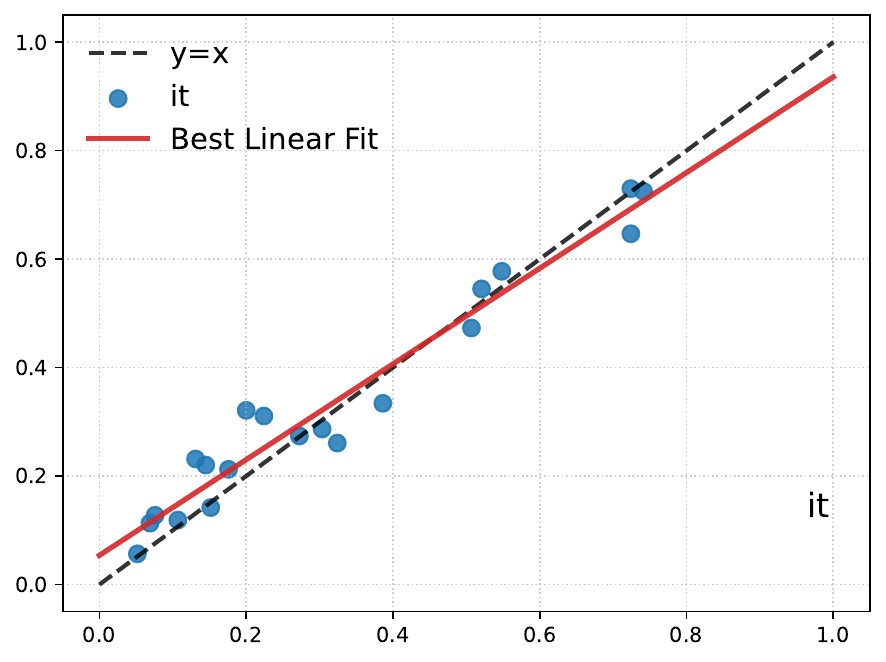} &
    \includegraphics[width=0.25\textwidth]{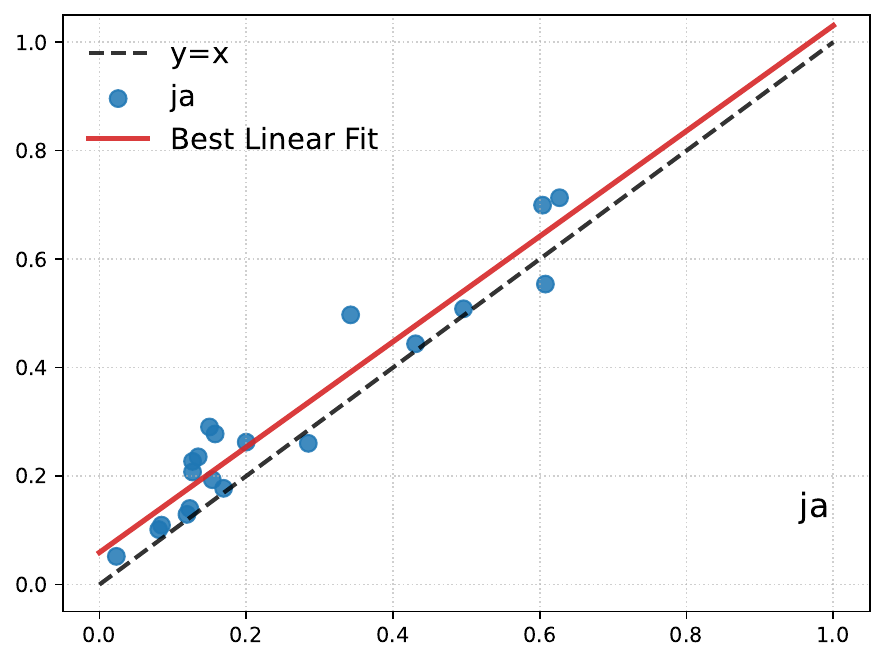} \\
    & \footnotesize (g) Indonesian (id) & \footnotesize (h) Italian (it) & \footnotesize (i) Japanese (ja) \\[0.4cm]

    & 
    \includegraphics[width=0.25\textwidth]{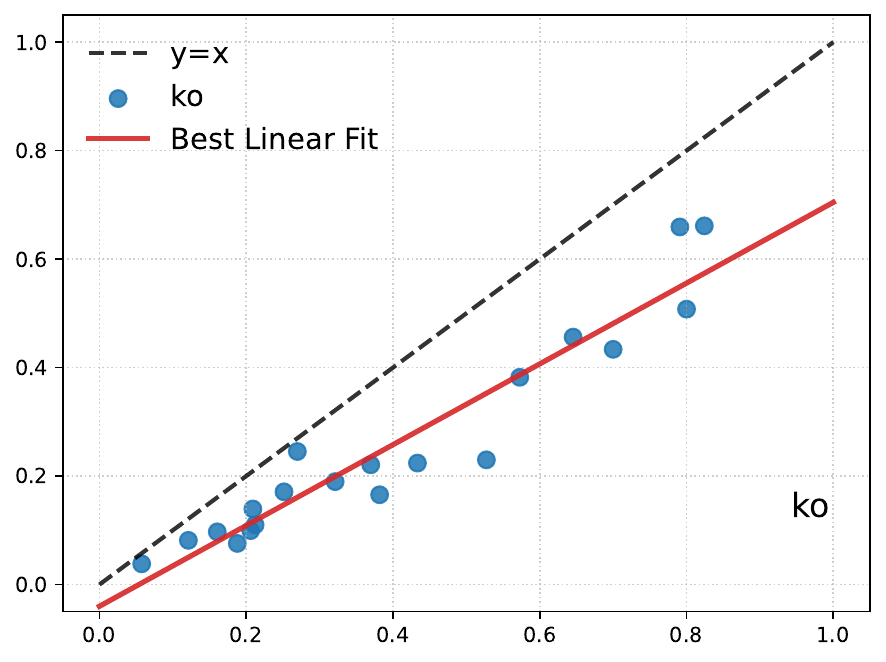} &
    \includegraphics[width=0.25\textwidth]{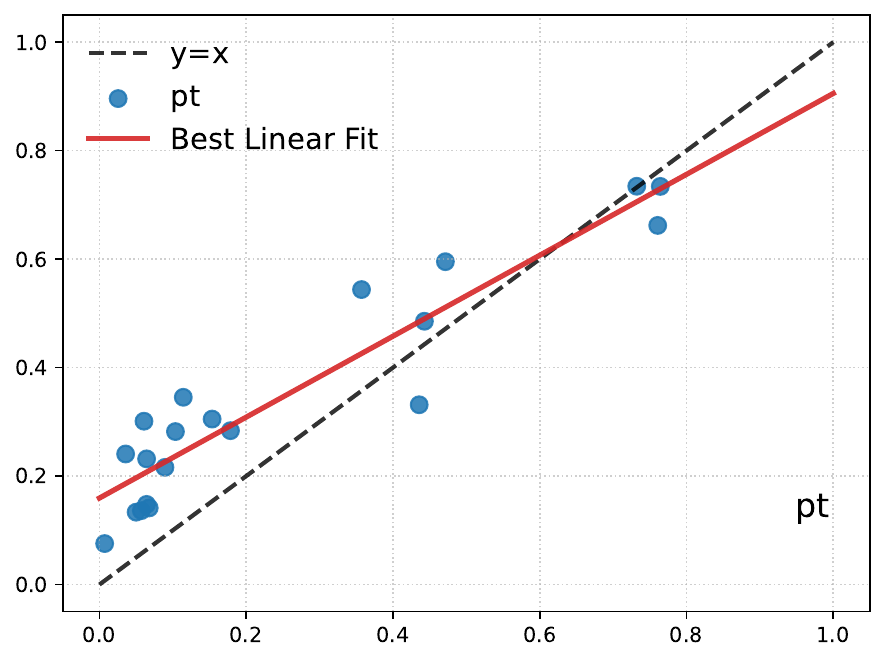} &
    \includegraphics[width=0.25\textwidth]{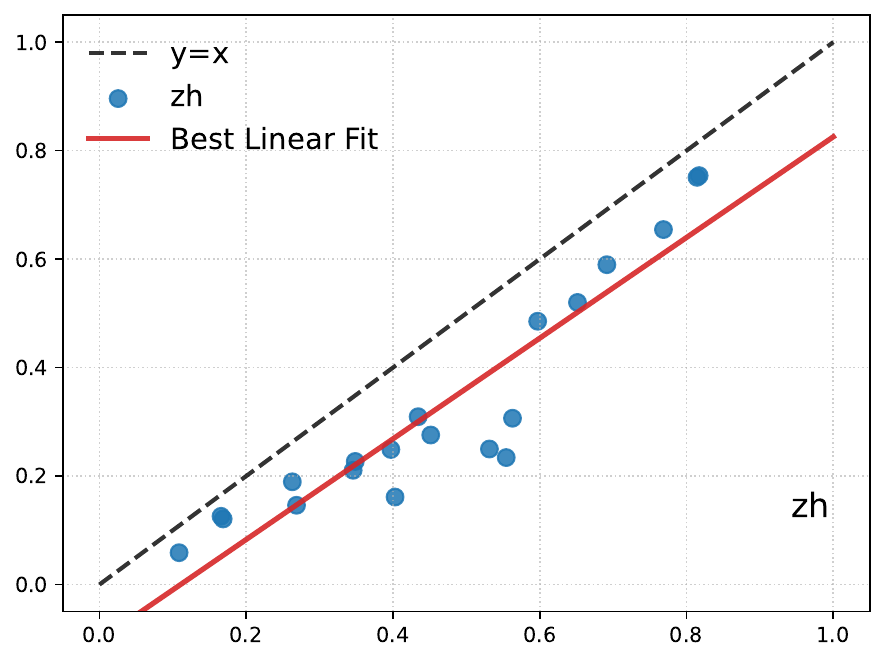} \\
    & \footnotesize (j) Korean (ko) & \footnotesize (k) Portuguese (pt) & \footnotesize (l) Chinese (zh) \\[0.4cm]
    
    & \multicolumn{3}{c}{\textbf{Source Accuracy}} \\
    
    \caption{Detailed language-by-language breakdown of target accuracy versus source accuracy curves across the ECLeKTic dataset.}
    \label{fig:eclektic_target_plots}
\end{longtable}
\end{center}

\twocolumn

\clearpage

\onecolumn
\paragraph{MMLU-ProX-Lite}
\begin{center}
\begin{longtable}{c ccc}
    \multirow{24}{*}{\rotatebox{90}{\textbf{Target Accuracy}}} &
    \includegraphics[width=0.25\textwidth]{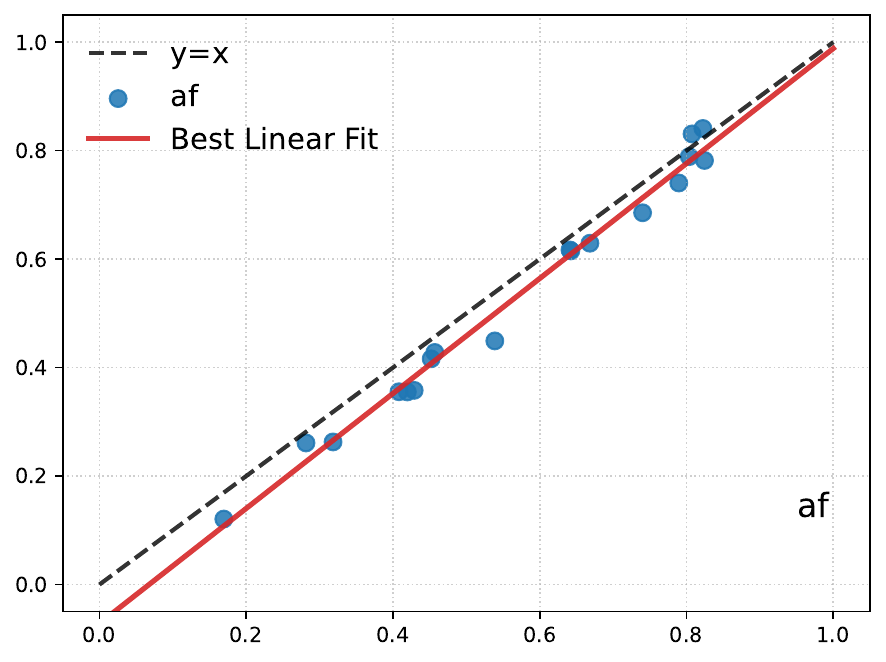} &
    \includegraphics[width=0.25\textwidth]{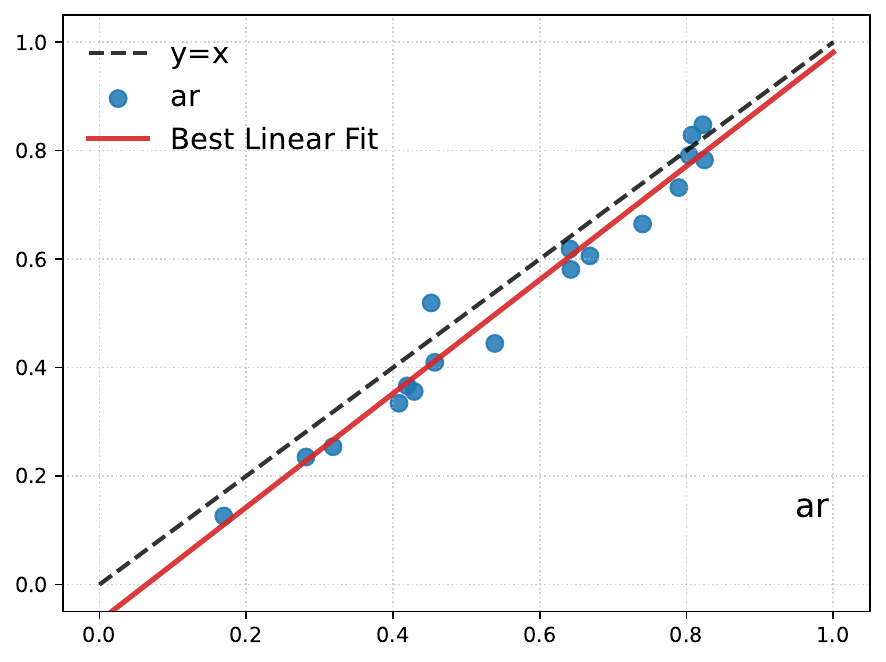} &
    \includegraphics[width=0.25\textwidth]{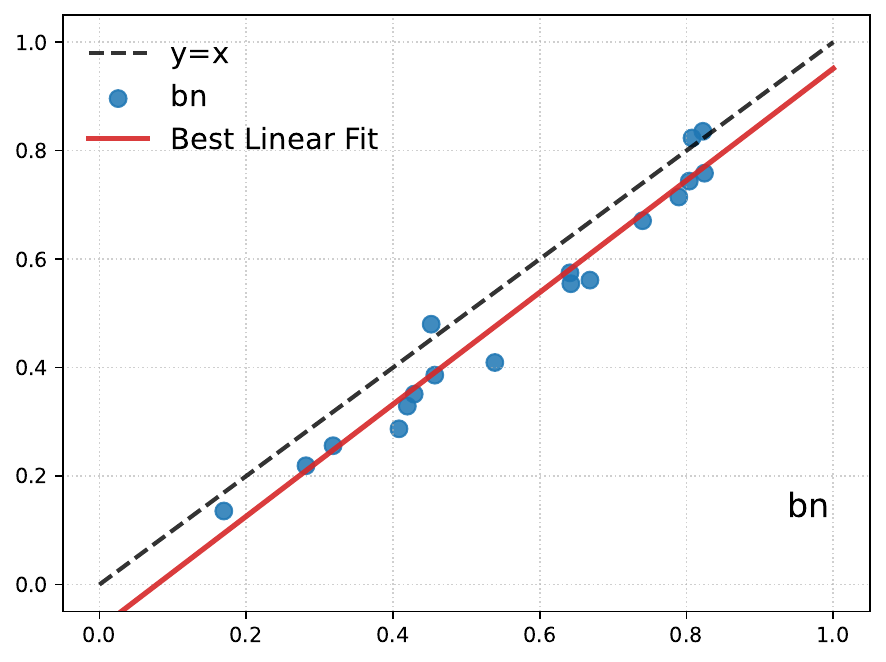} \\ &
    \footnotesize (a) Afrikaans (af) & \footnotesize (b) Arabic (ar) & \footnotesize (c) Bengali (bn) \\[0.4cm]

    &
    \includegraphics[width=0.25\textwidth]{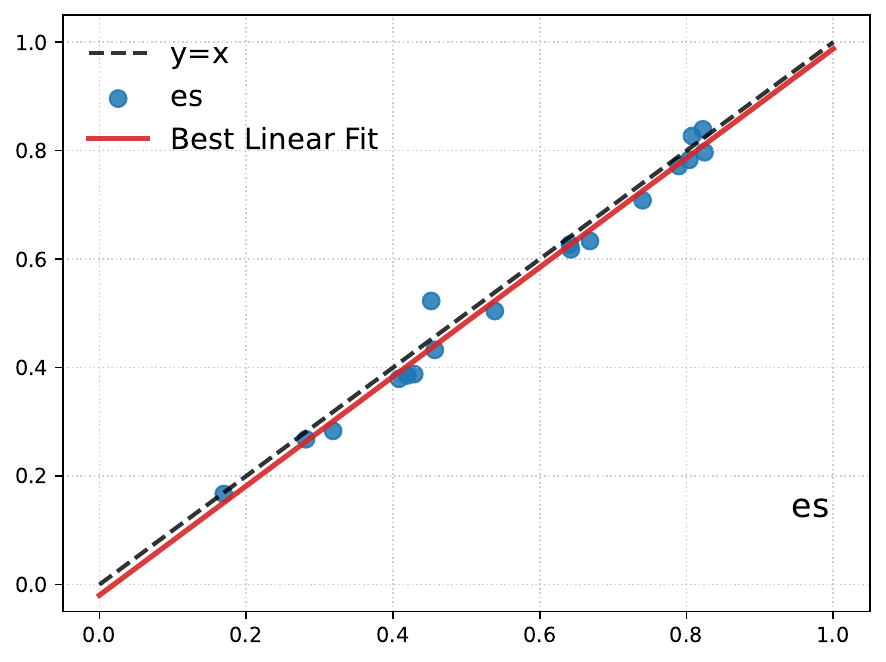} &
    \includegraphics[width=0.25\textwidth]{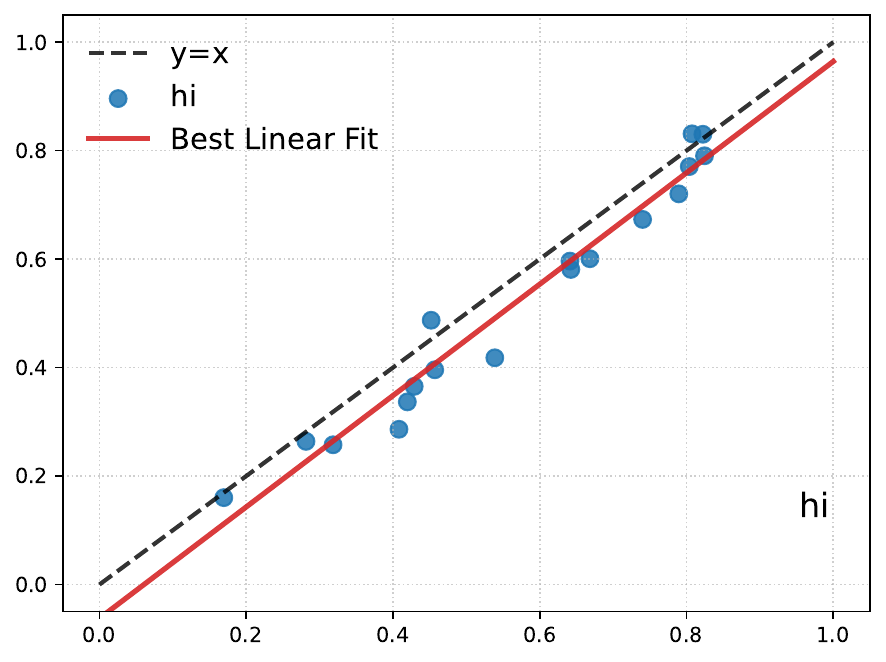} &
    \includegraphics[width=0.25\textwidth]{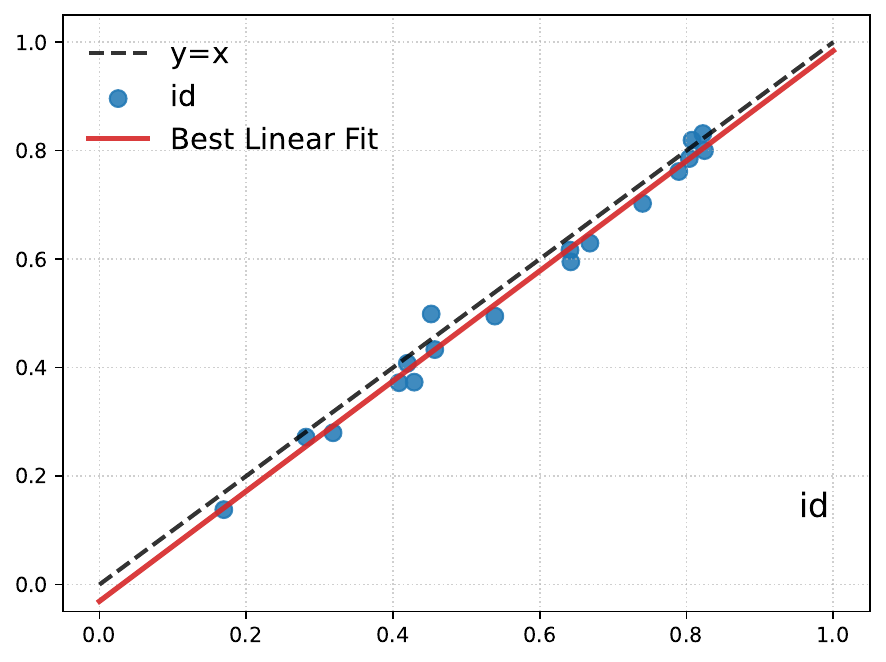} \\ &
    \footnotesize (d) Spanish (es) & \footnotesize (e) Hindi (hi) & \footnotesize (f) Indonesian (id) \\[0.4cm]

    &
    \includegraphics[width=0.25\textwidth]{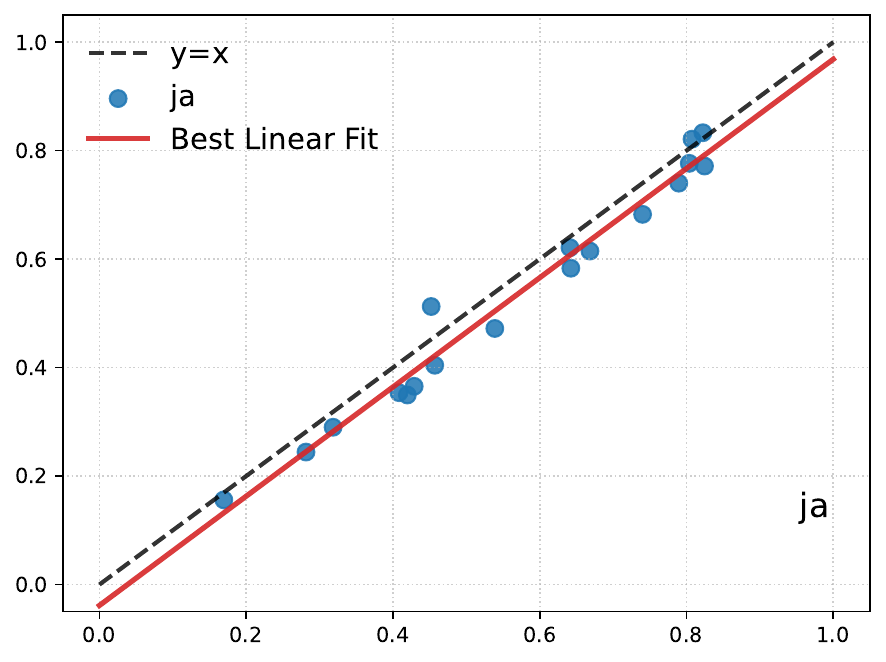} &
    \includegraphics[width=0.25\textwidth]{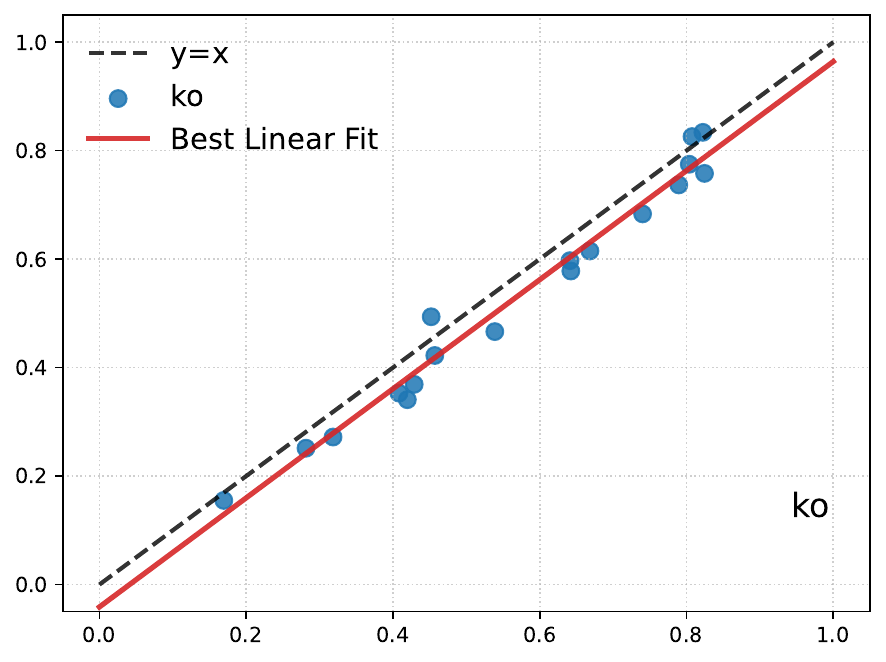} &
    \includegraphics[width=0.25\textwidth]{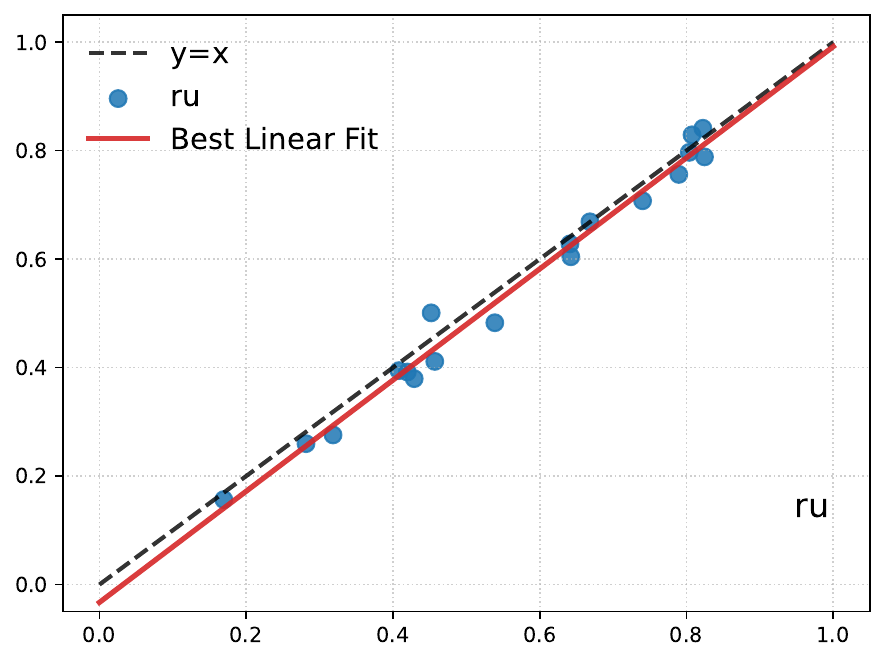} \\ &
    \footnotesize (g) Japanese (ja) & \footnotesize (h) Korean (ko) & \footnotesize (i) Russian (ru) \\[0.4cm]

    &
    \includegraphics[width=0.25\textwidth]{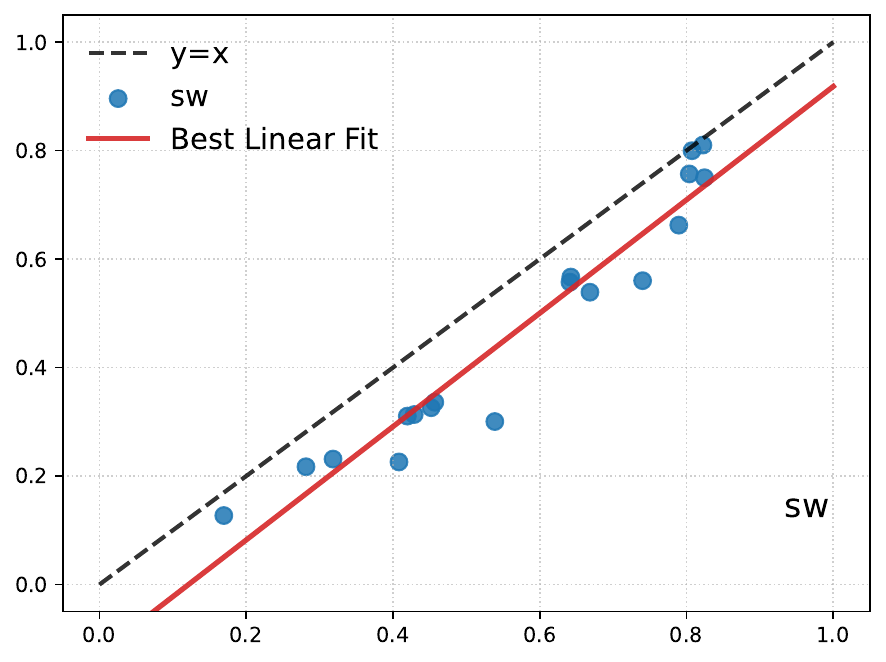} &
    \includegraphics[width=0.25\textwidth]{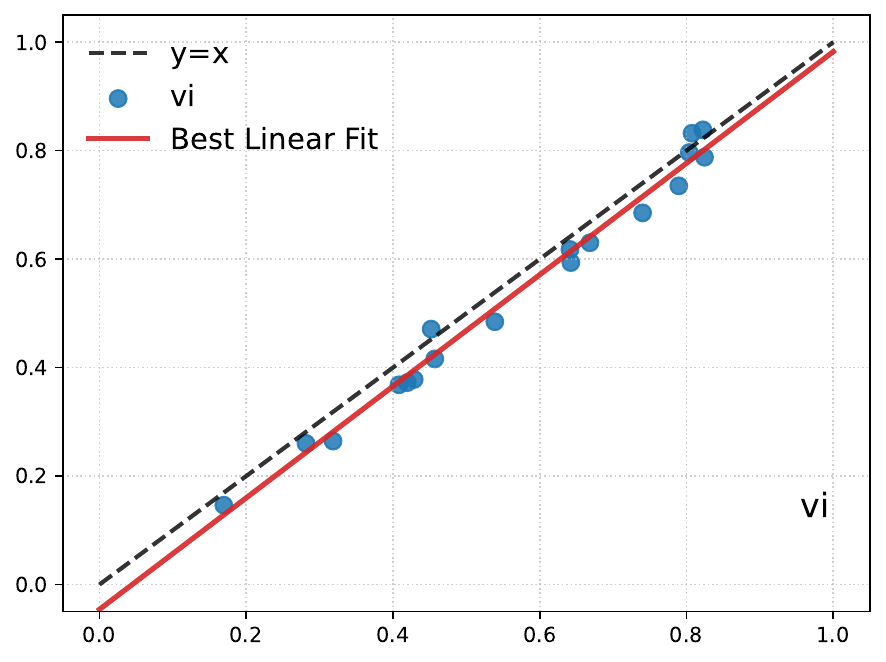} &
    \includegraphics[width=0.25\textwidth]{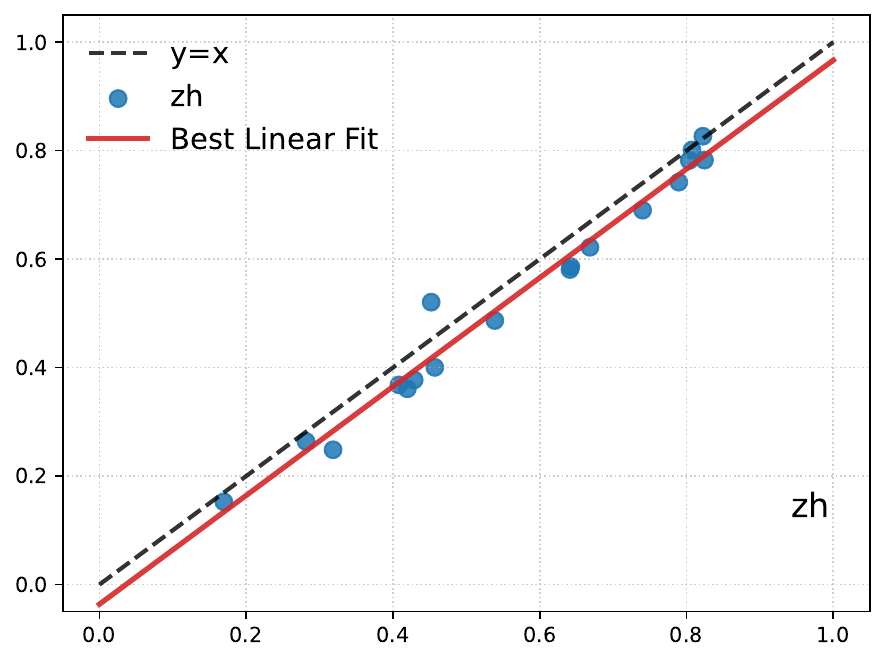} \\ &
    \footnotesize (j) Swahili (sw) & \footnotesize (k) Vietnamese (vi) & \footnotesize (l) Chinese (zh) \\
    
    & \multicolumn{3}{c}{\textbf{Source Accuracy}} \\
    
    \caption{Detailed language-by-language breakdown of target accuracy versus source accuracy curves across the MMLU-ProX-Lite dataset.}

    \label{fig:mmlu_target_plots}
\end{longtable}
\end{center}
\twocolumn
\clearpage
\onecolumn
\paragraph{MGSMv2}
\begin{center}
\begin{longtable}{c ccc}
    \multirow{24}{*}{\rotatebox{90}{\textbf{Target Accuracy}}} & \includegraphics[width=0.25\textwidth]{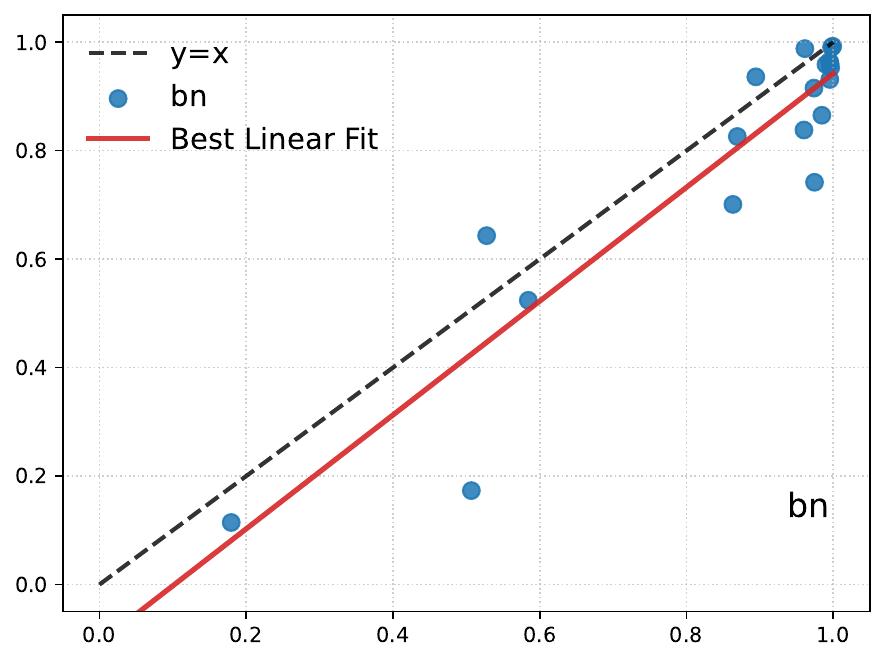} &
    \includegraphics[width=0.25\textwidth]{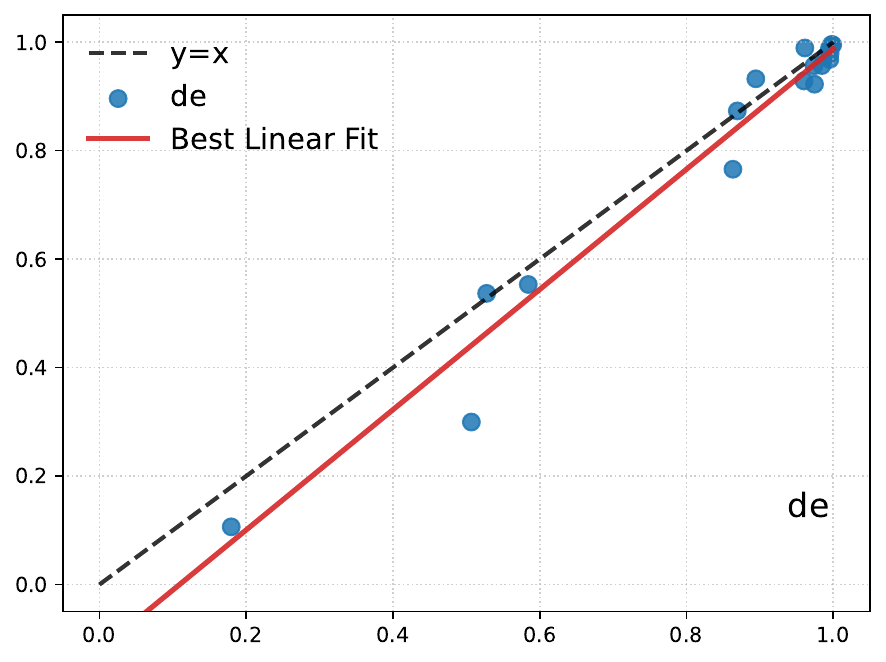} &
    \includegraphics[width=0.25\textwidth]{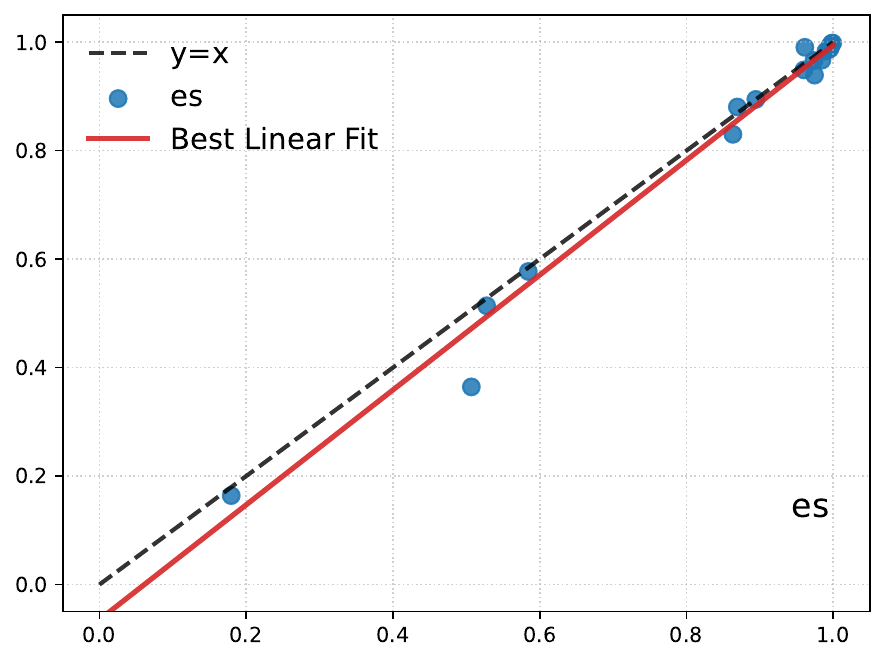} \\ &
    \footnotesize (a) Bengali (bn) & \footnotesize (b) German (de) & \footnotesize (c) Spanish (es) \\[0.4cm]

    &
    \includegraphics[width=0.25\textwidth]{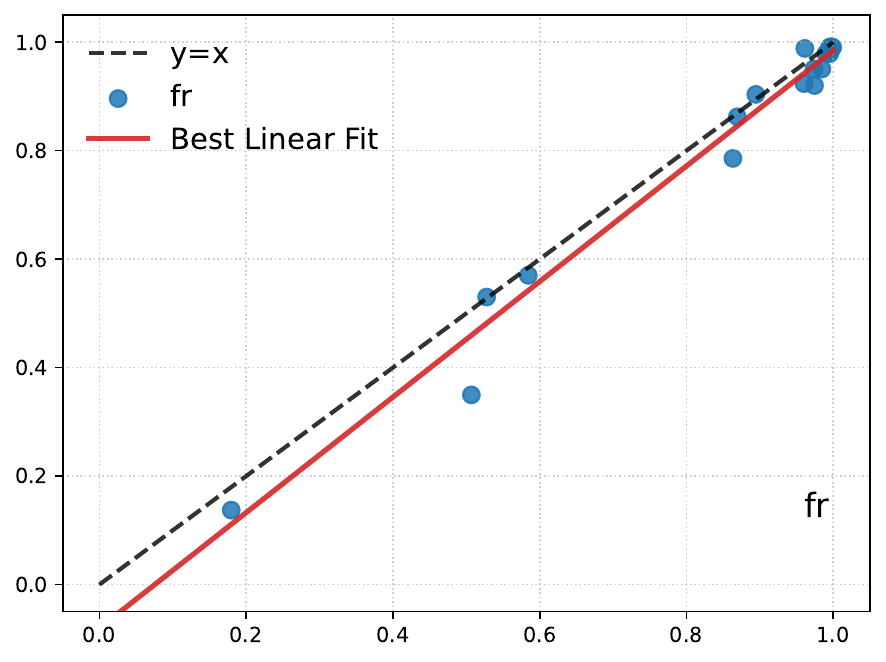} &
    \includegraphics[width=0.25\textwidth]{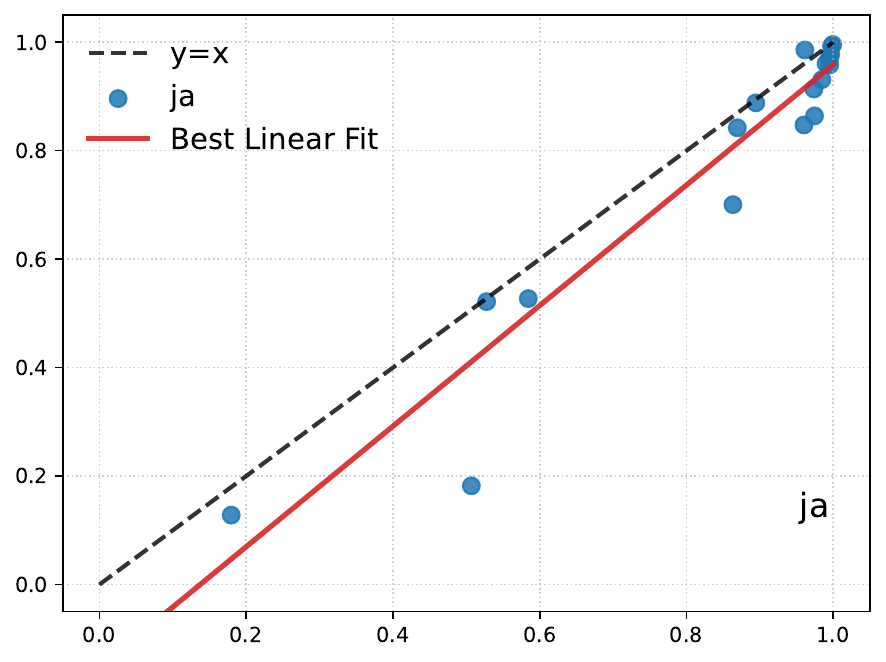} &
    \includegraphics[width=0.25\textwidth]{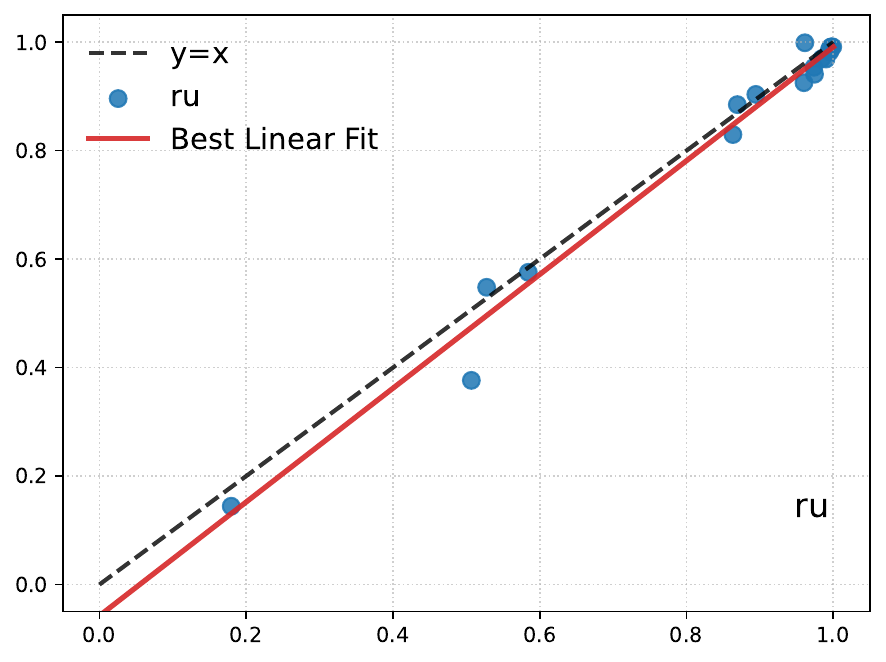} \\ &
    \footnotesize (d) French (fr) & \footnotesize (e) Japanese (ja) & \footnotesize (f) Russian (ru) \\[0.4cm]

    &
    \includegraphics[width=0.25\textwidth]{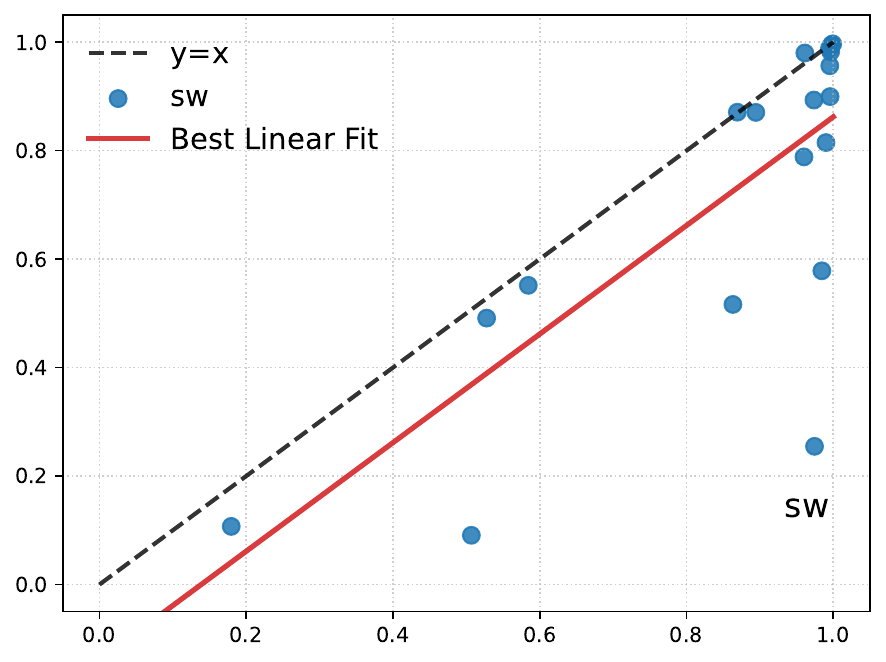} &
    \includegraphics[width=0.25\textwidth]{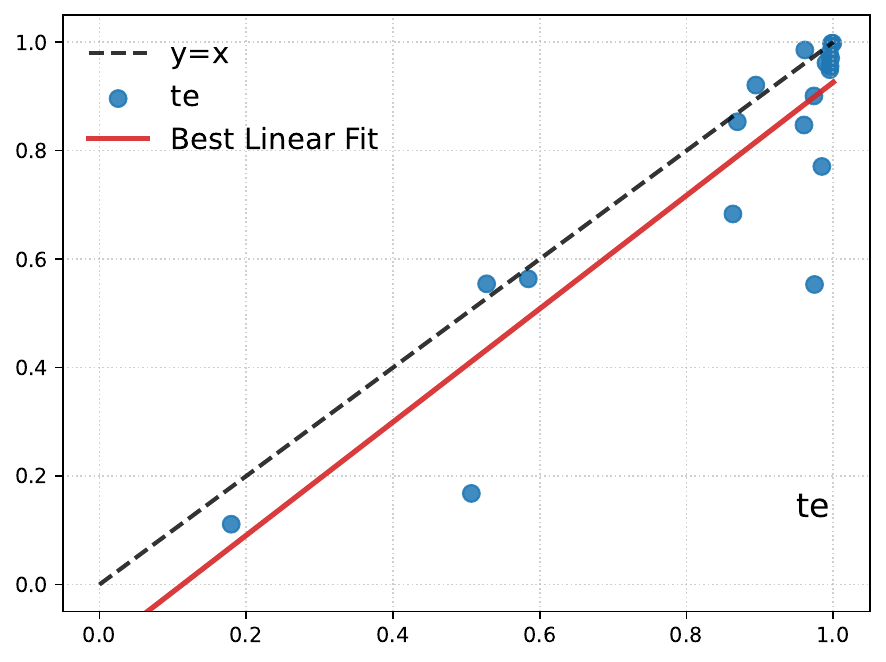} &
    \includegraphics[width=0.25\textwidth]{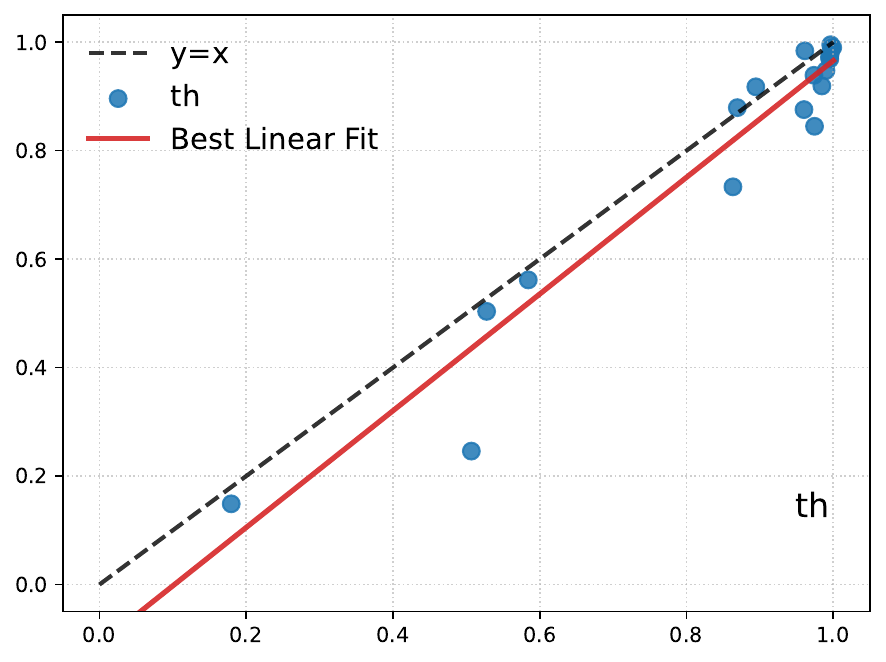} \\ &
    \footnotesize (g) Swahili (sw) & \footnotesize (h) Telugu (te) & \footnotesize (i) Thai (th) \\[0.4cm]

    &
    \includegraphics[width=0.25\textwidth]{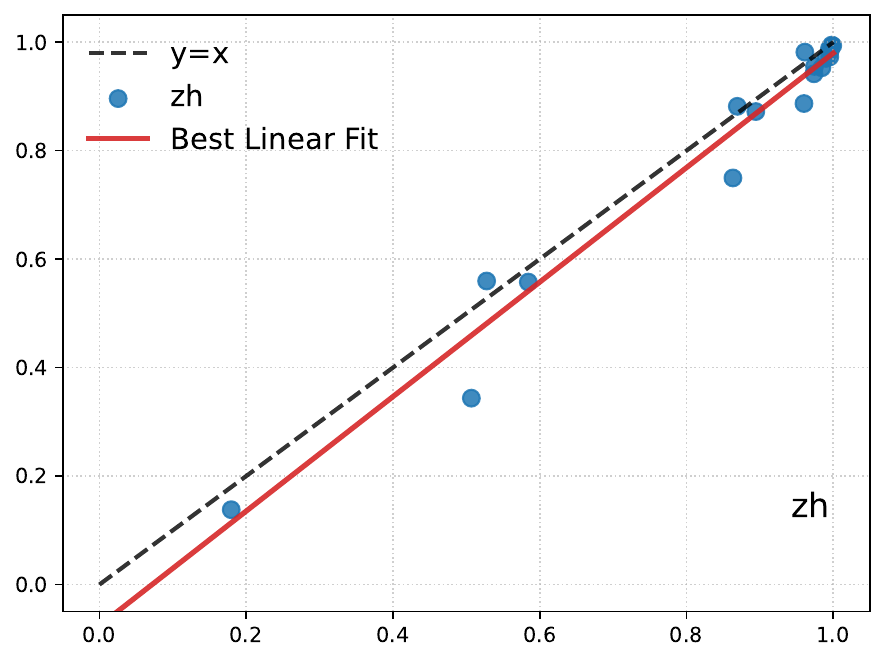} &
     &  \\ &
    \footnotesize (j) Chinese (zh) &  &  \\
    
    & \multicolumn{3}{c}{\textbf{Source Accuracy}} \\
    
    \caption{Detailed language-by-language breakdown of target accuracy versus source accuracy curves across the MGSMv2 dataset.}

    \label{fig:mgsmv2_target_plots}
\end{longtable}
\end{center}
\twocolumn

\clearpage

\begin{table*}[h]
    \centering
    
    \resizebox{0.75\textwidth}{!}{ 
    \begin{tabular}{l cccccc}
        \toprule
        \multirow{2}{*}{\textbf{Model}} & \multicolumn{2}{c}{\textbf{Target Accuracy (\%) ($\uparrow$)}} & \multicolumn{2}{c}{\textbf{XLT Gap (\%) ($\downarrow$)}} & \multicolumn{2}{c}{\textbf{HAT Score ($\uparrow$)}} \\
        \cmidrule(lr){2-3} \cmidrule(lr){4-5} \cmidrule(lr){6-7}
        & Estimate & 95\% CI & Estimate & 95\% CI & Estimate & 95\% CI \\
        \midrule
        
        \textsc{Gemini-2.5-Flash}               & 44.5 & [41.1, 48.1] & 6.4 & [4.3, 8.5] & 81.7 & [78.6, 84.7] \\
        \begin{tabular}{@{}l@{}}\textsc{Gemini-2.5-Flash} \\ \textsc{Thinking-Off}\end{tabular} & 29.4 & [26.5, 32.5] & 13.3 & [10.5, 16.3] & 60.4 & [55.9, 64.8] \\
        \addlinespace
        \midrule
        \textsc{Gemini-3-Flash}                 & 60.5 & [56.8, 64.3] & 8.0 & [5.6, 10.4] & 84.6 & [82.1, 87.1] \\
        \textsc{Gemini-3-Flash Low}             & 60.2 & [56.5, 64.0] & 8.9 & [6.5, 11.2] & 84.0 & [81.6, 86.7] \\
        \textsc{Gemini-3-Flash Minimal}         & 52.3 & [48.9, 56.1] & 13.4 & [10.2, 16.3] & 74.2 & [70.9, 77.5] \\
        \addlinespace
        \midrule
        
        \textsc{Gemma-3-1B}                     & 4.6 & [3.5, 6.0] & 0.8 & [-0.3, 2.2] & 48.2 & [34.9, 61.0] \\
        \textsc{Gemma-3-4B}                     & 9.6 & [7.7, 11.8] & 1.9 & [0.3, 4.0] & 56.4 & [47.0, 65.9] \\
        \textsc{Gemma-3-12B}                    & 15.2 & [12.8, 17.9] & 2.1 & [-0.2, 4.3] & 64.1 & [56.8, 71.4] \\
        \textsc{Gemma-3-27B}                    & 19.8 & [17.2, 23.0] & 6.5 & [4.1, 8.9] & 63.8 & [58.4, 69.6] \\
        \addlinespace
        \midrule
        
        \textsc{Qwen-3-4B}                      & 10.8 & [8.9, 13.1] & 2.9 & [1.1, 5.1] & 56.5 & [47.8, 64.9] \\
        \textsc{Qwen-3-30B-A3B}                 & 16.6 & [14.1, 19.5] & 2.8 & [0.7, 5.1] & 65.9 & [59.7, 72.6] \\
        \addlinespace
        \midrule
        
        \textsc{GPT-OSS-20B}                    & 18.0 & [15.3, 21.2] & 1.6 & [0.2, 3.2] & 80.1 & [74.7, 85.3] \\
        \textsc{GPT-OSS-120B}                   & 25.4 & [22.4, 28.7] & 2.4 & [0.6, 4.1] & 82.2 & [77.9, 86.1] \\
        \addlinespace
        \midrule
        
        \textsc{Gemma-4-E2B-IT}                 & 9.0 & [7.2, 11.0] & 2.8 & [0.9, 4.9] & 51.8 & [42.4, 61.3] \\
        \textsc{Gemma-4-E4B-IT}                 & 10.7 & [8.7, 13.0] & 5.0 & [3.1, 6.8] & 58.4 & [51.2, 65.7] \\
        \textsc{Gemma-4-26B-A4B-IT}             & 20.3 & [17.4, 23.3] & 5.7 & [3.0, 8.3] & 63.9 & [57.9, 69.6] \\
        \textsc{Gemma-4-31B-IT}                 & 22.6 & [19.5, 26.0] & 5.5 & [3.1, 8.0] & 68.6 & [62.9, 74.1] \\
        \addlinespace
        \midrule
        
        \textsc{Claude-Haiku-4.5}               & 20.8 & [17.8, 23.9] & 8.4 & [5.9, 11.3] & 59.8 & [53.8, 65.7] \\
        \textsc{Claude-Sonnet-4.6}              & 44.1 & [40.2, 48.0] & 8.7 & [5.6, 11.5] & 74.7 & [70.9, 78.5] \\
        \textsc{Claude-Opus-4.7}                & 45.2 & [41.6, 49.1] & 10.3 & [7.4, 13.5] & 73.5 & [69.7, 77.2] \\
        
        \bottomrule
    \end{tabular}
    }
    \caption{\textbf{Comprehensive evaluation results for \eclektic.} Complete estimates and 95\% confidence intervals for average target accuracy, XLT gap, and HAT score across all tested model configurations.}
    \label{tab:eclektic_full_metrics}
\end{table*}

\begin{table*}[htbp!]
    \centering
    
    \resizebox{0.75\textwidth}{!}{ 
    \begin{tabular}{l cccccc}
        \toprule
        \multirow{2}{*}{\textbf{Model}} & \multicolumn{2}{c}{\textbf{Target Accuracy (\%) ($\uparrow$)}} & \multicolumn{2}{c}{\textbf{XLT Gap (\%) ($\downarrow$)}} & \multicolumn{2}{c}{\textbf{HAT Score ($\uparrow$)}} \\
        \cmidrule(lr){2-3} \cmidrule(lr){4-5} \cmidrule(lr){6-7}
        & Estimate & 95\% CI & Estimate & 95\% CI & Estimate & 95\% CI \\
        \midrule
        \textsc{Gemini-2.5-Flash}               & 73.4 & [73.3, 75.5] & -1.9  & [-3.4, -0.5]   & 97.5 & [96.2, 98.8] \\
        \begin{tabular}{@{}l@{}}\textsc{Gemini-2.5-Flash} \\ \textsc{Thinking-Off}\end{tabular} & 61.6     & [60.1, 62.2]           & 5.9   & [4.5, 7.4]           & 88.9	 &[87.1, 90.3]           \\
        \addlinespace
        \midrule
        \textsc{Gemini-3-Flash}                 & 82.1 & [81.6, 82.6] & -1.4 & [-2.6, -0.3] & 98.8 & [97.8, 99.9] \\
        \textsc{Gemini-3-Flash Low}             & 83.0 & [82.5, 83.5] & -0.8 & [-2.0, 0.4]  & 98.3 & [97.3, 99.3] \\
        \textsc{Gemini-3-Flash Minimal}         & 77.2 & [76.6, 77.8] & 3.2  & [1.9, 4.6]   & 93.5 & [92.1, 94.8] \\
        \addlinespace
        \midrule
        
        \textsc{Gemma-3-1B}                     & 14.2 & [13.7, 14.7] & 2.7  & [0.8, 4.7]  & 58.4 & [52.3, 64.1] \\
        \textsc{Gemma-3-4B}                     & 24.6 & [23.9, 25.2] & 3.6  & [1.5, 5.6]  & 67.5 & [63.1, 71.7] \\
        \textsc{Gemma-3-12B}                    & 34.9 & [34.3, 35.6] & 7.0  & [4.7, 9.2]  & 71.6 & [68.2, 74.7] \\
        \textsc{Gemma-3-27B}                    & 39.4 & [38.6, 40.1] & 6.4  & [4.3, 8.2]  & 77.2 & [74.4, 80.1] \\
        \addlinespace
        \midrule
        
        \textsc{Qwen-3-4B}                      & 32.7 & [32.0, 33.4] & 8.1  & [6.0, 10.3] & 69.3 & [65.8, 72.6] \\
        \textsc{Qwen-3-30B-A3B}                 & 43.4 & [42.7, 44.2] & 10.5 & [8.5, 12.4] & 74.8 & [71.9, 77.6] \\
        \addlinespace
        \midrule
        
        \textsc{GPT-OSS-20B}                    & 66.3 & [65.7, 66.9] & 7.7  & [6.5, 9.0]  & 88.9 & [87.6, 90.3] \\
        \textsc{GPT-OSS-120B}                   & 72.3 & [71.8, 73.0] & 6.6  & [5.4, 8.0]  & 90.6 & [89.2, 92.0] \\
        \addlinespace
        \midrule
        
        \textsc{Gemma-4-E2B-IT}                 & 26.1 & [25.5, 26.7] & 5.7  & [3.2, 8.6]  & 59.3 & [54.7, 63.5] \\
        \textsc{Gemma-4-E4B-IT}                 & 35.8 & [35.2, 36.5] & 7.1  & [5.1, 9.0]  & 74.8 & [71.6, 77.8] \\
        \textsc{Gemma-4-26B-A4B-IT}             & 48.7 & [48.0, 49.5] & 4.8  & [2.6, 7.2]  & 78.7 & [76.2, 81.1] \\
        \textsc{Gemma-4-31B-IT}                 & 59.4 & [58.6, 60.1] & 4.8  & [2.9, 6.8]  & 84.7 & [82.5, 86.7] \\
        \addlinespace
        \midrule
        
        \textsc{Claude-Haiku-4.5}               & 44.6 & [43.9, 45.3] & 0.6  & [-1.9, 3.1] & 76.5 & [73.5, 79.2] \\
        \textsc{Claude-Sonnet-4.6}              & 59.5 & [58.7, 60.2] & 7.4  & [5.7, 9.0]  & 85.4 & [83.6, 87.3] \\
        \textsc{Claude-Opus-4.7}                & 76.7 & [76.2, 77.5] & 5.8  & [3.8, 7.3]  & 89.3 & [87.8, 90.9] \\
        \addlinespace
        
        \bottomrule
    \end{tabular}
    }
    \caption{\textbf{Comprehensive evaluation results for \mmlu.} Complete estimates and 95\% confidence intervals for average target accuracy, XLT gap, and HAT score across all tested model configurations.}
    \label{tab:mmlu_full_metrics}
\end{table*}

\begin{table*}[h]
    \centering
    
    \resizebox{0.75\textwidth}{!}{ 
    \begin{tabular}{l cccccc}
        \toprule
        \multirow{2}{*}{\textbf{Model}} & \multicolumn{2}{c}{\textbf{Target Accuracy (\%) ($\uparrow$)}} & \multicolumn{2}{c}{\textbf{XLT Gap (\%) ($\downarrow$)}} & \multicolumn{2}{c}{\textbf{HAT Score ($\uparrow$)}} \\
        \cmidrule(lr){2-3} \cmidrule(lr){4-5} \cmidrule(lr){6-7}
        & Estimate & 95\% CI & Estimate & 95\% CI & Estimate & 95\% CI \\
        \midrule
        
        \textsc{Gemini-2.5-Flash}               & 98.2 & [97.7, 98.6] & 1.3  & [0.8, 1.9]   & 98.7 & [98.1, 99.2] \\
        \begin{tabular}{@{}l@{}}\textsc{Gemini-2.5-Flash} \\ \textsc{Thinking-Off}\end{tabular} & 55.6 & [53.9, 57.5] & 2.8  & [-0.1, 5.4]  & 89.8 & [86.9, 92.4] \\
        \addlinespace
        \midrule
        \textsc{Gemini-3-Flash}                 & 99.3 & [99.1, 99.6] & 0.4  & [0.1, 0.8]   & 99.6 & [99.3, 99.9] \\
        \textsc{Gemini-3-Flash Low}             & 99.4 & [99.2, 99.6] & 0.5  & [0.2, 1.0]   & 99.5 & [99.1, 99.8] \\
        \textsc{Gemini-3-Flash Minimal}         & 86.5 & [85.4, 87.7] & 0.4  & [-1.8, 2.5]  & 96.2 & [94.8, 97.5] \\
        \addlinespace
        \midrule
        
        \textsc{Gemma-3-1B}                     & 25.9 & [24.6, 27.4] & 24.7 & [21.3, 28.3] & 47.7 & [43.7, 51.8] \\
        \textsc{Gemma-3-4B}                     & 72.9 & [71.5, 74.3] & 13.4 & [10.8, 15.9] & 82.6 & [80.3, 85.0] \\
        \textsc{Gemma-3-12B}                    & 88.1 & [87.1, 89.1] & 7.9  & [6.3, 9.7]   & 91.3 & [89.7, 92.8] \\
        \textsc{Gemma-3-27B}                    & 93.3 & [92.6, 94.1] & 4.1  & [2.7, 5.4]   & 95.3 & [94.0, 96.5] \\
        \addlinespace
        \midrule
        
        \textsc{Qwen-3-4B}                      & 79.4 & [77.9, 80.8] & 18.1 & [16.2, 20.1] & 81.3 & [79.4, 83.2] \\
        \textsc{Qwen-3-30B-A3B}                 & 88.6 & [87.6, 89.6] & 9.8  & [8.2, 11.4]  & 89.8 & [88.4, 91.3] \\
        \addlinespace
        \midrule
        
        \textsc{GPT-OSS-20B}                    & 95.2 & [94.6, 95.8] & 3.8  & [3.0, 4.6]   & 96.1 & [95.4, 96.9] \\
        \textsc{GPT-OSS-120B}                   & 97.0 & [96.4, 97.5] & 2.6  & [2.0, 3.2]   & 97.4 & [96.8, 98.0] \\
        \addlinespace
        \midrule
        
        \textsc{Gemma-4-E2B-IT}                 & 13.0 & [11.9, 14.1] & 5.0  & [2.2, 8.0]   & 53.7 & [45.1, 62.6] \\
        \textsc{Gemma-4-E4B-IT}                 & 17.5 & [16.3, 18.8] & 4.4  & [2.3, 6.8]           & 68.0 & [60.3, 74.1] \\
        \textsc{Gemma-4-26B-A4B-IT}             & 96.5 & [95.9, 97.1] & 3.0  & [1.9, 4.2]   & 96.8 & [95.7, 97.7] \\
        \textsc{Gemma-4-31B-IT}                 & 98.3 & [97.8, 98.6] & 1.4  & [0.6, 2.1]   & 98.4 & [97.8, 99.0] \\
        \addlinespace
        \midrule
        
        \textsc{Claude-Haiku-4.5}               & 54.0 & [52.3, 55.9] & -1.3 & [-4.4, 2.2]  & 86.0 & [82.1, 89.6] \\
        \textsc{Claude-Sonnet-4.6}              & 98.7 & [98.3, 99.1] & -2.6 & [-4.9, -0.5] & 99.4 & [98.7, 100.0] \\
        \textsc{Claude-Opus-4.7}                & 90.4 & [89.4, 91.5] & -1.0 & [-2.8, 1.1]  & 97.3 & [96.0, 98.4] \\
        \addlinespace
        
        \bottomrule
    \end{tabular}
    }
    \caption{\textbf{Comprehensive evaluation results for \newmgsm.} Complete estimates and 95\% confidence intervals for average target accuracy, XLT Gap, and HAT score across all tested model configurations.}
    \label{tab:mgsm2_full_metrics}
\end{table*}

\clearpage

\setlength{\tabcolsep}{1pt} 
\onecolumn
\subsection{HAT Plots for Different Datasets}
\subsubsection{ECLeKTic HAT Plots}
Detailed evaluation curves and HAT score transfer profiles for the ECLeKTic dataset are provided below.
\begin{longtable}{>{\centering\arraybackslash}p{0.24\textwidth} >{\centering\arraybackslash}p{0.24\textwidth} >{\centering\arraybackslash}p{0.24\textwidth} >{\centering\arraybackslash}p{0.24\textwidth}}
\includegraphics[width=\linewidth]{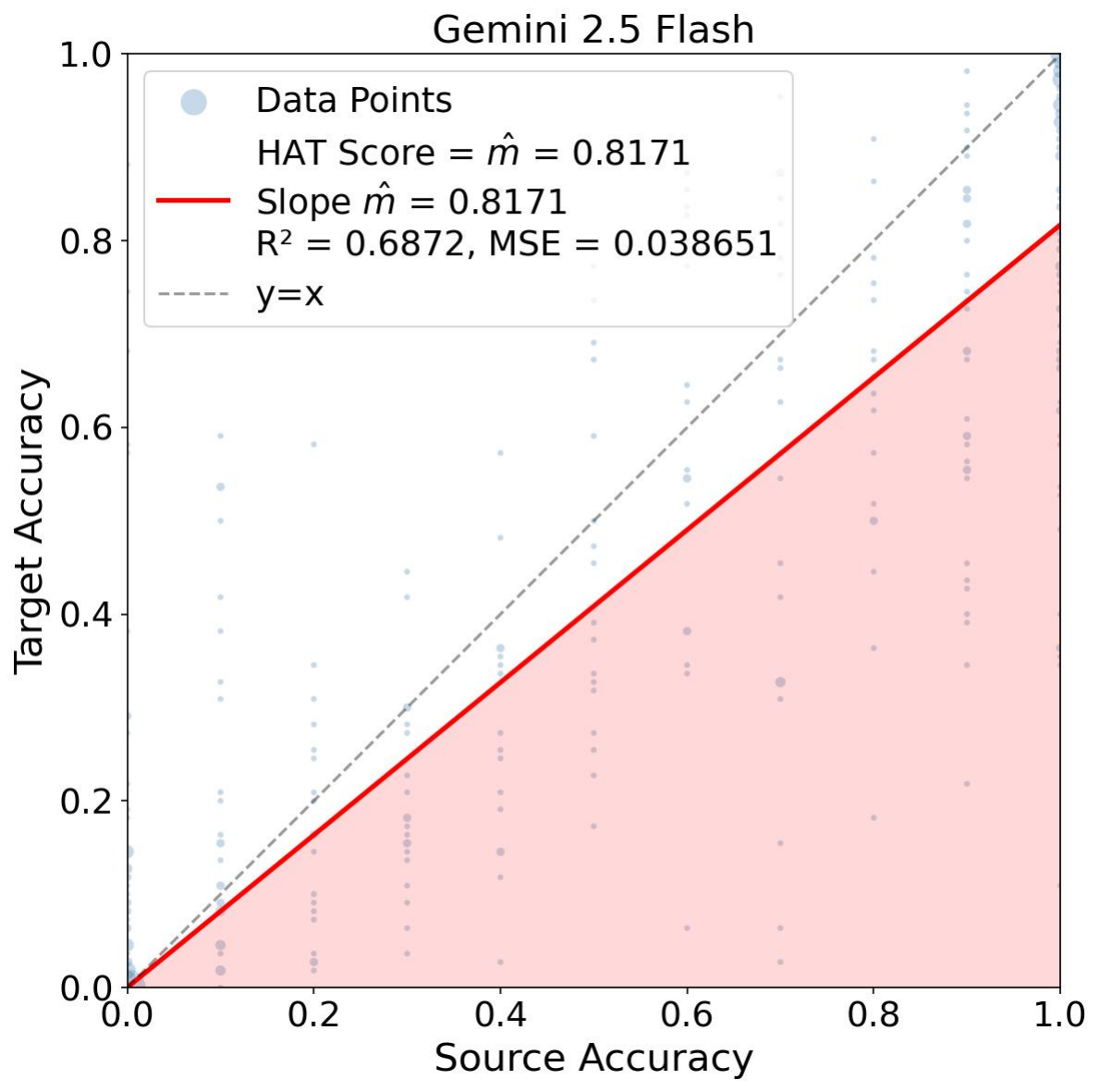} & \includegraphics[width=\linewidth]{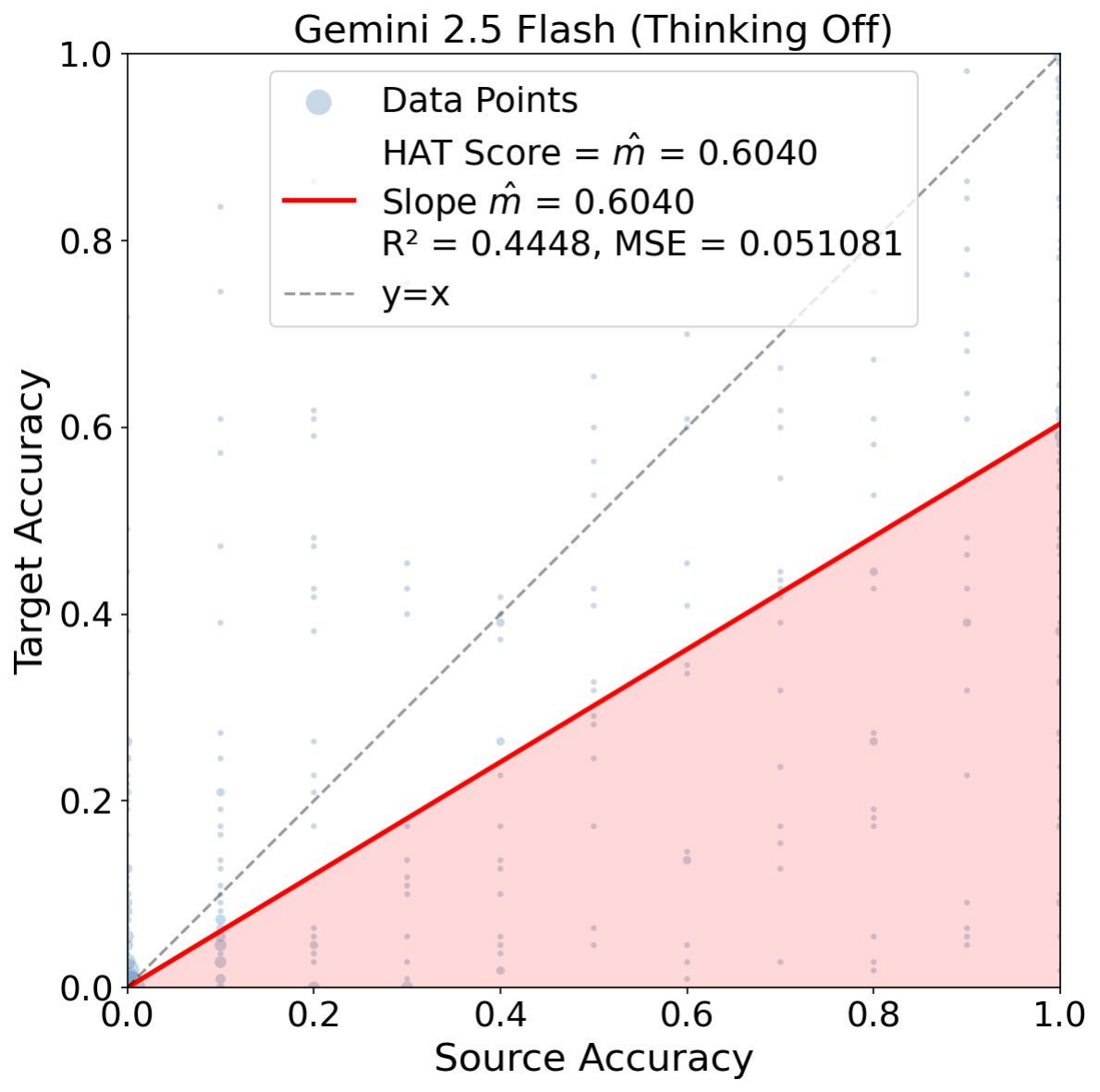} & \includegraphics[width=\linewidth]{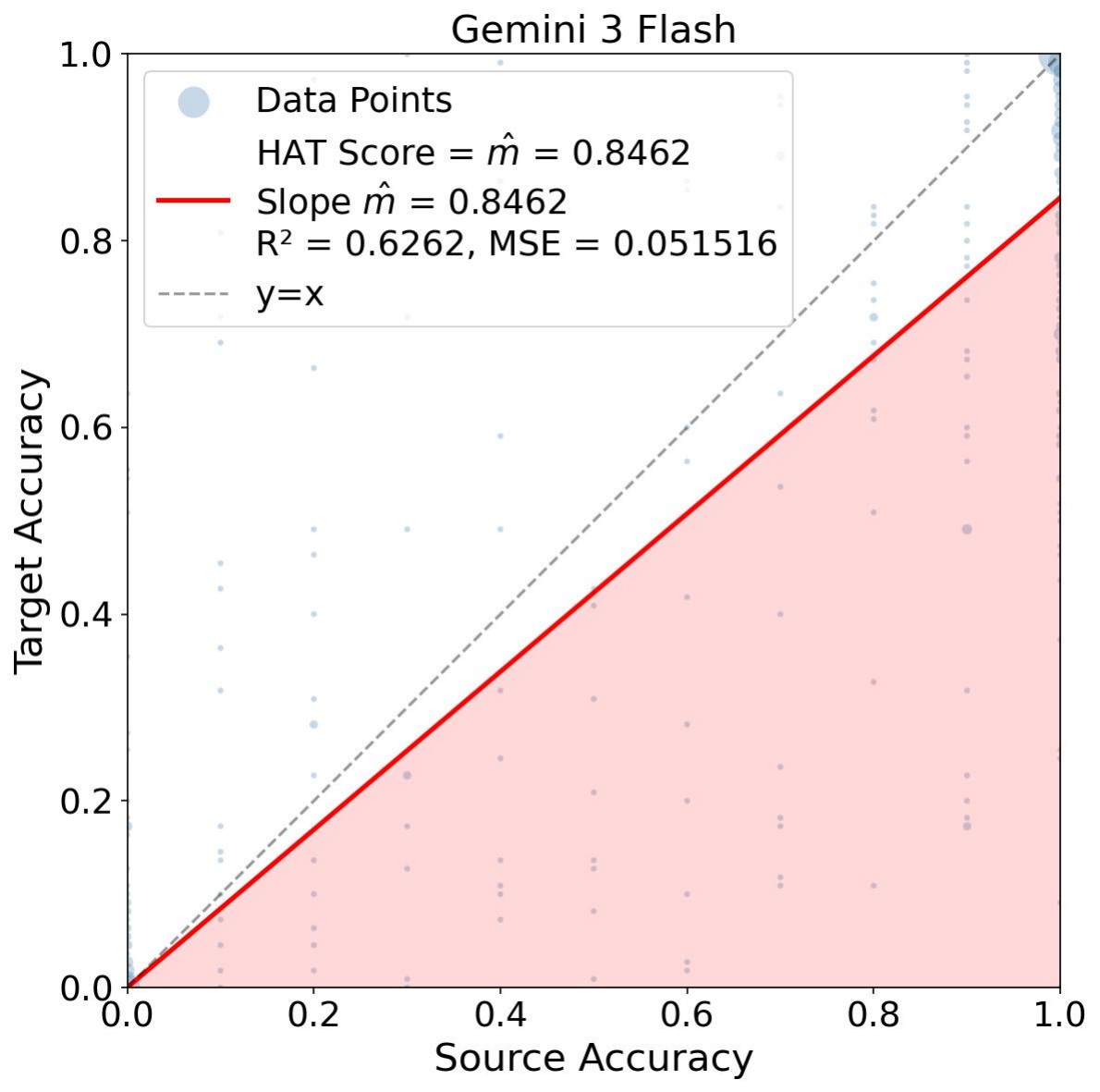} & \includegraphics[width=\linewidth]{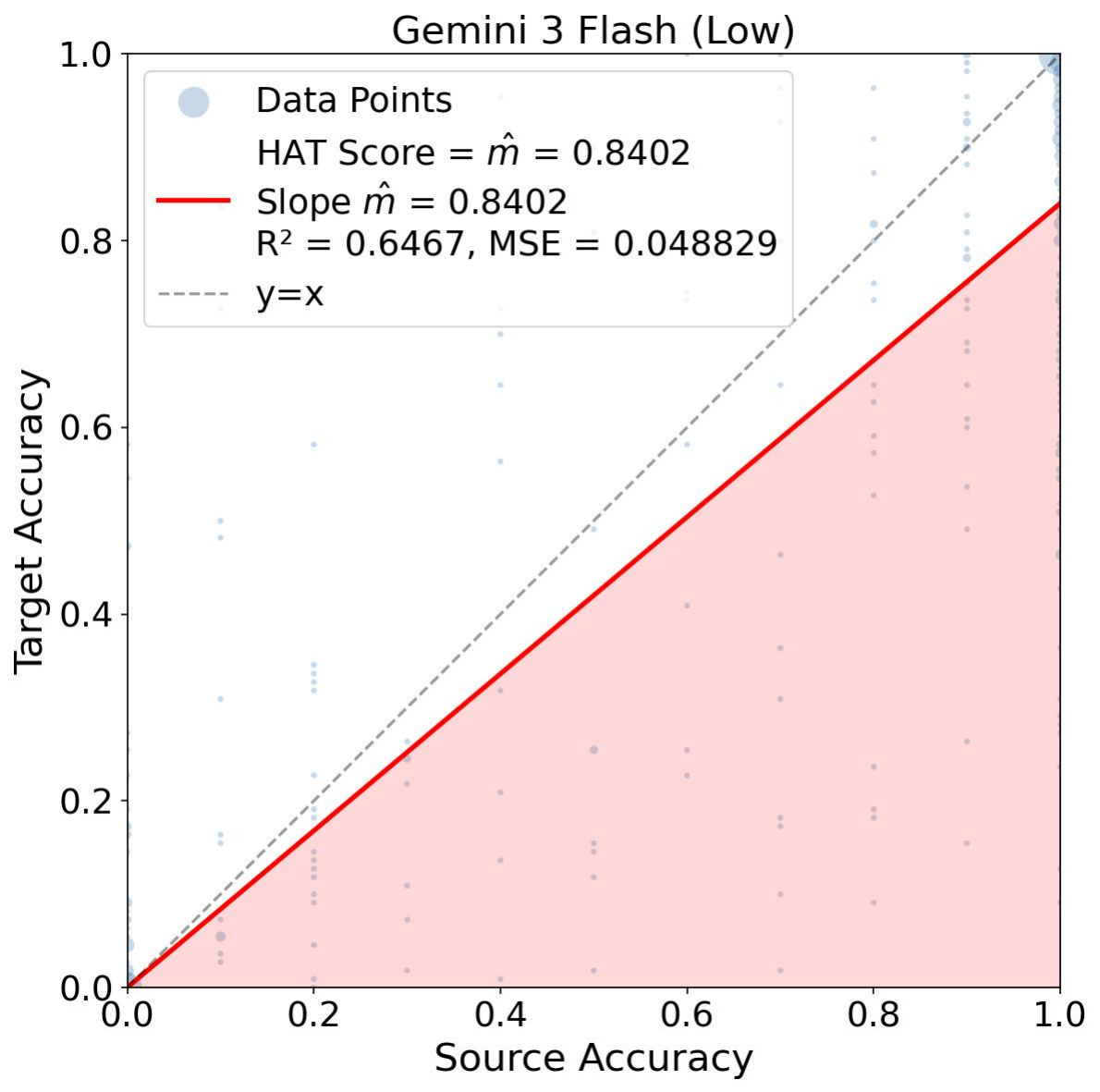} \\
\includegraphics[width=\linewidth]{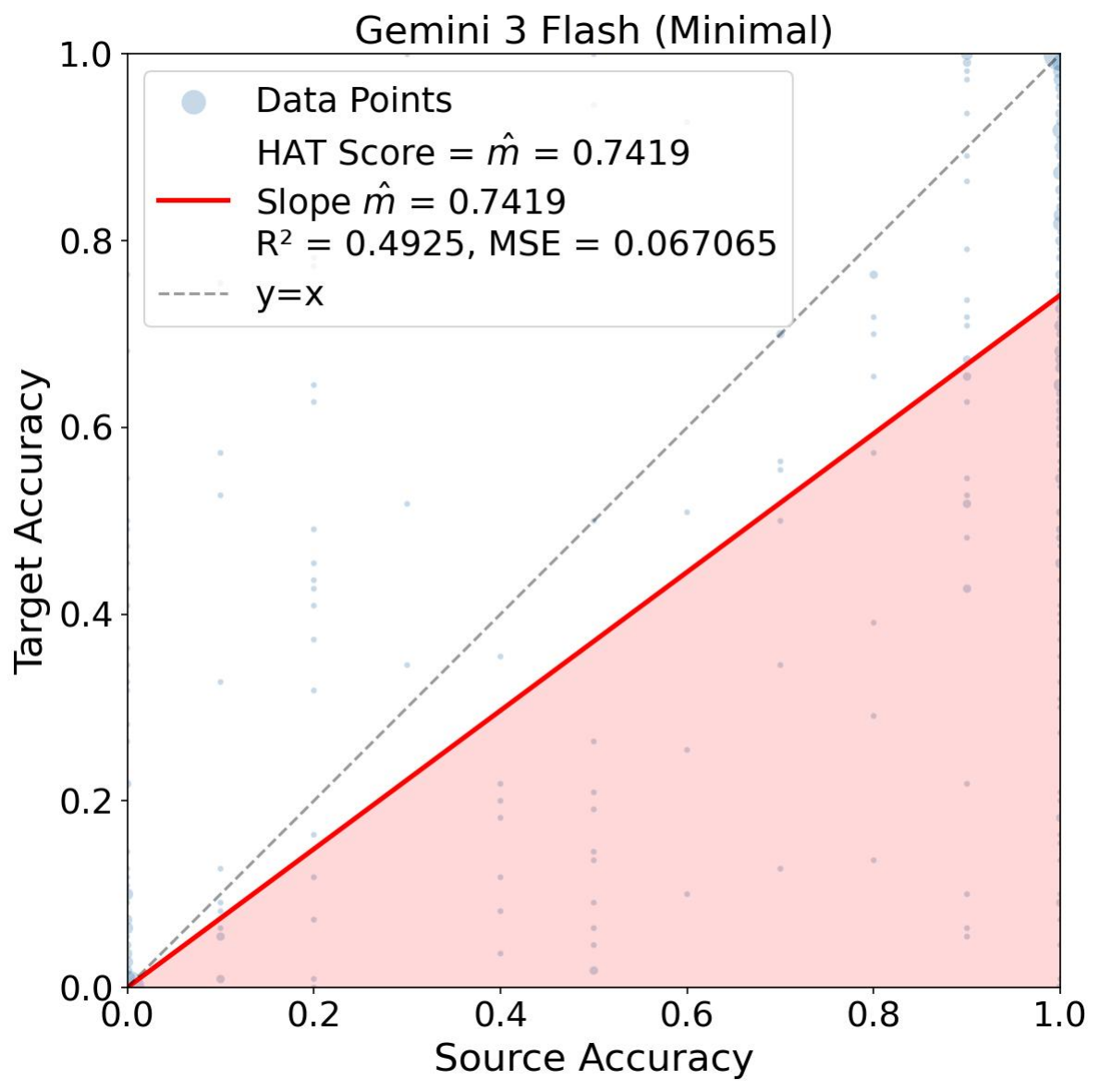} & \includegraphics[width=\linewidth]{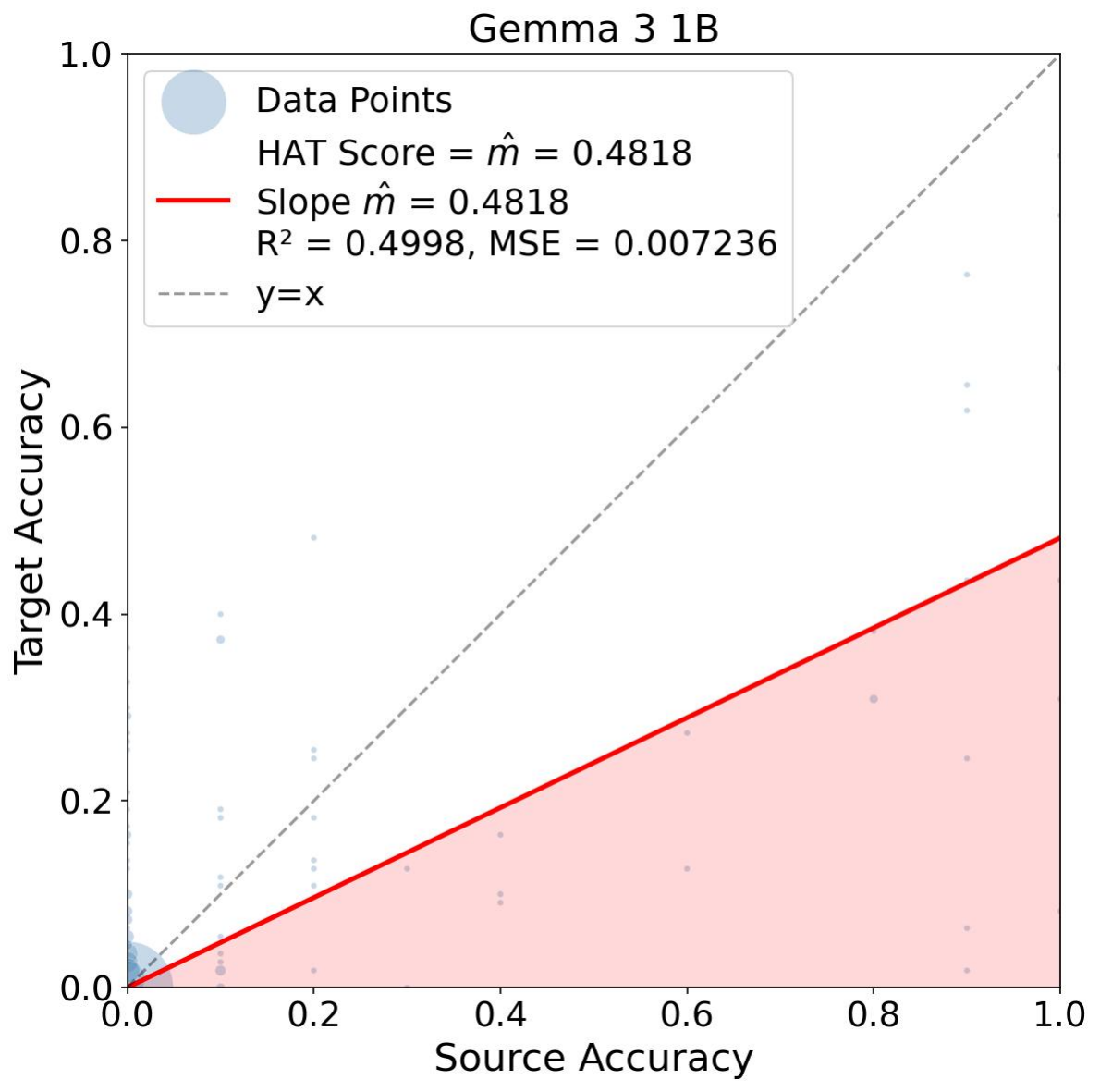} & \includegraphics[width=\linewidth]{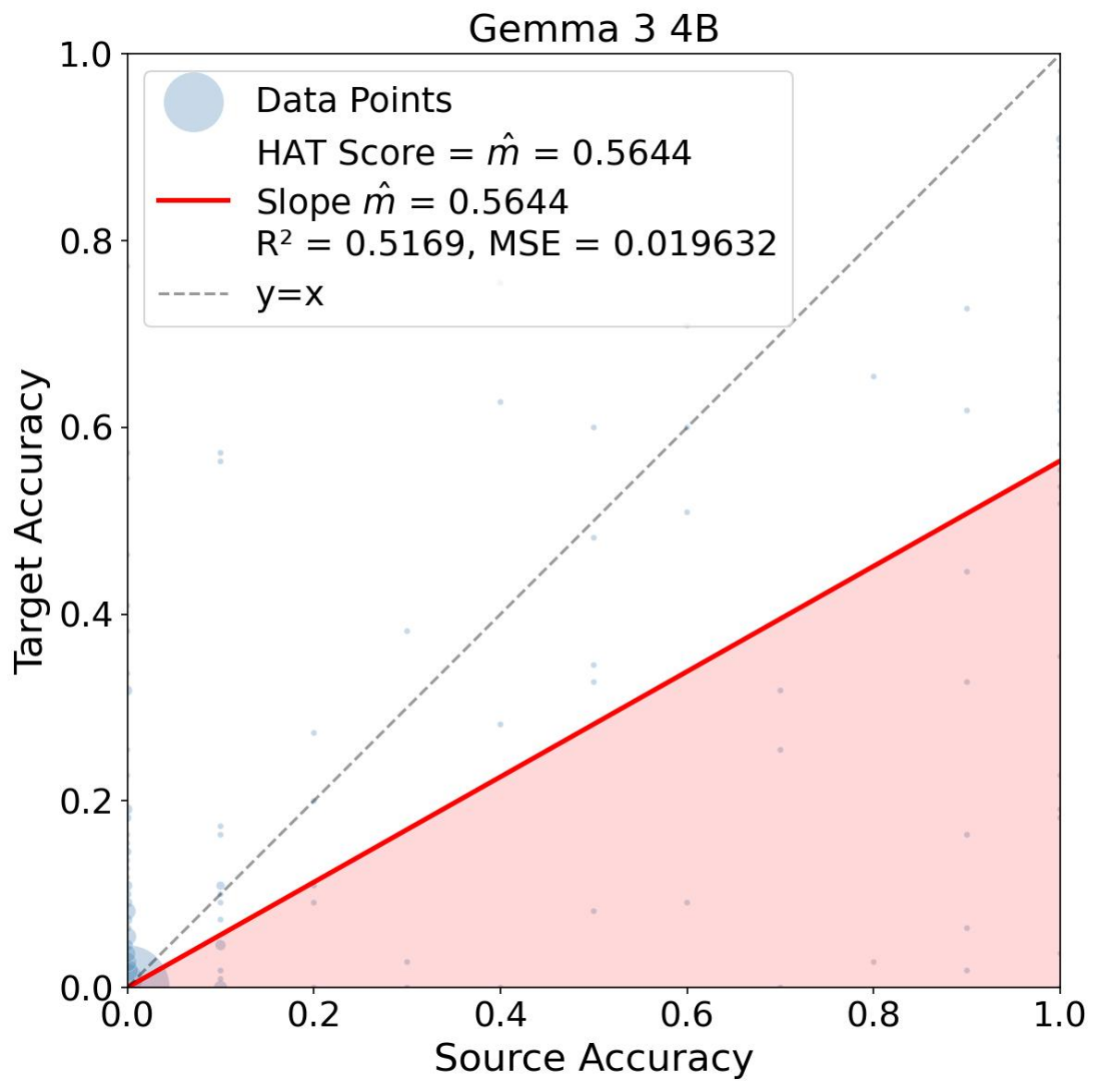} & \includegraphics[width=\linewidth]{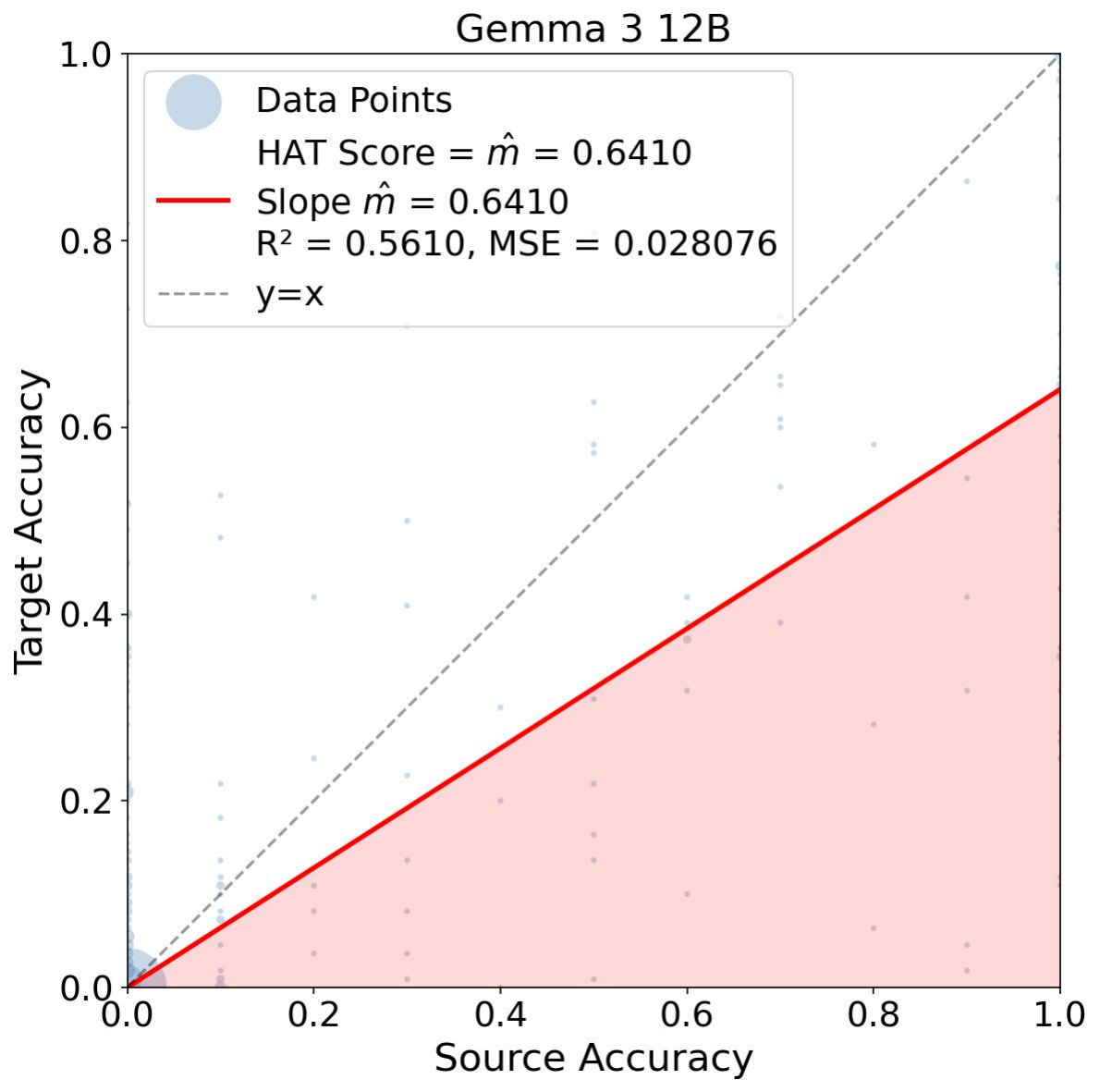} \\
\includegraphics[width=\linewidth]{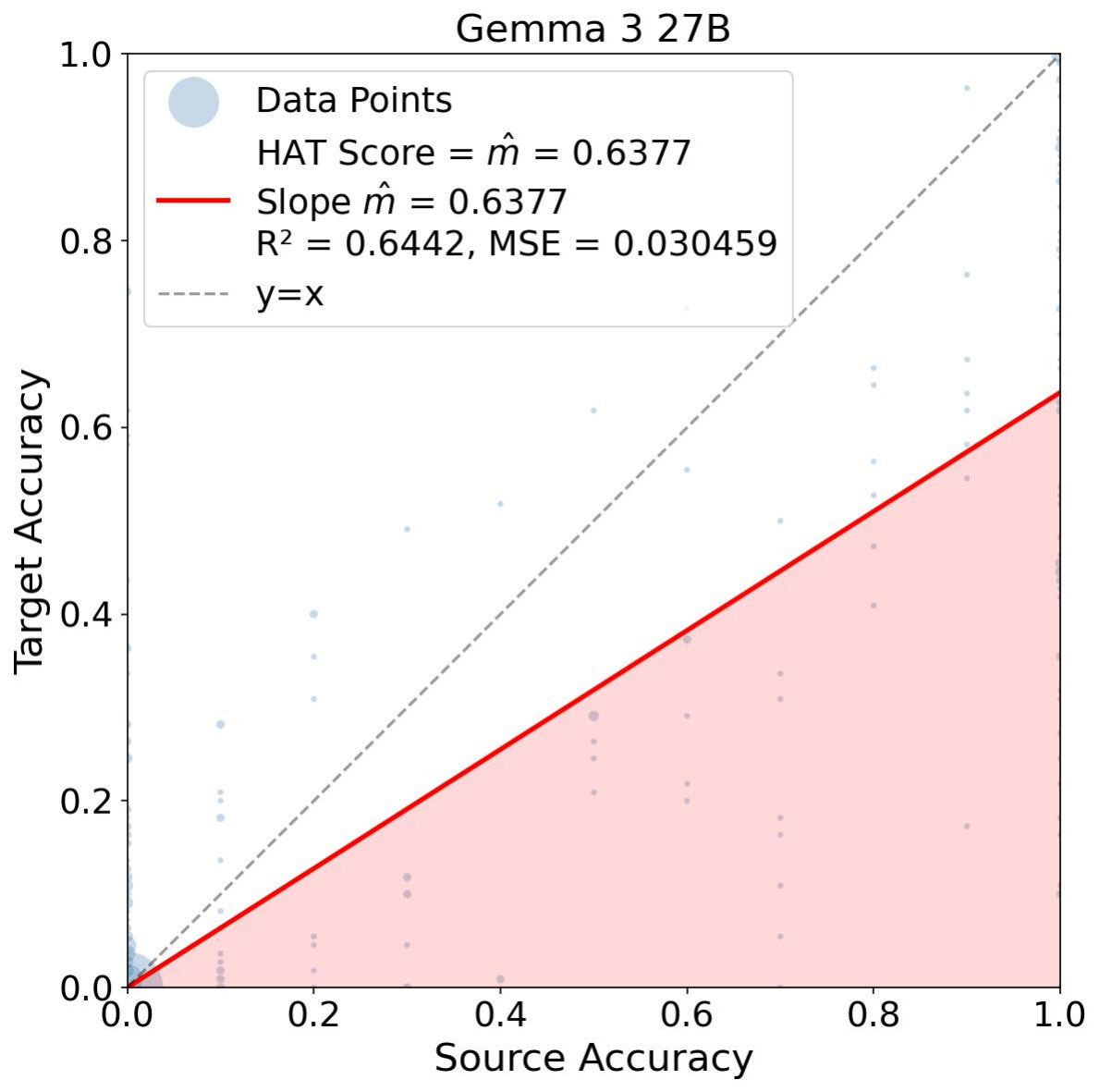} & \includegraphics[width=\linewidth]{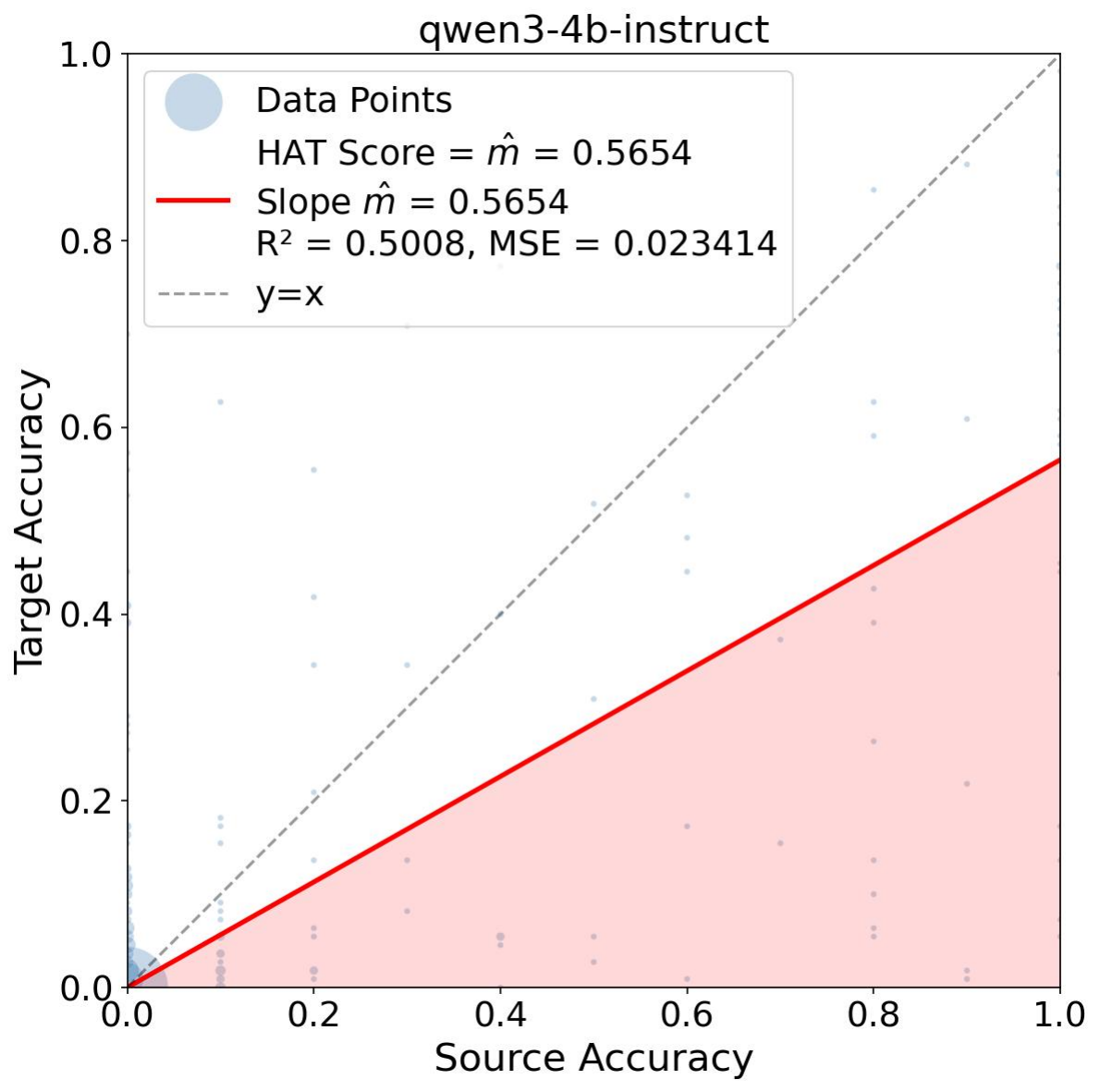} & \includegraphics[width=\linewidth]{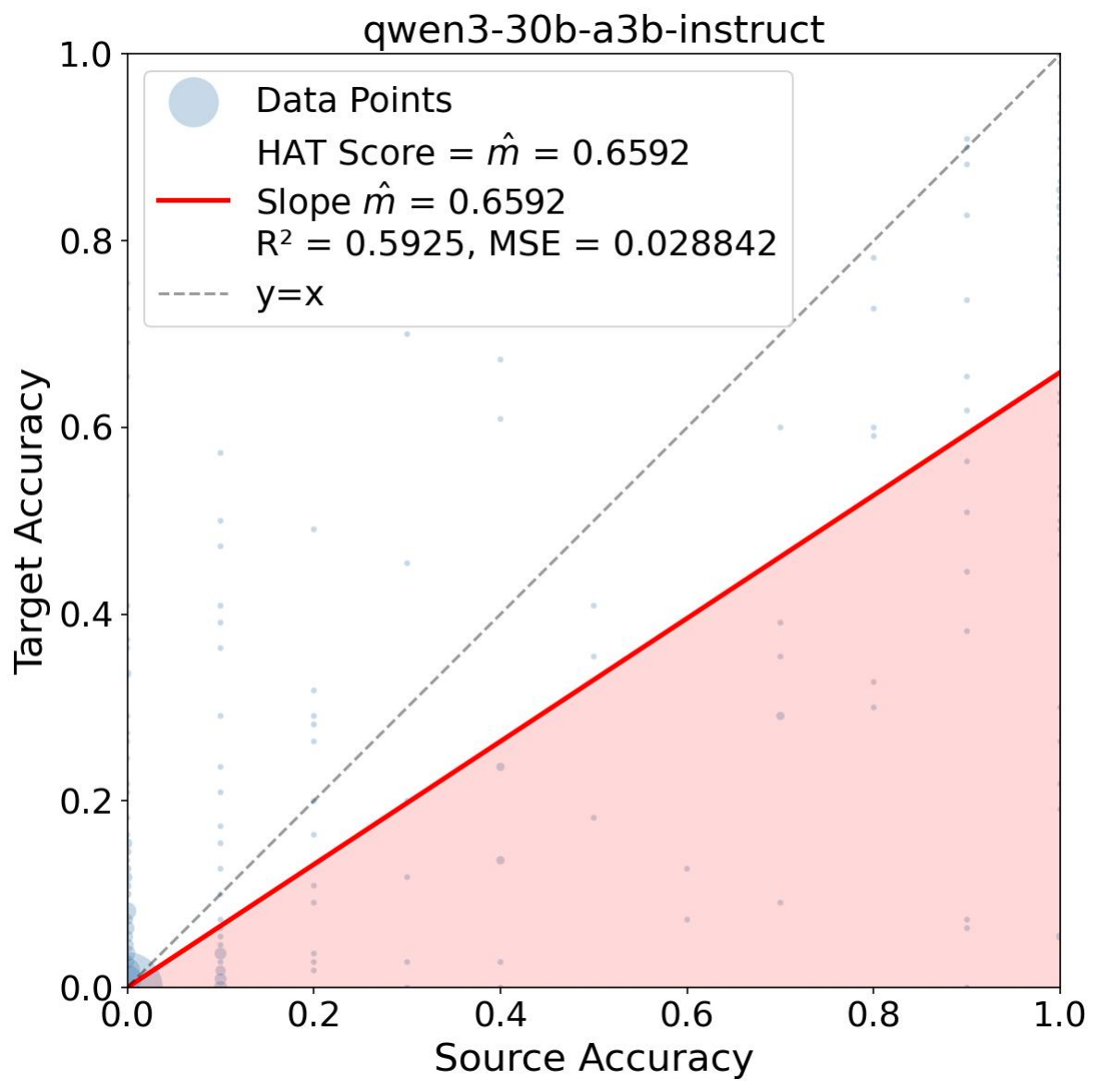} & \includegraphics[width=\linewidth]{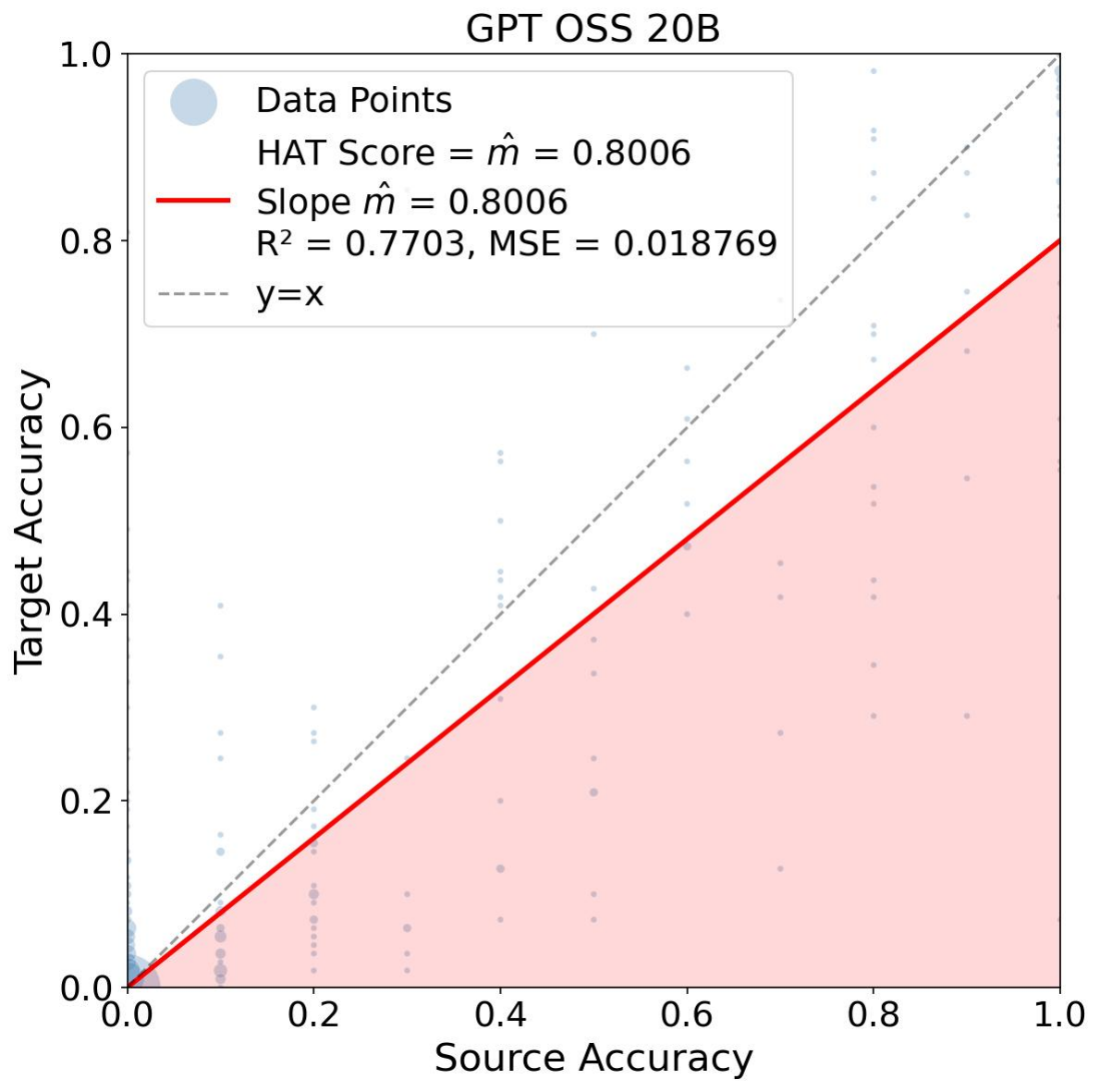} \\
\includegraphics[width=\linewidth]{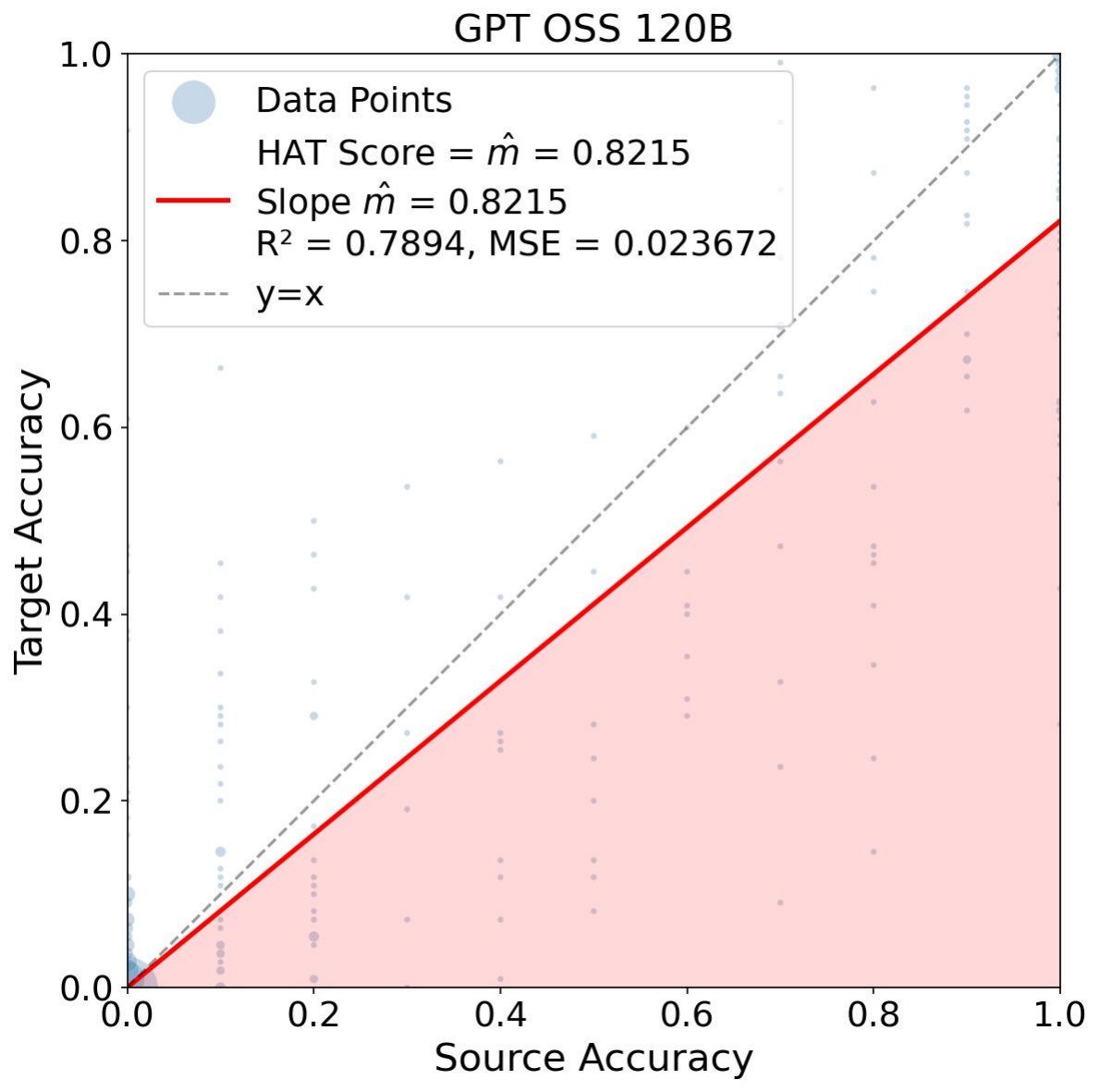} & \includegraphics[width=\linewidth]{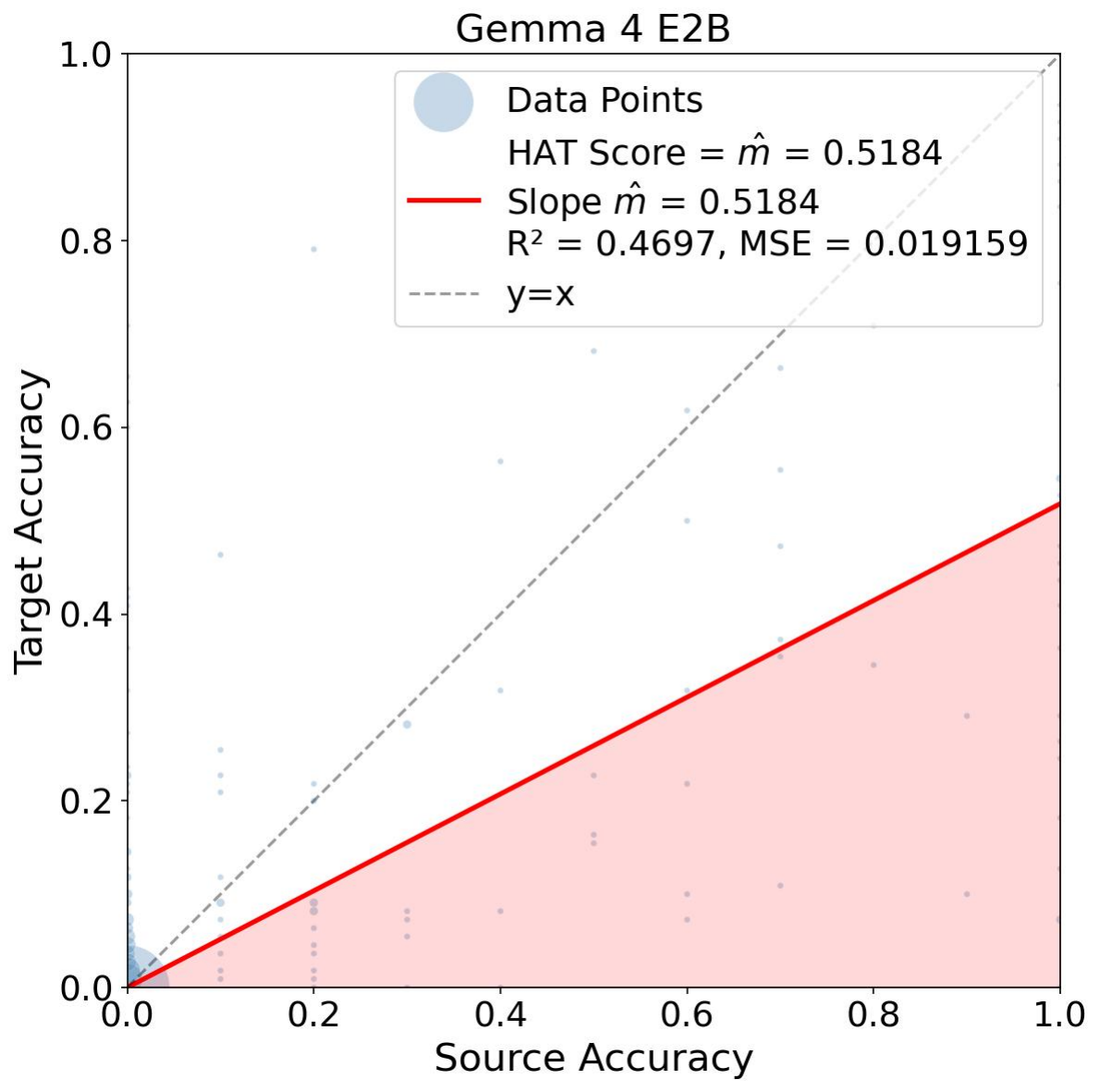} & \includegraphics[width=\linewidth]{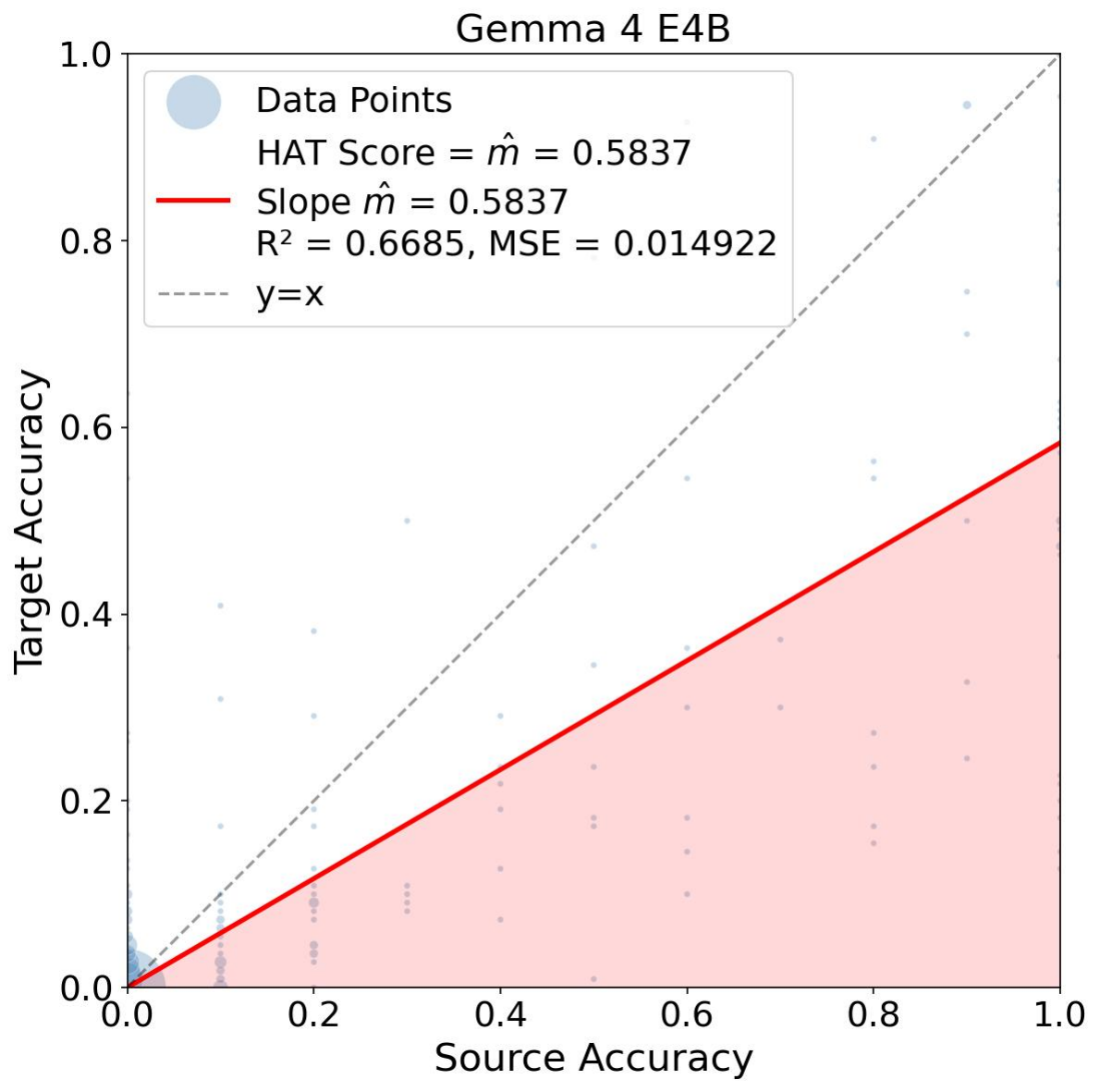} & \includegraphics[width=\linewidth]{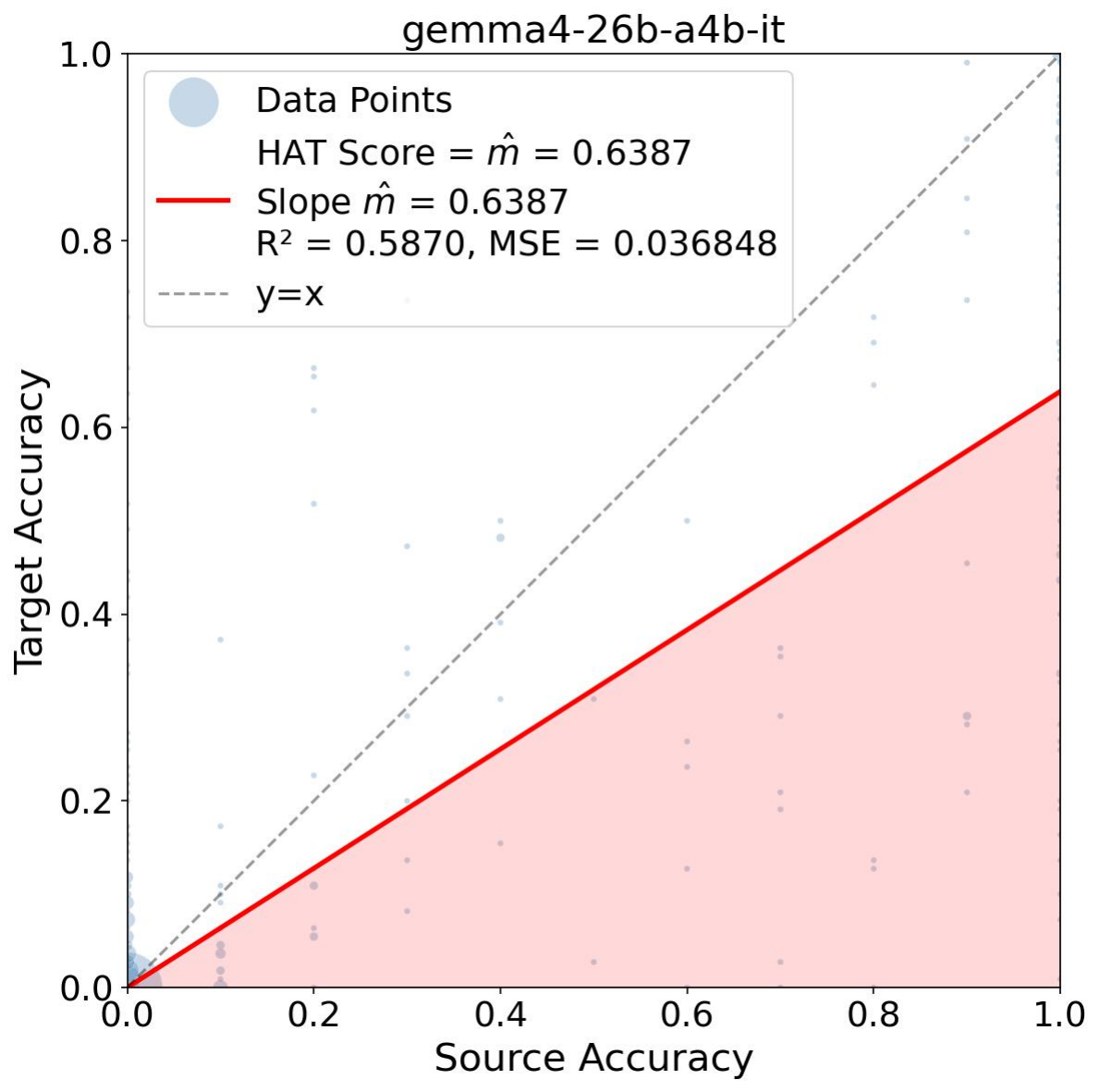} \\
\includegraphics[width=\linewidth]{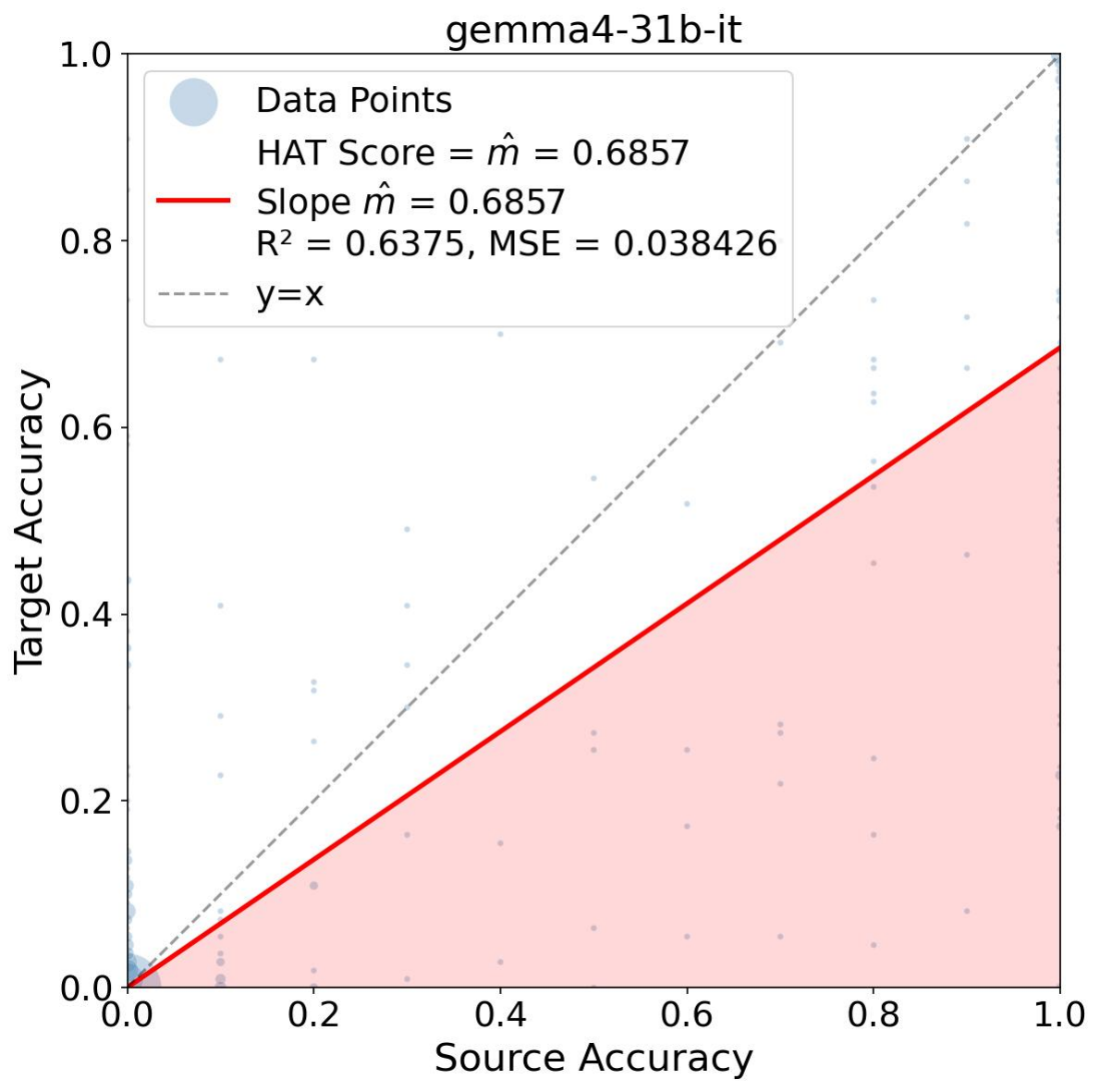} & \includegraphics[width=\linewidth]{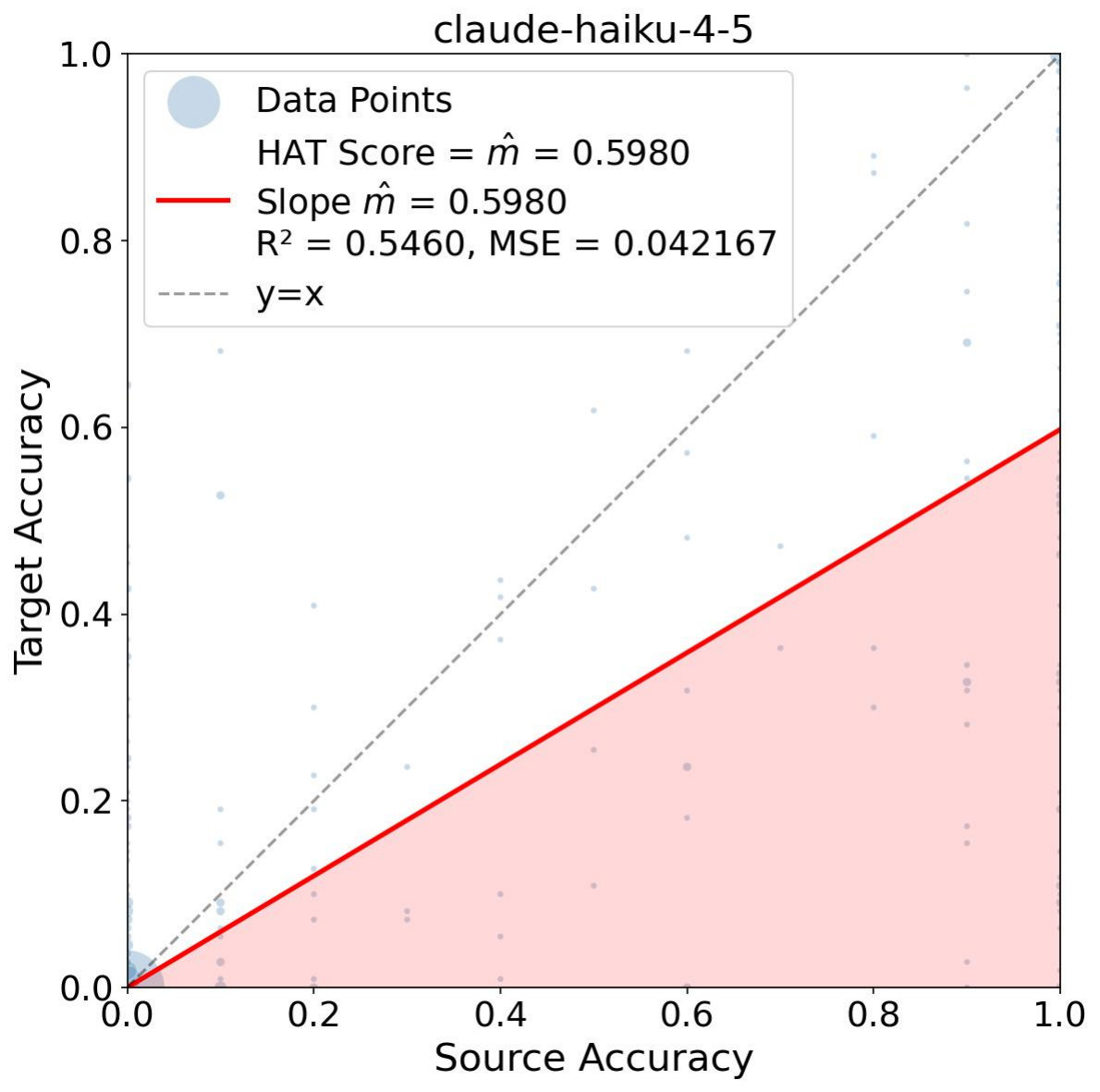} & \includegraphics[width=\linewidth]{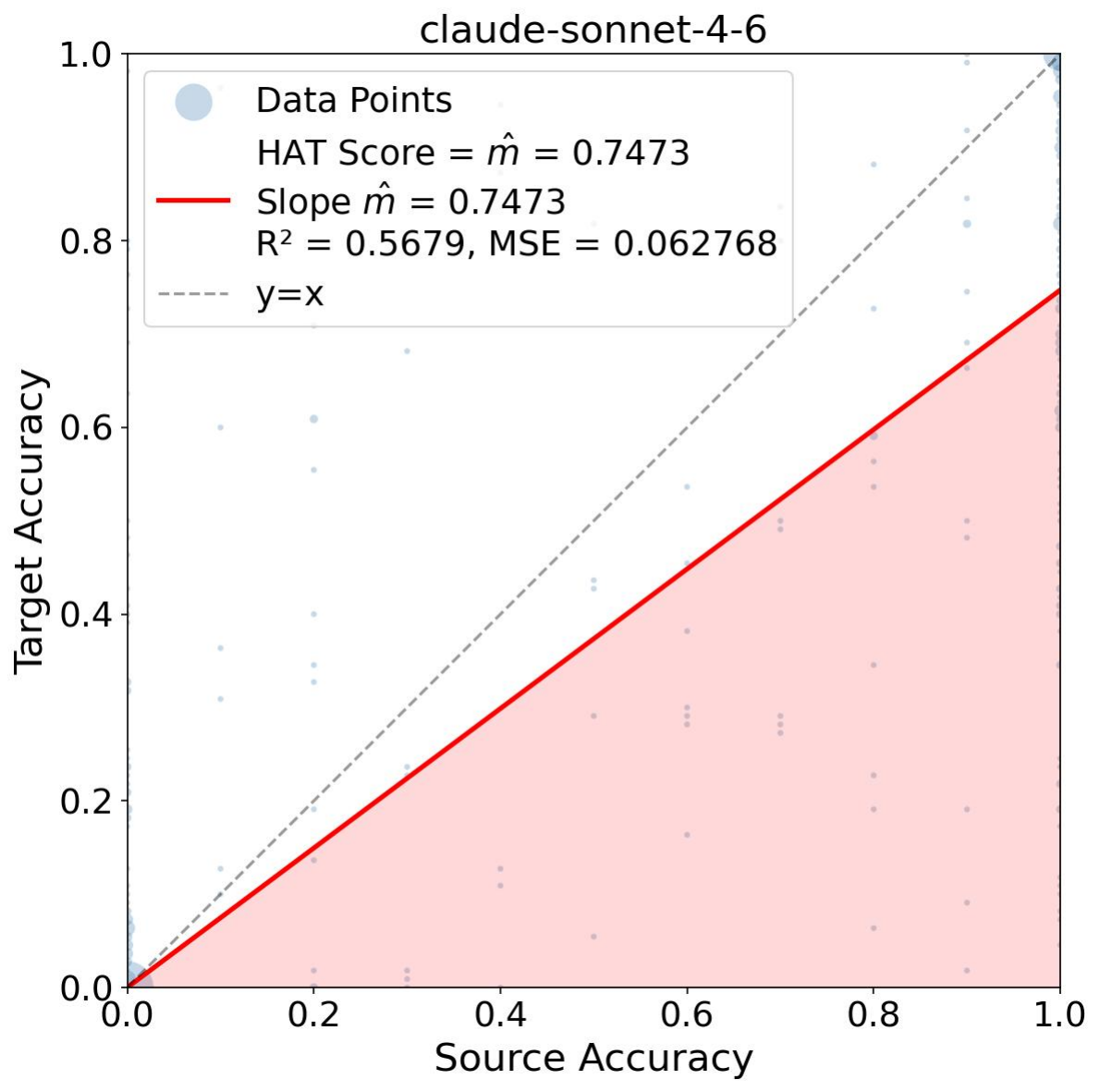} & \includegraphics[width=\linewidth]{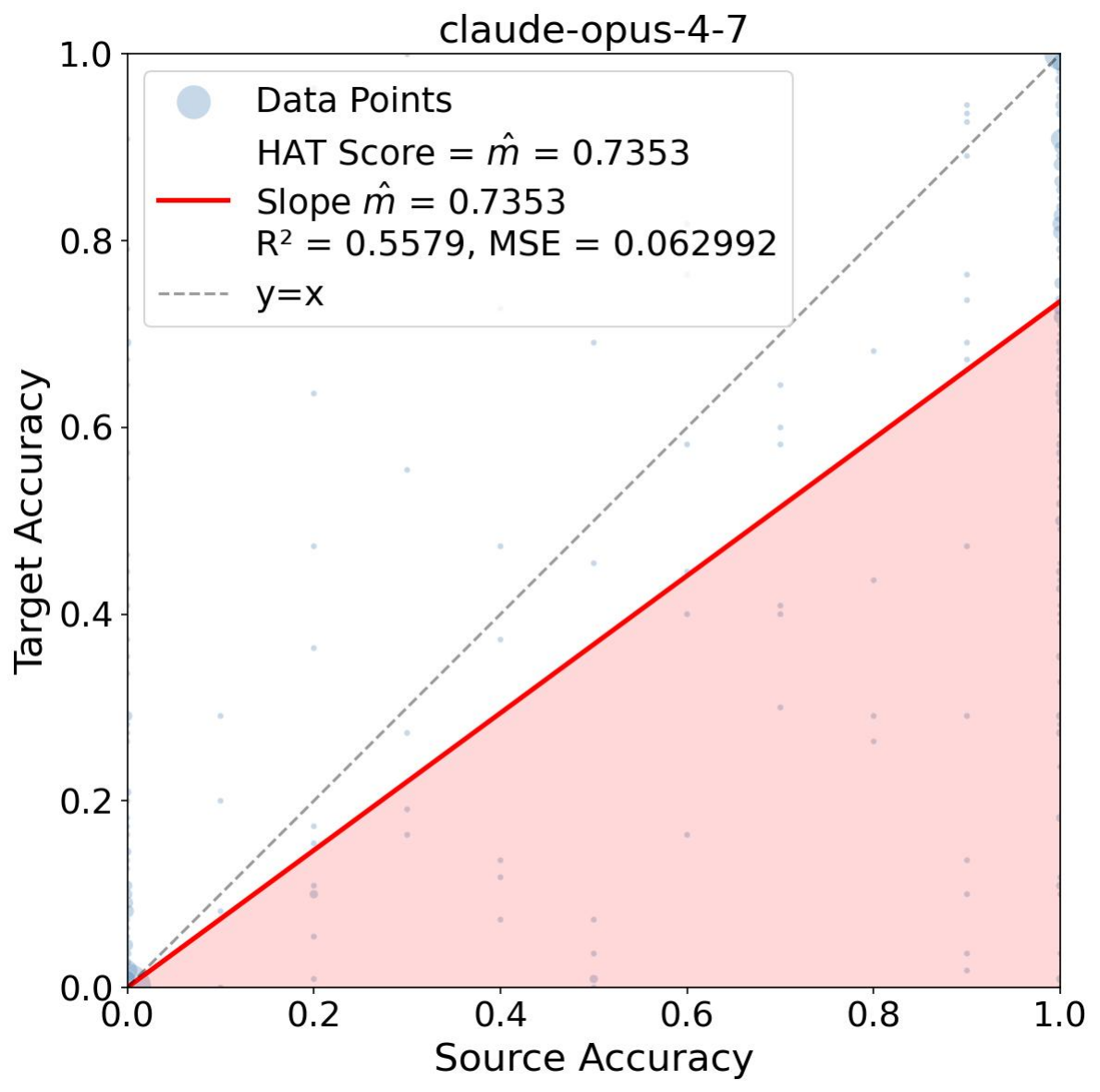} \\
\caption{HAT profiles for ECLeKTic.}
\end{longtable}
\twocolumn
\clearpage
\setlength{\tabcolsep}{1pt} 
\onecolumn
\subsubsection{MGSMv2 HAT Plots}
Detailed evaluation curves and HAT score transfer profiles for the MGSMv2 dataset are provided below.
\begin{longtable}{>{\centering\arraybackslash}p{0.24\textwidth} >{\centering\arraybackslash}p{0.24\textwidth} >{\centering\arraybackslash}p{0.24\textwidth} >{\centering\arraybackslash}p{0.24\textwidth}}
\includegraphics[width=\linewidth]{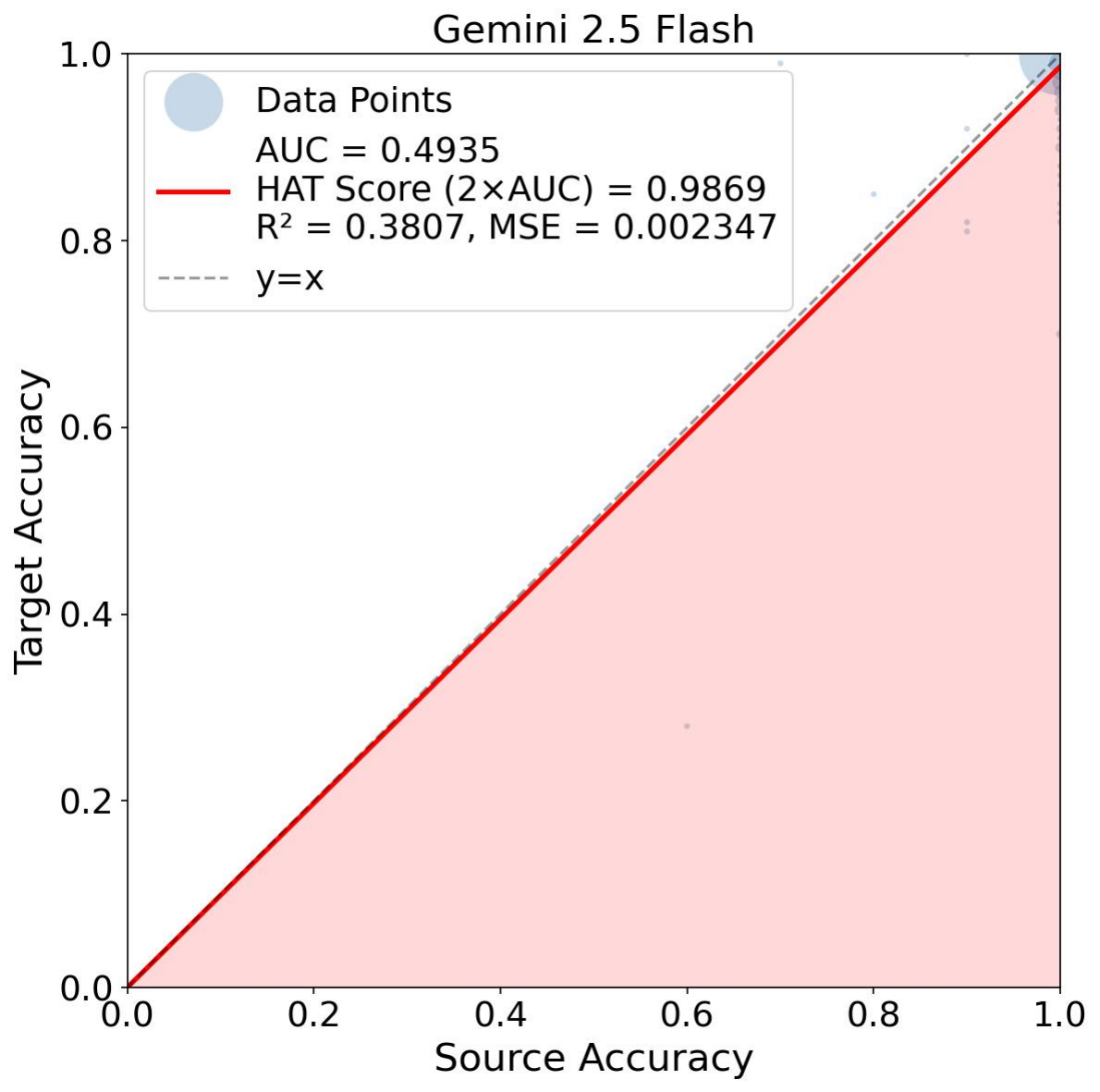} & \includegraphics[width=\linewidth]{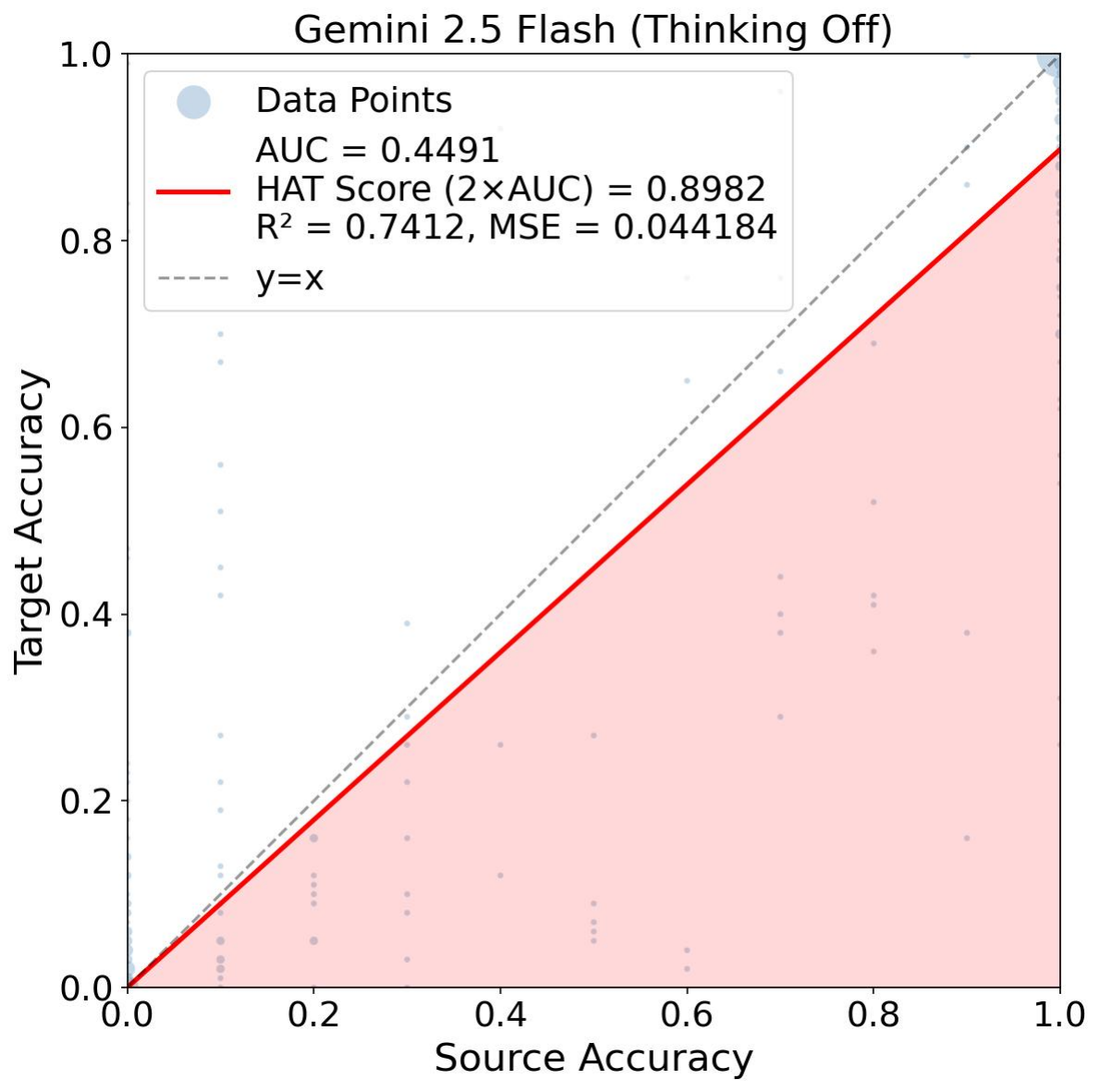} & \includegraphics[width=\linewidth]{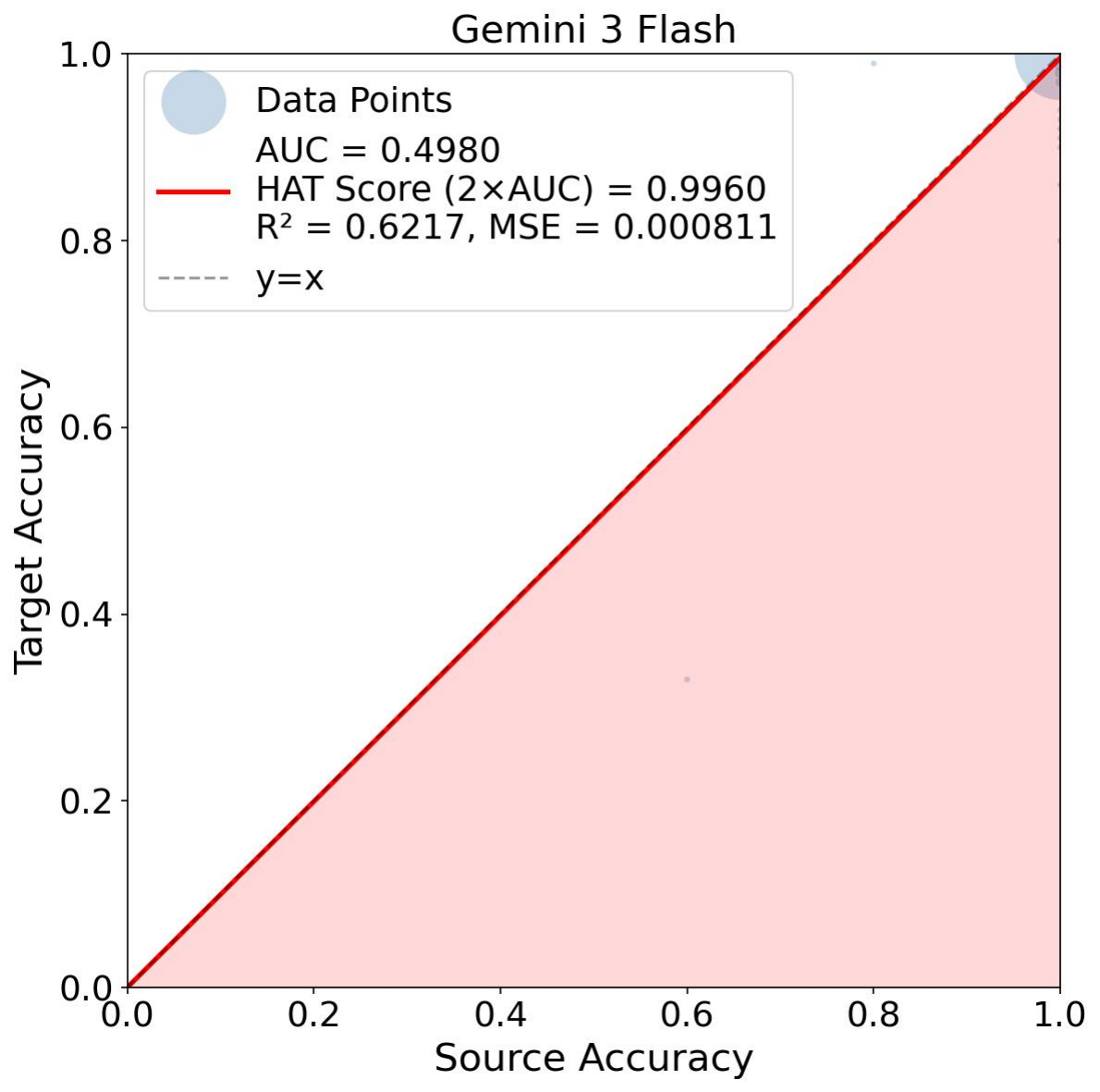} & \includegraphics[width=\linewidth]{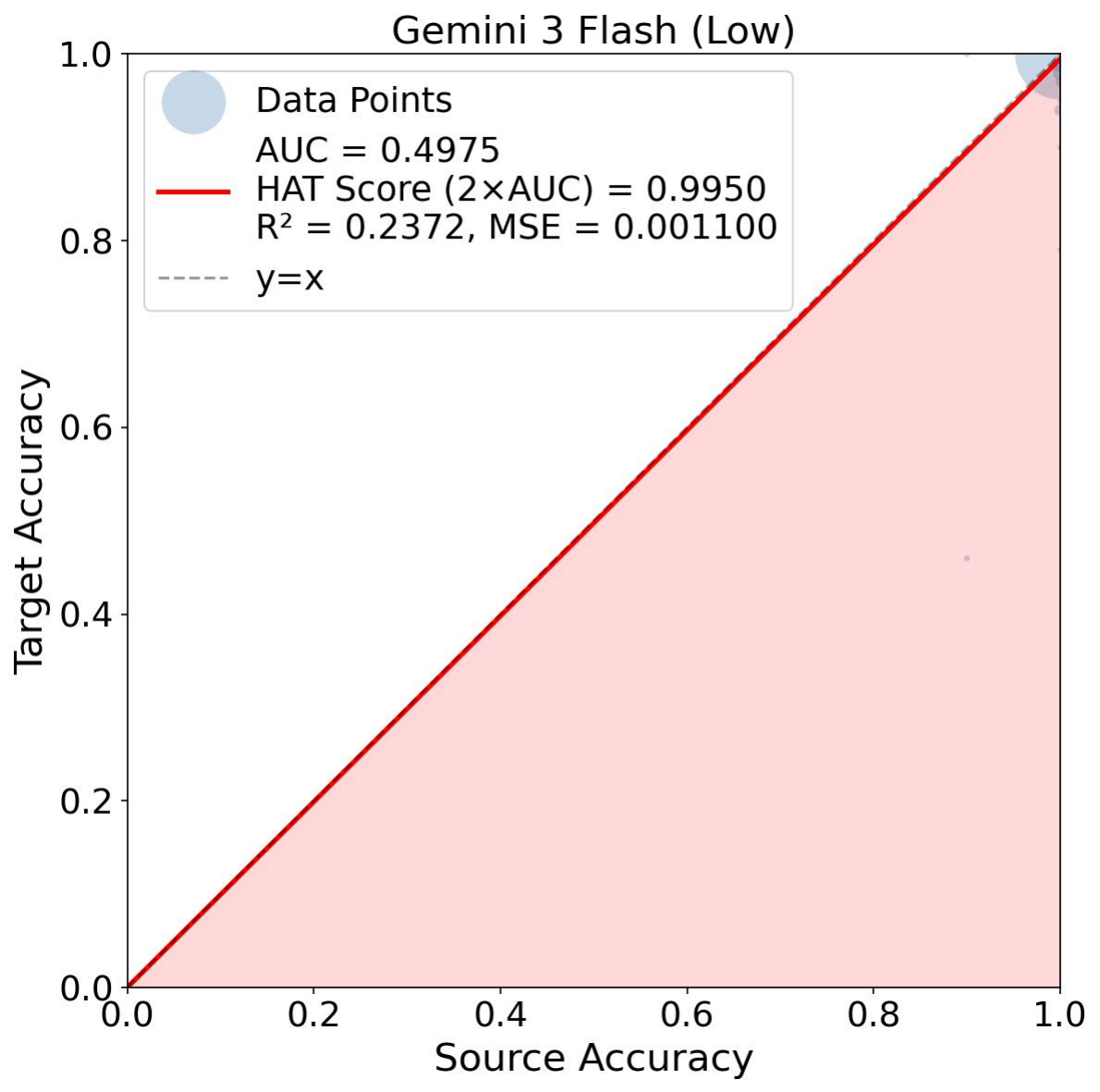} \\
\includegraphics[width=\linewidth]{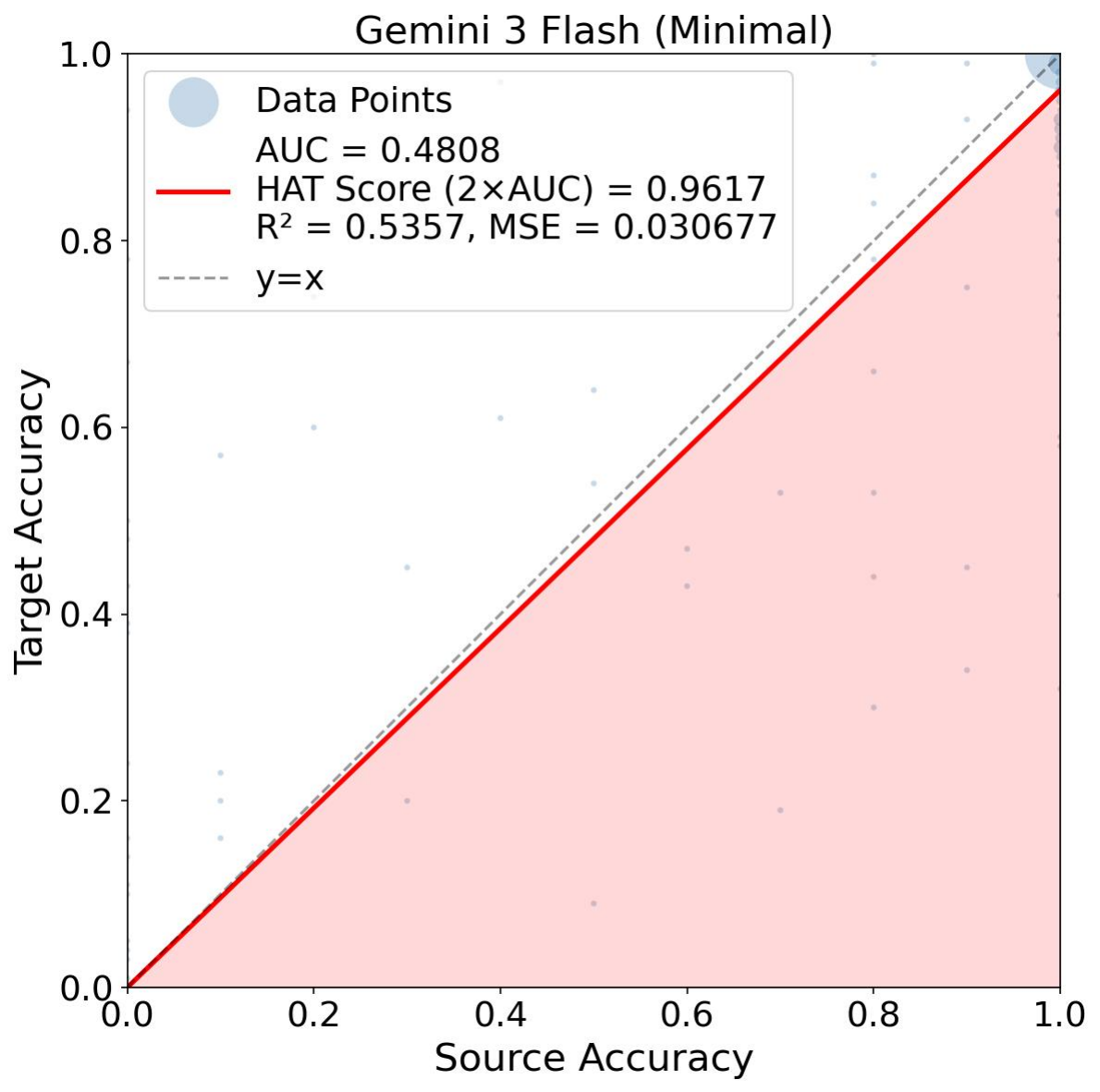} & \includegraphics[width=\linewidth]{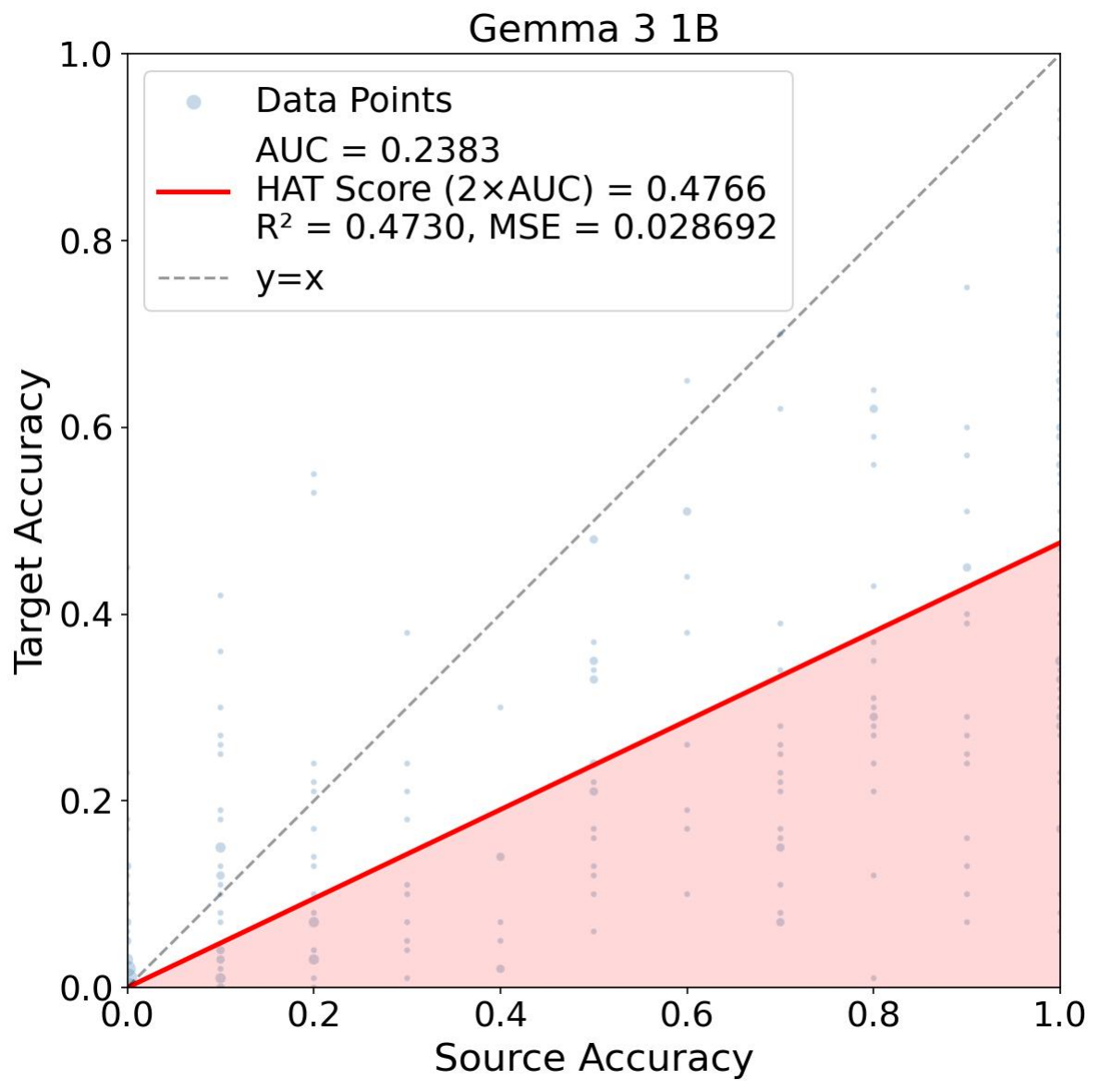} & \includegraphics[width=\linewidth]{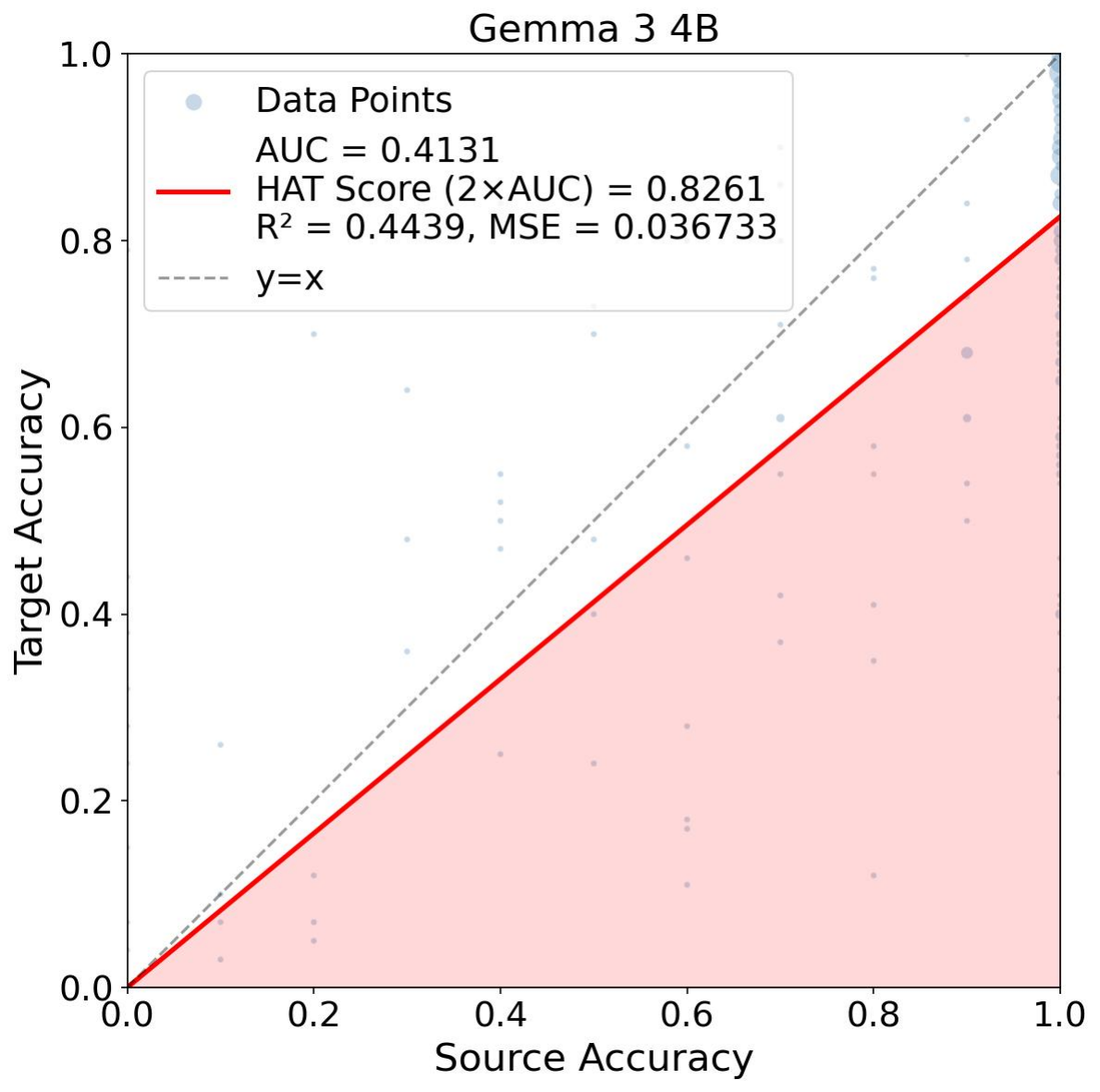} & \includegraphics[width=\linewidth]{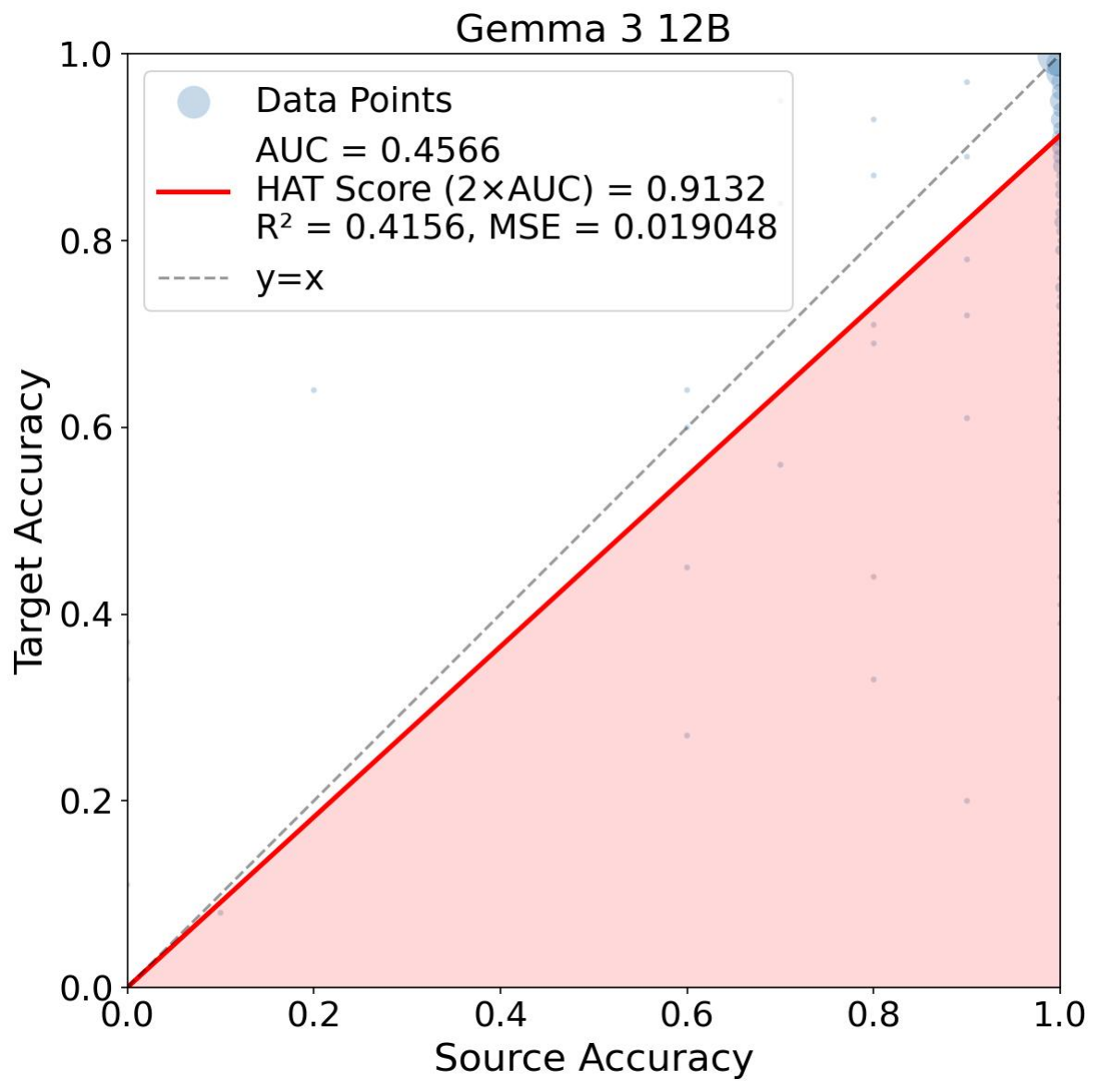} \\
\includegraphics[width=\linewidth]{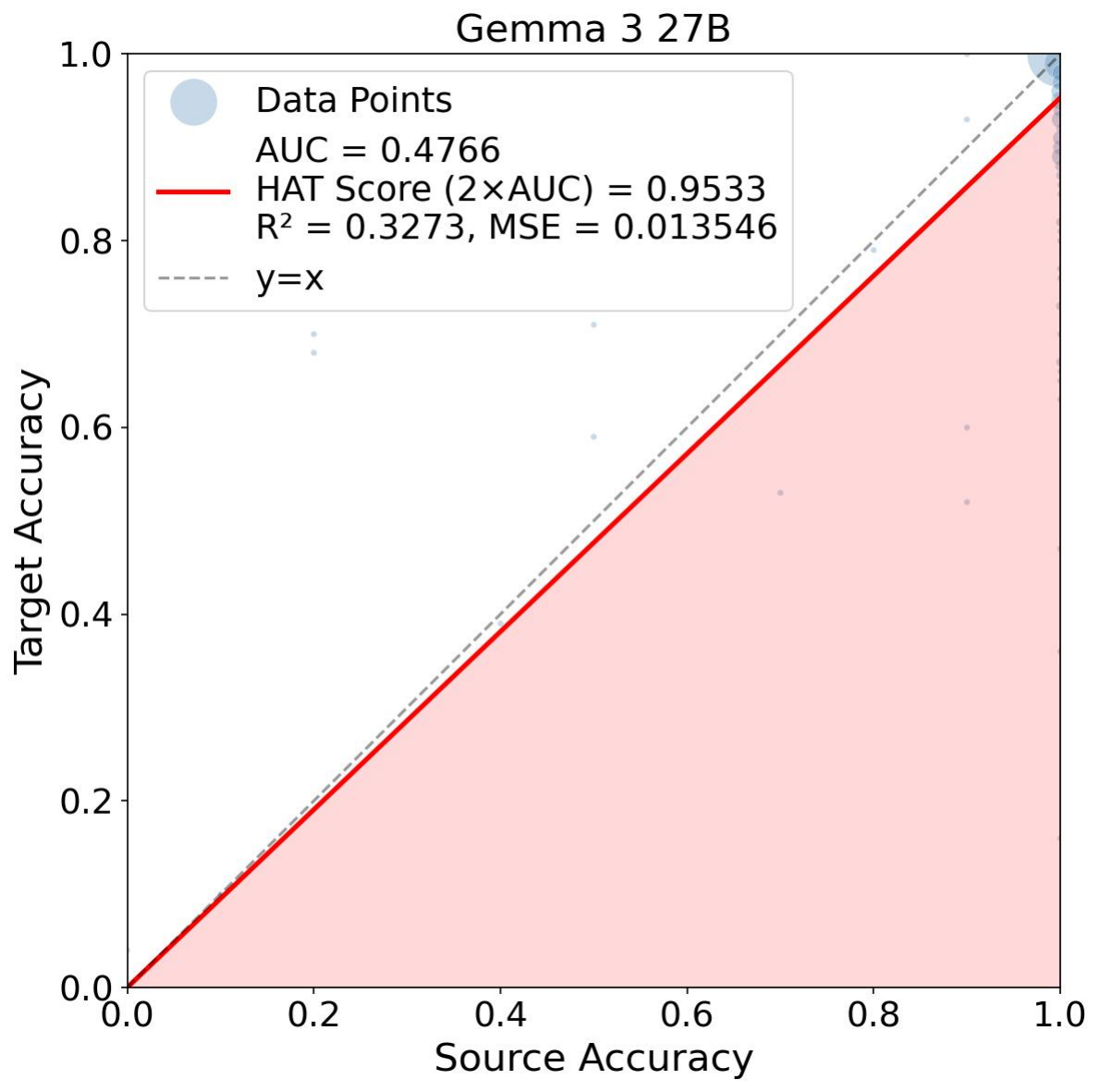} & \includegraphics[width=\linewidth]{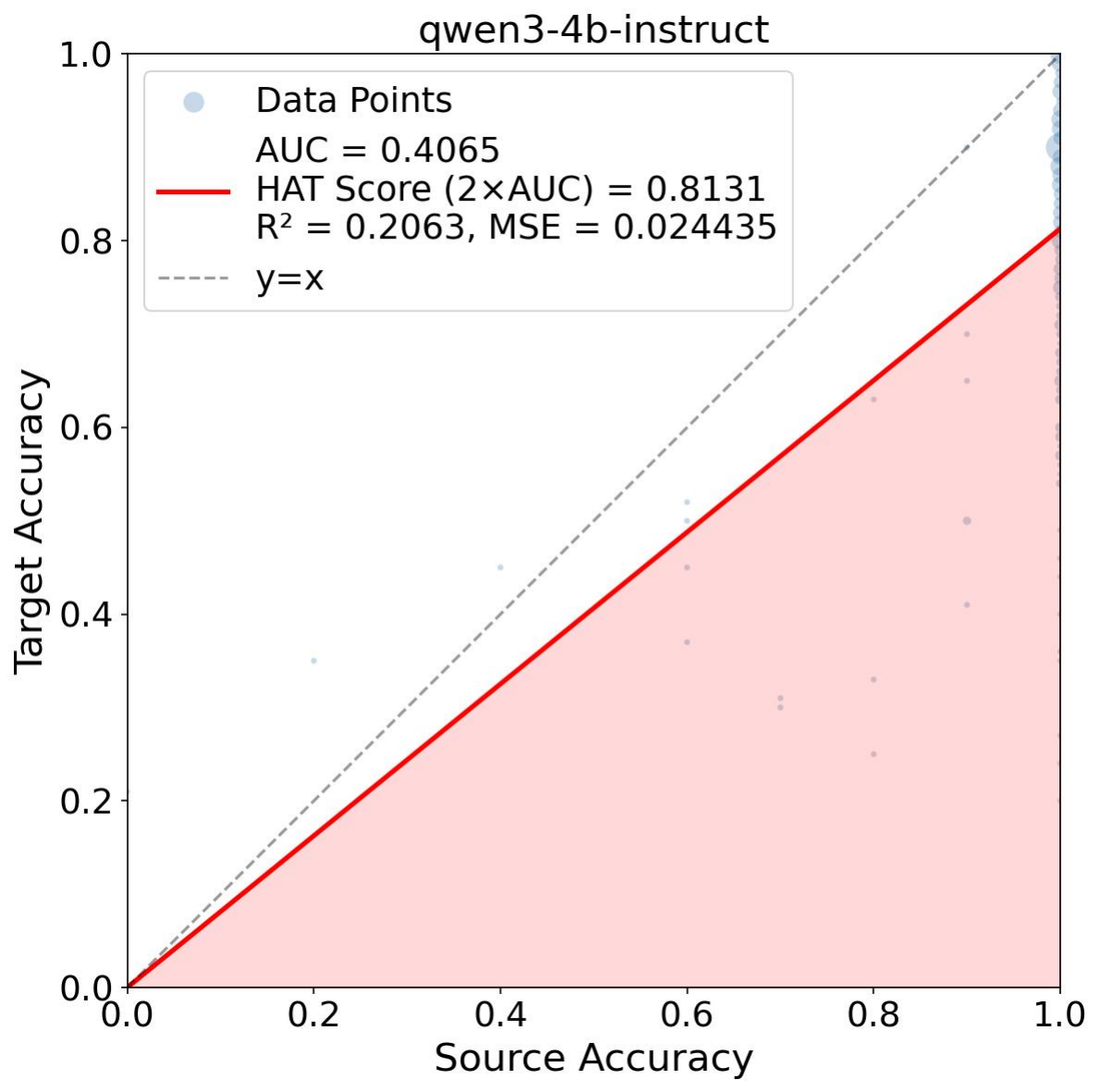} & \includegraphics[width=\linewidth]{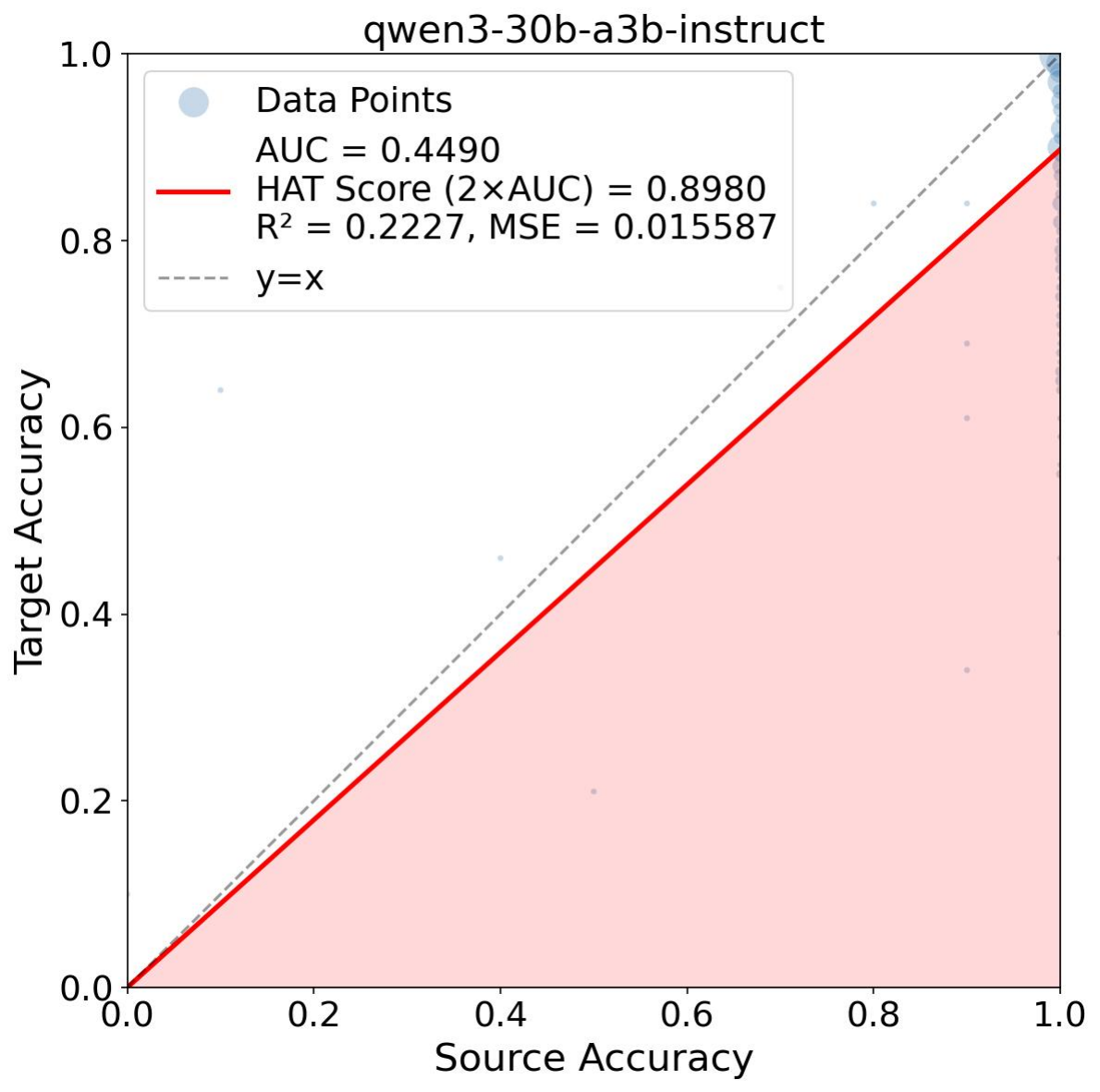} & \includegraphics[width=\linewidth]{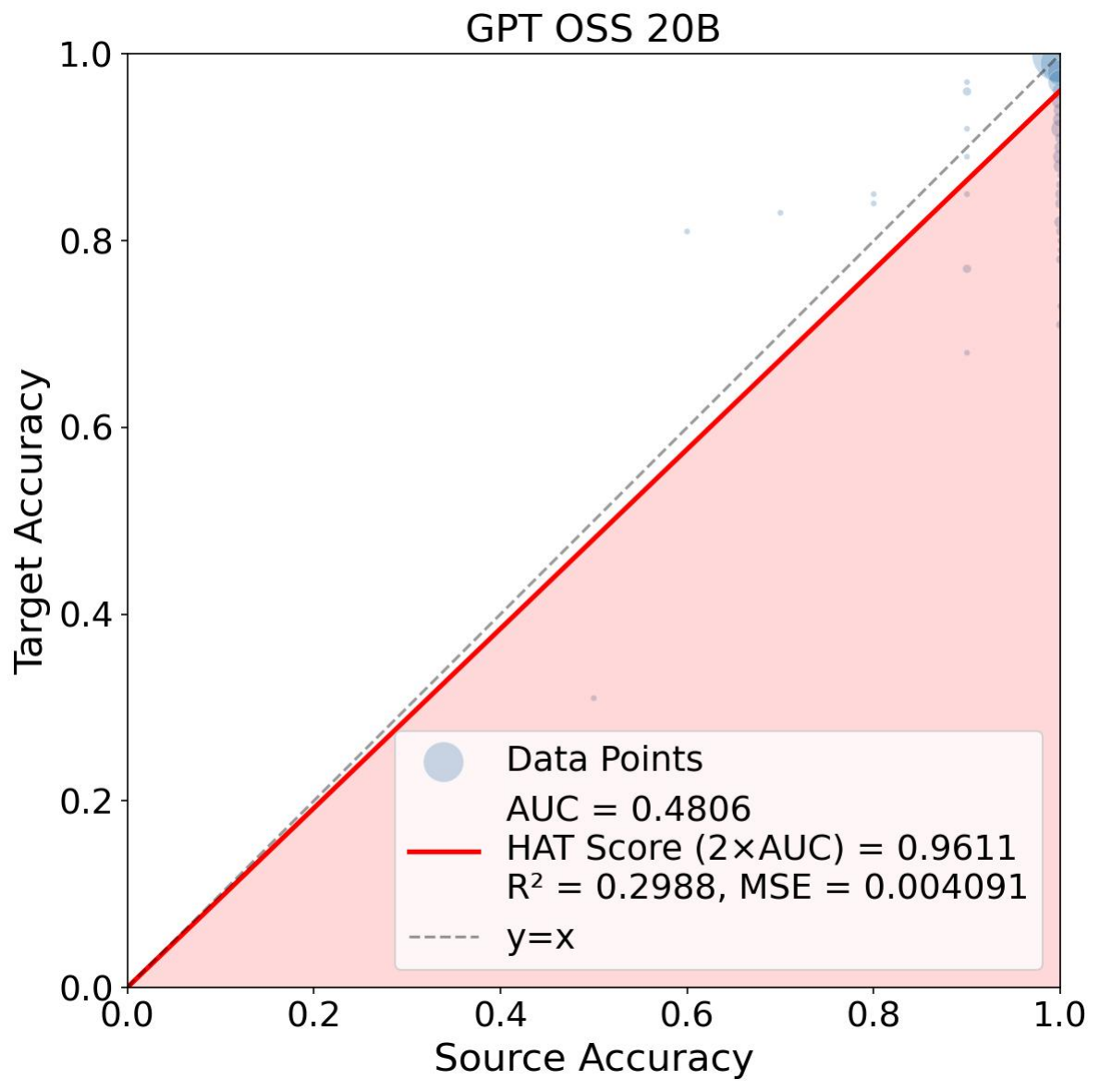} \\
\includegraphics[width=\linewidth]{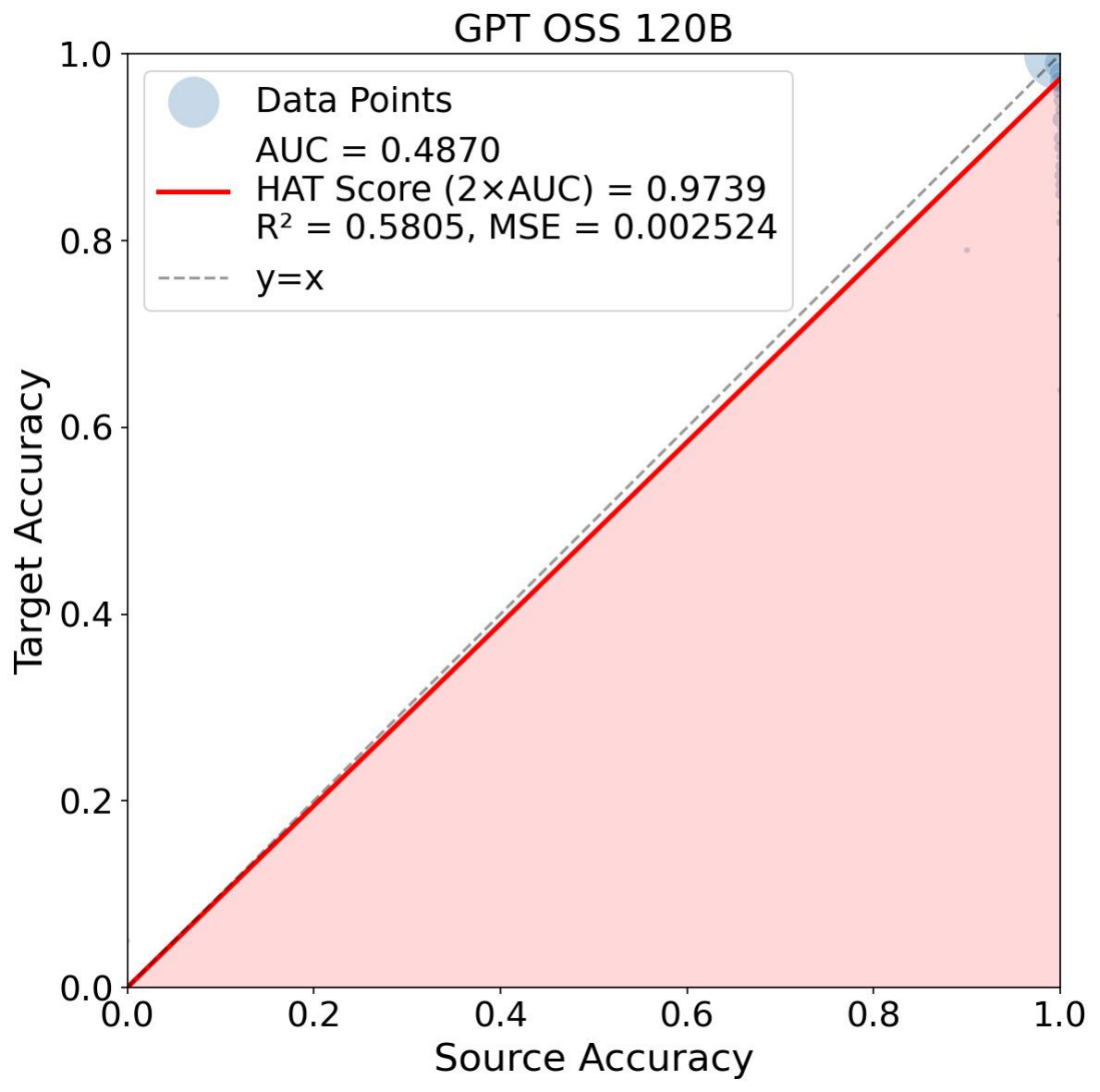} & \includegraphics[width=\linewidth]{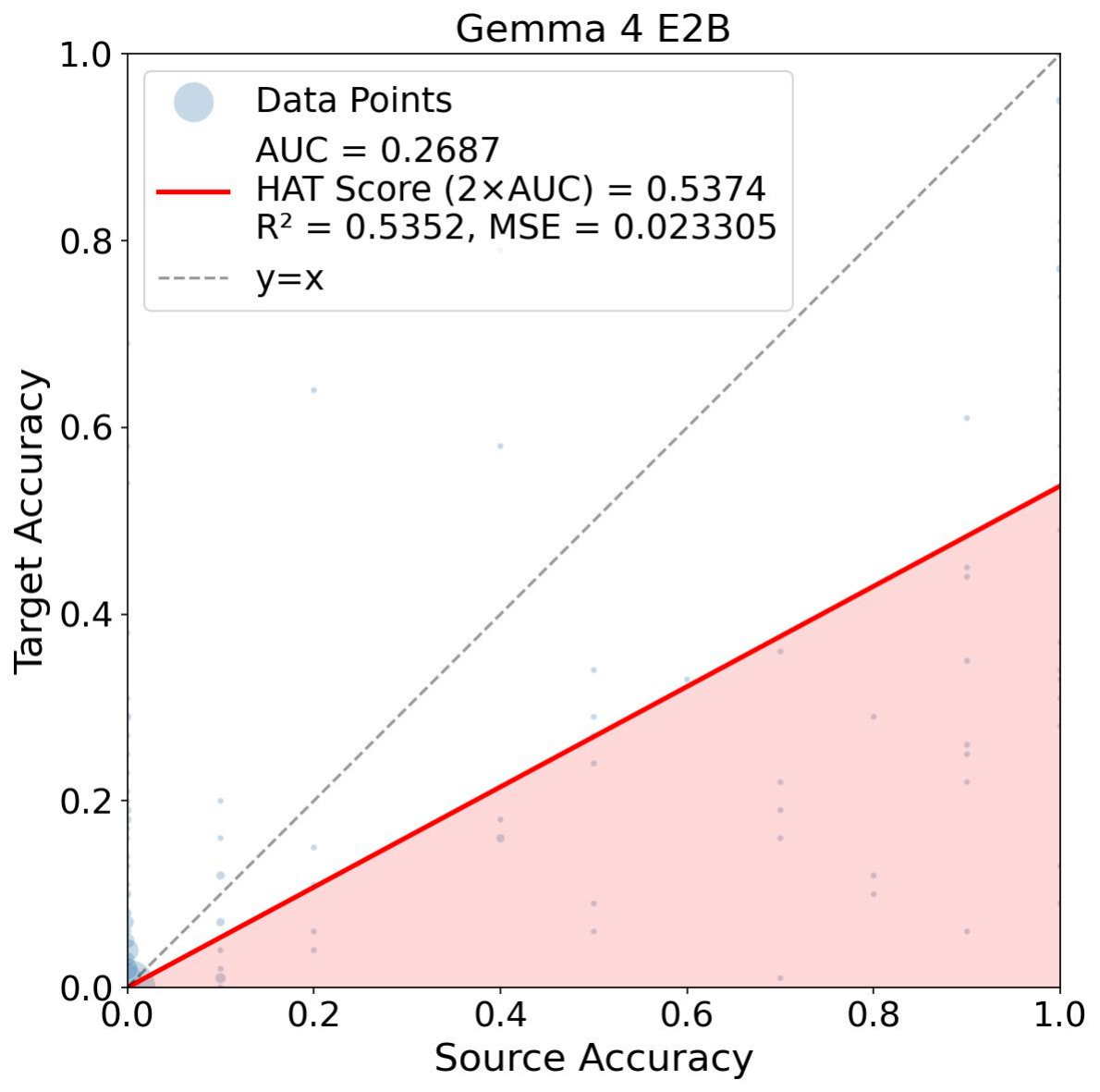} & \includegraphics[width=\linewidth]{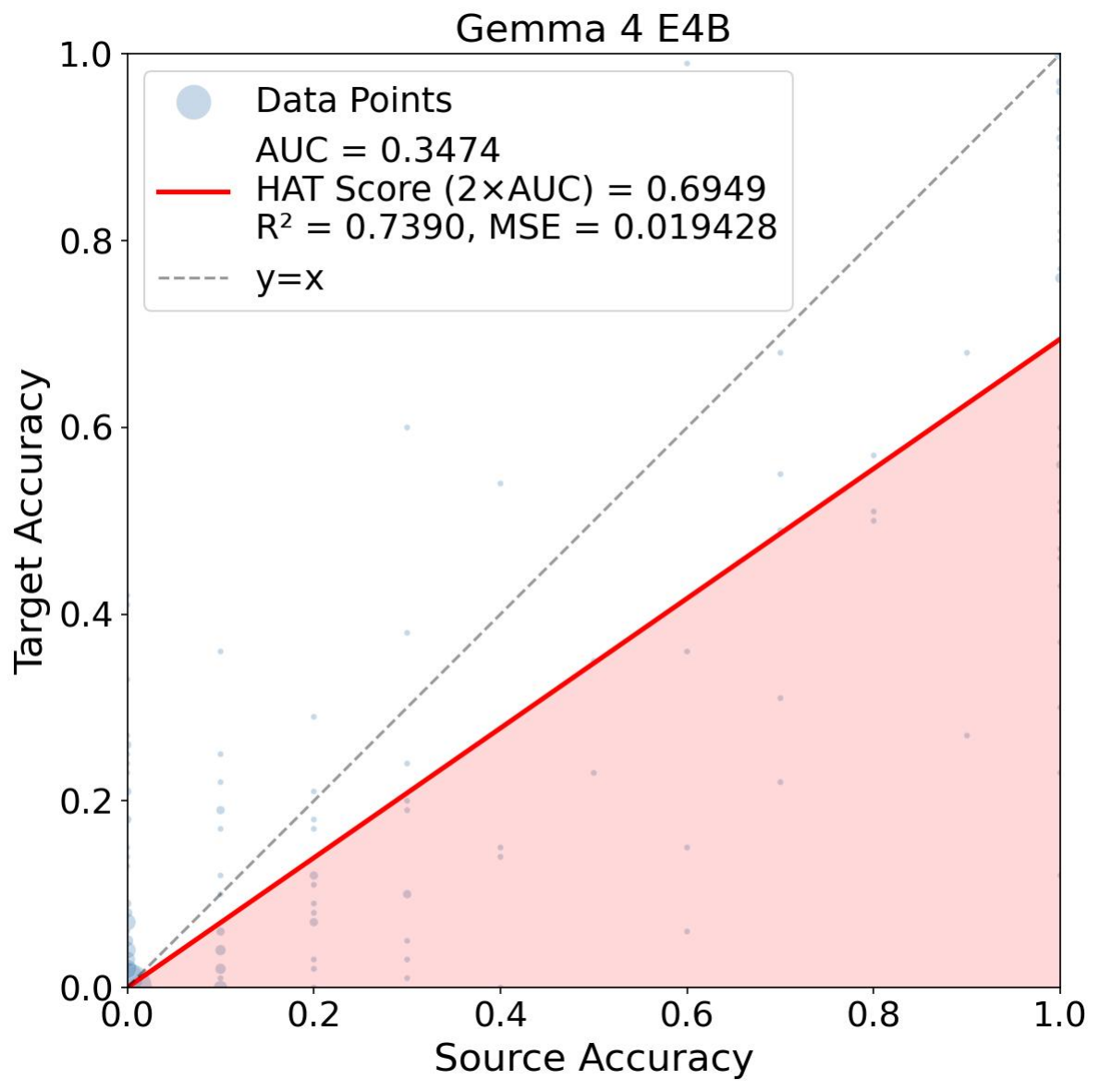} & \includegraphics[width=\linewidth]{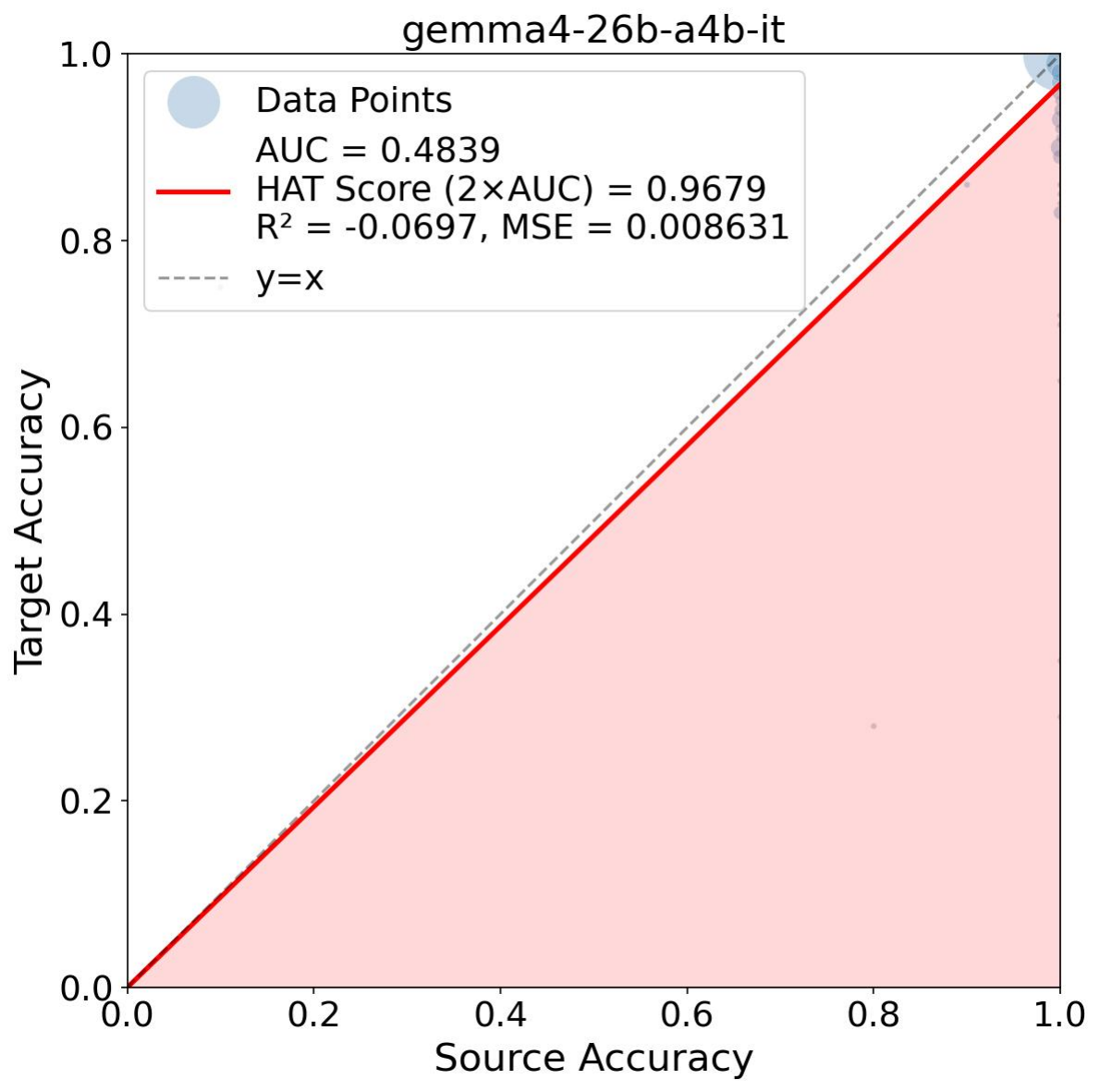} \\
\includegraphics[width=\linewidth]{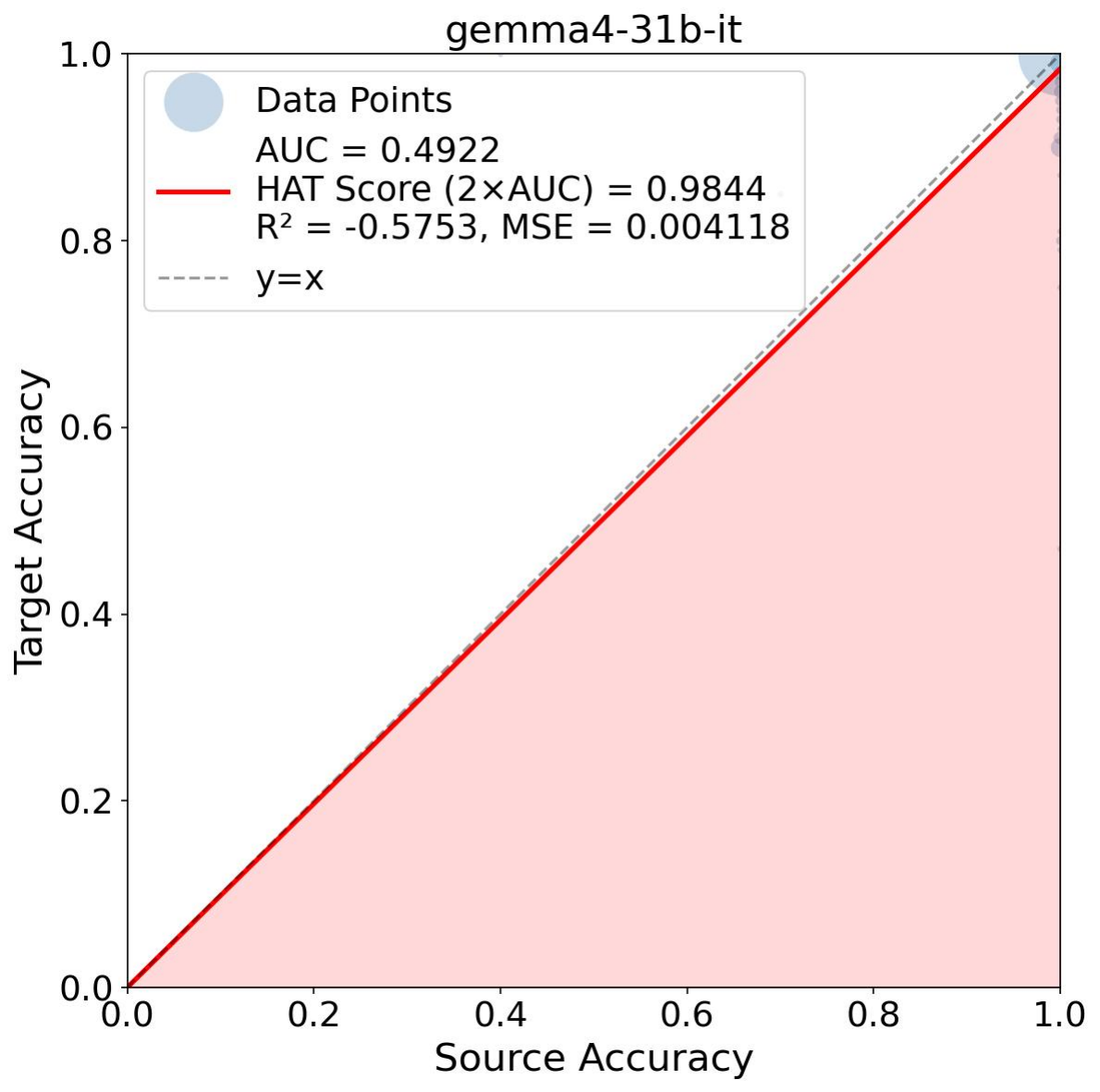} & \includegraphics[width=\linewidth]{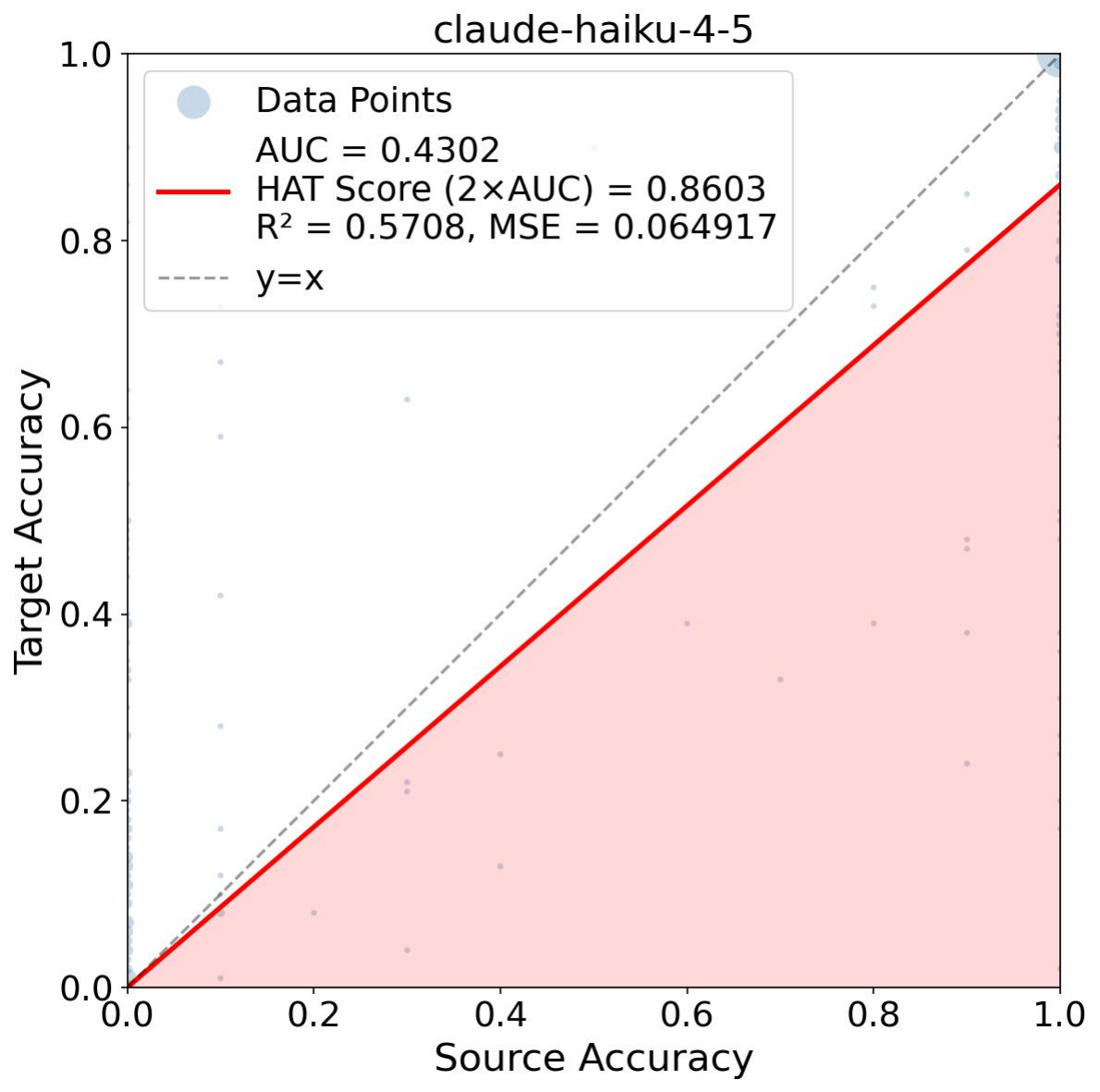} & \includegraphics[width=\linewidth]{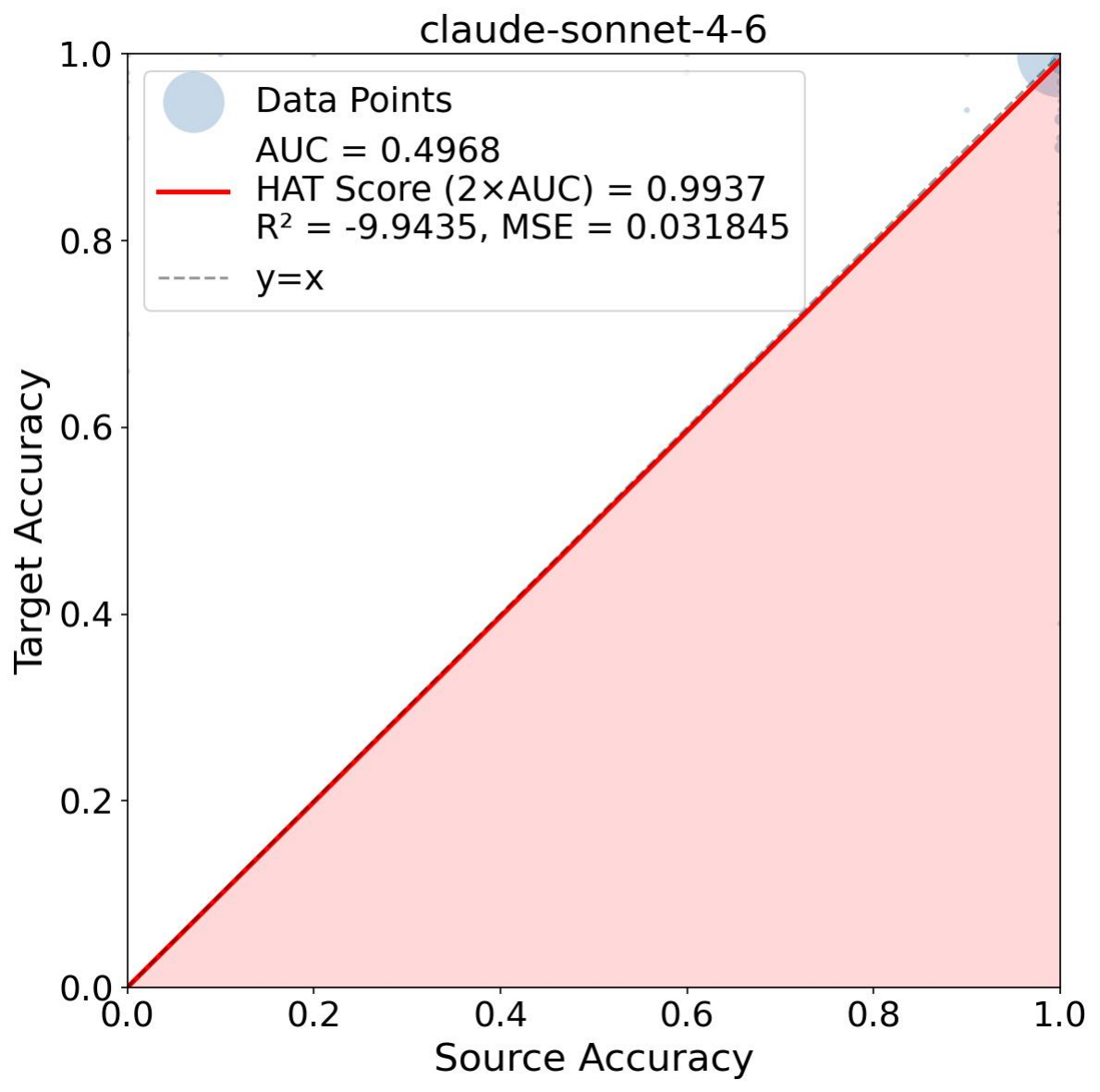} & \includegraphics[width=\linewidth]{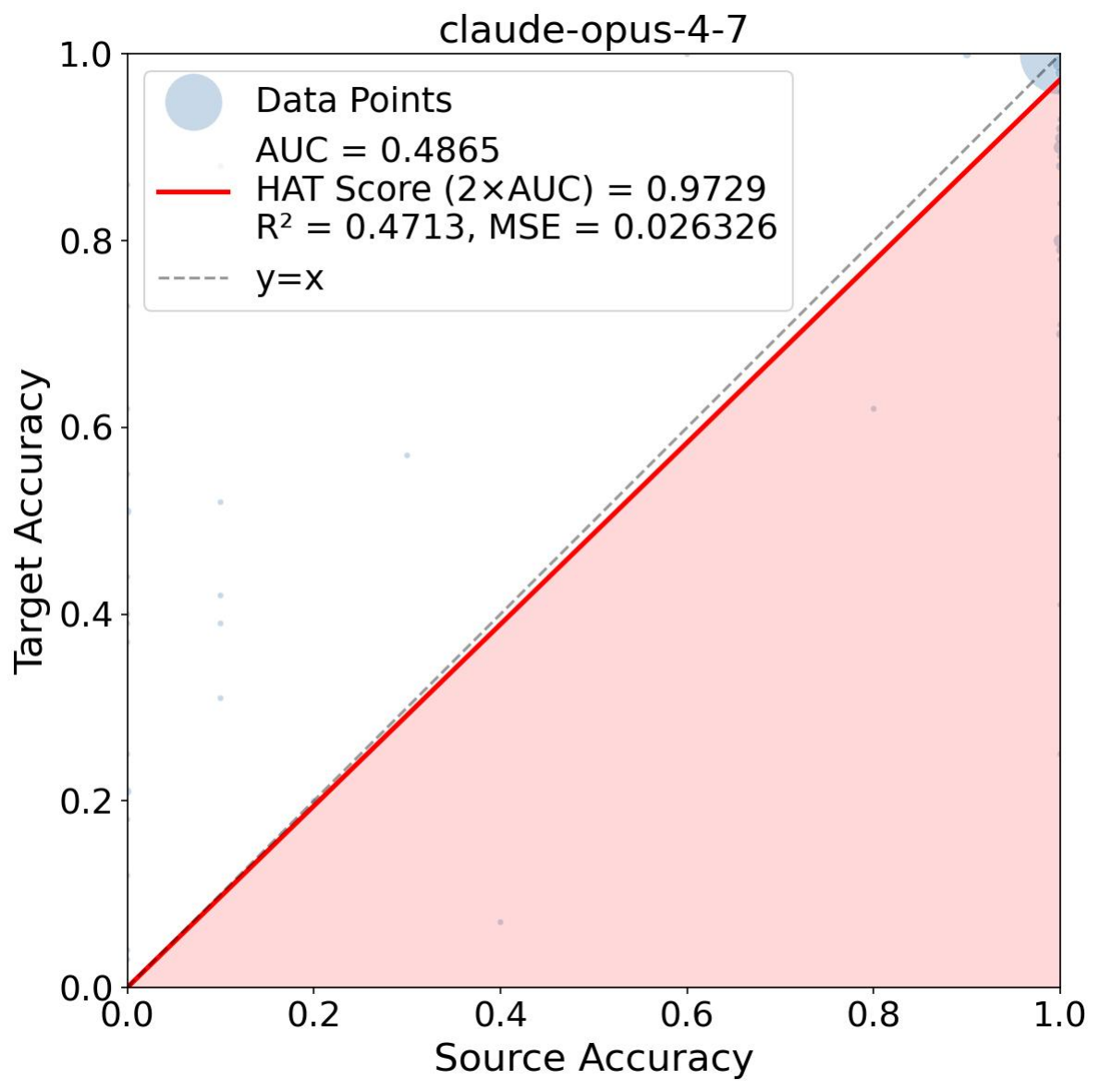} \\
\caption{HAT profiles for MGSMv2}
\end{longtable}
\twocolumn
\clearpage
\setlength{\tabcolsep}{1pt}
\onecolumn
\subsubsection{MMLU-ProX-Lite HAT Plots}
Detailed evaluation curves and HAT score transfer profiles for the MMLU-ProX-Lite dataset are provided below.
\begin{longtable}{>{\centering\arraybackslash}p{0.24\textwidth} >{\centering\arraybackslash}p{0.24\textwidth} >{\centering\arraybackslash}p{0.24\textwidth} >{\centering\arraybackslash}p{0.24\textwidth}}
\includegraphics[width=\linewidth]{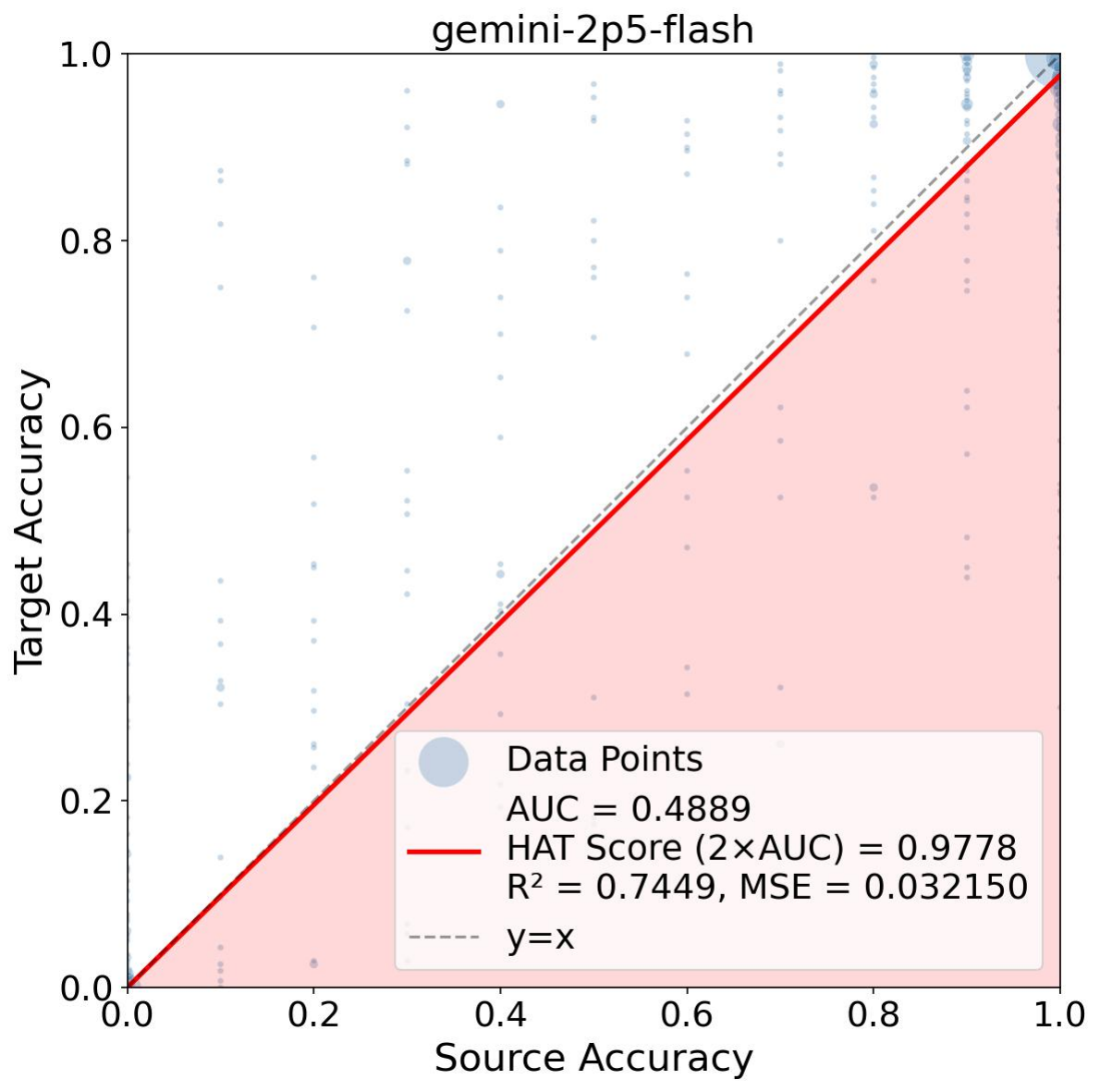} & \includegraphics[width=\linewidth]{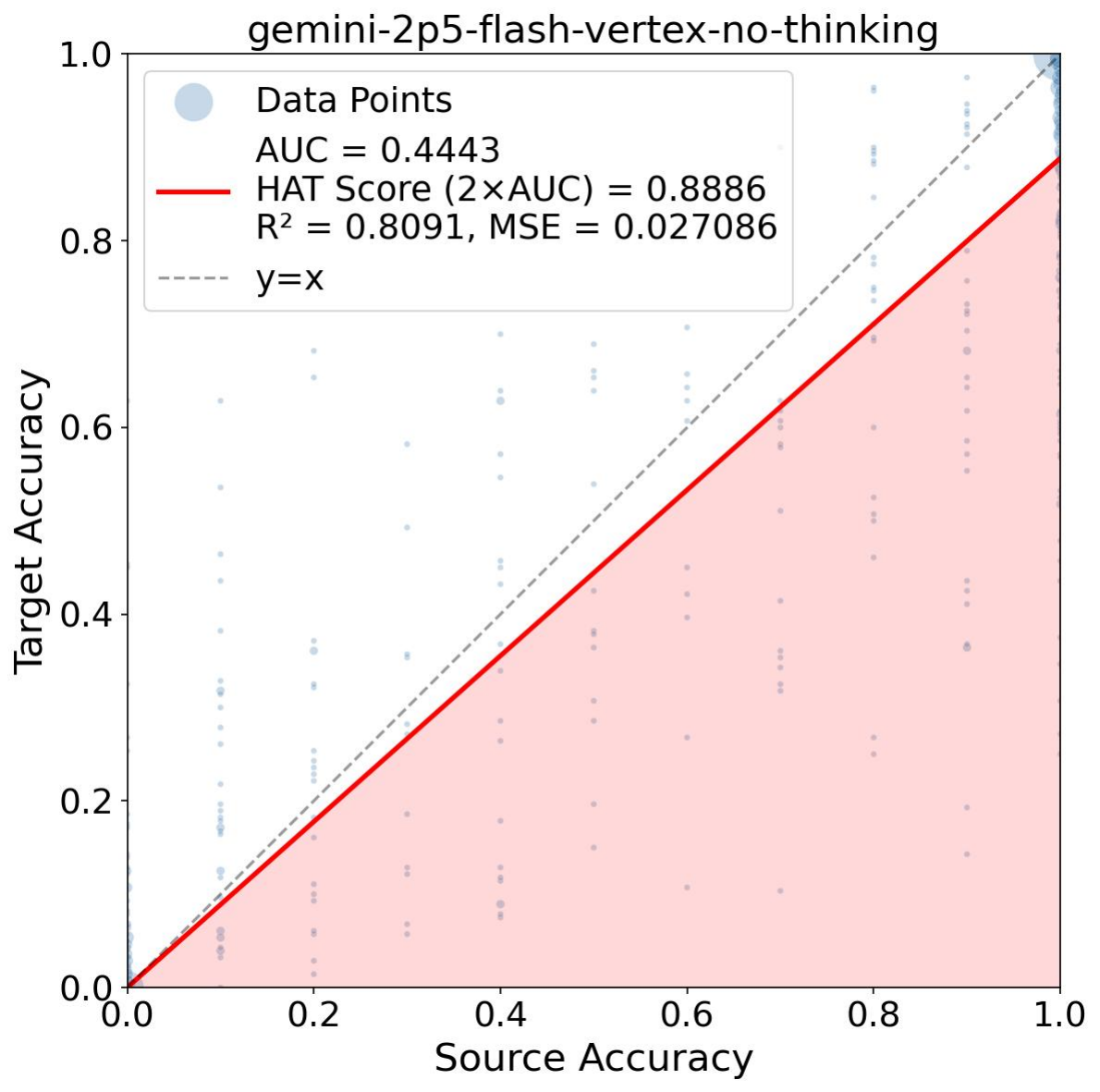} & \includegraphics[width=\linewidth]{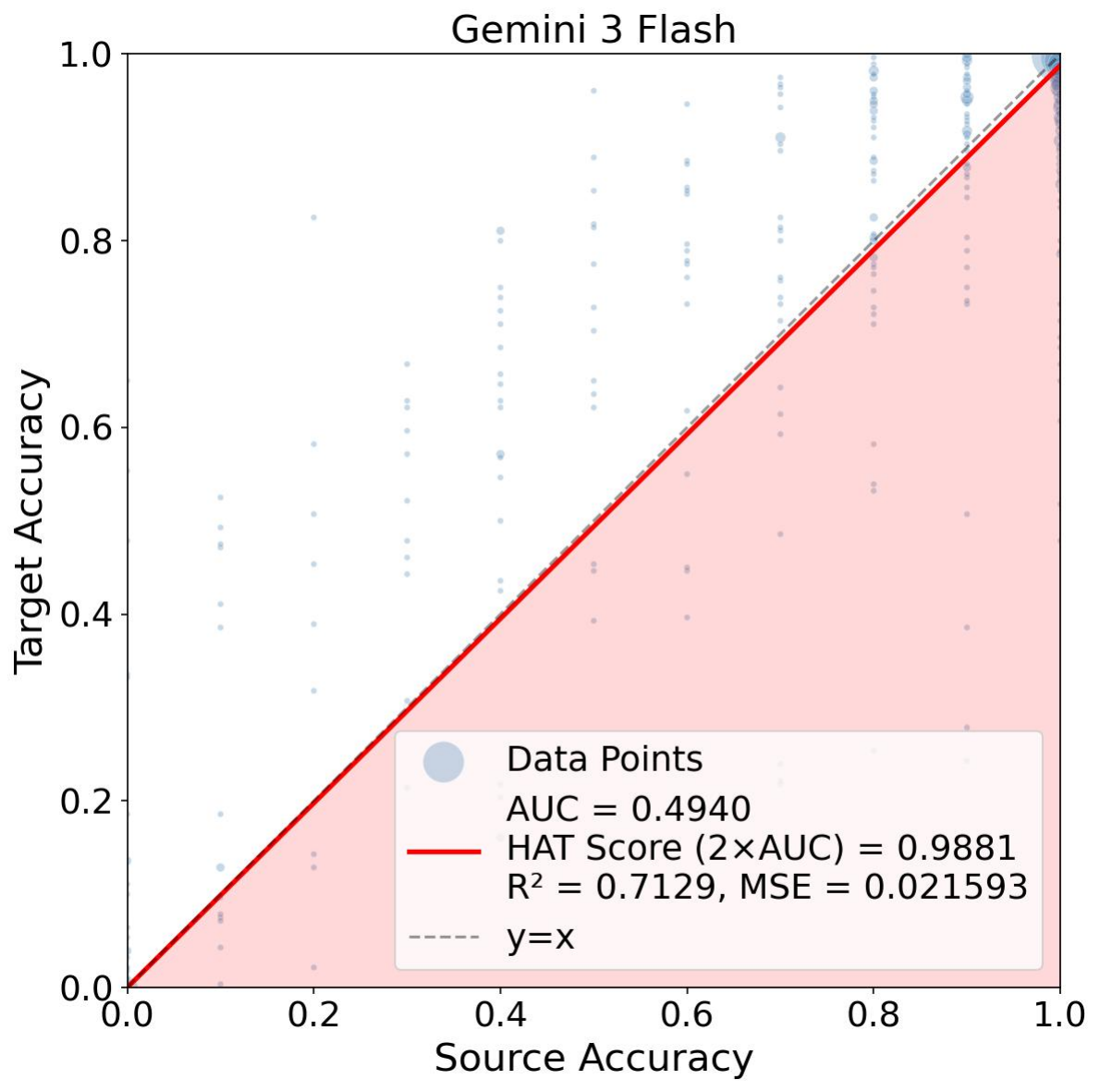} & \includegraphics[width=\linewidth]{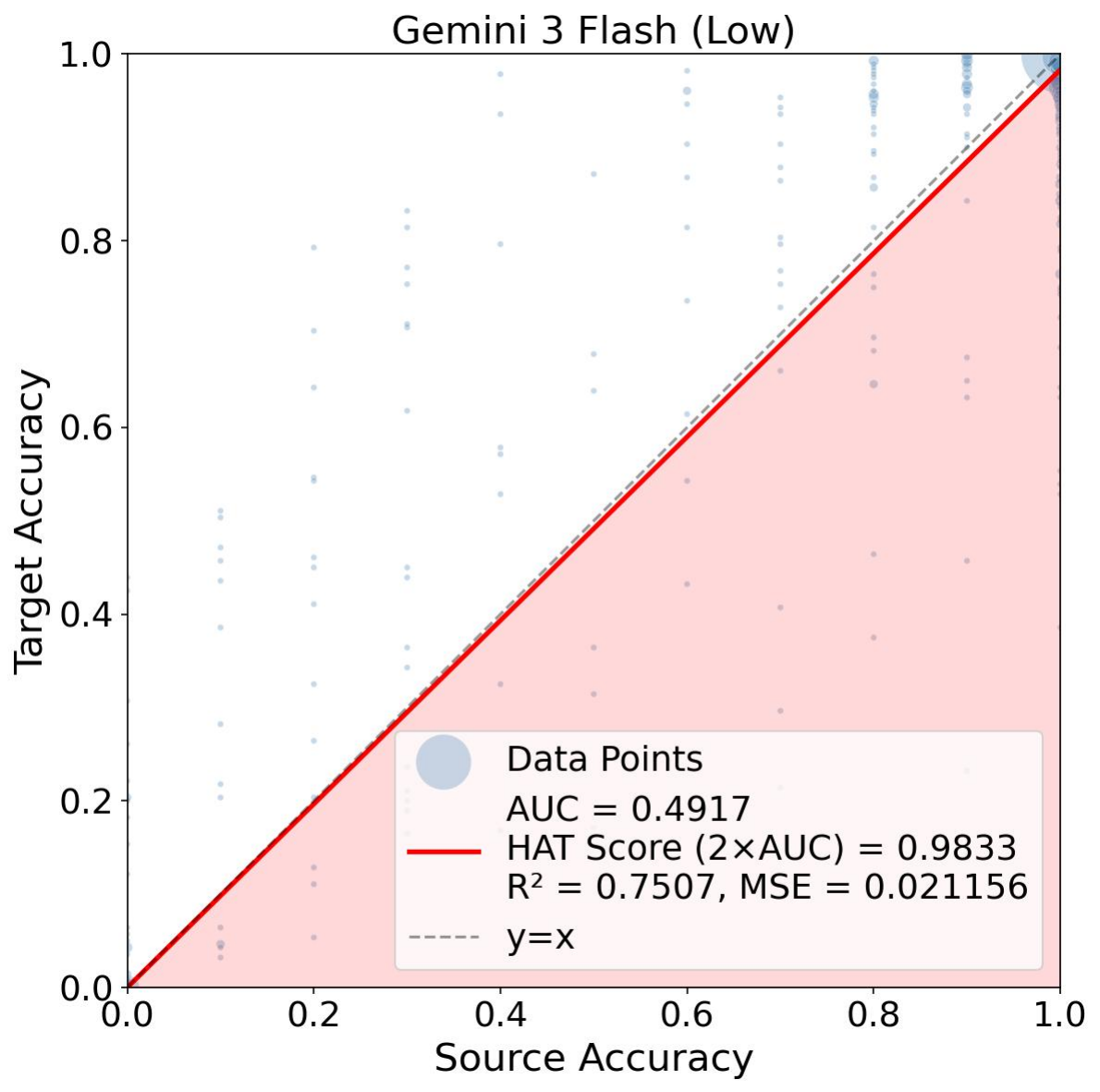} \\
\includegraphics[width=\linewidth]{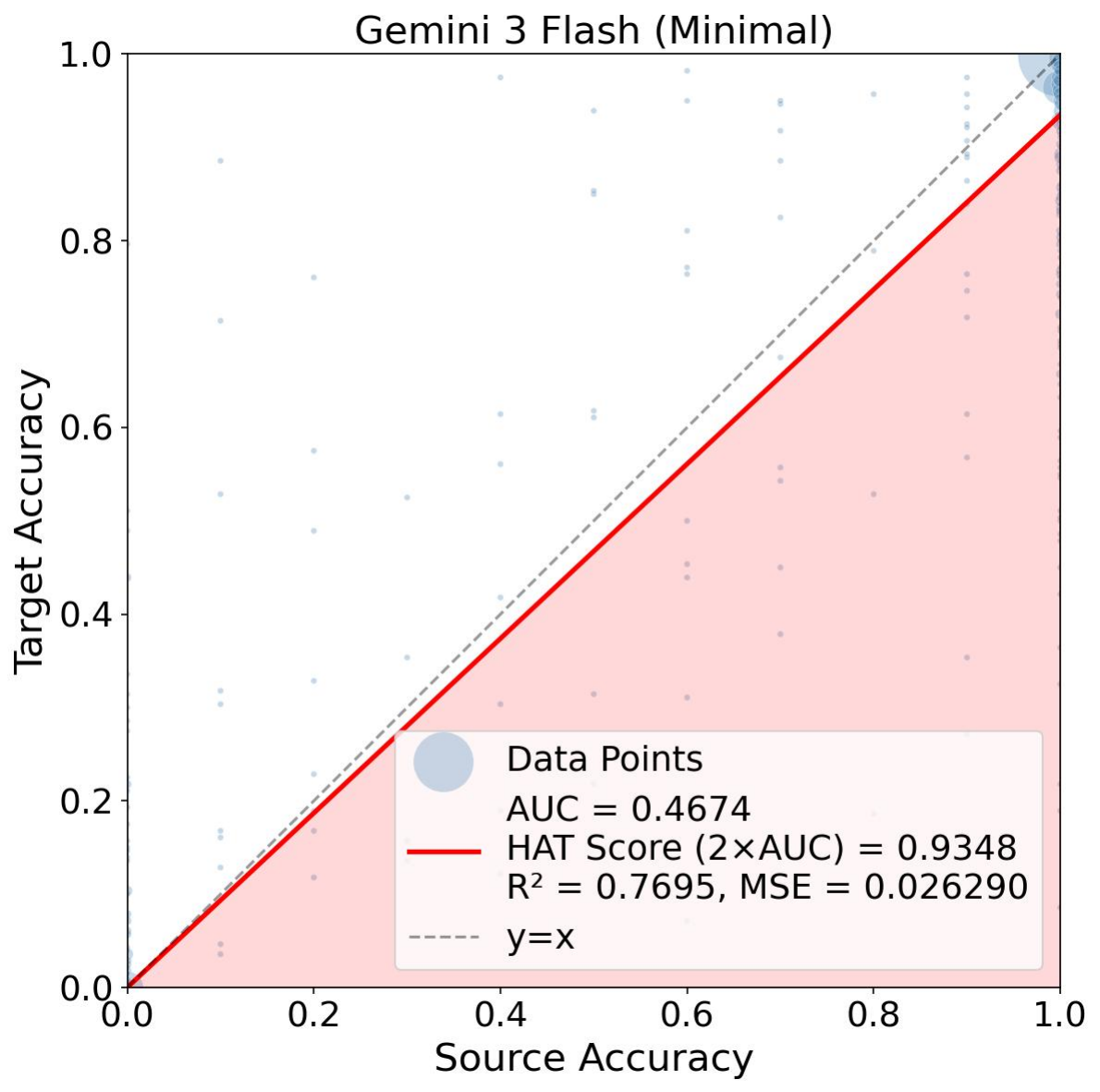} & \includegraphics[width=\linewidth]{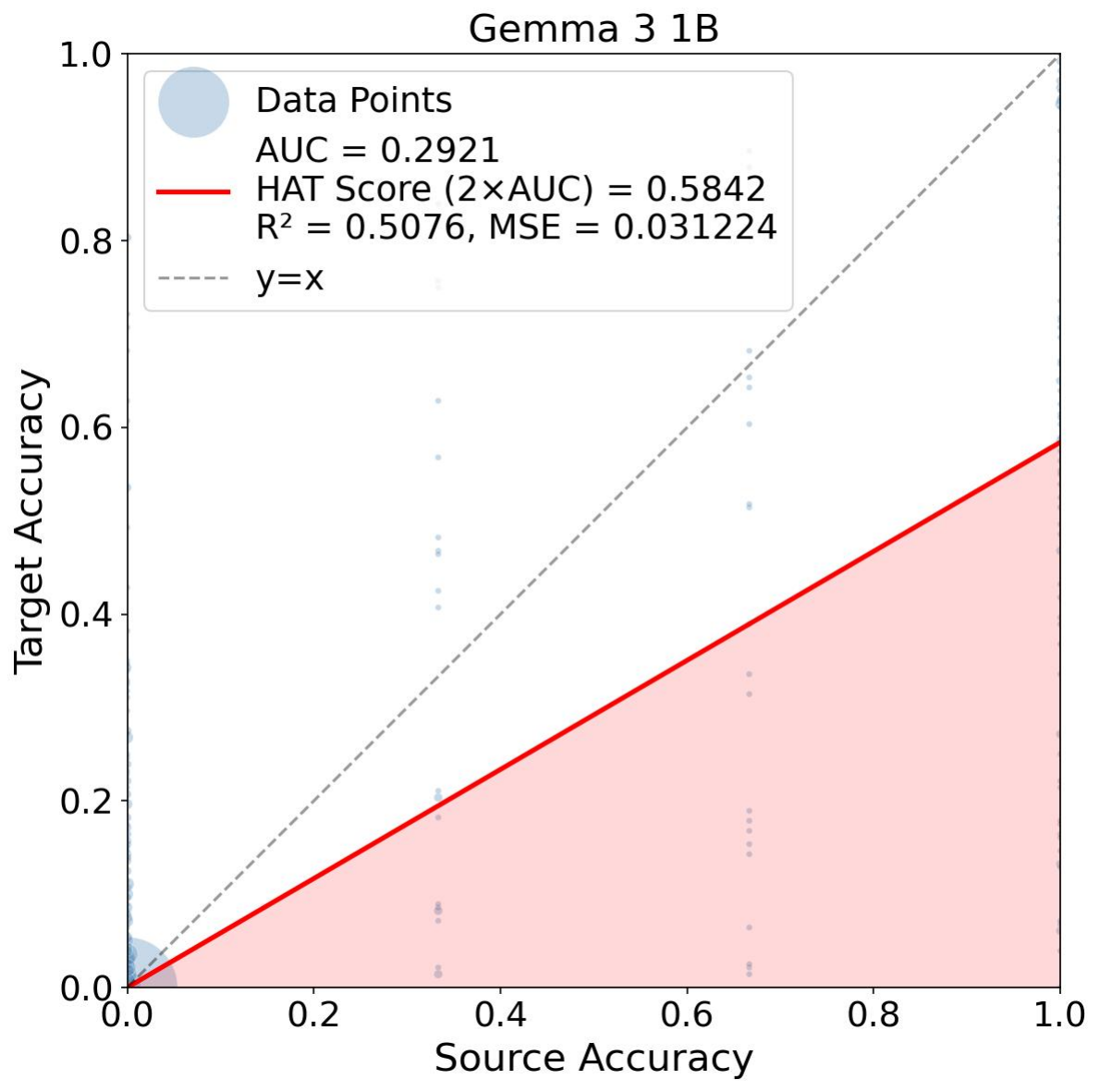} & \includegraphics[width=\linewidth]{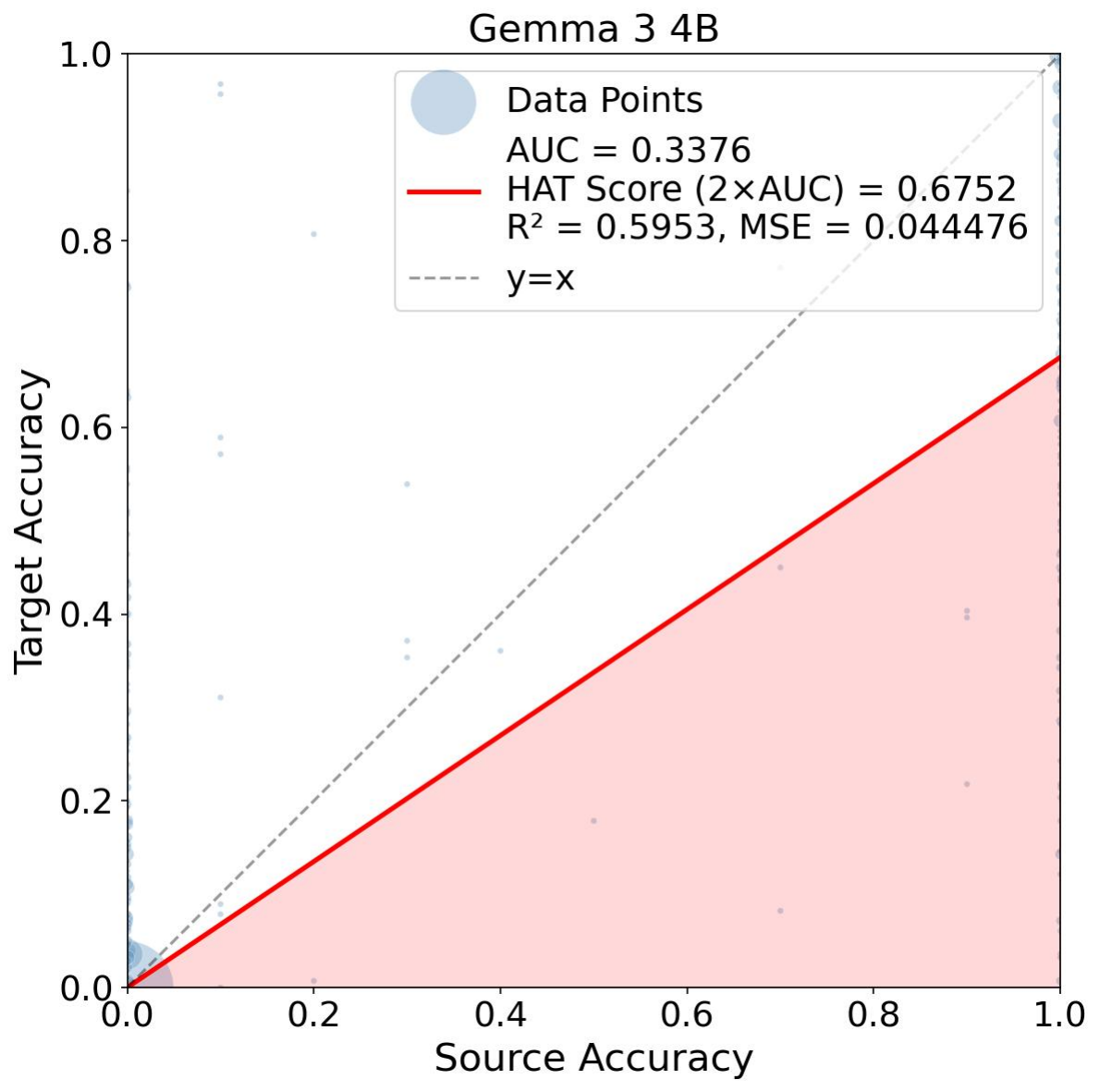} & \includegraphics[width=\linewidth]{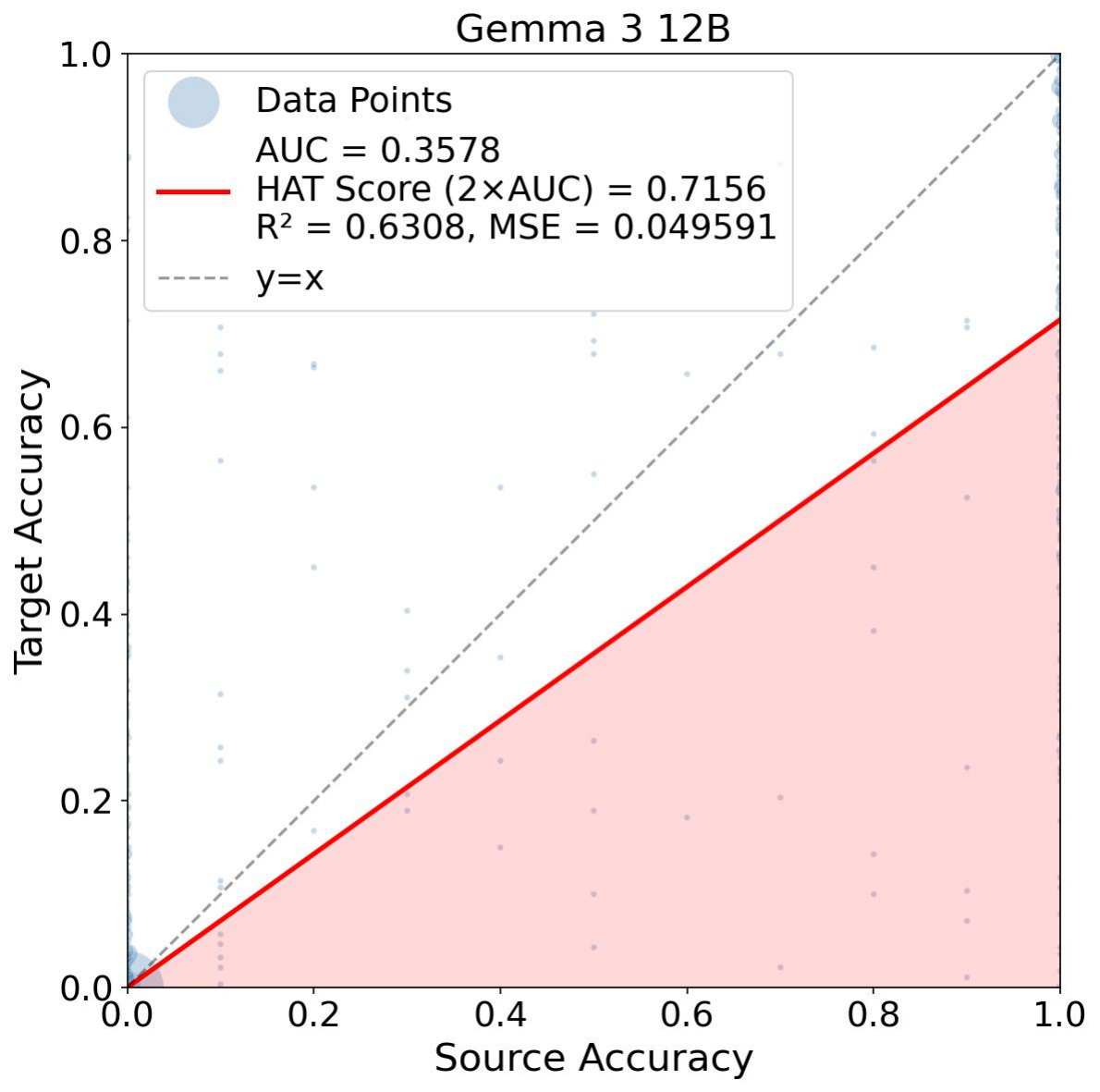} \\
\includegraphics[width=\linewidth]{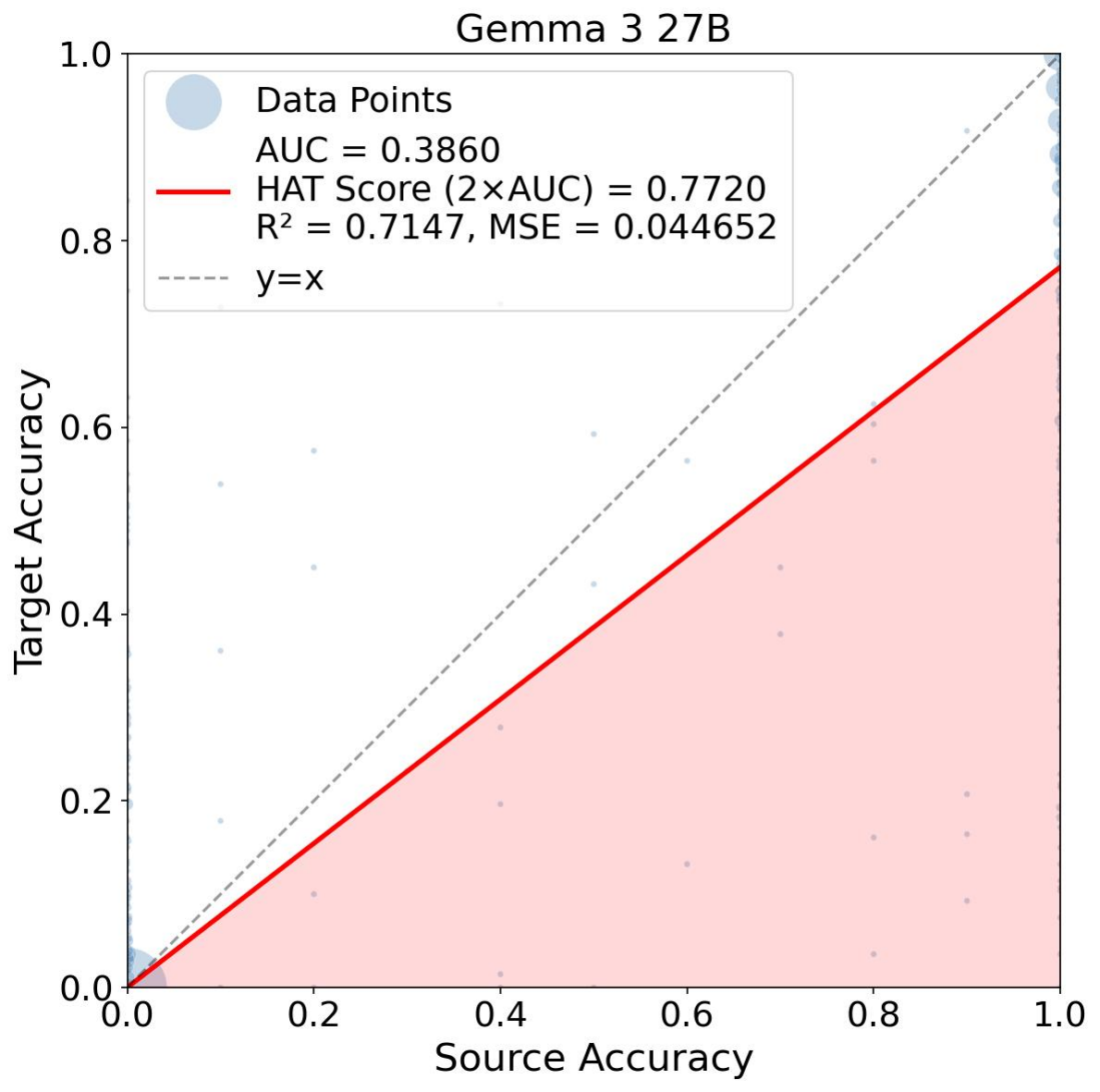} & \includegraphics[width=\linewidth]{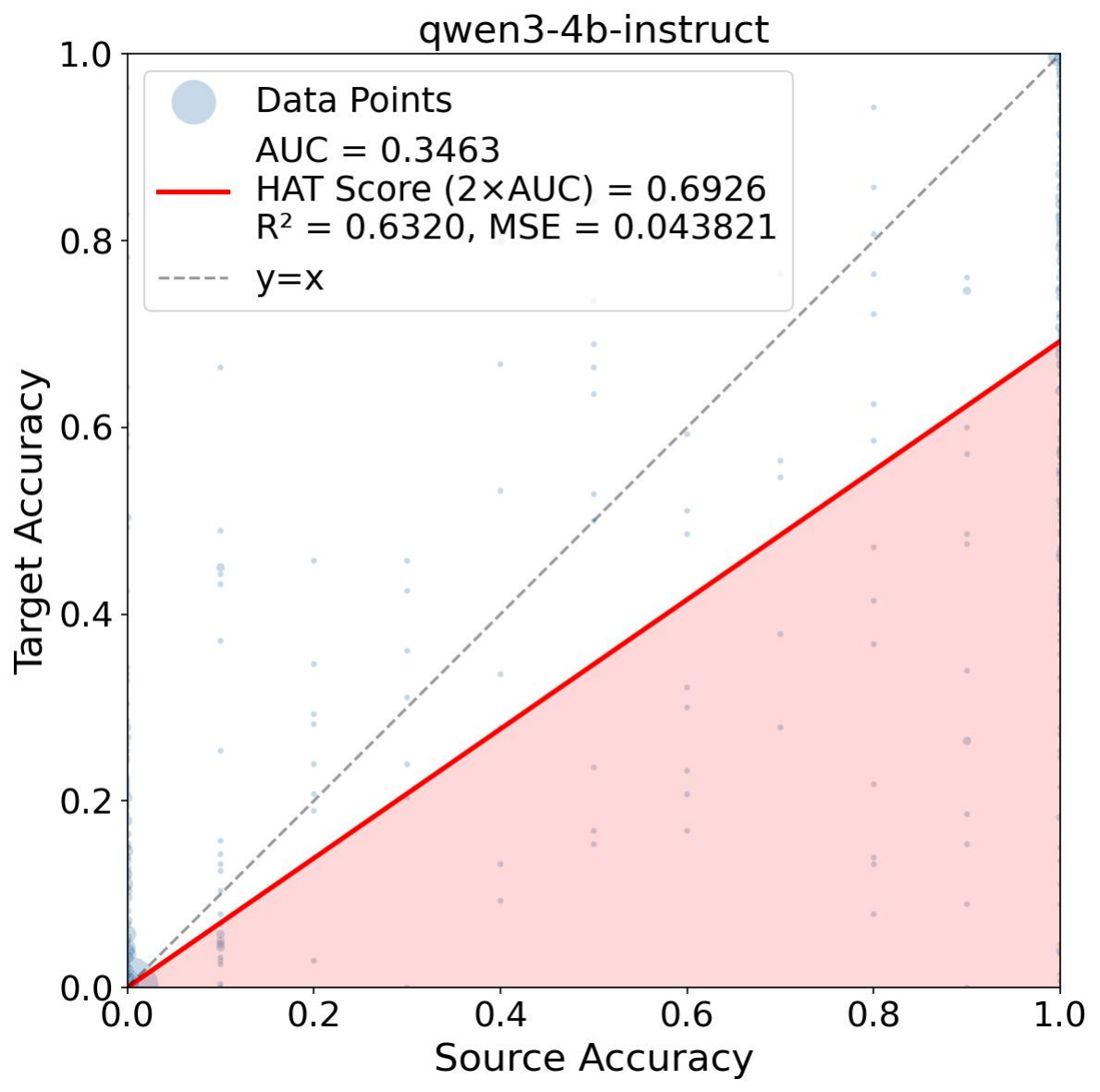} & \includegraphics[width=\linewidth]{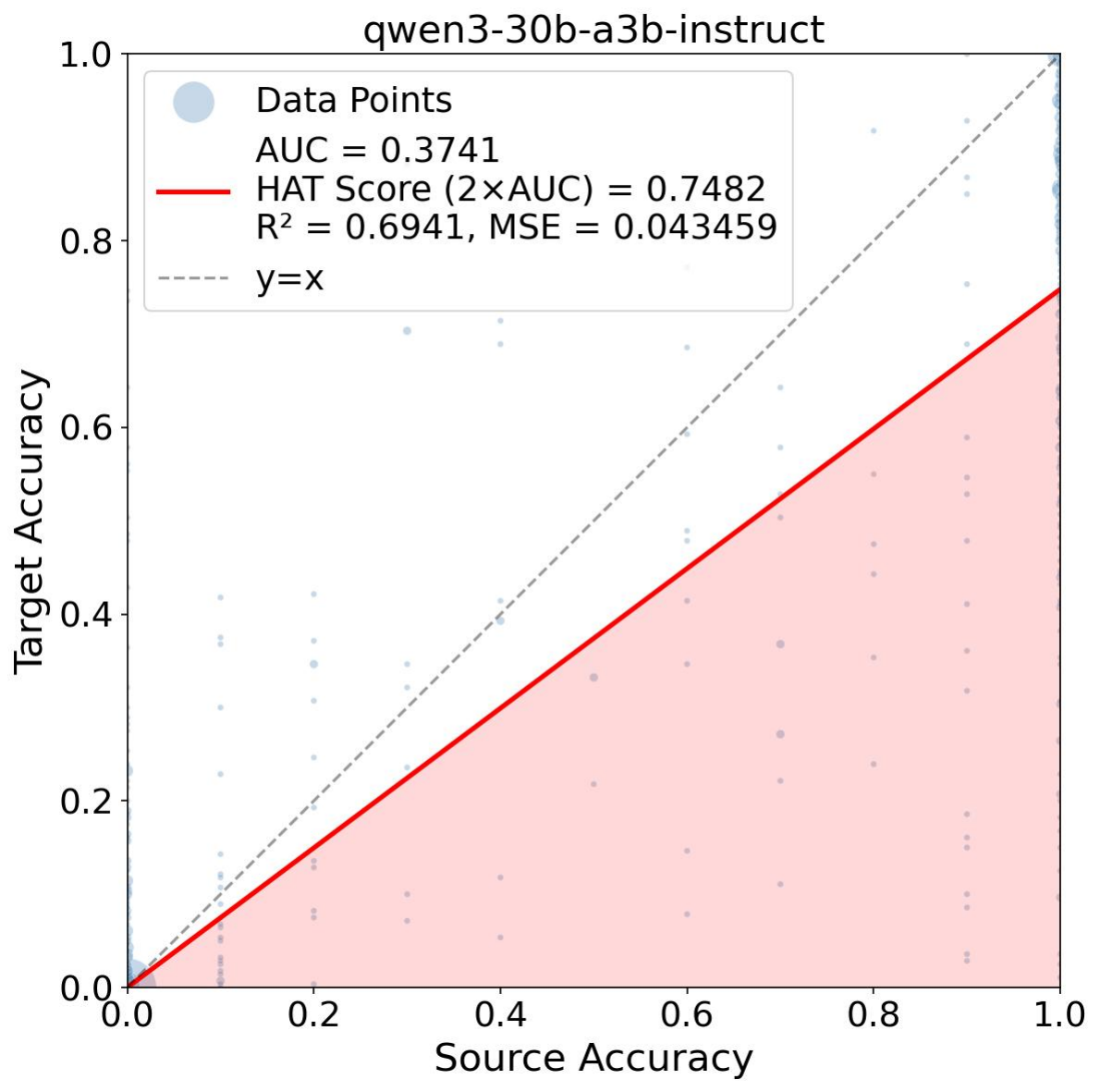} & \includegraphics[width=\linewidth]{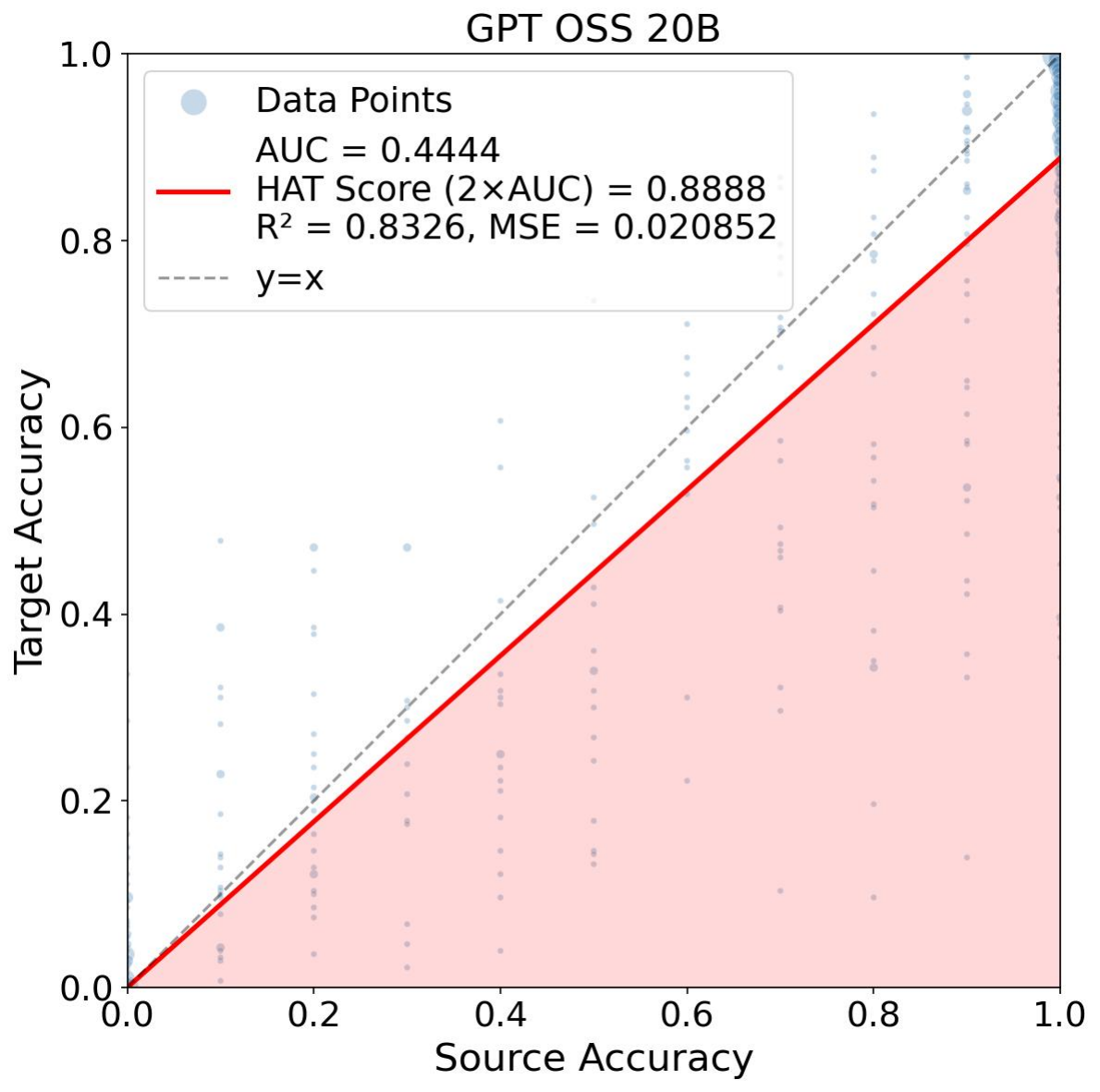} \\
\includegraphics[width=\linewidth]{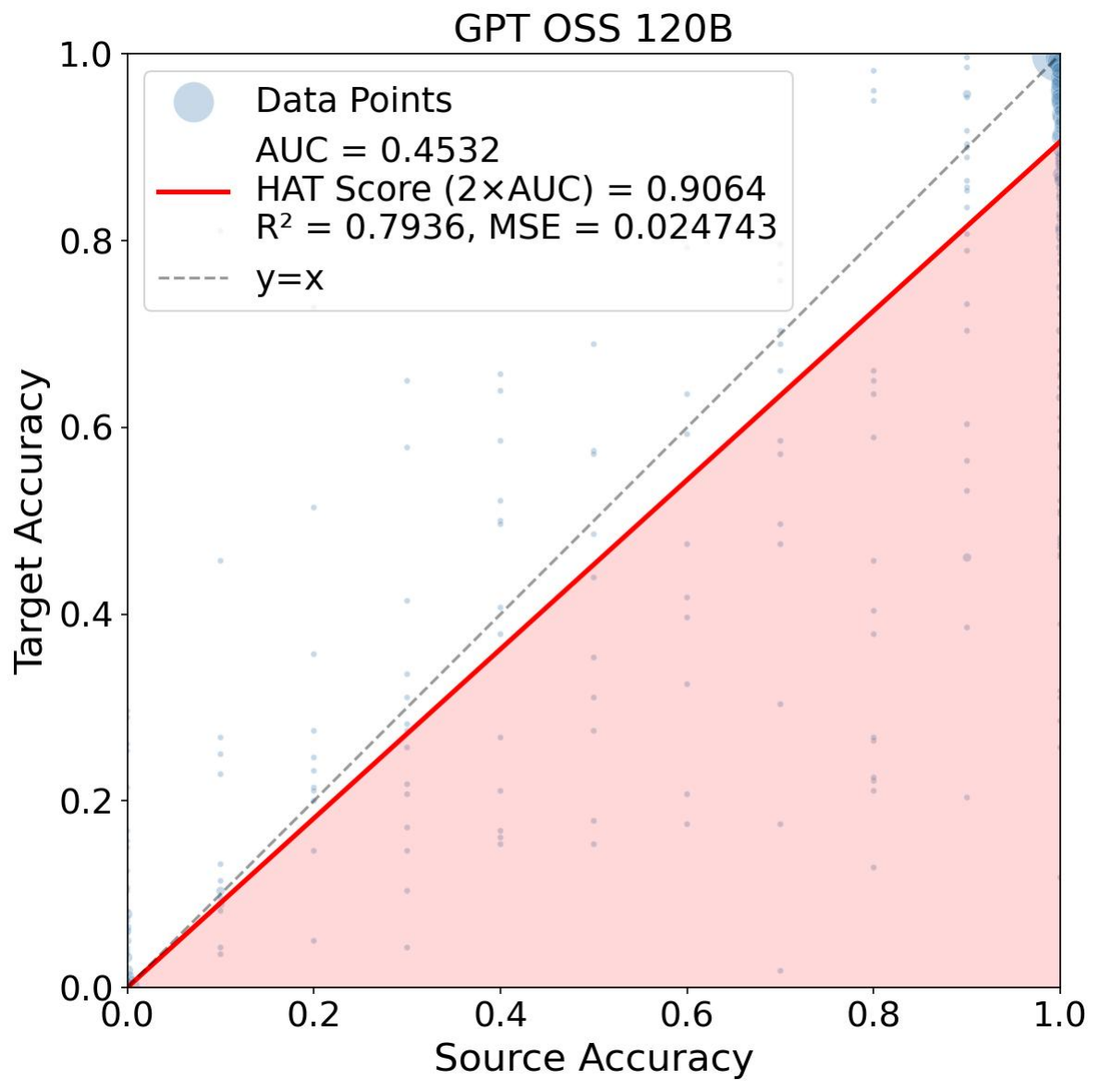} & \includegraphics[width=\linewidth]{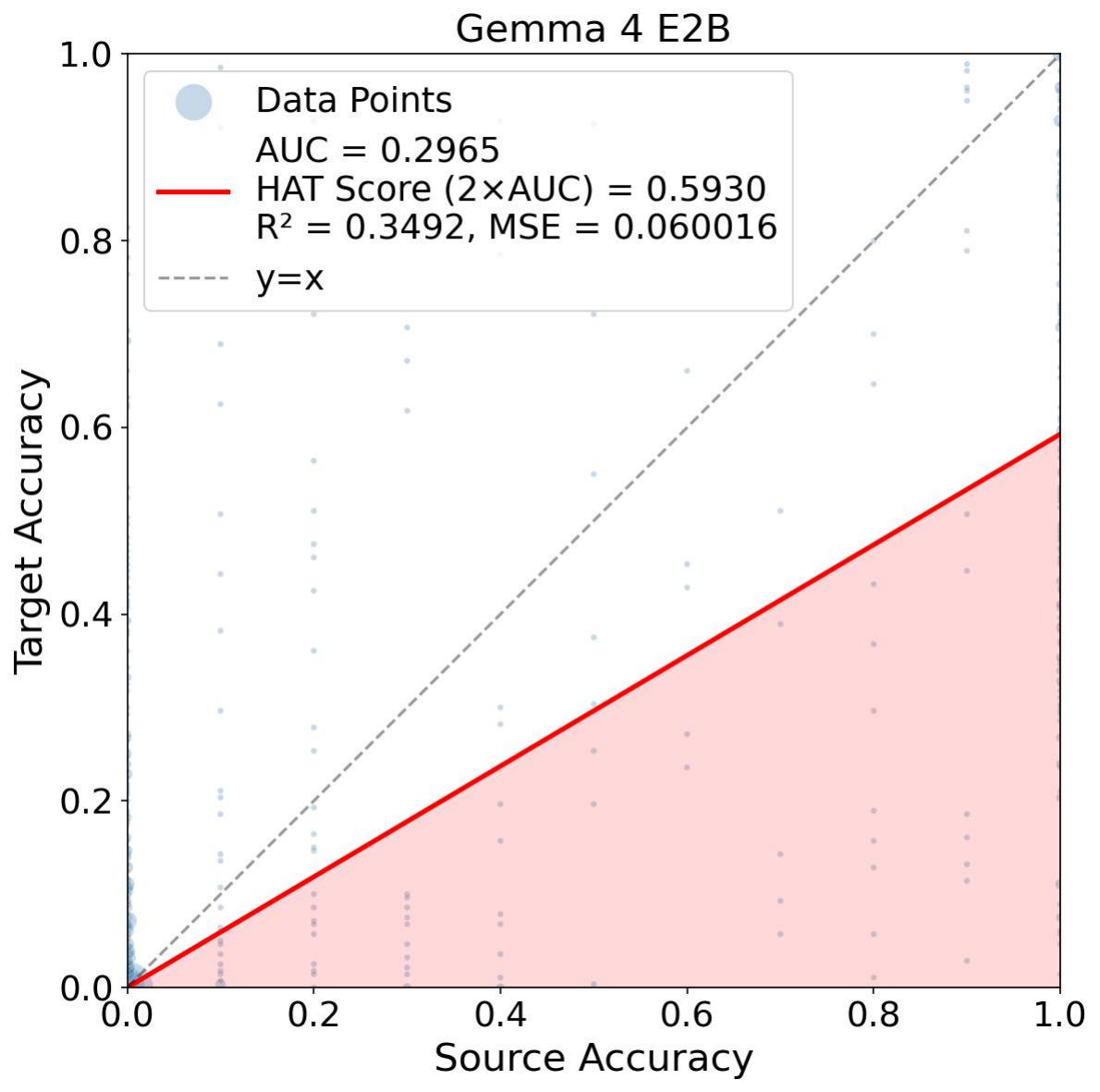} & \includegraphics[width=\linewidth]{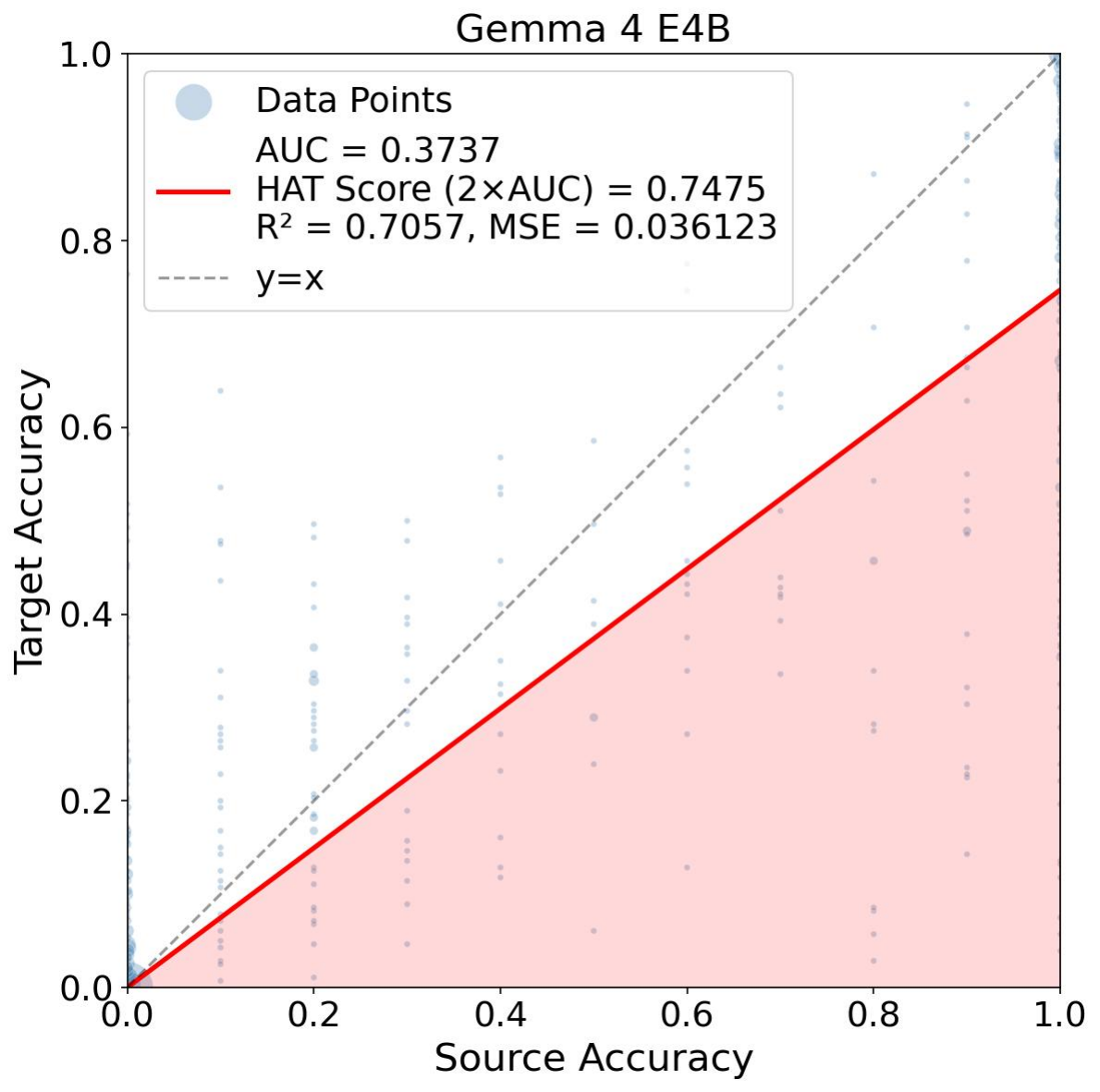} & \includegraphics[width=\linewidth]{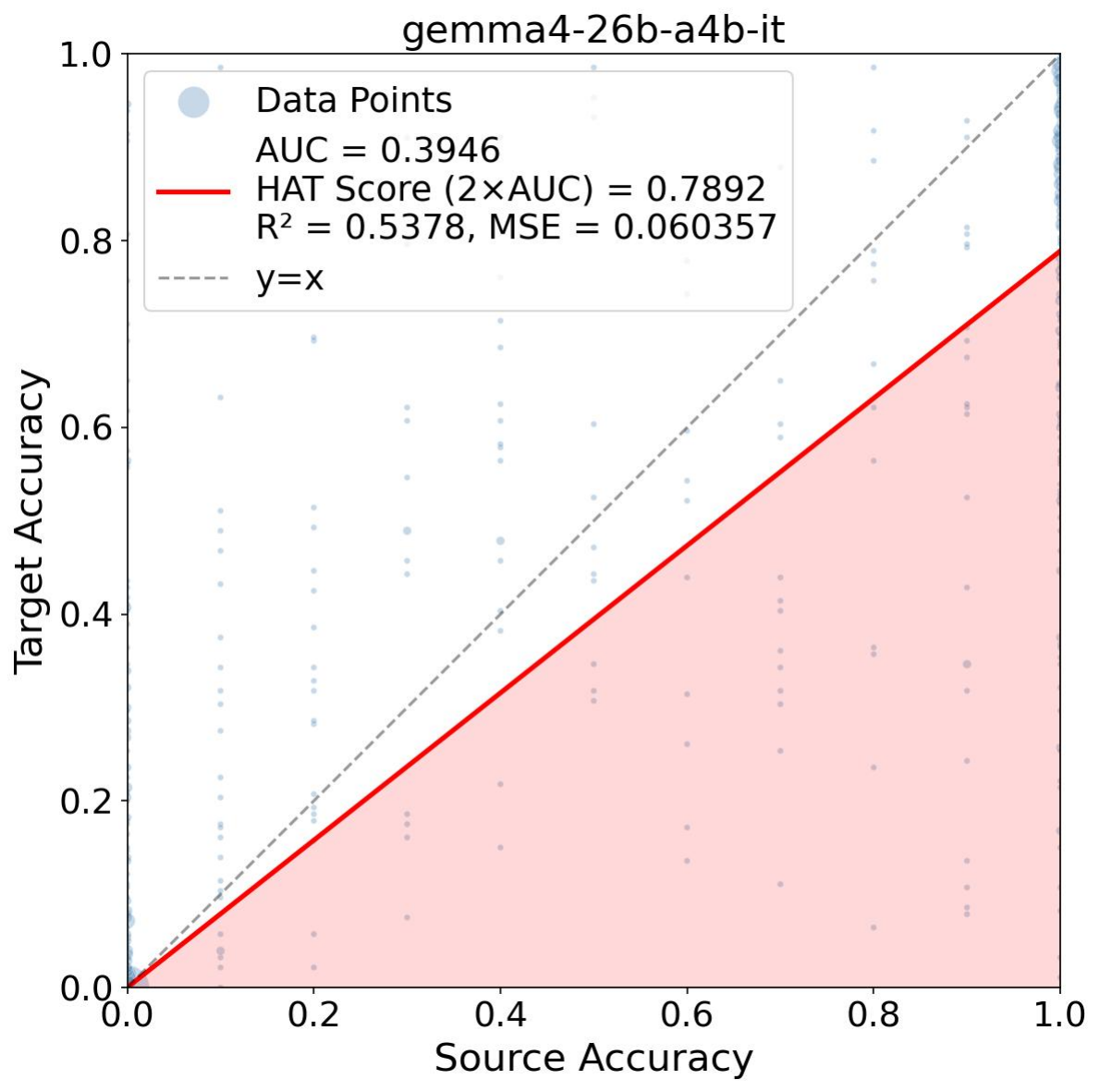} \\
\includegraphics[width=\linewidth]{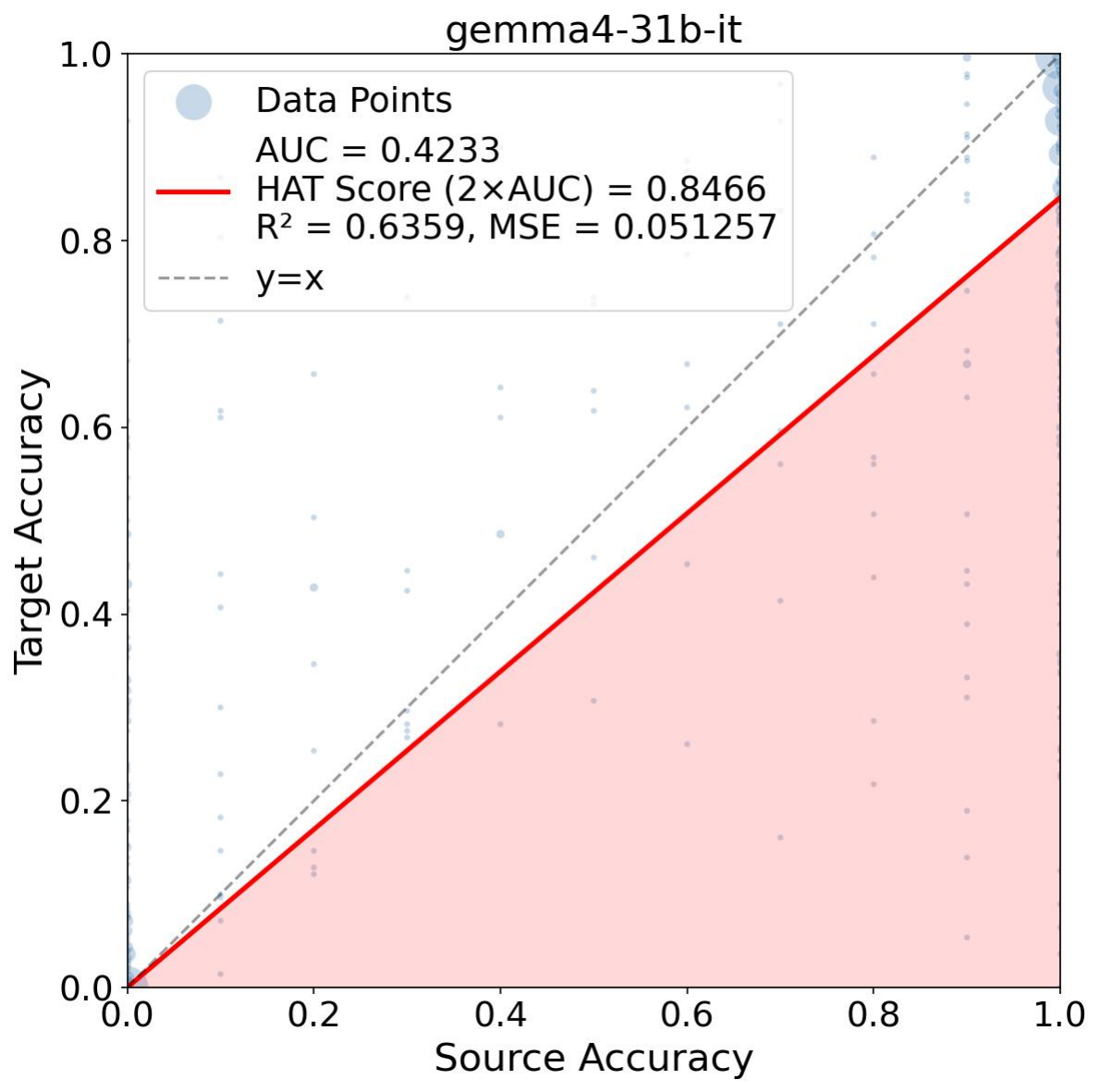} & \includegraphics[width=\linewidth]{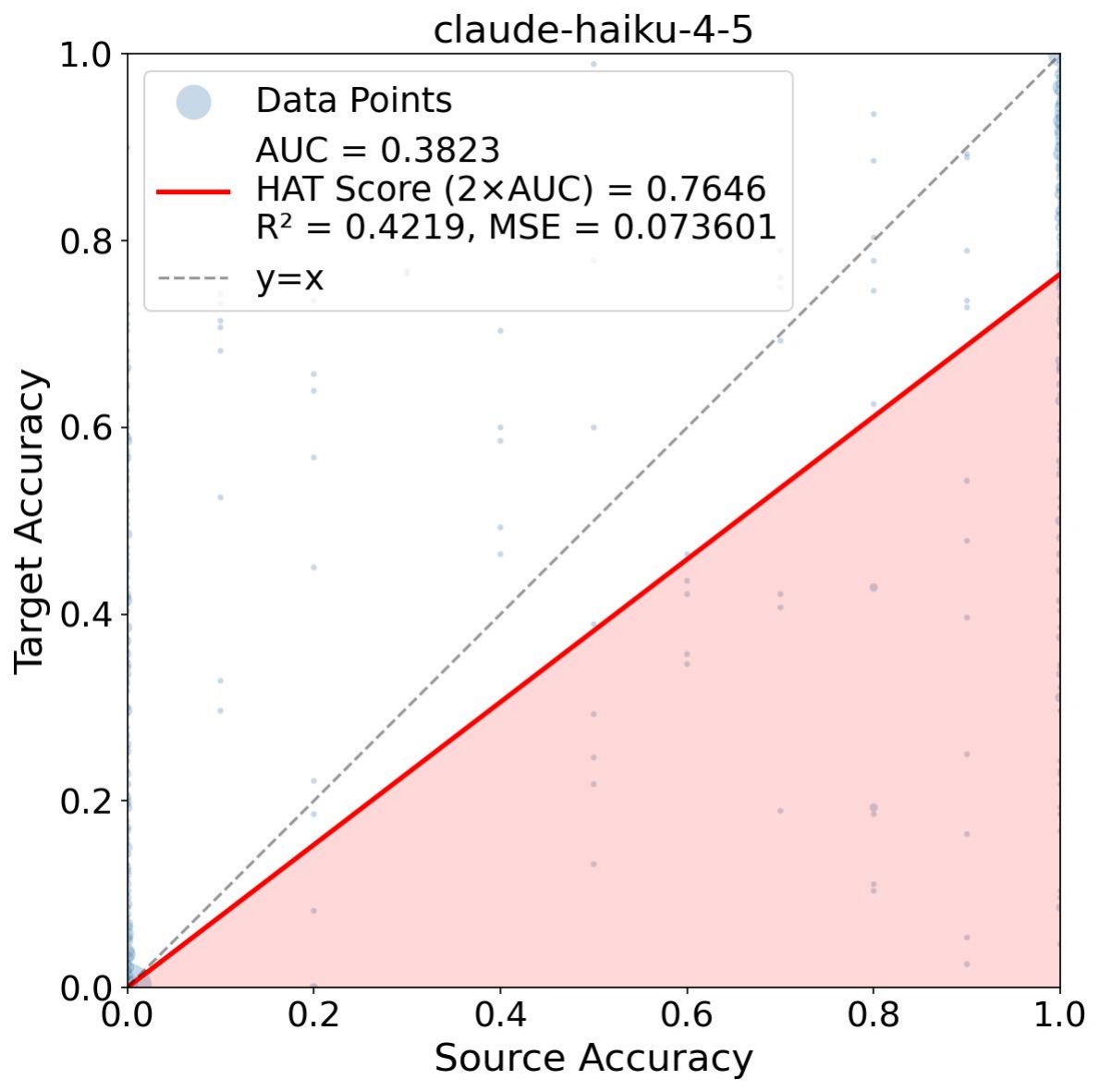} & \includegraphics[width=\linewidth]{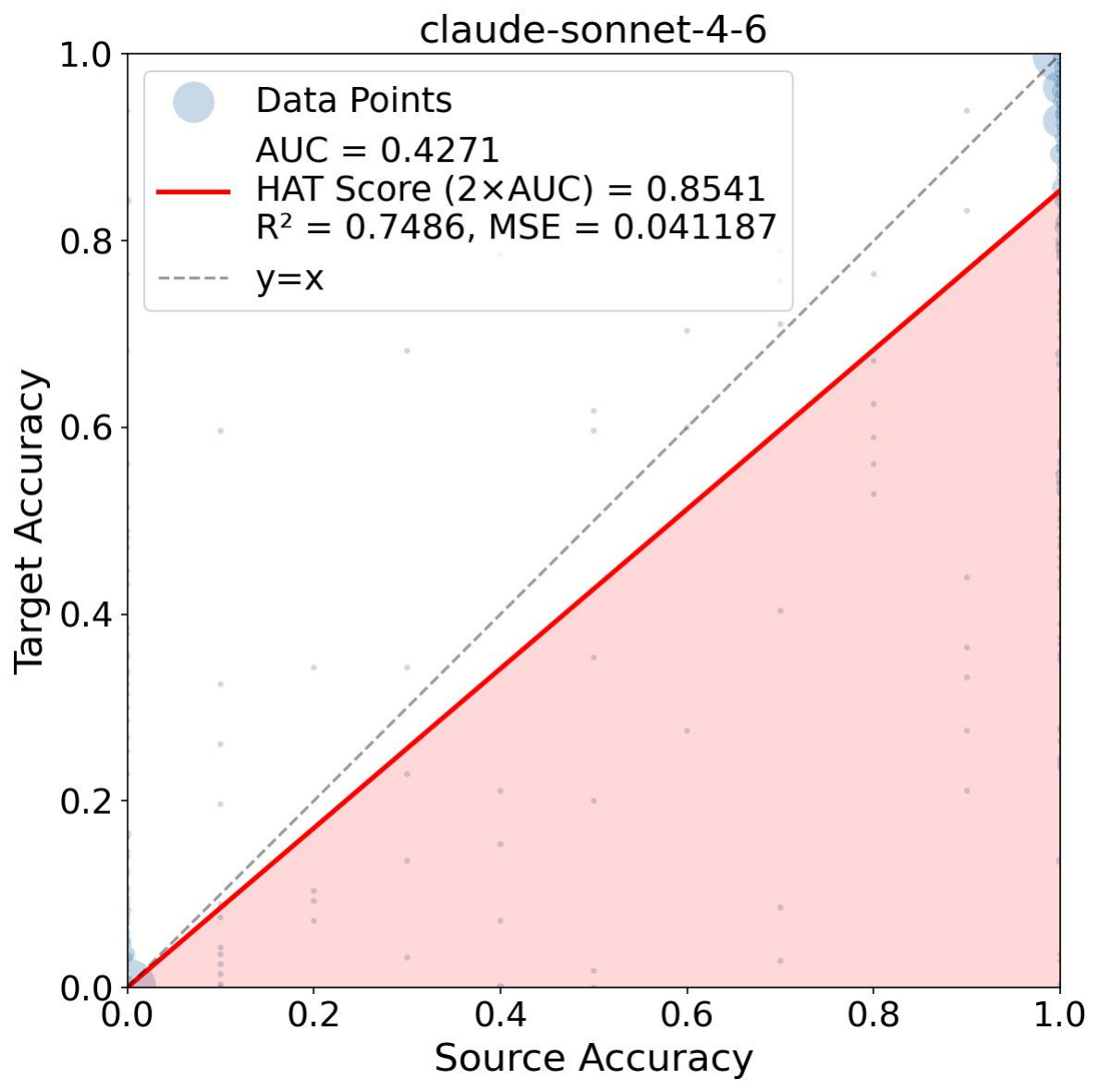} & \includegraphics[width=\linewidth]{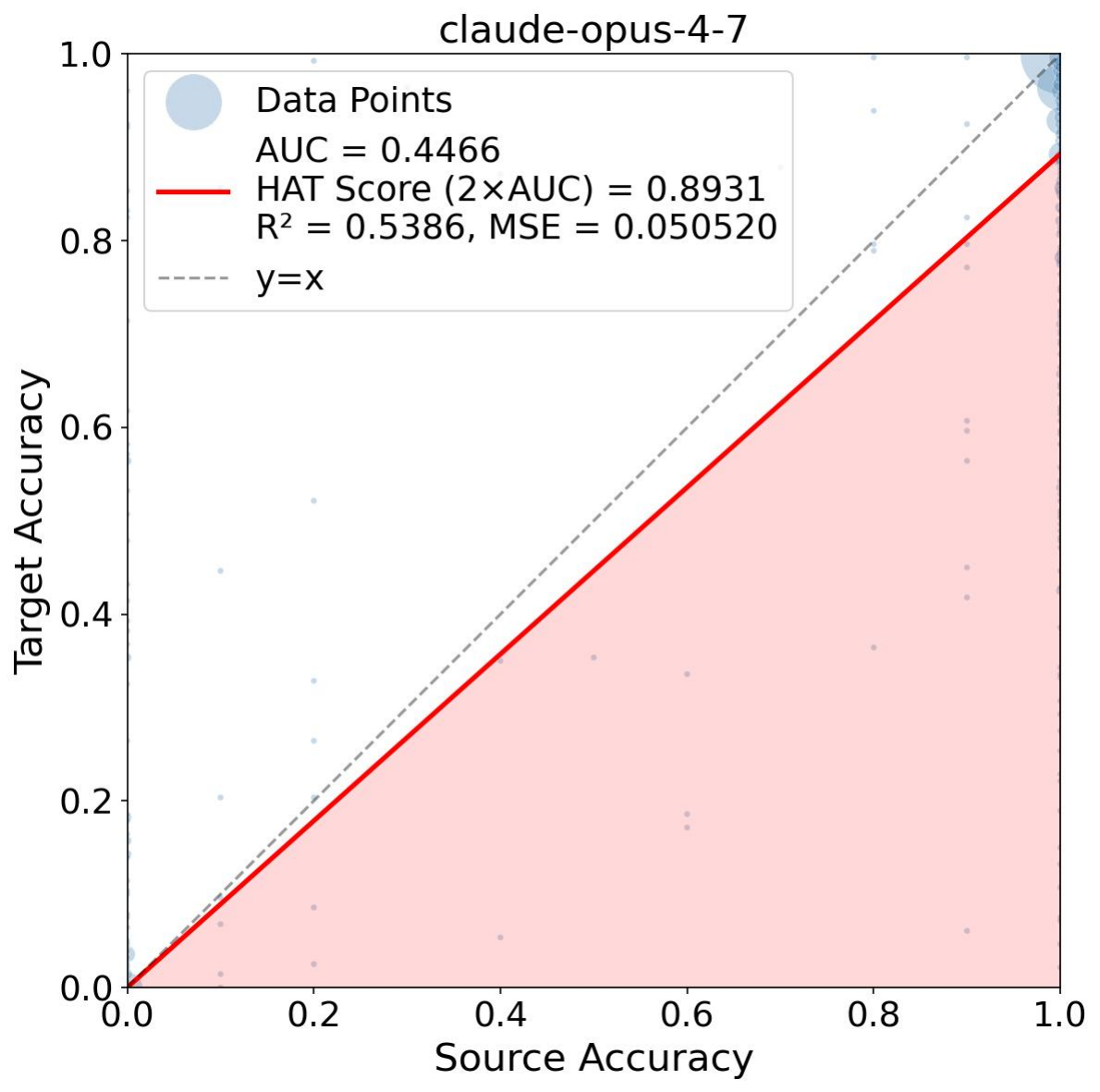} \\
\caption{HAT profiles for MMLU-ProX-Lite.}
\end{longtable}
\twocolumn
\clearpage

\begin{table*}[htbp]
    \centering
    \resizebox{0.6\textwidth}{!}{ 
    \begin{tabular}{l cccc}
        \toprule
        \multirow{2}{*}{\textbf{Model}} & \multicolumn{2}{c}{\textbf{ECLeKTic Transfer Score}} & \multicolumn{2}{c}{\textbf{HAT Score}} \\
        \cmidrule(lr){2-3} \cmidrule(lr){4-5}
        & Estimate & 95\% CI & Estimate & 95\% CI \\
        \midrule
        
        \textsc{Gemini-2.5-Flash}               & 66.0 & [63.6, 68.4] & 81.7 & [78.6, 84.7] \\
        \begin{tabular}{@{}l@{}}\textsc{Gemini-2.5-Flash} \\ \textsc{Thinking-Off}\end{tabular} & 44.9 & [42.5, 47.4] & 60.4 & [55.9, 64.8] \\
        \addlinespace
        \midrule
        
        \textsc{Gemini-3-Flash}                 & 78.0 & [76.5, 79.6] & 84.6 & [82.1, 87.1] \\
        \textsc{Gemini-3-Flash Low}             & 76.2 & [74.5, 77.9] & 84.0 & [81.6, 86.7] \\
        \textsc{Gemini-3-Flash Minimal}         & 68.0 & [66.1, 69.7] & 74.2 & [70.9, 77.5] \\
        \addlinespace
        \midrule
        
        \textsc{Gemma-3-1B}                     & 32.1 & [26.3, 38.0] & 48.2 & [34.9, 61.0] \\
        \textsc{Gemma-3-4B}                     & 45.8 & [41.4, 50.6] & 56.4 & [47.0, 65.9] \\
        \textsc{Gemma-3-12B}                    & 56.2 & [51.8, 60.5] & 64.1 & [56.8, 71.4] \\
        \textsc{Gemma-3-27B}                    & 58.8 & [55.1, 62.4] & 63.8 & [58.4, 69.6] \\
        \addlinespace
        \midrule
        
        \textsc{Qwen-3-4B}                      & 38.8 & [34.3, 43.1] & 56.5 & [47.8, 64.9] \\
        \textsc{Qwen-3-30B-A3B}                 & 48.9 & [45.3, 52.8] & 65.9 & [59.7, 72.6] \\
        \addlinespace
        \midrule
        
        \textsc{GPT-OSS-20B}                    & 53.5 & [49.4, 57.5] & 80.1 & [74.7, 85.3] \\
        \textsc{GPT-OSS-120B}                   & 61.0 & [58.0, 63.9] & 82.2 & [77.9, 86.1] \\
        \addlinespace
        \midrule
        
        \textsc{Gemma-4-E2B-IT}                 & 39.0 & [35.0, 43.2] & 51.8 & [42.4, 61.3] \\
        \textsc{Gemma-4-E4B-IT}                 & 43.9 & [40.0, 48.1] & 58.4 & [51.2, 65.7] \\
        \textsc{Gemma-4-26B-A4B-IT}             & 55.2 & [51.8, 58.8] & 63.9 & [57.9, 69.6] \\
        \textsc{Gemma-4-31B-IT}                 & 55.3 & [51.9, 58.6] & 68.6 & [62.9, 74.1] \\
        \addlinespace
        \midrule
        
        \textsc{Claude-Haiku-4.5}               & 51.1 & [48.2, 54.3] & 59.8 & [53.8, 65.7] \\
        \textsc{Claude-Sonnet-4.6}              & 70.0 & [67.9, 72.2] & 74.7 & [70.9, 78.5] \\
        \textsc{Claude-Opus-4.7}                & 67.0 & [64.9, 69.1] & 73.5 & [69.7, 77.2] \\
        
        \bottomrule
    \end{tabular}
    }
    \caption{ECLeKTic Transfer and HAT Scores across models.}
    \label{tab:combined-scores}
\end{table*}
\subsection{Empirical divergence of HAT score from transfer score}
\label{sec:comparison-eclektic}
\begin{figure}
    \centering
    \includegraphics[width=0.9\linewidth]{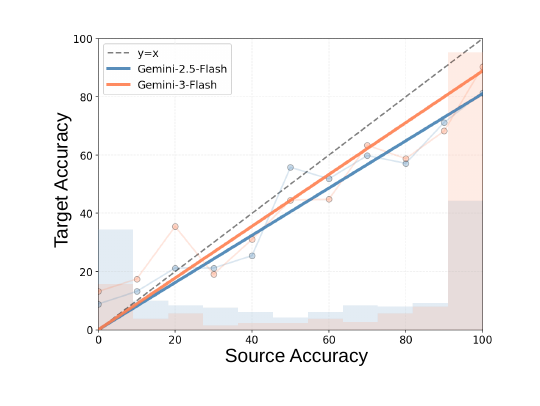}
    \caption{Gemini-2.5-Flash and Gemini-3-Flash have similar HAT Transfer profiles but their ECLeKTic transfer scores show significant improvements in XLT: 66\% to 78\%, likely due to confounding by improvements in source.}
    \label{fig:gemini_transfer_profile}
\end{figure}

Table~\ref{tab:combined-scores} contrasts transfer score and HAT Score side-by-side. The metrics are defined in more detail in Section~\ref{sec:proposed_xlt_metric}. We observe that the transfer score increases from 66\% for \textsc{Gemini-2.5-Flash} to 78\% for \textsc{Gemini-3-Flash}. Notwithstanding the large improvements to transfer suggested by the transfer score between the models, the transfer profiles of the two models shown in Figure~\ref{fig:gemini_transfer_profile} are nearly overlapping, which is better reflected by comparable HAT scores of 81.7 for \textsc{2.5-Flash} and 84.6 for \textsc{3-Flash}.

\subsection{Average thinking token counts}
\label{sec:thought_lens}
Table~\ref{tab:avg_tokens_per_split} presents the average thinking token counts for source and target splits across \eclektic, \newmgsm{} and \mmlu{} datasets for the Gemini models used in our work. We make the following observations: (1) Models have slightly longer thinking traces in target languages (15-20\% longer thinking on \eclektic{}), (2) Longer thinking des not trivially translate to better transfer; \textsc{Gemini-2.5-Flash} transfers slightly worse than \textsc{Gemini-3-Flash} on \mmlu{} despite twice the thinking trace length in the former.

\begin{table*}[h!]
    \centering
    
    \resizebox{0.75\textwidth}{!}{ 
    \begin{tabular}{ p{3.5cm} p{5cm} p{2.5cm} p{4cm} p{2cm} p{2cm}}
        \toprule
        \textbf{Dataset} & \textbf{Model} & \textbf{Split} & \textbf{Avg. Thinking Tokens} & \textbf{Proportion} & \textbf{English Fraction} \\
        \midrule
        
        \multirow{4}{*}{\textbf{\eclektic}} 
        & \textsc{Gemini-2.5-Flash} & source     & 276.1   & -     & 0.952 \\
        & \textsc{Gemini-2.5-Flash} & target     & 314.9   & 1.15x & 0.965 \\
        & \textsc{Gemini-3-Flash}   & source     & 264.5   & -     & 0.951 \\
        & \textsc{Gemini-3-Flash}   & target     & 315.9   & 1.19x & 0.959 \\
        \midrule
        
        \multirow{4}{*}{\textbf{MGSM2}} 
        & \textsc{Gemini-2.5-Flash} & source     & 269.0   & -     & 0.999 \\
        & \textsc{Gemini-2.5-Flash} & target     & 273.3   & 1.01x & 0.992 \\
        & \textsc{Gemini-3-Flash}   & source     & 283.2   & -     & 0.999 \\
        & \textsc{Gemini-3-Flash}   & target     & 298.0   & 1.05x & 0.990 \\
        \midrule
        
        \multirow{4}{*}{\textbf{MMLU-ProX-Lite}} 
        & \textsc{Gemini-2.5-Flash} & source     & 1,118.4 & -     & 0.991 \\
        & \textsc{Gemini-2.5-Flash} & target     & 1,258.2 & 1.13x & 0.970 \\
        & \textsc{Gemini-3-Flash}   & source     & 689.3   & -     & 0.989 \\
        & \textsc{Gemini-3-Flash}   & target     & 675.6   & 0.98x & 0.980 \\
        
        \bottomrule
    \end{tabular}
    }
    \caption{Average token counts and fraction of English tokens for source and target languages across \eclektic, \newmgsm{} and \mmlu{} datasets using Gemini-2.5-Flash and Gemini-3-Flash.}
    \label{tab:avg_tokens_per_split}
\end{table*}

%% file: main.bib
@misc{xuan2025mmluproxmultilingualbenchmarkadvanced,
      title={MMLU-ProX: A Multilingual Benchmark for Advanced Large Language Model Evaluation}, 
      author={Weihao Xuan and Rui Yang and Heli Qi and Qingcheng Zeng and Yunze Xiao and Aosong Feng and Dairui Liu and Yun Xing and Junjue Wang and Fan Gao and Jinghui Lu and Yuang Jiang and Huitao Li and Xin Li and Kunyu Yu and Ruihai Dong and Shangding Gu and Yuekang Li and Xiaofei Xie and Felix Juefei-Xu and Foutse Khomh and Osamu Yoshie and Qingyu Chen and Douglas Teodoro and Nan Liu and Randy Goebel and Lei Ma and Edison Marrese-Taylor and Shijian Lu and Yusuke Iwasawa and Yutaka Matsuo and Irene Li},
      year={2025},
      eprint={2503.10497},
      archivePrefix={arXiv},
      primaryClass={cs.CL},
      url={https://arxiv.org/abs/2503.10497}, 
}

@misc{miller2021accuracy,
      title={Accuracy on the Line: On the Strong Correlation Between Out-of-Distribution and In-Distribution Generalization}, 
      author={John Miller and Rohan Taori and Aditi Raghunathan and Shiori Sagawa and Pang Wei Koh and Vaishaal Shankar and Percy Liang and Yair Carmon and Ludwig Schmidt},
      year={2021},
      eprint={2107.04649},
      archivePrefix={arXiv},
      primaryClass={cs.LG},
      url={https://arxiv.org/abs/2107.04649}, 
}

@misc{singh2025openai,
      title={OpenAI GPT-5 System Card}, 
      author={Aaditya Singh and Adam Fry and Adam Perelman and Adam Tart and Adi Ganesh and Ahmed El-Kishky and Aidan McLaughlin and Aiden Low and AJ Ostrow and Akhila Ananthram and Akshay Nathan and Alan Luo and Alec Helyar and Aleksander Madry and Aleksandr Efremov and Aleksandra Spyra and Alex Baker-Whitcomb and Alex Beutel and Alex Karpenko and Alex Makelov and Alex Neitz and Alex Wei and Alexandra Barr and Alexandre Kirchmeyer and Alexey Ivanov and Alexi Christakis and Alistair Gillespie and Allison Tam and Ally Bennett and Alvin Wan and Alyssa Huang and Amy McDonald Sandjideh and Amy Yang and Ananya Kumar and Andre Saraiva and Andrea Vallone and Andrei Gheorghe and Andres Garcia Garcia and Andrew Braunstein and Andrew Liu and Andrew Schmidt and Andrey Mereskin and Andrey Mishchenko and Andy Applebaum and Andy Rogerson and Ann Rajan and Annie Wei and Anoop Kotha and Anubha Srivastava and Anushree Agrawal and Arun Vijayvergiya and Ashley Tyra and Ashvin Nair and Avi Nayak and Ben Eggers and Bessie Ji and Beth Hoover and Bill Chen and Blair Chen and Boaz Barak and Borys Minaiev and Botao Hao and Bowen Baker and Brad Lightcap and Brandon McKinzie and Brandon Wang and Brendan Quinn and Brian Fioca and Brian Hsu and Brian Yang and Brian Yu and Brian Zhang and Brittany Brenner and Callie Riggins Zetino and Cameron Raymond and Camillo Lugaresi and Carolina Paz and Cary Hudson and Cedric Whitney and Chak Li and Charles Chen and Charlotte Cole and Chelsea Voss and Chen Ding and Chen Shen and Chengdu Huang and Chris Colby and Chris Hallacy and Chris Koch and Chris Lu and Christina Kaplan and Christina Kim and CJ Minott-Henriques and Cliff Frey and Cody Yu and Coley Czarnecki and Colin Reid and Colin Wei and Cory Decareaux and Cristina Scheau and Cyril Zhang and Cyrus Forbes and Da Tang and Dakota Goldberg and Dan Roberts and Dana Palmie and Daniel Kappler and Daniel Levine and Daniel Wright and Dave Leo and David Lin and David Robinson and Declan Grabb and Derek Chen and Derek Lim and Derek Salama and Dibya Bhattacharjee and Dimitris Tsipras and Dinghua Li and Dingli Yu and DJ Strouse and Drew Williams and Dylan Hunn and Ed Bayes and Edwin Arbus and Ekin Akyurek and Elaine Ya Le and Elana Widmann and Eli Yani and Elizabeth Proehl and Enis Sert and Enoch Cheung and Eri Schwartz and Eric Han and Eric Jiang and Eric Mitchell and Eric Sigler and Eric Wallace and Erik Ritter and Erin Kavanaugh and Evan Mays and Evgenii Nikishin and Fangyuan Li and Felipe Petroski Such and Filipe de Avila Belbute Peres and Filippo Raso and Florent Bekerman and Foivos Tsimpourlas and Fotis Chantzis and Francis Song and Francis Zhang and Gaby Raila and Garrett McGrath and Gary Briggs and Gary Yang and Giambattista Parascandolo and Gildas Chabot and Grace Kim and Grace Zhao and Gregory Valiant and Guillaume Leclerc and Hadi Salman and Hanson Wang and Hao Sheng and Haoming Jiang and Haoyu Wang and Haozhun Jin and Harshit Sikchi and Heather Schmidt and Henry Aspegren and Honglin Chen and Huida Qiu and Hunter Lightman and Ian Covert and Ian Kivlichan and Ian Silber and Ian Sohl and Ibrahim Hammoud and Ignasi Clavera and Ikai Lan and Ilge Akkaya and Ilya Kostrikov and Irina Kofman and Isak Etinger and Ishaan Singal and Jackie Hehir and Jacob Huh and Jacqueline Pan and Jake Wilczynski and Jakub Pachocki and James Lee and James Quinn and Jamie Kiros and Janvi Kalra and Jasmyn Samaroo and Jason Wang and Jason Wolfe and Jay Chen and Jay Wang and Jean Harb and Jeffrey Han and Jeffrey Wang and Jennifer Zhao and Jeremy Chen and Jerene Yang and Jerry Tworek and Jesse Chand and Jessica Landon and Jessica Liang and Ji Lin and Jiancheng Liu and Jianfeng Wang and Jie Tang and Jihan Yin and Joanne Jang and Joel Morris and Joey Flynn and Johannes Ferstad and Johannes Heidecke and John Fishbein and John Hallman and Jonah Grant and Jonathan Chien and Jonathan Gordon and Jongsoo Park and Jordan Liss and Jos Kraaijeveld and Joseph Guay and Joseph Mo and Josh Lawson and Josh McGrath and Joshua Vendrow and Joy Jiao and Julian Lee and Julie Steele and Julie Wang and Junhua Mao and Kai Chen and Kai Hayashi and Kai Xiao and Kamyar Salahi and Kan Wu and Karan Sekhri and Karan Sharma and Karan Singhal and Karen Li and Kenny Nguyen and Keren Gu-Lemberg and Kevin King and Kevin Liu and Kevin Stone and Kevin Yu and Kristen Ying and Kristian Georgiev and Kristie Lim and Kushal Tirumala and Kyle Miller and Lama Ahmad and Larry Lv and Laura Clare and Laurance Fauconnet and Lauren Itow and Lauren Yang and Laurentia Romaniuk and Leah Anise and Lee Byron and Leher Pathak and Leon Maksin and Leyan Lo and Leyton Ho and Li Jing and Liang Wu and Liang Xiong and Lien Mamitsuka and Lin Yang and Lindsay McCallum and Lindsey Held and Liz Bourgeois and Logan Engstrom and Lorenz Kuhn and Louis Feuvrier and Lu Zhang and Lucas Switzer and Lukas Kondraciuk and Lukasz Kaiser and Manas Joglekar and Mandeep Singh and Mandip Shah and Manuka Stratta and Marcus Williams and Mark Chen and Mark Sun and Marselus Cayton and Martin Li and Marvin Zhang and Marwan Aljubeh and Matt Nichols and Matthew Haines and Max Schwarzer and Mayank Gupta and Meghan Shah and Melody Y. Guan and Melody Huang and Meng Dong and Mengqing Wang and Mia Glaese and Micah Carroll and Michael Lampe and Michael Malek and Michael Sharman and Michael Zhang and Michele Wang and Michelle Pokrass and Mihai Florian and Mikhail Pavlov and Miles Wang and Ming Chen and Mingxuan Wang and Minnia Feng and Mo Bavarian and Molly Lin and Moose Abdool and Mostafa Rohaninejad and Nacho Soto and Natalie Staudacher and Natan LaFontaine and Nathan Marwell and Nelson Liu and Nick Preston and Nick Turley and Nicklas Ansman and Nicole Blades and Nikil Pancha and Nikita Mikhaylin and Niko Felix and Nikunj Handa and Nishant Rai and Nitish Keskar and Noam Brown and Ofir Nachum and Oleg Boiko and Oleg Murk and Olivia Watkins and Oona Gleeson and Pamela Mishkin and Patryk Lesiewicz and Paul Baltescu and Pavel Belov and Peter Zhokhov and Philip Pronin and Phillip Guo and Phoebe Thacker and Qi Liu and Qiming Yuan and Qinghua Liu and Rachel Dias and Rachel Puckett and Rahul Arora and Ravi Teja Mullapudi and Raz Gaon and Reah Miyara and Rennie Song and Rishabh Aggarwal and RJ Marsan and Robel Yemiru and Robert Xiong and Rohan Kshirsagar and Rohan Nuttall and Roman Tsiupa and Ronen Eldan and Rose Wang and Roshan James and Roy Ziv and Rui Shu and Ruslan Nigmatullin and Saachi Jain and Saam Talaie and Sam Altman and Sam Arnesen and Sam Toizer and Sam Toyer and Samuel Miserendino and Sandhini Agarwal and Sarah Yoo and Savannah Heon and Scott Ethersmith and Sean Grove and Sean Taylor and Sebastien Bubeck and Sever Banesiu and Shaokyi Amdo and Shengjia Zhao and Sherwin Wu and Shibani Santurkar and Shiyu Zhao and Shraman Ray Chaudhuri and Shreyas Krishnaswamy and Shuaiqi and Xia and Shuyang Cheng and Shyamal Anadkat and Simón Posada Fishman and Simon Tobin and Siyuan Fu and Somay Jain and Song Mei and Sonya Egoian and Spencer Kim and Spug Golden and SQ Mah and Steph Lin and Stephen Imm and Steve Sharpe and Steve Yadlowsky and Sulman Choudhry and Sungwon Eum and Suvansh Sanjeev and Tabarak Khan and Tal Stramer and Tao Wang and Tao Xin and Tarun Gogineni and Taya Christianson and Ted Sanders and Tejal Patwardhan and Thomas Degry and Thomas Shadwell and Tianfu Fu and Tianshi Gao and Timur Garipov and Tina Sriskandarajah and Toki Sherbakov and Tomek Korbak and Tomer Kaftan and Tomo Hiratsuka and Tongzhou Wang and Tony Song and Tony Zhao and Troy Peterson and Val Kharitonov and Victoria Chernova and Vineet Kosaraju and Vishal Kuo and Vitchyr Pong and Vivek Verma and Vlad Petrov and Wanning Jiang and Weixing Zhang and Wenda Zhou and Wenlei Xie and Wenting Zhan and Wes McCabe and Will DePue and Will Ellsworth and Wulfie Bain and Wyatt Thompson and Xiangning Chen and Xiangyu Qi and Xin Xiang and Xinwei Shi and Yann Dubois and Yaodong Yu and Yara Khakbaz and Yifan Wu and Yilei Qian and Yin Tat Lee and Yinbo Chen and Yizhen Zhang and Yizhong Xiong and Yonglong Tian and Young Cha and Yu Bai and Yu Yang and Yuan Yuan and Yuanzhi Li and Yufeng Zhang and Yuguang Yang and Yujia Jin and Yun Jiang and Yunyun Wang and Yushi Wang and Yutian Liu and Zach Stubenvoll and Zehao Dou and Zheng Wu and Zhigang Wang},
      year={2026},
      eprint={2601.03267},
      archivePrefix={arXiv},
      primaryClass={cs.CL},
      url={https://arxiv.org/abs/2601.03267}, 
}

@report{cohere2024languagegap,
  title      = {The AI Language Gap},
  author     = {{Cohere Labs}},
  institution= {Cohere Labs},
  year       = {2024},
  url        = {https://cohere.com/research/papers/the-ai-language-gap.pdf}
}

@misc{hu2025quantifying,
      title={Quantifying Language Disparities in Multilingual Large Language Models}, 
      author={Songbo Hu and Ivan Vulić and Anna Korhonen},
      year={2025},
      eprint={2508.17162},
      archivePrefix={arXiv},
      primaryClass={cs.CL},
      url={https://arxiv.org/abs/2508.17162}, 
}

@inproceedings{pires-etal-2019-multilingual,
    title = "How Multilingual is Multilingual {BERT}?",
    author = "Pires, Telmo  and
      Schlinger, Eva  and
      Garrette, Dan",
    editor = "Korhonen, Anna  and
      Traum, David  and
      M{\`a}rquez, Llu{\'i}s",
    booktitle = "Proceedings of the 57th Annual Meeting of the Association for Computational Linguistics",
    month = jul,
    year = "2019",
    address = "Florence, Italy",
    publisher = "Association for Computational Linguistics",
    url = "https://aclanthology.org/P19-1493/",
    doi = "10.18653/v1/P19-1493",
    pages = "4996--5001"
}

@inproceedings{philippy-etal-2023-towards,
    title = "Towards a Common Understanding of Contributing Factors for Cross-Lingual Transfer in Multilingual Language Models: A Review",
    author = "Philippy, Fred  and
      Guo, Siwen  and
      Haddadan, Shohreh",
    editor = "Rogers, Anna  and
      Boyd-Graber, Jordan  and
      Okazaki, Naoaki",
    booktitle = "Proceedings of the 61st Annual Meeting of the Association for Computational Linguistics (Volume 1: Long Papers)",
    month = jul,
    year = "2023",
    address = "Toronto, Canada",
    publisher = "Association for Computational Linguistics",
    url = "https://aclanthology.org/2023.acl-long.323/",
    doi = "10.18653/v1/2023.acl-long.323",
    pages = "5877--5891"
}

@misc{salaudeen2024imagenot,
      title={ImageNot: A contrast with ImageNet preserves model rankings}, 
      author={Olawale Salaudeen and Moritz Hardt},
      year={2025},
      eprint={2404.02112},
      archivePrefix={arXiv},
      primaryClass={cs.LG},
      url={https://arxiv.org/abs/2404.02112}, 
}

@inproceedings{objectnet,
 author = {Barbu, Andrei and Mayo, David and Alverio, Julian and Luo, William and Wang, Christopher and Gutfreund, Dan and Tenenbaum, Josh and Katz, Boris},
 booktitle = {Advances in Neural Information Processing Systems},
 editor = {H. Wallach and H. Larochelle and A. Beygelzimer and F. d\textquotesingle Alch\'{e}-Buc and E. Fox and R. Garnett},
 pages = {},
 publisher = {Curran Associates, Inc.},
 title = {ObjectNet: A large-scale bias-controlled dataset for pushing the limits of object recognition models},
 url = {https://proceedings.neurips.cc/paper_files/paper/2019/file/97af07a14cacba681feacf3012730892-Paper.pdf},
 volume = {32},
 year = {2019}
}

@misc{piratla2025rethinking,
      title={Rethinking Cross-lingual Gaps from a Statistical Viewpoint}, 
      author={Vihari Piratla and Purvam Jain and Darshan Singh and Partha Talukdar and Trevor Cohn},
      year={2025},
      eprint={2510.15551},
      archivePrefix={arXiv},
      primaryClass={cs.CL},
      url={https://arxiv.org/abs/2510.15551}, 
}

@misc{goldman2025eclekticnovelchallengeset,
      title={ECLeKTic: a Novel Challenge Set for Evaluation of Cross-Lingual Knowledge Transfer}, 
      author={Omer Goldman and Uri Shaham and Dan Malkin and Sivan Eiger and Avinatan Hassidim and Yossi Matias and Joshua Maynez and Adi Mayrav Gilady and Jason Riesa and Shruti Rijhwani and Laura Rimell and Idan Szpektor and Reut Tsarfaty and Matan Eyal},
      year={2025},
      eprint={2502.21228},
      archivePrefix={arXiv},
      primaryClass={cs.CL},
      url={https://arxiv.org/abs/2502.21228}, 
}

@article{bootstrap_efron,
 ISSN = {00905364, 21688966},
 URL = {http://www.jstor.org/stable/2958830},
 author = {B. Efron},
 journal = {The Annals of Statistics},
 number = {1},
 pages = {1--26},
 publisher = {Institute of Mathematical Statistics},
 title = {Bootstrap Methods: Another Look at the Jackknife},
 urldate = {2026-05-19},
 volume = {7},
 year = {1979}
}

@misc{gemmateam2025gemma3technicalreport,
      title={Gemma 3 Technical Report}, 
      author={Gemma Team and Aishwarya Kamath and Johan Ferret and Shreya Pathak and Nino Vieillard and Ramona Merhej and Sarah Perrin and Tatiana Matejovicova and Alexandre Ramé and Morgane Rivière and Louis Rouillard and Thomas Mesnard and Geoffrey Cideron and Jean-bastien Grill and Sabela Ramos and Edouard Yvinec and Michelle Casbon and Etienne Pot and Ivo Penchev and Gaël Liu and Francesco Visin and Kathleen Kenealy and Lucas Beyer and Xiaohai Zhai and Anton Tsitsulin and Robert Busa-Fekete and Alex Feng and Noveen Sachdeva and Benjamin Coleman and Yi Gao and Basil Mustafa and Iain Barr and Emilio Parisotto and David Tian and Matan Eyal and Colin Cherry and Jan-Thorsten Peter and Danila Sinopalnikov and Surya Bhupatiraju and Rishabh Agarwal and Mehran Kazemi and Dan Malkin and Ravin Kumar and David Vilar and Idan Brusilovsky and Jiaming Luo and Andreas Steiner and Abe Friesen and Abhanshu Sharma and Abheesht Sharma and Adi Mayrav Gilady and Adrian Goedeckemeyer and Alaa Saade and Alex Feng and Alexander Kolesnikov and Alexei Bendebury and Alvin Abdagic and Amit Vadi and András György and André Susano Pinto and Anil Das and Ankur Bapna and Antoine Miech and Antoine Yang and Antonia Paterson and Ashish Shenoy and Ayan Chakrabarti and Bilal Piot and Bo Wu and Bobak Shahriari and Bryce Petrini and Charlie Chen and Charline Le Lan and Christopher A. Choquette-Choo and CJ Carey and Cormac Brick and Daniel Deutsch and Danielle Eisenbud and Dee Cattle and Derek Cheng and Dimitris Paparas and Divyashree Shivakumar Sreepathihalli and Doug Reid and Dustin Tran and Dustin Zelle and Eric Noland and Erwin Huizenga and Eugene Kharitonov and Frederick Liu and Gagik Amirkhanyan and Glenn Cameron and Hadi Hashemi and Hanna Klimczak-Plucińska and Harman Singh and Harsh Mehta and Harshal Tushar Lehri and Hussein Hazimeh and Ian Ballantyne and Idan Szpektor and Ivan Nardini and Jean Pouget-Abadie and Jetha Chan and Joe Stanton and John Wieting and Jonathan Lai and Jordi Orbay and Joseph Fernandez and Josh Newlan and Ju-yeong Ji and Jyotinder Singh and Kat Black and Kathy Yu and Kevin Hui and Kiran Vodrahalli and Klaus Greff and Linhai Qiu and Marcella Valentine and Marina Coelho and Marvin Ritter and Matt Hoffman and Matthew Watson and Mayank Chaturvedi and Michael Moynihan and Min Ma and Nabila Babar and Natasha Noy and Nathan Byrd and Nick Roy and Nikola Momchev and Nilay Chauhan and Noveen Sachdeva and Oskar Bunyan and Pankil Botarda and Paul Caron and Paul Kishan Rubenstein and Phil Culliton and Philipp Schmid and Pier Giuseppe Sessa and Pingmei Xu and Piotr Stanczyk and Pouya Tafti and Rakesh Shivanna and Renjie Wu and Renke Pan and Reza Rokni and Rob Willoughby and Rohith Vallu and Ryan Mullins and Sammy Jerome and Sara Smoot and Sertan Girgin and Shariq Iqbal and Shashir Reddy and Shruti Sheth and Siim Põder and Sijal Bhatnagar and Sindhu Raghuram Panyam and Sivan Eiger and Susan Zhang and Tianqi Liu and Trevor Yacovone and Tyler Liechty and Uday Kalra and Utku Evci and Vedant Misra and Vincent Roseberry and Vlad Feinberg and Vlad Kolesnikov and Woohyun Han and Woosuk Kwon and Xi Chen and Yinlam Chow and Yuvein Zhu and Zichuan Wei and Zoltan Egyed and Victor Cotruta and Minh Giang and Phoebe Kirk and Anand Rao and Kat Black and Nabila Babar and Jessica Lo and Erica Moreira and Luiz Gustavo Martins and Omar Sanseviero and Lucas Gonzalez and Zach Gleicher and Tris Warkentin and Vahab Mirrokni and Evan Senter and Eli Collins and Joelle Barral and Zoubin Ghahramani and Raia Hadsell and Yossi Matias and D. Sculley and Slav Petrov and Noah Fiedel and Noam Shazeer and Oriol Vinyals and Jeff Dean and Demis Hassabis and Koray Kavukcuoglu and Clement Farabet and Elena Buchatskaya and Jean-Baptiste Alayrac and Rohan Anil and Dmitry and Lepikhin and Sebastian Borgeaud and Olivier Bachem and Armand Joulin and Alek Andreev and Cassidy Hardin and Robert Dadashi and Léonard Hussenot},
      year={2025},
      eprint={2503.19786},
      archivePrefix={arXiv},
      primaryClass={cs.CL},
      url={https://arxiv.org/abs/2503.19786}, 
}

@software{gemma42026,
  author = {{Google DeepMind}},
  title = {Gemma 4: Open Models Family},
  year = {2026},
  url = {https://deepmind.google/models/gemma/gemma-4/},
  version = {4.0}
}

@misc{yang2025qwen3technicalreport,
      title={Qwen3 Technical Report}, 
      author={An Yang and Anfeng Li and Baosong Yang and Beichen Zhang and Binyuan Hui and Bo Zheng and Bowen Yu and Chang Gao and Chengen Huang and Chenxu Lv and Chujie Zheng and Dayiheng Liu and Fan Zhou and Fei Huang and Feng Hu and Hao Ge and Haoran Wei and Huan Lin and Jialong Tang and Jian Yang and Jianhong Tu and Jianwei Zhang and Jianxin Yang and Jiaxi Yang and Jing Zhou and Jingren Zhou and Junyang Lin and Kai Dang and Keqin Bao and Kexin Yang and Le Yu and Lianghao Deng and Mei Li and Mingfeng Xue and Mingze Li and Pei Zhang and Peng Wang and Qin Zhu and Rui Men and Ruize Gao and Shixuan Liu and Shuang Luo and Tianhao Li and Tianyi Tang and Wenbiao Yin and Xingzhang Ren and Xinyu Wang and Xinyu Zhang and Xuancheng Ren and Yang Fan and Yang Su and Yichang Zhang and Yinger Zhang and Yu Wan and Yuqiong Liu and Zekun Wang and Zeyu Cui and Zhenru Zhang and Zhipeng Zhou and Zihan Qiu},
      year={2025},
      eprint={2505.09388},
      archivePrefix={arXiv},
      primaryClass={cs.CL},
      url={https://arxiv.org/abs/2505.09388}, 
}

@misc{openai2025gptoss120bgptoss20bmodel,
      title={gpt-oss-120b and gpt-oss-20b Model Card}, 
      author={OpenAI and : and Sandhini Agarwal and Lama Ahmad and Jason Ai and Sam Altman and Andy Applebaum and Edwin Arbus and Rahul K. Arora and Yu Bai and Bowen Baker and Haiming Bao and Boaz Barak and Ally Bennett and Tyler Bertao and Nivedita Brett and Eugene Brevdo and Greg Brockman and Sebastien Bubeck and Che Chang and Kai Chen and Mark Chen and Enoch and others},
      year={2025},
      eprint={2508.10925},
      archivePrefix={arXiv},
      primaryClass={cs.CL},
      url={https://arxiv.org/abs/2508.10925}, 
}

@misc{comanici2025gemini,
      title={Gemini 2.5: Pushing the Frontier with Advanced Reasoning, Multimodality, Long Context, and Next Generation Agentic Capabilities}, 
      author={Gheorghe Comanici and Eric Bieber and Mike Schaekermann and Ice Pasupat and Noveen Sachdeva and Inderjit Dhillon and Marcel Blistein and Ori Ram and Dan Zhang and Evan Rosen and Luke Marris and Sam Petulla and Colin Gaffney and Asaf Aharoni and Nathan Lintz and Tiago Cardal Pais and Henrik Jacobsson and Idan Szpektor and Nan-Jiang Jiang and Krishna Haridasan and Ahmed Omran and Nikunj Saunshi and Dara Bahri and Gaurav Mishra and Eric Chu and Toby Boyd and Brad Hekman and Aaron Parisi and Chaoyi Zhang and Kornraphop Kawintiranon and Tania Bedrax-Weiss and Oliver Wang and Ya Xu and Ollie Purkiss and Uri Mendlovic and Ilaï Deutel and Nam Nguyen and Adam Langley and Flip Korn and Lucia Rossazza and Alexandre Ramé and Sagar Waghmare and Helen Miller and Nathan Byrd and Ashrith Sheshan and Raia Hadsell and Sangnie Bhardwaj and Pawel Janus and Tero Rissa and Dan Horgan and Alvin Abdagic and Lior Belenki and James Allingham and Anima Singh and Theo Guidroz and Srivatsan Srinivasan and Herman Schmit and Kristen Chiafullo and Andre Elisseeff and Nilpa Jha and Prateek Kolhar and Leonard Berrada and Frank Ding and Xiance Si and Shrestha Basu Mallick and Franz Och and Sofia Erell and Eric Ni and Tejasi Latkar and Sherry Yang and Petar Sirkovic and Ziqiang Feng and Robert Leland and Rachel Hornung and Gang Wu and Charles Blundell and Hamidreza Alvari and Po-Sen Huang and Cathy Yip and Sanja Deur and Li Liu and Gabriela Surita and Pablo Duque and Dima Damen and Johnson Jia and Arthur Guez and Markus Mircea and Animesh Sinha and Alberto Magni and Paweł Stradomski and Tal Marian and Vlado Galić and Wenhu Chen and Hisham Husain and Achintya Singhal and Dominik Grewe and François-Xavier Aubet and Shuang Song and Lorenzo Blanco and Leland Rechis and Lewis Ho and Rich Munoz and Kelvin Zheng and Jessica Hamrick and Kevin Mather and Hagai Taitelbaum and Eliza Rutherford and Yun Lei and Kuangyuan Chen and Anand Shukla and Erica Moreira and Eric Doi and Berivan Isik and Nir Shabat and Dominika Rogozińska and Kashyap Kolipaka and Jason Chang and Eugen Vušak and Srinivasan Venkatachary and Shadi Noghabi and Tarun Bharti and Younghoon Jun and Aleksandr Zaks and Simon Green and Jeshwanth Challagundla and William Wong and Muqthar Mohammad and Dean Hirsch and Yong Cheng and Iftekhar Naim and Lev Proleev and Damien Vincent and Aayush Singh and Maxim Krikun and Dilip Krishnan and Zoubin Ghahramani and Aviel Atias and Rajeev Aggarwal and Christo Kirov and Dimitrios Vytiniotis and Christy Koh and Alexandra Chronopoulou and Pawan Dogra and Vlad-Doru Ion and Gladys Tyen and Jason Lee and Felix Weissenberger and Trevor Strohman and Ashwin Balakrishna and Jack Rae and Marko Velic and Raoul de Liedekerke and Oded Elyada and Wentao Yuan and Canoee Liu and Lior Shani and Sergey Kishchenko and Bea Alessio and Yandong Li and Richard Song and Sam Kwei and Orion Jankowski and Aneesh Pappu and Youhei Namiki and Yenai Ma and Nilesh Tripuraneni and Colin Cherry and Marissa Ikonomidis and Yu-Cheng Ling and Colin Ji and Beka Westberg and Auriel Wright and Da Yu and David Parkinson and Swaroop Ramaswamy and Jerome Connor and Soheil Hassas Yeganeh and Snchit Grover and George Kenwright and Lubo Litchev and Chris Apps and Alex Tomala and Felix Halim and Alex Castro-Ros and Zefei Li and Anudhyan Boral and Pauline Sho and Michal Yarom and Eric Malmi and David Klinghoffer and Rebecca Lin and Alan Ansell and Pradeep Kumar S and Shubin Zhao and Siqi Zuo and Adam Santoro and Heng-Tze Cheng and Solomon Demmessie and Yuchi Liu and Nicole Brichtova and Allie Culp and Nathaniel Braun and Dan Graur and Will Ng and Nikhil Mehta and Aaron Phillips and Patrik Sundberg and Varun Godbole and Fangyu Liu and Yash Katariya and David Rim and Mojtaba Seyedhosseini and Sean Ammirati and Jonas Valfridsson and Mahan Malihi and Timothy Knight and Andeep Toor and Thomas Lampe and Abe Ittycheriah and Lewis Chiang and Chak Yeung and Alexandre Fréchette and Jinmeng Rao and Huisheng Wang and Himanshu Srivastava and Richard Zhang and Rocky Rhodes and Ariel Brand and Dean Weesner and Ilya Figotin and Felix Gimeno and Rachana Fellinger and Pierre Marcenac and José Leal and Eyal Marcus and Victor Cotruta and Rodrigo Cabrera and Sheryl Luo and Dan Garrette and Vera Axelrod and Sorin Baltateanu and David Barker and Dongkai Chen and Horia Toma and Ben Ingram and Jason Riesa and Chinmay Kulkarni and Yujing Zhang and Hongbin Liu and Chao Wang and Martin Polacek and Will Wu and Kai Hui and Adrian N Reyes and Yi Su and Megan Barnes and Ishaan Malhi and Anfal Siddiqui and Qixuan Feng and Mihai Damaschin and Daniele Pighin and Andreas Steiner and Samuel Yang and Ramya Sree Boppana and Simeon Ivanov and Arun Kandoor and Aditya Shah and Asier Mujika and Da Huang and Christopher A. Choquette-Choo and Mohak Patel and Tianhe Yu and Toni Creswell and Jerry and Liu and Catarina Barros and Yasaman Razeghi and Aurko Roy and Phil Culliton and Binbin Xiong and Jiaqi Pan and Thomas Strohmann and Tolly Powell and Babi Seal and Doug DeCarlo and Pranav Shyam and Kaan Katircioglu and Xuezhi Wang and Cassidy Hardin and Immanuel Odisho and Josef Broder and Oscar Chang and Arun Nair and Artem Shtefan and Maura O'Brien and Manu Agarwal and Sahitya Potluri and Siddharth Goyal and Amit Jhindal and Saksham Thakur and Yury Stuken and James Lyon and Kristina Toutanova and Fangxiaoyu Feng and Austin Wu and Ben Horn and Alek Wang and Alex Cullum and Gabe Taubman and Disha Shrivastava and Chongyang Shi and Hamish Tomlinson and Roma Patel and Tao Tu and Ada Maksutaj Oflazer and Francesco Pongetti and Mingyao Yang and Adrien Ali Taïga and Vincent Perot and Nuo Wang Pierse and Feng Han and Yoel Drori and Iñaki Iturrate and Ayan Chakrabarti and Legg Yeung and Dave Dopson and Yi-ting Chen and Apoorv Kulshreshtha and Tongfei Guo and Philip Pham and Tal Schuster and Junquan Chen and Alex Polozov and Jinwei Xing and Huanjie Zhou and Praneeth Kacham and Doron Kukliansky and Antoine Miech and Sergey Yaroshenko and Ed Chi and Sholto Douglas and Hongliang Fei and Mathieu Blondel and Preethi Myla and Lior Madmoni and Xing Wu and Daniel Keysers and Kristian Kjems and Isabela Albuquerque and Lijun Yu and Joel D'sa and Michelle Plantan and Vlad Ionescu and Jaume Sanchez Elias and Abhirut Gupta and Manish Reddy Vuyyuru and Fred Alcober and Tong Zhou and Kaiyang Ji and Florian Hartmann and Subha Puttagunta and Hugo Song and Ehsan Amid and Anca Stefanoiu and Andrew Lee and Paul Pucciarelli and Emma Wang and Amit Raul and Slav Petrov and Isaac Tian and Valentin Anklin and Nana Nti and Victor Gomes and Max Schumacher and Grace Vesom and Alex Panagopoulos and Konstantinos Bousmalis and Daniel Andor and Josh Jacob and Yuan Zhang and Bill Rosgen and Matija Kecman and Matthew Tung and Alexandra Belias and Noah Goodman and Paul Covington and Brian Wieder and Nikita Saxena and Elnaz Davoodi and Muhuan Huang and Sharath Maddineni and Vincent Roulet and Folawiyo Campbell-Ajala and Pier Giuseppe Sessa and Xintian and Wu and Guangda Lai and Paul Collins and Alex Haig and Vytenis Sakenas and Xiaowei Xu and Marissa Giustina and Laurent El Shafey and Pichi Charoenpanit and Shefali Garg and Joshua Ainslie and Boone Severson and Montse Gonzalez Arenas and Shreya Pathak and Sujee Rajayogam and Jie Feng and Michiel Bakker and Sheng Li and Nevan Wichers and Jamie Rogers and Xinyang Geng and Yeqing Li and Rolf Jagerman and Chao Jia and Nadav Olmert and David Sharon and Matthew Mauger and Sandeep Mariserla and Hongxu Ma and Megha Mohabey and Kyuyeun Kim and Alek Andreev and Scott Pollom and Juliette Love and Vihan Jain and Priyanka Agrawal and Yannick Schroecker and Alisa Fortin and Manfred Warmuth and Ji Liu and Andrew Leach and Irina Blok and Ganesh Poomal Girirajan and Roee Aharoni and Benigno Uria and Andrei Sozanschi and Dan Goldberg and Lucian Ionita and Marco Tulio Ribeiro and Martin Zlocha and Vighnesh Birodkar and Sami Lachgar and Liangzhe Yuan and Himadri Choudhury and Matt Ginsberg and Fei Zheng and Gregory Dibb and Emily Graves and Swachhand Lokhande and Gabriel Rasskin and George-Cristian Muraru and Corbin Quick and Sandeep Tata and Pierre Sermanet and Aditya Chawla and Itay Karo and Yan Wang and Susan Zhang and Orgad Keller and Anca Dragan and Guolong Su and Ian Chou and Xi Liu and Yiqing Tao and Shruthi Prabhakara and Marc Wilson and Ruibo Liu and Shibo Wang and Georgie Evans and David Du and Alfonso Castaño and Gautam Prasad and Mona El Mahdy and Sebastian Gerlach and Machel Reid and Jarrod Kahn and Amir Zait and Thanumalayan Sankaranarayana Pillai and Thatcher Ulrich and Guanyu Wang and Jan Wassenberg and Efrat Farkash and Kiran Yalasangi and Congchao Wang and Maria Bauza and Simon Bucher and Ting Liu and Jun Yan and Gary Leung and Vikas Sindhwani and Parker Barnes and Avi Singh and Ivan Jurin and Jichuan Chang and Niket Kumar Bhumihar and Sivan Eiger and Gui Citovsky and Ben Withbroe and Zhang Li and Siyang Xue and Niccolò Dal Santo and Georgi Stoyanov and Yves Raimond and Steven Zheng and Yilin Gao and Vít Listík and Sławek Kwasiborski and Rachel Saputro and Adnan Ozturel and Ganesh Mallya and Kushal Majmundar and Ross West and Paul Caron and Jinliang Wei and Lluis Castrejon and Sharad Vikram and Deepak Ramachandran and Nikhil Dhawan and Jiho Park and Sara Smoot and George van den Driessche and Yochai Blau and Chase Malik and Wei Liang and Roy Hirsch and Cicero Nogueira dos Santos and Eugene Weinstein and Aäron van den Oord and Sid Lall and Nicholas FitzGerald and Zixuan Jiang and Xuan Yang and Dale Webster and Ali Elqursh and Aedan Pope and Georges Rotival and David Raposo and Wanzheng Zhu and Jeff Dean and Sami Alabed and Dustin Tran and Arushi Gupta and Zach Gleicher and Jessica Austin and Edouard Rosseel and Megh Umekar and Dipanjan Das and Yinghao Sun and Kai Chen and Karolis Misiunas and Xiang Zhou and Yixian Di and Alyssa Loo and Josh Newlan and Bo Li and Vinay Ramasesh and Ying Xu and Alex Chen and Sudeep Gandhe and Radu Soricut and Nikita Gupta and Shuguang Hu and Seliem El-Sayed and Xavier Garcia and Idan Brusilovsky and Pu-Chin Chen and Andrew Bolt and Lu Huang and Alex Gurney and Zhiying Zhang and Alexander Pritzel and Jarek Wilkiewicz and Bryan Seybold and Bhargav Kanagal Shamanna and Felix Fischer and Josef Dean and Karan Gill and Ross Mcilroy and Abhishek Bhowmick and Jeremy Selier and Antoine Yang and Derek Cheng and Vladimir Magay and Jie Tan and Dhriti Varma and Christian Walder and Tomas Kocisky and Ryo Nakashima and Paul Natsev and Mike Kwong and Ionel Gog and Chiyuan Zhang and Sander Dieleman and Thomas Jimma and Andrey Ryabtsev and Siddhartha Brahma and David Steiner and Dayou Du and Ante Žužul and Mislav Žanić and Mukund Raghavachari and Willi Gierke and Zeyu Zheng and Dessie Petrova and Yann Dauphin and Yuchuan Liu and Ido Kessler and Steven Hand and Chris Duvarney and Seokhwan Kim and Hyo Lee and Léonard Hussenot and Jeffrey Hui and Josh Smith and Deepali Jain and Jiawei Xia and Gaurav Singh Tomar and Keyvan Amiri and Du Phan and Fabian Fuchs and Tobias Weyand and Nenad Tomasev and Alexandra Cordell and Xin Liu and Jonathan Mallinson and Pankaj Joshi and Andy Crawford and Arun Suggala and Steve Chien and Nick Fernando and Mariella Sanchez-Vargas and Duncan Williams and Phil Crone and Xiyang Luo and Igor Karpov and Jyn Shan and Terry Thurk and Robin Strudel and Paul Voigtlaender and Piyush Patil and Tim Dozat and Ali Khodaei and Sahil Singla and Piotr Ambroszczyk and Qiyin Wu and Yifan Chang and Brian Roark and Chaitra Hegde and Tianli Ding and Angelos Filos and Zhongru Wu and André Susano Pinto and Shuang Liu and Saarthak Khanna and Aditya Pandey and Siobhan Mcloughlin and Qiujia Li and Sam Haves and Allan Zhou and Elena Buchatskaya and Isabel Leal and Peter de Boursac and Nami Akazawa and Nina Anderson and Terry Chen and Krishna Somandepalli and Chen Liang and Sheela Goenka and Stephanie Winkler and Alexander Grushetsky and Yifan Ding and Jamie Smith and Fan Ye and Jordi Pont-Tuset and Eric Li and Ruichao Li and Tomer Golany and Dawid Wegner and Tao Jiang and Omer Barak and Yuan Shangguan and Eszter Vértes and Renee Wong and Jörg Bornschein and Alex Tudor and Michele Bevilacqua and Tom Schaul and Ankit Singh Rawat and Yang Zhao and Kyriakos Axiotis and Lei Meng and Cory McLean and Jonathan Lai and Jennifer Beattie and Nate Kushman and Yaxin Liu and Blair Kutzman and Fiona Lang and Jingchen Ye and Praneeth Netrapalli and Pushkar Mishra and Myriam Khan and Megha Goel and Rob Willoughby and David Tian and Honglei Zhuang and JD Chen and Zak Tsai and Tasos Kementsietsidis and Arjun Khare and James Keeling and Keyang Xu and Nathan Waters and Florent Altché and Ashok Popat and Bhavishya Mittal and David Saxton and Dalia El Badawy and Michael Mathieu and Zheng Zheng and Hao Zhou and Nishant Ranka and Richard Shin and Qingnan Duan and Tim Salimans and Ioana Mihailescu and Uri Shaham and Ming-Wei Chang and Yannis Assael and Nishanth Dikkala and Martin Izzard and Vincent Cohen-Addad and Cat Graves and Vlad Feinberg and Grace Chung and DJ Strouse and Danny Karmon and Sahand Sharifzadeh and Zoe Ashwood and Khiem Pham and Jon Blanton and Alex Vasiloff and Jarred Barber and Mark Geller and Aurick Zhou and Fedir Zubach and Tzu-Kuo Huang and Lei Zhang and Himanshu Gupta and Matt Young and Julia Proskurnia and Ronny Votel and Valentin Gabeur and Gabriel Barcik and Aditya Tripathi and Hongkun Yu and Geng Yan and Beer Changpinyo and Filip Pavetić and Amy Coyle and Yasuhisa Fujii and Jorge Gonzalez Mendez and Tianhao Zhou and Harish Rajamani and Blake Hechtman and Eddie Cao and Da-Cheng Juan and Yi-Xuan Tan and Valentin Dalibard and Yilun Du and Natalie Clay and Kaisheng Yao and Wenhao Jia and Dimple Vijaykumar and Yuxiang Zhou and Xinyi Bai and Wei-Chih Hung and Steven Pecht and Georgi Todorov and Nikhil Khadke and Pramod Gupta and Preethi Lahoti and Arnaud Autef and Karthik Duddu and James Lee-Thorp and Alexander Bykovsky and Tautvydas Misiunas and Sebastian Flennerhag and Santhosh Thangaraj and Jed McGiffin and Zack Nado and Markus Kunesch and Andreas Noever and Amir Hertz and Marco Liang and Victor Stone and Evan Palmer and Samira Daruki and Arijit Pramanik and Siim Põder and Austin Kyker and Mina Khan and Evgeny Sluzhaev and Marvin Ritter and Avraham Ruderman and Wenlei Zhou and Chirag Nagpal and Kiran Vodrahalli and George Necula and Paul Barham and Ellie Pavlick and Jay Hartford and Izhak Shafran and Long Zhao and Maciej Mikuła and Tom Eccles and Hidetoshi Shimokawa and Kanav Garg and Luke Vilnis and Hanwen Chen and Ilia Shumailov and Kuang-Huei Lee and Abdelrahman Abdelhamed and Meiyan Xie and Vered Cohen and Ester Hlavnova and Dan Malkin and Chawin Sitawarin and James Lottes and Pauline Coquinot and Tianli Yu and Sandeep Kumar and Jingwei Zhang and Aroma Mahendru and Zafarali Ahmed and James Martens and Tao Chen and Aviel Boag and Daiyi Peng and Coline Devin and Arseniy Klimovskiy and Mary Phuong and Danny Vainstein and Jin Xie and Bhuvana Ramabhadran and Nathan Howard and Xinxin Yu and Gitartha Goswami and Jingyu Cui and Sam Shleifer and Mario Pinto and Chih-Kuan Yeh and Ming-Hsuan Yang and Sara Javanmardi and Dan Ethier and Chace Lee and Jordi Orbay and Suyog Kotecha and Carla Bromberg and Pete Shaw and James Thornton and Adi Gerzi Rosenthal and Shane Gu and Matt Thomas and Ian Gemp and Aditya Ayyar and Asahi Ushio and Aarush Selvan and Joel Wee and Chenxi Liu and Maryam Majzoubi and Weiren Yu and Jake Abernethy and Tyler Liechty and Renke Pan and Hoang Nguyen and Qiong and Hu and Sarah Perrin and Abhinav Arora and Emily Pitler and Weiyi Wang and Kaushik Shivakumar and Flavien Prost and Ben Limonchik and Jing Wang and Yi Gao and Timothee Cour and Shyamal Buch and Huan Gui and Maria Ivanova and Philipp Neubeck and Kelvin Chan and Lucy Kim and Huizhong Chen and Naman Goyal and Da-Woon Chung and Lu Liu and Yao Su and Anastasia Petrushkina and Jiajun Shen and Armand Joulin and Yuanzhong Xu and Stein Xudong Lin and Yana Kulizhskaya and Ciprian Chelba and Shobha Vasudevan and Eli Collins and Vasilisa Bashlovkina and Tony Lu and Doug Fritz and Jongbin Park and Yanqi Zhou and Chen Su and Richard Tanburn and Mikhail Sushkov and Mitchelle Rasquinha and Jinning Li and Jennifer Prendki and Yiming Li and Pallavi LV and Shriya Sharma and Hen Fitoussi and Hui Huang and Andrew Dai and Phuong Dao and Mike Burrows and Henry Prior and Danfeng Qin and Golan Pundak and Lars Lowe Sjoesund and Art Khurshudov and Zhenkai Zhu and Albert Webson and Elizabeth Kemp and Tat Tan and Saurabh Agrawal and Susie Sargsyan and Liqun Cheng and Jim Stephan and Tom Kwiatkowski and David Reid and Arunkumar Byravan and Assaf Hurwitz Michaely and Nicolas Heess and Luowei Zhou and Sonam Goenka and Viral Carpenter and Anselm Levskaya and Bo Wang and Reed Roberts and Rémi Leblond and Sharat Chikkerur and Stav Ginzburg and Max Chang and Robert Riachi and Chuqiao and Xu and Zalán Borsos and Michael Pliskin and Julia Pawar and Morgane Lustman and Hannah Kirkwood and Ankit Anand and Aditi Chaudhary and Norbert Kalb and Kieran Milan and Sean Augenstein and Anna Goldie and Laurel Prince and Karthik Raman and Yanhua Sun and Vivian Xia and Aaron Cohen and Zhouyuan Huo and Josh Camp and Seher Ellis and Lukas Zilka and David Vilar Torres and Lisa Patel and Sho Arora and Betty Chan and Jonas Adler and Kareem Ayoub and Jacky Liang and Fayaz Jamil and Jiepu Jiang and Simon Baumgartner and Haitian Sun and Yael Karov and Yaroslav Akulov and Hui Zheng and Irene Cai and Claudio Fantacci and James Rubin and Alex Rav Acha and Mengchao Wang and Nina D'Souza and Rohit Sathyanarayana and Shengyang Dai and Simon Rowe and Andrey Simanovsky and Omer Goldman and Yuheng Kuang and Xiaoyue Pan and Andrew Rosenberg and Tania Rojas-Esponda and Praneet Dutta and Amy Zeng and Irina Jurenka and Greg Farquhar and Yamini Bansal and Shariq Iqbal and Becca Roelofs and Ga-Young Joung and Parker Beak and Changwan Ryu and Ryan Poplin and Yan Wu and Jean-Baptiste Alayrac and Senaka Buthpitiya and Olaf Ronneberger and Caleb Habtegebriel and Wei Li and Paul Cavallaro and Aurora Wei and Guy Bensky and Timo Denk and Harish Ganapathy and Jeff Stanway and Pratik Joshi and Francesco Bertolini and Jessica Lo and Olivia Ma and Zachary Charles and Geta Sampemane and Himanshu Sahni and Xu Chen and Harry Askham and David Gaddy and Peter Young and Jiewen Tan and Matan Eyal and Arthur Bražinskas and Li Zhong and Zhichun Wu and Mark Epstein and Kai Bailey and Andrew Hard and Kamyu Lee and Sasha Goldshtein and Alex Ruiz and Mohammed Badawi and Matthias Lochbrunner and JK Kearns and Ashley Brown and Fabio Pardo and Theophane Weber and Haichuan Yang and Pan-Pan Jiang and Berkin Akin and Zhao Fu and Marcus Wainwright and Chi Zou and Meenu Gaba and Pierre-Antoine Manzagol and Wendy Kan and Yang Song and Karina Zainullina and Rui Lin and Jeongwoo Ko and Salil Deshmukh and Apoorv Jindal and James Svensson and Divya Tyam and Heri Zhao and Christine Kaeser-Chen and Scott Baird and Pooya Moradi and Jamie Hall and Qiuchen Guo and Vincent Tsang and Bowen Liang and Fernando Pereira and Suhas Ganesh and Ivan Korotkov and Jakub Adamek and Sridhar Thiagarajan and Vinh Tran and Charles Chen and Chris Tar and Sanil Jain and Ishita Dasgupta and Taylan Bilal and David Reitter and Kai Zhao and Giulia Vezzani and Yasmin Gehman and Pulkit Mehta and Lauren Beltrone and Xerxes Dotiwalla and Sergio Guadarrama and Zaheer Abbas and Stefani Karp and Petko Georgiev and Chun-Sung Ferng and Marc Brockschmidt and Liqian Peng and Christoph Hirnschall and Vikas Verma and Yingying Bi and Ying Xiao and Avigail Dabush and Kelvin Xu and Phil Wallis and Randall Parker and Qifei Wang and Yang Xu and Ilkin Safarli and Dinesh Tewari and Yin Zhang and Seungyeon Kim and Andrea Gesmundo and Mackenzie Thomas and Sergey Levi and Ahmed Chowdhury and Kanishka Rao and Peter Garst and Sam Conway-Rahman and Helen Ran and Kay McKinney and Zhisheng Xiao and Wenhao Yu and Rohan Agrawal and Axel Stjerngren and Catalin Ionescu and Jingjing Chen and Vivek Sharma and Justin Chiu and Fei Liu and Ken Franko and Clayton Sanford and Xingyu Cai and Paul Michel and Sanjay Ganapathy and Jane Labanowski and Zachary Garrett and Ben Vargas and Sean Sun and Bryan Gale and Thomas Buschmann and Guillaume Desjardins and Nimesh Ghelani and Palak Jain and Mudit Verma and Chulayuth Asawaroengchai and Julian Eisenschlos and Jitendra Harlalka and Hideto Kazawa and Don Metzler and Joshua Howland and Ying Jian and Jake Ades and Viral Shah and Tynan Gangwani and Seungji Lee and Roman Ring and Steven M. Hernandez and Dean Reich and Amer Sinha and Ashutosh Sathe and Joe Kovac and Ashleah Gill and Ajay Kannan and Andrea D'olimpio and Martin Sevenich and Jay Whang and Been Kim and Khe Chai Sim and Jilin Chen and Jiageng Zhang and Shuba Lall and Yossi Matias and Bill Jia and Abe Friesen and Sara Nasso and Ashish Thapliyal and Bryan Perozzi and Ting Yu and Anna Shekhawat and Safeen Huda and Peter Grabowski and Eric Wang and Ashwin Sreevatsa and Hilal Dib and Mehadi Hassen and Parker Schuh and Vedrana Milutinovic and Chris Welty and Michael Quinn and Ali Shah and Bangju Wang and Gabe Barth-Maron and Justin Frye and Natalie Axelsson and Tao Zhu and Yukun Ma and Irene Giannoumis and Hanie Sedghi and Chang Ye and Yi Luan and Kevin Aydin and Bilva Chandra and Vivek Sampathkumar and Ronny Huang and Victor Lavrenko and Ahmed Eleryan and Zhi Hong and Steven Hansen and Sara Mc Carthy and Bidisha Samanta and Domagoj Ćevid and Xin Wang and Fangtao Li and Michael Voznesensky and Matt Hoffman and Andreas Terzis and Vikash Sehwag and Gil Fidel and Luheng He and Mu Cai and Yanzhang He and Alex Feng and Martin Nikoltchev and Samrat Phatale and Jason Chase and Rory Lawton and Ming Zhang and Tom Ouyang and Manuel Tragut and Mehdi Hafezi Manshadi and Arjun Narayanan and Jiaming Shen and Xu Gao and Tolga Bolukbasi and Nick Roy and Xin Li and Daniel Golovin and Liviu Panait and Zhen Qin and Guangxing Han and Thomas Anthony and Sneha Kudugunta and Viorica Patraucean and Aniket Ray and Xinyun Chen and Xiaochen Yang and Tanuj Bhatia and Pranav Talluri and Alex Morris and Andrija Ražnatović and Bethanie Brownfield and James An and Sheng Peng and Patrick Kane and Ce Zheng and Nico Duduta and Joshua Kessinger and James Noraky and Siqi Liu and Keran Rong and Petar Veličković and Keith Rush and Alex Goldin and Fanny Wei and Shiva Mohan Reddy Garlapati and Caroline Pantofaru and Okwan Kwon and Jianmo Ni and Eric Noland and Julia Di Trapani and Françoise Beaufays and Abhijit Guha Roy and Yinlam Chow and Aybuke Turker and Geoffrey Cideron and Lantao Mei and Jon Clark and Qingyun Dou and Matko Bošnjak and Ralph Leith and Yuqing Du and Amir Yazdanbakhsh and Milad Nasr and Chester Kwak and Suraj Satishkumar Sheth and Alex Kaskasoli and Ankesh Anand and Balaji Lakshminarayanan and Sammy Jerome and David Bieber and Chun-Te Chu and Alexandre Senges and Tianxiao Shen and Mukund Sridhar and Ndaba Ndebele and Benjamin Beyret and Shakir Mohamed and Mia Chen and Markus Freitag and Jiaxian Guo and Luyang Liu and Paul Roit and Heng Chen and Shen Yan and Tom Stone and JD Co-Reyes and Jeremy Cole and Salvatore Scellato and Shekoofeh Azizi and Hadi Hashemi and Alicia Jin and Anand Iyer and Marcella Valentine and András György and Arun Ahuja and Daniel Hernandez Diaz and Chen-Yu Lee and Nathan Clement and Weize Kong and Drew Garmon and Ishaan Watts and Kush Bhatia and Khyatti Gupta and Matt Miecnikowski and Hugo Vallet and Ankur Taly and Edward Loper and Saket Joshi and James Atwood and Jo Chick and Mark Collier and Fotis Iliopoulos and Ryan Trostle and Beliz Gunel and Ramiro Leal-Cavazos and Arnar Mar Hrafnkelsson and Michael Guzman and Xiaoen Ju and Andy Forbes and Jesse Emond and Kushal Chauhan and Ben Caine and Li Xiao and Wenjun Zeng and Alexandre Moufarek and Daniel Murphy and Maya Meng and Nitish Gupta and Felix Riedel and Anil Das and Elijah Lawal and Shashi Narayan and Tiberiu Sosea and James Swirhun and Linda Friso and Behnam Neyshabur and Jing Lu and Sertan Girgin and Michael Wunder and Edouard Yvinec and Aroonalok Pyne and Victor Carbune and Shruti Rijhwani and Yang Guo and Tulsee Doshi and Anton Briukhov and Max Bain and Ayal Hitron and Xuanhui Wang and Ashish Gupta and Ke Chen and Cosmo Du and Weiyang Zhang and Dhruv Shah and Arjun Akula and Max Dylla and Ashyana Kachra and Weicheng Kuo and Tingting Zou and Lily Wang and Luyao Xu and Jifan Zhu and Justin Snyder and Sachit Menon and Orhan Firat and Igor Mordatch and Yuan Yuan and Natalia Ponomareva and Rory Blevins and Lawrence Moore and Weijun Wang and Phil Chen and Martin Scholz and Artur Dwornik and Jason Lin and Sicheng Li and Diego Antognini and Te I and Xiaodan Song and Matt Miller and Uday Kalra and Adam Raveret and Oscar Akerlund and Felix Wu and Andrew Nystrom and Namrata Godbole and Tianqi Liu and Hannah DeBalsi and Jewel Zhao and Buhuang Liu and Avi Caciularu and Lauren Lax and Urvashi Khandelwal and Victoria Langston and Eric Bailey and Silvio Lattanzi and Yufei Wang and Neel Kovelamudi and Sneha Mondal and Guru Guruganesh and Nan Hua and Ofir Roval and Paweł Wesołowski and Rishikesh Ingale and Jonathan Halcrow and Tim Sohn and Christof Angermueller and Bahram Raad and Eli Stickgold and Eva Lu and Alec Kosik and Jing Xie and Timothy Lillicrap and Austin Huang and Lydia Lihui Zhang and Dominik Paulus and Clement Farabet and Alex Wertheim and Bing Wang and Rishabh Joshi and Chu-ling Ko and Yonghui Wu and Shubham Agrawal and Lily Lin and XiangHai Sheng and Peter Sung and Tyler Breland-King and Christina Butterfield and Swapnil Gawde and Sumeet Singh and Qiao Zhang and Raj Apte and Shilpa Shetty and Adrian Hutter and Tao Li and Elizabeth Salesky and Federico Lebron and Jonni Kanerva and Michela Paganini and Arthur Nguyen and Rohith Vallu and Jan-Thorsten Peter and Sarmishta Velury and David Kao and Jay Hoover and Anna Bortsova and Colton Bishop and Shoshana Jakobovits and Alessandro Agostini and Alekh Agarwal and Chang Liu and Charles Kwong and Sasan Tavakkol and Ioana Bica and Alex Greve and Anirudh GP and Jake Marcus and Le Hou and Tom Duerig and Rivka Moroshko and Dave Lacey and Andy Davis and Julien Amelot and Guohui Wang and Frank Kim and Theofilos Strinopoulos and Hui Wan and Charline Le Lan and Shankar Krishnan and Haotian Tang and Peter Humphreys and Junwen Bai and Idan Heimlich Shtacher and Diego Machado and Chenxi Pang and Ken Burke and Dangyi Liu and Renga Aravamudhan and Yue Song and Ed Hirst and Abhimanyu Singh and Brendan Jou and Liang Bai and Francesco Piccinno and Chuyuan Kelly Fu and Robin Alazard and Barak Meiri and Daniel Winter and Charlie Chen and Mingda Zhang and Jens Heitkaemper and John Lambert and Jinhyuk Lee and Alexander Frömmgen and Sergey Rogulenko and Pranav Nair and Paul Niemczyk and Anton Bulyenov and Bibo Xu and Hadar Shemtov and Morteza Zadimoghaddam and Serge Toropov and Mateo Wirth and Hanjun Dai and Sreenivas Gollapudi and Daniel Zheng and Alex Kurakin and Chansoo Lee and Kalesha Bullard and Nicolas Serrano and Ivana Balazevic and Yang Li and Johan Schalkwyk and Mark Murphy and Mingyang Zhang and Kevin Sequeira and Romina Datta and Nishant Agrawal and Charles Sutton and Nithya Attaluri and Mencher Chiang and Wael Farhan and Gregory Thornton and Kate Lin and Travis Choma and Hung Nguyen and Kingshuk Dasgupta and Dirk Robinson and Iulia Comşa and Michael Riley and Arjun Pillai and Basil Mustafa and Ben Golan and Amir Zandieh and Jean-Baptiste Lespiau and Billy Porter and David Ross and Sujeevan Rajayogam and Mohit Agarwal and Subhashini Venugopalan and Bobak Shahriari and Qiqi Yan and Hao Xu and Taylor Tobin and Pavel Dubov and Hongzhi Shi and Adrià Recasens and Anton Kovsharov and Sebastian Borgeaud and Lucio Dery and Shanthal Vasanth and Elena Gribovskaya and Linhai Qiu and Mahdis Mahdieh and Wojtek Skut and Elizabeth Nielsen and CJ Zheng and Adams Yu and Carrie Grimes Bostock and Shaleen Gupta and Aaron Archer and Chris Rawles and Elinor Davies and Alexey Svyatkovskiy and Tomy Tsai and Yoni Halpern and Christian Reisswig and Bartek Wydrowski and Bo Chang and Joan Puigcerver and Mor Hazan Taege and Jian Li and Eva Schnider and Xinjian Li and Dragos Dena and Yunhan Xu and Umesh Telang and Tianze Shi and Heiga Zen and Kyle Kastner and Yeongil Ko and Neesha Subramaniam and Aviral Kumar and Pete Blois and Zhuyun Dai and John Wieting and Yifeng Lu and Yoel Zeldes and Tian Xie and Anja Hauth and Alexandru Ţifrea and Yuqi Li and Sam El-Husseini and Dan Abolafia and Howard Zhou and Wen Ding and Sahra Ghalebikesabi and Carlos Guía and Andrii Maksai and Ágoston Weisz and Sercan Arik and Nick Sukhanov and Aga Świetlik and Xuhui Jia and Luo Yu and Weiyue Wang and Mark Brand and Dawn Bloxwich and Sean Kirmani and Zhe Chen and Alec Go and Pablo Sprechmann and Nithish Kannen and Alen Carin and Paramjit Sandhu and Isabel Edkins and Leslie Nooteboom and Jai Gupta and Loren Maggiore and Javad Azizi and Yael Pritch and Pengcheng Yin and Mansi Gupta and Danny Tarlow and Duncan Smith and Desi Ivanov and Mohammad Babaeizadeh and Ankita Goel and Satish Kambala and Grace Chu and Matej Kastelic and Michelle Liu and Hagen Soltau and Austin Stone and Shivani Agrawal and Min Kim and Kedar Soparkar and Srinivas Tadepalli and Oskar Bunyan and Rachel Soh and Arvind Kannan and DY Kim and Blake JianHang Chen and Afief Halumi and Sudeshna Roy and Yulong Wang and Olcan Sercinoglu and Gena Gibson and Sijal Bhatnagar and Motoki Sano and Daniel von Dincklage and Qingchun Ren and Blagoj Mitrevski and Mirek Olšák and Jennifer She and Carl Doersch and Jilei and Wang and Bingyuan Liu and Qijun Tan and Tamar Yakar and Tris Warkentin and Alex Ramirez and Carl Lebsack and Josh Dillon and Rajiv Mathews and Tom Cobley and Zelin Wu and Zhuoyuan Chen and Jon Simon and Swaroop Nath and Tara Sainath and Alexei Bendebury and Ryan Julian and Bharath Mankalale and Daria Ćurko and Paulo Zacchello and Adam R. Brown and Kiranbir Sodhia and Heidi Howard and Sergi Caelles and Abhinav Gupta and Gareth Evans and Anna Bulanova and Lesley Katzen and Roman Goldenberg and Anton Tsitsulin and Joe Stanton and Benoit Schillings and Vitaly Kovalev and Corey Fry and Rushin Shah and Kuo Lin and Shyam Upadhyay and Cheng Li and Soroush Radpour and Marcello Maggioni and Jing Xiong and Lukas Haas and Jenny Brennan and Aishwarya Kamath and Nikolay Savinov and Arsha Nagrani and Trevor Yacovone and Ryan Kappedal and Kostas Andriopoulos and Li Lao and YaGuang Li and Grigory Rozhdestvenskiy and Kazuma Hashimoto and Andrew Audibert and Sophia Austin and Daniel Rodriguez and Anian Ruoss and Garrett Honke and Deep Karkhanis and Xi Xiong and Qing Wei and James Huang and Zhaoqi Leng and Vittal Premachandran and Stan Bileschi and Georgios Evangelopoulos and Thomas Mensink and Jay Pavagadhi and Denis Teplyashin and Paul Chang and Linting Xue and Garrett Tanzer and Sally Goldman and Kaushal Patel and Shixin Li and Jeremy Wiesner and Ivy Zheng and Ian Stewart-Binks and Jie Han and Zhi Li and Liangchen Luo and Karel Lenc and Mario Lučić and Fuzhao Xue and Ryan Mullins and Alexey Guseynov and Chung-Ching Chang and Isaac Galatzer-Levy and Adam Zhang and Garrett Bingham and Grace Hu and Ale Hartman and Yue Ma and Jordan Griffith and Alex Irpan and Carey Radebaugh and Summer Yue and Lijie Fan and Victor Ungureanu and Christina Sorokin and Hannah Teufel and Peiran Li and Rohan Anil and Dimitris Paparas and Todd Wang and Chu-Cheng Lin and Hui Peng and Megan Shum and Goran Petrovic and Demetra Brady and Richard Nguyen and Klaus Macherey and Zhihao Li and Harman Singh and Madhavi Yenugula and Mariko Iinuma and Xinyi Chen and Kavya Kopparapu and Alexey Stern and Shachi Dave and Chandu Thekkath and Florence Perot and Anurag Kumar and Fangda Li and Yang Xiao and Matthew Bilotti and Mohammad Hossein Bateni and Isaac Noble and Lisa Lee and Amelio Vázquez-Reina and Julian Salazar and Xiaomeng Yang and Boyu Wang and Ela Gruzewska and Anand Rao and Sindhu Raghuram and Zheng Xu and Eyal Ben-David and Jieru Mei and Sid Dalmia and Zhaoyi Zhang and Yuchen Liu and Gagan Bansal and Helena Pankov and Steven Schwarcz and Andrea Burns and Christine Chan and Sumit Sanghai and Ricky Liang and Ethan Liang and Antoine He and Amy Stuart and Arun Narayanan and Yukun Zhu and Christian Frank and Bahar Fatemi and Amit Sabne and Oran Lang and Indro Bhattacharya and Shane Settle and Maria Wang and Brendan McMahan and Andrea Tacchetti and Livio Baldini Soares and Majid Hadian and Serkan Cabi and Timothy Chung and Nikita Putikhin and Gang Li and Jeremy Chen and Austin Tarango and Henryk Michalewski and Mehran Kazemi and Hussain Masoom and Hila Sheftel and Rakesh Shivanna and Archita Vadali and Ramona Comanescu and Doug Reid and Joss Moore and Arvind Neelakantan and Michaël Sander and Jonathan Herzig and Aviv Rosenberg and Mostafa Dehghani and JD Choi and Michael Fink and Reid Hayes and Eric Ge and Shitao Weng and Chia-Hua Ho and John Karro and Kalpesh Krishna and Lam Nguyen Thiet and Amy Skerry-Ryan and Daniel Eppens and Marco Andreetto and Navin Sarma and Silvano Bonacina and Burcu Karagol Ayan and Megha Nawhal and Zhihao Shan and Mike Dusenberry and Shantanu Thakoor and Sagar Gubbi and Duc Dung Nguyen and Reut Tsarfaty and Samuel Albanie and Jovana Mitrović and Meet Gandhi and Bo-Juen Chen and Alessandro Epasto and Georgi Stephanov and Ye Jin and Samuel Gehman and Aida Amini and Jack Weber and Feryal Behbahani and Shawn Xu and Miltos Allamanis and Xi Chen and Myle Ott and Claire Sha and Michal Jastrzebski and Hang Qi and David Greene and Xinyi Wu and Abodunrinwa Toki and Daniel Vlasic and Jane Shapiro and Ragha Kotikalapudi and Zhe Shen and Takaaki Saeki and Sirui Xie and Albin Cassirer and Shikhar Bharadwaj and Tatsuya Kiyono and Srinadh Bhojanapalli and Elan Rosenfeld and Sam Ritter and Jieming Mao and João Gabriel Oliveira and Zoltan Egyed and Bernd Bandemer and Emilio Parisotto and Keisuke Kinoshita and Juliette Pluto and Petros Maniatis and Steve Li and Yaohui Guo and Golnaz Ghiasi and Jean Tarbouriech and Srimon Chatterjee and Julie Jin and Katrina and Xu and Jennimaria Palomaki and Séb Arnold and Madhavi Sewak and Federico Piccinini and Mohit Sharma and Ben Albrecht and Sean Purser-haskell and Ashwin Vaswani and Chongyan Chen and Matheus Wisniewski and Qin Cao and John Aslanides and Nguyet Minh Phu and Maximilian Sieb and Lauren Agubuzu and Anne Zheng and Daniel Sohn and Marco Selvi and Anders Andreassen and Krishan Subudhi and Prem Eruvbetine and Oliver Woodman and Tomas Mery and Sebastian Krause and Xiaoqi Ren and Xiao Ma and Jincheng Luo and Dawn Chen and Wei Fan and Henry Griffiths and Christian Schuler and Alice Li and Shujian Zhang and Jean-Michel Sarr and Shixin Luo and Riccardo Patana and Matthew Watson and Dani Naboulsi and Michael Collins and Sailesh Sidhwani and Emiel Hoogeboom and Sharon Silver and Emily Caveness and Xiaokai Zhao and Mikel Rodriguez and Maxine Deines and Libin Bai and Patrick Griffin and Marco Tagliasacchi and Emily Xue and Spandana Raj Babbula and Bo Pang and Nan Ding and Gloria Shen and Elijah Peake and Remi Crocker and Shubha Srinivas Raghvendra and Danny Swisher and Woohyun Han and Richa Singh and Ling Wu and Vladimir Pchelin and Tsendsuren Munkhdalai and Dana Alon and Geoff Bacon and Efren Robles and Jannis Bulian and Melvin Johnson and George Powell and Felipe Tiengo Ferreira and Yaoyiran Li and Frederik Benzing and Mihajlo Velimirović and Hubert Soyer and William Kong and Tony and Nguyên and Zhen Yang and Jeremiah Liu and Joost van Amersfoort and Daniel Gillick and Baochen Sun and Nathalie Rauschmayr and Katie Zhang and Serena Zhan and Tao Zhou and Alexey Frolov and Chengrun Yang and Denis Vnukov and Louis Rouillard and Hongji Li and Amol Mandhane and Nova Fallen and Rajesh Venkataraman and Clara Huiyi Hu and Jennifer Brennan and Jenny Lee and Jerry Chang and Martin Sundermeyer and Zhufeng Pan and Rosemary Ke and Simon Tong and Alex Fabrikant and William Bono and Jindong Gu and Ryan Foley and Yiran Mao and Manolis Delakis and Dhruva Bhaswar and Roy Frostig and Nick Li and Avital Zipori and Cath Hope and Olga Kozlova and Swaroop Mishra and Josip Djolonga and Craig Schiff and Majd Al Merey and Eleftheria Briakou and Peter Morgan and Andy Wan and Avinatan Hassidim and RJ Skerry-Ryan and Kuntal Sengupta and Mary Jasarevic and Praveen Kallakuri and Paige Kunkle and Hannah Brennan and Tom Lieber and Hassan Mansoor and Julian Walker and Bing Zhang and Annie Xie and Goran Žužić and Adaeze Chukwuka and Alex Druinsky and Donghyun Cho and Rui Yao and Ferjad Naeem and Shiraz Butt and Eunyoung Kim and Zhipeng Jia and Mandy Jordan and Adam Lelkes and Mark Kurzeja and Sophie Wang and James Zhao and Andrew Over and Abhishek Chakladar and Marcel Prasetya and Neha Jha and Sriram Ganapathy and Yale Cong and Prakash Shroff and Carl Saroufim and Sobhan Miryoosefi and Mohamed Hammad and Tajwar Nasir and Weijuan Xi and Yang Gao and Young Maeng and Ben Hora and Chin-Yi Cheng and Parisa Haghani and Yoad Lewenberg and Caden Lu and Martin Matysiak and Naina Raisinghani and Huiyu Wang and Lexi Baugher and Rahul Sukthankar and Minh Giang and John Schultz and Noah Fiedel and Minmin Chen and Cheng-Chun Lee and Tapomay Dey and Hao Zheng and Shachi Paul and Celine Smith and Andy Ly and Yicheng Wang and Rishabh Bansal and Bartek Perz and Susanna Ricco and Stasha Blank and Vaishakh Keshava and Deepak Sharma and Marvin Chow and Kunal Lad and Komal Jalan and Simon Osindero and Craig Swanson and Jacob Scott and Anastasija Ilić and Xiaowei Li and Siddhartha Reddy Jonnalagadda and Afzal Shama Soudagar and Yan Xiong and Bat-Orgil Batsaikhan and Daniel Jarrett and Naveen Kumar and Maulik Shah and Matt Lawlor and Austin Waters and Mark Graham and Rhys May and Sabela Ramos and Sandra Lefdal and Zeynep Cankara and Nacho Cano and Brendan O'Donoghue and Jed Borovik and Frederick Liu and Jordan Grimstad and Mahmoud Alnahlawi and Katerina Tsihlas and Tom Hudson and Nikolai Grigorev and Yiling Jia and Terry Huang and Tobenna Peter Igwe and Sergei Lebedev and Xiaodan Tang and Igor Krivokon and Frankie Garcia and Melissa Tan and Eric Jia and Peter Stys and Shikhar Vashishth and Yu Liang and Balaji Venkatraman and Chenjie Gu and Anastasios Kementsietsidis and Chen Zhu and Junehyuk Jung and Yunfei Bai and Mohammad Javad Hosseini and Faruk Ahmed and Aditya Gupta and Xin Yuan and Shereen Ashraf and Shitij Nigam and Gautam Vasudevan and Pranjal Awasthi and Adi Mayrav Gilady and Zelda Mariet and Ramy Eskander and Haiguang Li and Hexiang Hu and Guillermo Garrido and Philippe Schlattner and George Zhang and Rohun Saxena and Petar Dević and Kritika Muralidharan and Ashwin Murthy and Yiqian Zhou and Min Choi and Arissa Wongpanich and Zhengdong Wang and Premal Shah and Yuntao Xu and Yiling Huang and Stephen Spencer and Alice Chen and James Cohan and Junjie Wang and Jonathan Tompson and Junru Wu and Ruba Haroun and Haiqiong Li and Blanca Huergo and Fan Yang and Tongxin Yin and James Wendt and Michael Bendersky and Rahma Chaabouni and Javier Snaider and Johan Ferret and Abhishek Jindal and Tara Thompson and Andrew Xue and Will Bishop and Shubham Milind Phal and Archit Sharma and Yunhsuan Sung and Prabakar Radhakrishnan and Mo Shomrat and Reeve Ingle and Roopali Vij and Justin Gilmer and Mihai Dorin Istin and Sam Sobell and Yang Lu and Emily Nottage and Dorsa Sadigh and Jeremiah Willcock and Tingnan Zhang and Steve Xu and Sasha Brown and Katherine Lee and Gary Wang and Yun Zhu and Yi Tay and Cheolmin Kim and Audrey Gutierrez and Abhanshu Sharma and Yongqin Xian and Sungyong Seo and Claire Cui and Elena Pochernina and Cip Baetu and Krzysztof Jastrzębski and Mimi Ly and Mohamed Elhawaty and Dan Suh and Eren Sezener and Pidong Wang and Nancy Yuen and George Tucker and Jiahao Cai and Zuguang Yang and Cindy Wang and Alex Muzio and Hai Qian and Jae Yoo and Derek Lockhart and Kevin R. McKee and Mandy Guo and Malika Mehrotra and Artur Mendonça and Sanket Vaibhav Mehta and Sherry Ben and Chetan Tekur and Jiaqi Mu and Muye Zhu and Victoria Krakovna and Hongrae Lee and AJ Maschinot and Sébastien Cevey and HyunJeong Choe and Aijun Bai and Hansa Srinivasan and Derek Gasaway and Nick Young and Patrick Siegler and Dan Holtmann-Rice and Vihari Piratla and Kate Baumli and Roey Yogev and Alex Hofer and Hado van Hasselt and Svetlana Grant and Yuri Chervonyi and David Silver and Andrew Hogue and Ayushi Agarwal and Kathie Wang and Preeti Singh and Four Flynn and Josh Lipschultz and Robert David and Lizzetth Bellot and Yao-Yuan Yang and Long Le and Filippo Graziano and Kate Olszewska and Kevin Hui and Akanksha Maurya and Nikos Parotsidis and Weijie Chen and Tayo Oguntebi and Joe Kelley and Anirudh Baddepudi and Johannes Mauerer and Gregory Shaw and Alex Siegman and Lin Yang and Shravya Shetty and Subhrajit Roy and Yunting Song and Wojciech Stokowiec and Ryan Burnell and Omkar Savant and Robert Busa-Fekete and Jin Miao and Samrat Ghosh and Liam MacDermed and Phillip Lippe and Mikhail Dektiarev and Zach Behrman and Fabian Mentzer and Kelvin Nguyen and Meng Wei and Siddharth Verma and Chris Knutsen and Sudeep Dasari and Zhipeng Yan and Petr Mitrichev and Xingyu Wang and Virat Shejwalkar and Jacob Austin and Srinivas Sunkara and Navneet Potti and Yan Virin and Christian Wright and Gaël Liu and Oriana Riva and Etienne Pot and Greg Kochanski and Quoc Le and Gargi Balasubramaniam and Arka Dhar and Yuguo Liao and Adam Bloniarz and Divyansh Shukla and Elizabeth Cole and Jong Lee and Sheng Zhang and Sushant Kafle and Siddharth Vashishtha and Parsa Mahmoudieh and Grace Chen and Raphael Hoffmann and Pranesh Srinivasan and Agustin Dal Lago and Yoav Ben Shalom and Zi Wang and Michael Elabd and Anuj Sharma and Junhyuk Oh and Suraj Kothawade and Maigo Le and Marianne Monteiro and Shentao Yang and Kaiz Alarakyia and Robert Geirhos and Diana Mincu and Håvard Garnes and Hayato Kobayashi and Soroosh Mariooryad and Kacper Krasowiak and Zhixin and Lai and Shibl Mourad and Mingqiu Wang and Fan Bu and Ophir Aharoni and Guanjie Chen and Abhimanyu Goyal and Vadim Zubov and Ankur Bapna and Elahe Dabir and Nisarg Kothari and Kay Lamerigts and Nicola De Cao and Jeremy Shar and Christopher Yew and Nitish Kulkarni and Dre Mahaarachchi and Mandar Joshi and Zhenhai Zhu and Jared Lichtarge and Yichao Zhou and Hannah Muckenhirn and Vittorio Selo and Oriol Vinyals and Peter Chen and Anthony Brohan and Vaibhav Mehta and Sarah Cogan and Ruth Wang and Ty Geri and Wei-Jen Ko and Wei Chen and Fabio Viola and Keshav Shivam and Lisa Wang and Madeleine Clare Elish and Raluca Ada Popa and Sébastien Pereira and Jianqiao Liu and Raphael Koster and Donnie Kim and Gufeng Zhang and Sayna Ebrahimi and Partha Talukdar and Yanyan Zheng and Petra Poklukar and Ales Mikhalap and Dale Johnson and Anitha Vijayakumar and Mark Omernick and Matt Dibb and Ayush Dubey and Qiong Hu and Apurv Suman and Vaibhav Aggarwal and Ilya Kornakov and Fei Xia and Wing Lowe and Alexey Kolganov and Ted Xiao and Vitaly Nikolaev and Steven Hemingray and Bonnie Li and Joana Iljazi and Mikołaj Rybiński and Ballie Sandhu and Peggy Lu and Thang Luong and Rodolphe Jenatton and Vineetha Govindaraj and Hui and Li and Gabriel Dulac-Arnold and Wonpyo Park and Henry Wang and Abhinit Modi and Jean Pouget-Abadie and Kristina Greller and Rahul Gupta and Robert Berry and Prajit Ramachandran and Jinyu Xie and Liam McCafferty and Jianling Wang and Kilol Gupta and Hyeontaek Lim and Blaž Bratanič and Andy Brock and Ilia Akolzin and Jim Sproch and Dan Karliner and Duhyeon Kim and Adrian Goedeckemeyer and Noam Shazeer and Cordelia Schmid and Daniele Calandriello and Parul Bhatia and Krzysztof Choromanski and Ceslee Montgomery and Dheeru Dua and Ana Ramalho and Helen King and Yue Gao and Lynn Nguyen and David Lindner and Divya Pitta and Oleaser Johnson and Khalid Salama and Diego Ardila and Michael Han and Erin Farnese and Seth Odoom and Ziyue Wang and Xiangzhuo Ding and Norman Rink and Ray Smith and Harshal Tushar Lehri and Eden Cohen and Neera Vats and Tong He and Parthasarathy Gopavarapu and Adam Paszke and Miteyan Patel and Wouter Van Gansbeke and Lucia Loher and Luis Castro and Maria Voitovich and Tamara von Glehn and Nelson George and Simon Niklaus and Zach Eaton-Rosen and Nemanja Rakićević and Erik Jue and Sagi Perel and Carrie Zhang and Yuval Bahat and Angéline Pouget and Zhi Xing and Fantine Huot and Ashish Shenoy and Taylor Bos and Vincent Coriou and Bryan Richter and Natasha Noy and Yaqing Wang and Santiago Ontanon and Siyang Qin and Gleb Makarchuk and Demis Hassabis and Zhuowan Li and Mandar Sharma and Kumaran Venkatesan and Iurii Kemaev and Roxanne Daniel and Shiyu Huang and Saloni Shah and Octavio Ponce and Warren and Chen and Manaal Faruqui and Jialin Wu and Slavica Andačić and Szabolcs Payrits and Daniel McDuff and Tom Hume and Yuan Cao and MH Tessler and Qingze Wang and Yinan Wang and Ivor Rendulic and Eirikur Agustsson and Matthew Johnson and Tanya Lando and Andrew Howard and Sri Gayatri Sundara Padmanabhan and Mayank Daswani and Andrea Banino and Michael Kilgore and Jonathan Heek and Ziwei Ji and Alvaro Caceres and Conglong Li and Nora Kassner and Alexey Vlaskin and Zeyu Liu and Alex Grills and Yanhan Hou and Roykrong Sukkerd and Gowoon Cheon and Nishita Shetty and Larisa Markeeva and Piotr Stanczyk and Tejas Iyer and Yuan Gong and Shawn Gao and Keerthana Gopalakrishnan and Tim Blyth and Malcolm Reynolds and Avishkar Bhoopchand and Misha Bilenko and Dero Gharibian and Vicky Zayats and Aleksandra Faust and Abhinav Singh and Min Ma and Hongyang Jiao and Sudheendra Vijayanarasimhan and Lora Aroyo and Vikas Yadav and Sarah Chakera and Ashwin Kakarla and Vilobh Meshram and Karol Gregor and Gabriela Botea and Evan Senter and Dawei Jia and Geza Kovacs and Neha Sharma and Sebastien Baur and Kai Kang and Yifan He and Lin Zhuo and Marija Kostelac and Itay Laish and Songyou Peng and Louis O'Bryan and Daniel Kasenberg and Girish Ramchandra Rao and Edouard Leurent and Biao Zhang and Sage Stevens and Ana Salazar and Ye Zhang and Ivan Lobov and Jake Walker and Allen Porter and Morgan Redshaw and Han Ke and Abhishek Rao and Alex Lee and Hoi Lam and Michael Moffitt and Jaeyoun Kim and Siyuan Qiao and Terry Koo and Robert Dadashi and Xinying Song and Mukund Sundararajan and Peng Xu and Chizu Kawamoto and Yan Zhong and Clara Barbu and Apoorv Reddy and Mauro Verzetti and Leon Li and George Papamakarios and Hanna Klimczak-Plucińska and Mary Cassin and Koray Kavukcuoglu and Rigel Swavely and Alain Vaucher and Jeffrey Zhao and Ross Hemsley and Michael Tschannen and Heming Ge and Gaurav Menghani and Yang Yu and Natalie Ha and Wei He and Xiao Wu and Maggie Song and Rachel Sterneck and Stefan Zinke and Dan A. Calian and Annie Marsden and Alejandro Cruzado Ruiz and Matteo Hessel and Almog Gueta and Benjamin Lee and Brian Farris and Manish Gupta and Yunjie Li and Mohammad Saleh and Vedant Misra and Kefan Xiao and Piermaria Mendolicchio and Gavin Buttimore and Varvara Krayvanova and Nigamaa Nayakanti and Matthew Wiethoff and Yash Pande and Azalia Mirhoseini and Ni Lao and Jasmine Liu and Yiqing Hua and Angie Chen and Yury Malkov and Dmitry Kalashnikov and Shubham Gupta and Kartik Audhkhasi and Yuexiang Zhai and Sudhindra Kopalle and Prateek Jain and Eran Ofek and Clemens Meyer and Khuslen Baatarsukh and Hana Strejček and Jun Qian and James Freedman and Ricardo Figueira and Michal Sokolik and Olivier Bachem and Raymond Lin and Dia Kharrat and Chris Hidey and Pingmei Xu and Dennis Duan and Yin Li and Muge Ersoy and Richard Everett and Kevin Cen and Rebeca Santamaria-Fernandez and Amir Taubenfeld and Ian Mackinnon and Linda Deng and Polina Zablotskaia and Shashank Viswanadha and Shivanker Goel and Damion Yates and Yunxiao Deng and Peter Choy and Mingqing Chen and Abhishek Sinha and Alex Mossin and Yiming Wang and Arthur Szlam and Susan Hao and Paul Kishan Rubenstein and Metin Toksoz-Exley and Miranda Aperghis and Yin Zhong and Junwhan Ahn and Michael Isard and Olivier Lacombe and Florian Luisier and Chrysovalantis Anastasiou and Yogesh Kalley and Utsav Prabhu and Emma Dunleavy and Shaan Bijwadia and Justin Mao-Jones and Kelly Chen and Rama Pasumarthi and Emily Wood and Adil Dostmohamed and Nate Hurley and Jiri Simsa and Alicia Parrish and Mantas Pajarskas and Matt Harvey and Ondrej Skopek and Yony Kochinski and Javier Rey and Verena Rieser and Denny Zhou and Sun Jae Lee and Trilok Acharya and Guowang Li and Joe Jiang and Xiaofan Zhang and Bryant Gipson and Ethan Mahintorabi and Marco Gelmi and Nima Khajehnouri and Angel Yeh and Kayi Lee and Loic Matthey and Leslie Baker and Trang Pham and Han Fu and Alex Pak and Prakhar Gupta and Cristina Vasconcelos and Adam Sadovsky and Brian Walker and Sissie Hsiao and Patrik Zochbauer and Andreea Marzoca and Noam Velan and Junhao Zeng and Gilles Baechler and Danny Driess and Divya Jain and Yanping Huang and Lizzie Tao and John Maggs and Nir Levine and Jon Schneider and Erika Gemzer and Samuel Petit and Shan Han and Zach Fisher and Dustin Zelle and Courtney Biles and Eugene Ie and Asya Fadeeva and Casper Liu and Juliana Vicente Franco and Adrian Collister and Hao Zhang and Renshen Wang and Ruizhe Zhao and Leandro Kieliger and Kurt Shuster and Rui Zhu and Boqing Gong and Lawrence Chan and Ruoxi Sun and Sujoy Basu and Roland Zimmermann and Jamie Hayes and Abhishek Bapna and Jasper Snoek and Weel Yang and Puranjay Datta and Jad Al Abdallah and Kevin Kilgour and Lu Li and SQ Mah and Yennie Jun and Morgane Rivière and Abhijit Karmarkar and Tammo Spalink and Tao Huang and Lucas Gonzalez and Duc-Hieu Tran and Averi Nowak and John Palowitch and Martin Chadwick and Ellie Talius and Harsh Mehta and Thibault Sellam and Philipp Fränken and Massimo Nicosia and Kyle He and Aditya Kini and David Amos and Sugato Basu and Harrison Jobe and Eleni Shaw and Qiantong Xu and Colin Evans and Daisuke Ikeda and Chaochao Yan and Larry Jin and Lun Wang and Sachin Yadav and Ilia Labzovsky and Ramesh Sampath and Ada Ma and Candice Schumann and Aditya Siddhant and Rohin Shah and John Youssef and Rishabh Agarwal and Natalie Dabney and Alessio Tonioni and Moran Ambar and Jing Li and Isabelle Guyon and Benny Li and David Soergel and Boya Fang and Georgi Karadzhov and Cristian Udrescu and Trieu Trinh and Vikas Raunak and Seb Noury and Dee Guo and Sonal Gupta and Mara Finkelstein and Denis Petek and Lihao Liang and Greg Billock and Pei Sun and David Wood and Yiwen Song and Xiaobin Yu and Tatiana Matejovicova and Regev Cohen and Kalyan Andra and David D'Ambrosio and Zhiwei Deng and Vincent Nallatamby and Ebrahim Songhori and Rumen Dangovski and Andrew Lampinen and Pankil Botadra and Adam Hillier and Jiawei Cao and Nagabhushan Baddi and Adhi Kuncoro and Toshihiro Yoshino and Ankit Bhagatwala and Marcáurelio Ranzato and Rylan Schaeffer and Tianlin Liu and Shuai Ye and Obaid Sarvana and John Nham and Chenkai Kuang and Isabel Gao and Jinoo Baek and Shubham Mittal and Ayzaan Wahid and Anita Gergely and Bin Ni and Josh Feldman and Carrie Muir and Pascal Lamblin and Wolfgang Macherey and Ethan Dyer and Logan Kilpatrick and Víctor Campos and Mukul Bhutani and Stanislav Fort and Yanif Ahmad and Aliaksei Severyn and Kleopatra Chatziprimou and Oleksandr Ferludin and Mason Dimarco and Aditya Kusupati and Joe Heyward and Dan Bahir and Kevin Villela and Katie Millican and Dror Marcus and Sanaz Bahargam and Caglar Unlu and Nicholas Roth and Zichuan Wei and Siddharth Gopal and Deepanway Ghoshal and Edward Lee and Sharon Lin and Jennie Lees and Dayeong Lee and Anahita Hosseini and Connie Fan and Seth Neel and Marcus Wu and Yasemin Altun and Honglong Cai and Enrique Piqueras and Josh Woodward and Alessandro Bissacco and Salem Haykal and Mahyar Bordbar and Prasha Sundaram and Sarah Hodkinson and Daniel Toyama and George Polovets and Austin Myers and Anu Sinha and Tomer Levinboim and Kashyap Krishnakumar and Rachita Chhaparia and Tatiana Sholokhova and Nitesh Bharadwaj Gundavarapu and Ganesh Jawahar and Haroon Qureshi and Jieru Hu and Nikola Momchev and Matthew Rahtz and Renjie Wu and Aishwarya P S and Kedar Dhamdhere and Meiqi Guo and Umang Gupta and Ali Eslami and Mariano Schain and Michiel Blokzijl and David Welling and Dave Orr and Levent Bolelli and Nicolas Perez-Nieves and Mikhail Sirotenko and Aman Prasad and Arjun Kar and Borja De Balle Pigem and Tayfun Terzi and Gellért Weisz and Dipankar Ghosh and Aditi Mavalankar and Dhruv Madeka and Kaspar Daugaard and Hartwig Adam and Viraj Shah and Dana Berman and Maggie Tran and Steven Baker and Ewa Andrejczuk and Grishma Chole and Ganna Raboshchuk and Mahdi Mirzazadeh and Thais Kagohara and Shimu Wu and Christian Schallhart and Bernett Orlando and Chen Wang and Alban Rrustemi and Hao Xiong and Hao Liu and Arpi Vezer and Nolan Ramsden and Shuo-yiin Chang and Sidharth Mudgal and Yan Li and Nino Vieillard and Yedid Hoshen and Farooq Ahmad and Ambrose Slone and Amy Hua and Natan Potikha and Mirko Rossini and Jon Stritar and Sushant Prakash and Zifeng Wang and Xuanyi Dong and Alireza Nazari and Efrat Nehoran and Kaan Tekelioglu and Yinxiao Li and Kartikeya Badola and Tom Funkhouser and Yuanzhen Li and Varun Yerram and Ramya Ganeshan and Daniel Formoso and Karol Langner and Tian Shi and Huijian Li and Yumeya Yamamori and Amayika Panda and Alaa Saade and Angelo Scorza Scarpati and Chris Breaux and CJ Carey and Zongwei Zhou and Cho-Jui Hsieh and Sophie Bridgers and Alena Butryna and Nishesh Gupta and Vaibhav Tulsyan and Sanghyun Woo and Evgenii Eltyshev and Will Grathwohl and Chanel Parks and Seth Benjamin and Rina Panigrahy and Shenil Dodhia and Daniel De Freitas and Chris Sauer and Will Song and Ferran Alet and Jackson Tolins and Cosmin Paduraru and Xingyi Zhou and Brian Albert and Zizhao Zhang and Lei Shu and Mudit Bansal and Sarah Nguyen and Amir Globerson and Owen Xiao and James Manyika and Tom Hennigan and Rong Rong and Josip Matak and Anton Bakalov and Ankur Sharma and Danila Sinopalnikov and Andrew Pierson and Stephen Roller and Geoff Brown and Mingcen Gao and Toshiyuki Fukuzawa and Amin Ghafouri and Kenny Vassigh and Iain Barr and Zhicheng Wang and Anna Korsun and Rajesh Jayaram and Lijie Ren and Tim Zaman and Samira Khan and Yana Lunts and Dan Deutsch and Dave Uthus and Nitzan Katz and Masha Samsikova and Amr Khalifa and Nikhil Sethi and Jiao Sun and Luming Tang and Uri Alon and Xianghong Luo and Dian Yu and Abhishek Nayyar and Bryce Petrini and Will Truong and Vincent Hellendoorn and Nikolai Chinaev and Chris Alberti and Wei Wang and Jingcao Hu and Vahab Mirrokni and Ananth Balashankar and Avia Aharon and Aahil Mehta and Ahmet Iscen and Joseph Kready and Lucas Manning and Anhad Mohananey and Yuankai Chen and Anshuman Tripathi and Allen Wu and Igor Petrovski and Dawsen Hwang and Martin Baeuml and Shreyas Chandrakaladharan and Yuan Liu and Rey Coaguila and Maxwell Chen and Sally Ma and Pouya Tafti and Susheel Tatineni and Terry Spitz and Jiayu Ye and Paul Vicol and Mihaela Rosca and Adrià Puigdomènech and Zohar Yahav and Sanjay Ghemawat and Hanzhao Lin and Phoebe Kirk and Zaid Nabulsi and Sergey Brin and Bernd Bohnet and Ken Caluwaerts and Aditya Srikanth Veerubhotla and Dan Zheng and Zihang Dai and Petre Petrov and Yichong Xu and Ramin Mehran and Zhuo Xu and Luisa Zintgraf and Jiho Choi and Spurthi Amba Hombaiah and Romal Thoppilan and Sashank Reddi and Lukasz Lew and Li Li and Kellie Webster and KP Sawhney and Lampros Lamprou and Siamak Shakeri and Mayank Lunayach and Jianmin Chen and Sumit Bagri and Alex Salcianu and Ying Chen and Yani Donchev and Charlotte Magister and Signe Nørly and Vitor Rodrigues and Tomas Izo and Hila Noga and Joe Zou and Thomas Köppe and Wenxuan Zhou and Kenton Lee and Xiangzhu Long and Danielle Eisenbud and Anthony Chen and Connor Schenck and Chi Ming To and Peilin Zhong and Emanuel Taropa and Minh Truong and Omer Levy and Danilo Martins and Zhiyuan Zhang and Christopher Semturs and Kelvin Zhang and Alex Yakubovich and Pol Moreno and Lara McConnaughey and Di Lu and Sam Redmond and Lotte Weerts and Yonatan Bitton and Tiziana Refice and Nicolas Lacasse and Arthur Conmy and Corentin Tallec and Julian Odell and Hannah Forbes-Pollard and Arkadiusz Socala and Jonathan Hoech and Pushmeet Kohli and Alanna Walton and Rui Wang and Mikita Sazanovich and Kexin Zhu and Andrei Kapishnikov and Rich Galt and Matthew Denton and Ben Murdoch and Caitlin Sikora and Kareem Mohamed and Wei Wei and Uri First and Tim McConnell and Luis C. Cobo and James Qin and Thi Avrahami and Daniel Balle and Yu Watanabe and Annie Louis and Adam Kraft and Setareh Ariafar and Yiming Gu and Eugénie Rives and Charles Yoon and Andrei Rusu and James Cobon-Kerr and Chris Hahn and Jiaming Luo and Yuvein and Zhu and Niharika Ahuja and Rodrigo Benenson and Raphaël Lopez Kaufman and Honglin Yu and Lloyd Hightower and Junlin Zhang and Darren Ni and Lisa Anne Hendricks and Gabby Wang and Gal Yona and Lalit Jain and Pablo Barrio and Surya Bhupatiraju and Siva Velusamy and Allan Dafoe and Sebastian Riedel and Tara Thomas and Zhe Yuan and Mathias Bellaiche and Sheena Panthaplackel and Klemen Kloboves and Sarthak Jauhari and Canfer Akbulut and Todor Davchev and Evgeny Gladchenko and David Madras and Aleksandr Chuklin and Tyrone Hill and Quan Yuan and Mukundan Madhavan and Luke Leonhard and Dylan Scandinaro and Qihang Chen and Ning Niu and Arthur Douillard and Bogdan Damoc and Yasumasa Onoe and Fabian Pedregosa and Fred Bertsch and Chas Leichner and Joseph Pagadora and Jonathan Malmaud and Sameera Ponda and Andy Twigg and Oleksii Duzhyi and Jingwei Shen and Miaosen Wang and Roopal Garg and Jing Chen and Utku Evci and Jonathan Lee and Leon Liu and Koji Kojima and Masa Yamaguchi and Arunkumar Rajendran and AJ Piergiovanni and Vinodh Kumar Rajendran and Marco Fornoni and Gabriel Ibagon and Harry Ragan and Sadh MNM Khan and John Blitzer and Andrew Bunner and Guan Sun and Takahiro Kosakai and Scott Lundberg and Ndidi Elue and Kelvin Guu and SK Park and Jane Park and Arunachalam Narayanaswamy and Chengda Wu and Jayaram Mudigonda and Trevor Cohn and Hairong Mu and Ravi Kumar and Laura Graesser and Yichi Zhang and Richard Killam and Vincent Zhuang and Mai Giménez and Wael Al Jishi and Ruy Ley-Wild and Alex Zhai and Kazuki Osawa and Diego Cedillo and Jialu Liu and Mayank Upadhyay and Marcin Sieniek and Roshan Sharma and Tom Paine and Anelia Angelova and Sravanti Addepalli and Carolina Parada and Kingshuk Majumder and Avery Lamp and Sanjiv Kumar and Xiang Deng and Artiom Myaskovsky and Tea Sabolić and Jeffrey Dudek and Sarah York and Félix de Chaumont Quitry and Jiazhong Nie and Dee Cattle and Alok Gunjan and Bilal Piot and Waleed Khawaja and Seojin Bang and Simon Wang and Siavash Khodadadeh and Raghavender R and Praynaa Rawlani and Richard Powell and Kevin Lee and Johannes Griesser and GS Oh and Cesar Magalhaes and Yujia Li and Simon Tokumine and Hadas Natalie Vogel and Dennis Hsu and Arturo BC and Disha Jindal and Matan Cohen and Zi Yang and Junwei Yuan and Dario de Cesare and Tony Bruguier and Jun Xu and Monica Roy and Alon Jacovi and Dan Belov and Rahul Arya and Phoenix Meadowlark and Shlomi Cohen-Ganor and Wenting Ye and Patrick Morris-Suzuki and Praseem Banzal and Gan Song and Pranavaraj Ponnuramu and Fred Zhang and George Scrivener and Salah Zaiem and Alif Raditya Rochman and Kehang Han and Badih Ghazi and Kate Lee and Shahar Drath and Daniel Suo and Antonious Girgis and Pradeep Shenoy and Duy Nguyen and Douglas Eck and Somit Gupta and Le Yan and Joao Carreira and Anmol Gulati and Ruoxin Sang and Daniil Mirylenka and Emma Cooney and Edward Chou and Mingyang Ling and Cindy Fan and Ben Coleman and Guilherme Tubone and Ravin Kumar and Jason Baldridge and Felix Hernandez-Campos and Angeliki Lazaridou and James Besley and Itay Yona and Neslihan Bulut and Quentin Wellens and AJ Pierigiovanni and Jasmine George and Richard Green and Pu Han and Connie Tao and Geoff Clark and Chong You and Abbas Abdolmaleki and Justin Fu and Tongzhou Chen and Ashwin Chaugule and Angad Chandorkar and Altaf Rahman and Will Thompson and Penporn Koanantakool and Mike Bernico and Jie Ren and Andrey Vlasov and Sergei Vassilvitskii and Maciej Kula and Yizhong Liang and Dahun Kim and Yangsibo Huang and Chengxi Ye and Dmitry Lepikhin and Wesley Helmholz},
      year={2025},
      eprint={2507.06261},
      archivePrefix={arXiv},
      primaryClass={cs.CL},
      url={https://arxiv.org/abs/2507.06261}, 
}

@misc{peter2025mindgapnottranslation,
      title={Mind the Gap... or Not? How Translation Errors and Evaluation Details Skew Multilingual Results}, 
      author={Jan-Thorsten Peter and David Vilar and Tobias Domhan and Dan Malkin and Markus Freitag},
      year={2025},
      eprint={2511.05162},
      archivePrefix={arXiv},
      primaryClass={cs.CL},
      url={https://arxiv.org/abs/2511.05162}, 
}

@misc{Google2025Gemini3,
  author       = {{Google DeepMind}},
  title        = {Gemini 3 [Large language model]},
  year         = {2025},
  url          = {https://blog.google/products-and-platforms/products/gemini/gemini-3/},
  howpublished = {Official Blog Release}
}

@online{anthropic_claude,
  author       = {{Anthropic}},
  title        = {Claude Family},
  year         = {2025},
  url          = {https://anthropic.com},
  note         = {Accessed: May 21, 2026}
}

@article{zhang2016understanding,
  title={Understanding deep learning requires rethinking generalization},
  author={Zhang, Chiyuan and Bengio, Samy and Hardt, Moritz and Recht, Benjamin and Vinyals, Oriol},
  journal={arXiv preprint arXiv:1611.03530},
  year={2016}
}

@misc{dominguez2025training,
      title={Training on the Test Task Confounds Evaluation and Emergence}, 
      author={Ricardo Dominguez-Olmedo and Florian E. Dorner and Moritz Hardt},
      year={2025},
      eprint={2407.07890},
      archivePrefix={arXiv},
      primaryClass={cs.CL},
      url={https://arxiv.org/abs/2407.07890}, 
}

@inproceedings{Artetxe_2020,
   title={On the Cross-lingual Transferability of Monolingual Representations},
   url={http://dx.doi.org/10.18653/v1/2020.acl-main.421},
   DOI={10.18653/v1/2020.acl-main.421},
   booktitle={Proceedings of the 58th Annual Meeting of the Association for Computational Linguistics},
   publisher={Association for Computational Linguistics},
   author={Artetxe, Mikel and Ruder, Sebastian and Yogatama, Dani},
   year={2020},
   pages={4623–4637} }

@misc{shi2022languagemodelsmultilingualchainofthought,
      title={Language Models are Multilingual Chain-of-Thought Reasoners}, 
      author={Freda Shi and Mirac Suzgun and Markus Freitag and Xuezhi Wang and Suraj Srivats and Soroush Vosoughi and Hyung Won Chung and Yi Tay and Sebastian Ruder and Denny Zhou and Dipanjan Das and Jason Wei},
      year={2022},
      eprint={2210.03057},
      archivePrefix={arXiv},
      primaryClass={cs.CL},
      url={https://arxiv.org/abs/2210.03057}, 
}

@article{10.1007/s10994-009-5152-4,
author = {Ben-David, Shai and Blitzer, John and Crammer, Koby and Kulesza, Alex and Pereira, Fernando and Vaughan, Jennifer Wortman},
title = {A theory of learning from different domains},
year = {2010},
issue_date = {May 2010},
publisher = {Kluwer Academic Publishers},
address = {USA},
volume = {79},
number = {1–2},
issn = {0885-6125},
url = {https://doi.org/10.1007/s10994-009-5152-4},
doi = {10.1007/s10994-009-5152-4},
journal = {Mach. Learn.},
month = may,
pages = {151–175},
numpages = {25}
}

@misc{mansour2023domainadaptationlearningbounds,
      title={Domain Adaptation: Learning Bounds and Algorithms}, 
      author={Yishay Mansour and Mehryar Mohri and Afshin Rostamizadeh},
      year={2023},
      eprint={0902.3430},
      archivePrefix={arXiv},
      primaryClass={cs.LG},
      url={https://arxiv.org/abs/0902.3430}, 
}

@misc{hu2025largelanguagemodelscrosslingual,
      title={Large Language Models Are Cross-Lingual Knowledge-Free Reasoners}, 
      author={Peng Hu and Sizhe Liu and Changjiang Gao and Xin Huang and Xue Han and Junlan Feng and Chao Deng and Shujian Huang},
      year={2025},
      eprint={2406.16655},
      archivePrefix={arXiv},
      primaryClass={cs.CL},
      url={https://arxiv.org/abs/2406.16655}, 
}

@misc{hu2020xtrememassivelymultilingualmultitask,
      title={XTREME: A Massively Multilingual Multi-task Benchmark for Evaluating Cross-lingual Generalization}, 
      author={Junjie Hu and Sebastian Ruder and Aditya Siddhant and Graham Neubig and Orhan Firat and Melvin Johnson},
      year={2020},
      eprint={2003.11080},
      archivePrefix={arXiv},
      primaryClass={cs.CL},
      url={https://arxiv.org/abs/2003.11080}, 
}

@misc{ifergan2024beneathsurfaceconsistencyexploring,
      title={Beneath the Surface of Consistency: Exploring Cross-lingual Knowledge Representation Sharing in LLMs}, 
      author={Maxim Ifergan and Leshem Choshen and Roee Aharoni and Idan Szpektor and Omri Abend},
      year={2024},
      eprint={2408.10646},
      archivePrefix={arXiv},
      primaryClass={cs.CL},
      url={https://arxiv.org/abs/2408.10646}, 
}

@misc{cobbe2021trainingverifierssolvemath,
      title={Training Verifiers to Solve Math Word Problems}, 
      author={Karl Cobbe and Vineet Kosaraju and Mohammad Bavarian and Mark Chen and Heewoo Jun and Lukasz Kaiser and Matthias Plappert and Jerry Tworek and Jacob Hilton and Reiichiro Nakano and Christopher Hesse and John Schulman},
      year={2021},
      eprint={2110.14168},
      archivePrefix={arXiv},
      primaryClass={cs.LG},
      url={https://arxiv.org/abs/2110.14168}, 
}

@misc{chi2021infoxlminformationtheoreticframeworkcrosslingual,
      title={InfoXLM: An Information-Theoretic Framework for Cross-Lingual Language Model Pre-Training}, 
      author={Zewen Chi and Li Dong and Furu Wei and Nan Yang and Saksham Singhal and Wenhui Wang and Xia Song and Xian-Ling Mao and Heyan Huang and Ming Zhou},
      year={2021},
      eprint={2007.07834},
      archivePrefix={arXiv},
      primaryClass={cs.CL},
      url={https://arxiv.org/abs/2007.07834}, 
}

@misc{wang2024probingemergencecrosslingualalignment,
      title={Probing the Emergence of Cross-lingual Alignment during LLM Training}, 
      author={Hetong Wang and Pasquale Minervini and Edoardo M. Ponti},
      year={2024},
      eprint={2406.13229},
      archivePrefix={arXiv},
      primaryClass={cs.CL},
      url={https://arxiv.org/abs/2406.13229}, 
}

@misc{yun2023xsnscrosslingualtransferprediction,
      title={X-SNS: Cross-Lingual Transfer Prediction through Sub-Network Similarity}, 
      author={Taejun Yun and Jinhyeon Kim and Deokyeong Kang and Seong Hoon Lim and Jihoon Kim and Taeuk Kim},
      year={2023},
      eprint={2310.17166},
      archivePrefix={arXiv},
      primaryClass={cs.CL},
      url={https://arxiv.org/abs/2310.17166}, 
}

@misc{han2025mubenchassessmentmultilingualcapabilities,
      title={MuBench: Assessment of Multilingual Capabilities of Large Language Models Across 61 Languages}, 
      author={Wenhan Han and Yifan Zhang and Zhixun Chen and Binbin Liu and Haobin Lin and Bingni Zhang and Taifeng Wang and Mykola Pechenizkiy and Meng Fang and Yin Zheng},
      year={2025},
      eprint={2506.19468},
      archivePrefix={arXiv},
      primaryClass={cs.CL},
      url={https://arxiv.org/abs/2506.19468}, 
}

@article{Xu_2025,
   title={A survey on multilingual large language models: corpora, alignment, and bias},
   volume={19},
   ISSN={2095-2236},
   url={http://dx.doi.org/10.1007/s11704-024-40579-4},
   DOI={10.1007/s11704-024-40579-4},
   number={11},
   journal={Frontiers of Computer Science},
   publisher={Springer Science and Business Media LLC},
   author={Xu, Yuemei and Hu, Ling and Zhao, Jiayi and Qiu, Zihan and Xu, Kexin and Ye, Yuqi and Gu, Hanwen},
   year={2025},
   month=Apr }

@misc{hengle2024multilingualneedlehaystackinvestigating,
      title={Multilingual Needle in a Haystack: Investigating Long-Context Behavior of Multilingual Large Language Models}, 
      author={Amey Hengle and Prasoon Bajpai and Soham Dan and Tanmoy Chakraborty},
      year={2024},
      eprint={2408.10151},
      archivePrefix={arXiv},
      primaryClass={cs.CL},
      url={https://arxiv.org/abs/2408.10151}, 
}

@misc{zhao2024largelanguagemodelshandle,
      title={How do Large Language Models Handle Multilingualism?}, 
      author={Yiran Zhao and Wenxuan Zhang and Guizhen Chen and Kenji Kawaguchi and Lidong Bing},
      year={2024},
      eprint={2402.18815},
      archivePrefix={arXiv},
      primaryClass={cs.CL},
      url={https://arxiv.org/abs/2402.18815}, 
}

@inproceedings{tang-etal-2024-language,
    title = "Language-Specific Neurons: The Key to Multilingual Capabilities in Large Language Models",
    author = "Tang, Tianyi  and
      Luo, Wenyang  and
      Huang, Haoyang  and
      Zhang, Dongdong  and
      Wang, Xiaolei  and
      Zhao, Xin  and
      Wei, Furu  and
      Wen, Ji-Rong",
    editor = "Ku, Lun-Wei  and
      Martins, Andre  and
      Srikumar, Vivek",
    booktitle = "Proceedings of the 62nd Annual Meeting of the Association for Computational Linguistics (Volume 1: Long Papers)",
    month = aug,
    year = "2024",
    address = "Bangkok, Thailand",
    publisher = "Association for Computational Linguistics",
    url = "https://aclanthology.org/2024.acl-long.309/",
    doi = "10.18653/v1/2024.acl-long.309",
    pages = "5701--5715"
}

@misc{gu2025surveyllmasajudge,
      title={A Survey on LLM-as-a-Judge}, 
      author={Jiawei Gu and Xuhui Jiang and Zhichao Shi and Hexiang Tan and Xuehao Zhai and Chengjin Xu and Wei Li and Yinghan Shen and Shengjie Ma and Honghao Liu and Saizhuo Wang and Kun Zhang and Yuanzhuo Wang and Wen Gao and Lionel Ni and Jian Guo},
      year={2025},
      eprint={2411.15594},
      archivePrefix={arXiv},
      primaryClass={cs.CL},
      url={https://arxiv.org/abs/2411.15594}, 
}

@misc{haas2026simpleqaverifiedreliablefactuality,
      title={SimpleQA Verified: A Reliable Factuality Benchmark to Measure Parametric Knowledge}, 
      author={Lukas Haas and Gal Yona and Giovanni D'Antonio and Sasha Goldshtein and Dipanjan Das},
      year={2026},
      eprint={2509.07968},
      archivePrefix={arXiv},
      primaryClass={cs.CL},
      url={https://arxiv.org/abs/2509.07968}, 
}

@misc{google_deepmind_gemma3_model_card_2025,
  author       = {{Google DeepMind}},
  title        = {{Gemma 3 model card}},
  year         = {2025},
  month        = aug,
  day          = {14},
  howpublished = {\url{https://ai.google.dev/gemma/docs/core/model_card_3}},
  note         = {Google AI for Developers model card}
}

@misc{hillebrandt_knowledge_cutoff_date_2026,
  author       = {Hillebrandt, Finn},
  title        = {Knowledge Cutoff Date: Definition \& Explanation},
  year         = {2026},
  month        = may,
  day          = {21},
  howpublished = {\url{https://www.gradually.ai/en/ai-glossary/knowledge-cutoff-date/}},
  note         = {Gradually.ai}
}

@misc{google_gemini_2_5_flash_2026,
  author       = {{Google}},
  title        = {{Gemini 2.5 Flash}},
  year         = {2026},
  month        = apr,
  day          = {28},
  howpublished = {\url{https://ai.google.dev/gemini-api/docs/models/gemini-2.5-flash}},
  note         = {Google AI for Developers documentation. Knowledge cutoff: January 2025}
}

@misc{google_gemini3_developer_guide_2026,
  author       = {{Google}},
  title        = {{Gemini 3 Developer Guide}},
  year         = {2026},
  month        = may,
  day          = {7},
  howpublished = {\url{https://ai.google.dev/gemini-api/docs/gemini-3}},
  note         = {Google AI for Developers documentation. Knowledge cutoff: January 2025}
}

@misc{google_deepmind_gemma4_model_card_2026,
  author       = {{Google DeepMind}},
  title        = {{Gemma 4 model card}},
  year         = {2026},
  month        = apr,
  day          = {17},
  howpublished = {\url{https://ai.google.dev/gemma/docs/core/model_card_4}},
  note         = {Google AI for Developers model card. Training data cutoff: January 2025}
}

@misc{anthropic_transparency_hub,
  author       = {{Anthropic}},
  title        = {{Anthropic's Transparency Hub}},
  year         = {n.d.},
  howpublished = {\url{https://www.anthropic.com/transparency}},
  note         = {Accessed: 2026-05-25}
}

@Article{         harris2020array,
 title         = {Array programming with {NumPy}},
 author        = {Charles R. Harris and K. Jarrod Millman and St{\'{e}}fan J.
                 van der Walt and Ralf Gommers and Pauli Virtanen and David
                 Cournapeau and Eric Wieser and Julian Taylor and Sebastian
                 Berg and Nathaniel J. Smith and Robert Kern and Matti Picus
                 and Stephan Hoyer and Marten H. van Kerkwijk and Matthew
                 Brett and Allan Haldane and Jaime Fern{\'{a}}ndez del
                 R{\'{i}}o and Mark Wiebe and Pearu Peterson and Pierre
                 G{\'{e}}rard-Marchant and Kevin Sheppard and Tyler Reddy and
                 Warren Weckesser and Hameer Abbasi and Christoph Gohlke and
                 Travis E. Oliphant},
 year          = {2020},
 month         = sep,
 journal       = {Nature},
 volume        = {585},
 number        = {7825},
 pages         = {357--362},
 doi           = {10.1038/s41586-020-2649-2},
 publisher     = {Springer Science and Business Media {LLC}},
 url           = {https://doi.org/10.1038/s41586-020-2649-2}
}

@inproceedings{taubench,
    title = {$\tau$-Bench: A Benchmark for Tool-agent-user Interaction in Real-world Domains},
    author = {Shunyu Yao and Noah Shinn and Pedram Razavi and Karthik Narasimhan},
    booktitle = {International Conference on Learning Representations (ICLR)},
    year = {2025}
}

@misc{wataoka2025selfpreferencebiasllmasajudge,
      title={Self-Preference Bias in LLM-as-a-Judge}, 
      author={Koki Wataoka and Tsubasa Takahashi and Ryokan Ri},
      year={2025},
      eprint={2410.21819},
      archivePrefix={arXiv},
      primaryClass={cs.CL},
      url={https://arxiv.org/abs/2410.21819}, 
}
